\documentclass{article}

\usepackage[nolist,nohyperlinks]{acronym}
\usepackage{jfrExamplee}
\usepackage{algorithm}
\usepackage{algorithmic}
\usepackage{graphicx}
\usepackage{subfig}
\usepackage{apalike}
\usepackage{setspace}
\usepackage{hyperref}
\usepackage[normalem]{ulem} 
\usepackage{amsmath}
\usepackage{amssymb,amsfonts,bm,bbm}
\usepackage{multicol}
\usepackage{comment}
\usepackage{bm}
\usepackage{color}
\usepackage{hyperref}
\usepackage{float}
\hypersetup{pdfborder = 0 0 0}
\usepackage{microtype} 
\usepackage[justification=justified,font=small, format=plain]{caption}
\usepackage{tabularx}
\usepackage{indentfirst}
\usepackage{tikz}
\usepackage{pgfgantt}
\usepackage{fixltx2e}
\usepackage{xcolor}
\usepackage{pgfplots}
\pgfplotsset{compat=1.12} 
\usetikzlibrary{calc, arrows,arrows.meta,positioning,graphs,automata, shapes, shapes.geometric, shapes.multipart}
\usetikzlibrary{tikzmark,arrows.meta}
\tikzset{
    >=stealth',
    defnode/.style={
           rectangle,
           rounded corners,
           draw=black, thick,
           minimum height=2em,
           text centered},
    defedge/.style={
           ->,
           thick,
           shorten <=2pt,
           shorten >=2pt,}
}
\tikzstyle{component} = [draw, fill=blue!20, align=center, text centered, rounded corners]
\tikzstyle{human_stage} = [draw, fill=green!15, minimum width=5em, text centered, rounded corners]
\pgfdeclarelayer{background}
\pgfdeclarelayer{foreground}
\pgfsetlayers{background,main,foreground}
\RequirePackage{suffix}
\RequirePackage{soul}
\usepackage{multirow}
\usepackage{makecell} 

\usepackage{xcolor}

\newcommand{\neweq}[2]{
\begin{equation}
 \label{e:#1}
 #2
\end{equation}
}

\WithSuffix\newcommand\neweq*[1]{
	$$
		#1
	$$
}

\makeatletter

\newcounter{oli}
\newcounter{olii}[oli]
\providecommand\theleveli{\@arabic\c@oli}%
\providecommand\thelevelii{\theleveli.\@arabic\c@olii}%
\makeatother

\RequirePackage{amsfonts}
\RequirePackage{amsmath}

\RequirePackage{graphicx}
\RequirePackage[verbose]{wrapfig}
\RequirePackage{float}
\RequirePackage{subfig}

\newcommand{\twofer}[6]{
	\begin{figure}[!htb]
			\centering
			\subfloat[#5]{{\includegraphics[width=.47\textwidth]{#1}\label{f:#3_l} }}%
			\hspace{10pt}
			\subfloat[#6]{{\includegraphics[width=.47\textwidth]{#2}\label{f:#3_r} }}%
			\caption{#4}
			\label{f:#3}
	\end{figure}
}

\newcommand{\twoferheight}[7]{
	\begin{figure}[!htb]
			\centering
			\subfloat[#5]{{\includegraphics[width=.47\textwidth, height=#7]{#1}\label{f:#3_l} }}%
			\hspace{10pt}
			\subfloat[#6]{{\includegraphics[width=.47\textwidth, height=#7]{#2}\label{f:#3_r} }}%
			\caption{#4}
			\label{f:#3}
	\end{figure}
}

\newcommand{\threefer}[5]{
	\begin{figure*}[!htb]
			\centering
			\subfloat[]{{\includegraphics[width=.32\textwidth]{#1}\label{f:#4_l} }}%
			\subfloat[]{{\includegraphics[width=.32\textwidth]{#2}\label{f:#4_c} }}%
			\subfloat[]{{\includegraphics[width=.32\textwidth]{#3}\label{f:#4_r} }}%
			\caption{#5}
			\label{f:#4}
	\end{figure*}
}

\newcommand{\sidebyside}[6]{
	\begin{figure}[!htb]
		\begin{minipage}[b]{0.48\linewidth}
			\centering
			\includegraphics[width=\textwidth]{#1}
			\caption{#3}
			\label{f:#2}
		\end{minipage}
		\hspace{0.02\linewidth}
		\begin{minipage}[b]{0.48\linewidth}
			\centering
			\includegraphics[width=\textwidth]{#4}
			\caption{#6}
			\label{f:#5}
		\end{minipage}
	\end{figure}
}

\newcommand{\centerfig}[3]{
	\begin{figure}[ht]
		{\centering
			\includegraphics[width=\textwidth]{#1}
			\caption{#3}
			\label{f:#2}
		}
	\end{figure}
}
\newcommand{\scaledfig}[4]{
	\begin{figure}[!ht]
		{\centering
			\includegraphics[width=#4\textwidth]{#1}
			\caption{#3}
			\vspace{-10pt}
			\label{f:#2}
		}
	\end{figure}
}

\newcommand{\tikzfigpos}[3]{
\begin{figure}[!htb]
	\begin{centering}
	#3
	\caption{#2}
	\label{f:#1}
    \end{centering}
\end{figure}
}

\newcommand{\fig}[1]{Figure \ref{f:#1}}

\def\fps@figure{htbp}
\def\fps@table{htbp}

\usepackage{colortbl}

\makeatletter

\newcounter{hypo}
\newcounter{subhypo}[hypo]
\makeatother

\definecolor{lgreen} {RGB}{180,210,100}
\definecolor{dblue}  {RGB}{20,66,129}
\definecolor{ddblue} {RGB}{11,36,69}
\definecolor{lred}   {RGB}{220,0,0}
\definecolor{nred}   {RGB}{224,0,0}
\definecolor{norange}{RGB}{230,120,20}
\definecolor{nyellow}{RGB}{255,221,0}
\definecolor{ngreen} {RGB}{98,158,31}
\definecolor{dgreen} {RGB}{78,138,21}
\definecolor{nblue}  {RGB}{28,130,185}
\definecolor{jblue}  {RGB}{20,50,100}

\definecolor{GreenYellow}       {RGB}{217, 229, 6} 	    
\definecolor{Yellow}            {RGB}{254, 223, 0} 	    
\definecolor{Goldenrod}         {RGB}{249, 214, 22} 	
\definecolor{Dandelion}         {RGB}{253, 200, 47} 	
\definecolor{Apricot}           {RGB}{255, 170, 123} 	
\definecolor{Peach}             {RGB}{255, 127, 69} 	
\definecolor{Melon}             {RGB}{255, 129, 141} 	
\definecolor{YellowOrange}      {RGB}{240, 171, 0} 	    
\definecolor{Orange}            {RGB}{255, 88, 0} 	    
\definecolor{BurntOrange}       {RGB}{199, 98, 43} 	    
\definecolor{Bittersweet}       {RGB}{189, 79, 25} 	    
\definecolor{RedOrange}         {RGB}{222, 56, 49} 	    
\definecolor{Mahogany}          {RGB}{152, 50, 34} 	    
\definecolor{Maroon}            {RGB}{152, 30, 50} 	    
\definecolor{BrickRed}          {RGB}{170, 39, 47} 	    
\definecolor{Red}               {RGB}{255, 0, 0}        
\definecolor{BrilliantRed}      {RGB}{237, 41, 57} 	    
\definecolor{OrangeRed}         {RGB}{231, 58, 0} 	    
\definecolor{RubineRed}         {RGB}{202, 0, 93}       
\definecolor{WildStrawberry}    {RGB}{203, 0, 68} 	    
\definecolor{Salmon}            {RGB}{250, 147, 171} 	
\definecolor{CarnationPink}     {RGB}{226, 110, 178} 	
\definecolor{Magenta}           {RGB}{255, 0, 144} 	    
\definecolor{VioletRed}         {RGB}{215, 31, 133} 	
\definecolor{Rhodamine}         {RGB}{224, 17, 157} 	
\definecolor{Mulberry}          {RGB}{163, 26, 126} 	
\definecolor{RedViolet}         {RGB}{161, 0, 107} 	    
\definecolor{Fuchsia}           {RGB}{155, 24, 137} 	
\definecolor{Lavender}          {RGB}{240, 146, 205} 	
\definecolor{Thistle}           {RGB}{222, 129, 211} 	
\definecolor{Orchid}            {RGB}{201, 102, 205} 	
\definecolor{DarkOrchid}        {RGB}{153, 50, 204} 	
\definecolor{Purple}            {RGB}{182, 52, 187} 	
\definecolor{Plum}              {RGB}{79, 50, 76} 	    
\definecolor{Violet}            {RGB}{75, 8, 161} 	    
\definecolor{RoyalPurple}       {RGB}{82, 35, 152} 	    
\definecolor{BlueViolet}        {RGB}{33, 7, 106} 	    
\definecolor{Periwinkle}        {RGB}{136, 132, 213} 	
\definecolor{CadetBlue}	  	    {RGB}{95, 158, 160} 	
\definecolor{CornflowerBlue}  	{RGB}{99, 177, 229} 	
\definecolor{MidnightBlue}	  	{RGB}{0, 65, 101} 	    
\definecolor{NavyBlue}          {RGB}{0, 70, 173}       
\definecolor{RoyalBlue}         {RGB}{0, 35, 102}       
\definecolor{Blue}              {RGB}{0, 24, 168}       
\definecolor{Cerulean}          {RGB}{0, 122, 201}      
\definecolor{Cyan}              {RGB}{0, 159, 218}      
\definecolor{ProcessBlue}       {RGB}{0, 136, 206}      
\definecolor{SkyBlue}           {RGB}{91, 198, 232}     

\definecolor{Turquoise}         {RGB}{0, 255, 239} 	    

\definecolor{TealBlue}          {RGB}{0, 124, 146} 	    
\definecolor{Aquamarine}        {RGB}{0, 148, 179} 	    
\definecolor{BlueGreen}         {RGB}{0, 154, 166} 	    
\definecolor{Emerald}           {RGB}{80, 200, 120} 	
\definecolor{JungleGreen}       {RGB}{0, 115, 99} 	    
\definecolor{SeaGreen}          {RGB}{0, 176, 146} 	    
\definecolor{Green}             {RGB}{0, 173, 131} 	    
\definecolor{ForestGreen}       {RGB}{0, 105, 60} 	    
\definecolor{PineGreen}         {RGB}{0, 98, 101} 	    
\definecolor{LimeGreen}         {RGB}{50, 205, 50} 	    
\definecolor{YellowGreen}       {RGB}{146, 212, 0} 	    
\definecolor{SpringGreen}       {RGB}{201, 221, 3} 	    
\definecolor{OliveGreen}        {RGB}{135, 136, 0} 	    
\definecolor{RawSienna}         {RGB}{149, 82, 20} 	    
\definecolor{Sepia}             {RGB}{98, 60, 27} 	    
\definecolor{Brown}             {RGB}{134, 67, 30}      
\definecolor{Tan}               {RGB}{210, 180, 140}	
\definecolor{Gray}              {RGB}{139, 141, 142} 	

\definecolor{Black}		  	    {RGB}{30, 30, 30}       
\definecolor{White}		  	    {RGB}{255, 255, 255}    

\title{Flexible Supervised Autonomy for Exploration in Subterranean Environments}

\author{
Harel Biggie$^*$\\
Computer Science\\
University of Colorado Boulder\\
\texttt{Harel.Biggie@colorado.edu}
\And
Eugene R. Rush$^*$\\
Mechanical Engineering\\
University of Colorado Boulder\\
\texttt{Eugene.Rush@colorado.edu}
\And
Danny G. Riley\\
Computer Science\\
University of Colorado Boulder\\
\texttt{Dan.Riley@colorado.edu}
\And
Shakeeb Ahmad\\
Aerospace Engineering Sciences\\
University of Colorado Boulder\\
\texttt{Shakeeb.Ahmad@colorado.edu}\\
\And
Michael T. Ohradzansky\\
Aerospace Engineering Sciences\\
University of Colorado Boulder\\
\texttt{Michael.Ohradzansky@colorado.edu}
\And
Kyle Harlow\\
Computer Science\\
University of Colorado Boulder\\
\texttt{Kyle.Harlow@colorado.edu}
\And
Michael J. Miles\\
Mechanical Engineering\\
University of Colorado Boulder\\
\texttt{Mike.Miles@colorado.edu}
\And
Daniel Torres\\
Computer Science\\
University of Colorado Boulder\\
\texttt{Daniel.TorresDominguez@colorado.edu}
\And
Steve McGuire\\
Electrical \& Computer Engineering\\
University of California Santa Cruz\\
\texttt{steve.mcguire@ucsc.edu}
\And
Eric W. Frew\\
Aerospace Engineering Sciences\\
University of Colorado Boulder\\
\texttt{Eric.Frew@colorado.edu}
\And
Christoffer Heckman\\
Computer Science\\
University of Colorado Boulder\\
\texttt{Christoffer.Heckman@colorado.edu}
\And
J.\ Sean Humbert\\
Mechanical Engineering\\
University of Colorado Boulder\\
\texttt{Sean.Humbert@colorado.edu}
}

\begin{document}

\maketitle

\begin{NoHyper}
\def\thefootnote{*}\footnotetext{These authors contributed equally to this work.}
\end{NoHyper}

\begin{abstract}

While the capabilities of autonomous systems have been steadily improving in recent years, these systems still struggle to rapidly explore previously unknown environments without the aid of GPS-assisted navigation. The DARPA Subterranean (SubT) Challenge aimed to fast track the development of autonomous exploration systems by evaluating their performance  in real-world underground search-and-rescue scenarios. Subterranean environments present a plethora of challenges for robotic systems, such as limited communications, complex topology, visually-degraded sensing, and harsh terrain. The presented solution enables long-term autonomy with minimal human supervision by combining a powerful and independent single-agent autonomy stack, with higher level mission management operating over a flexible mesh network. The autonomy suite deployed on quadruped and wheeled robots was fully independent, freeing the human supervision to loosely supervise the mission and make high-impact strategic decisions. We also discuss lessons learned from fielding our system at the SubT Final Event, relating to vehicle versatility, system adaptability, and re-configurable communications.

\end{abstract}


\section{Introduction}

Despite a myriad of developments in sensing, planning, control and state estimation over the last few decades, deploying robots in harsh subterranean environments for the purpose of rapid situational awareness presents a number of new challenges to robot autonomy. Traditionally, robots rely on a number of complex, interconnected sub-processes, such as localization, mapping, and planning, to navigate unknown environments. Maintaining accurate state estimates, a process critical to mapping and exploration, is exceptionally challenging in subterranean environments. GPS is unavailable for obtaining position estimates and visual-based localization methods can be affected by varied lighting conditions and environmental factors such as heavy dust, fog, or smoke. Subterranean environments, such as mines and caves, are often unstructured and contain hazardous obstacles, making navigation with ground vehicles challenging. Additionally, aerial vehicles can be exceptionally difficult to deploy in tight constrained underground spaces due to self-induced propeller wash. The DARPA Subterranean Challenge (SubT) \cite{DARPA2022} aimed to spark new developments in the areas of autonomy, perception, mobility, and networking in subterranean environments. In the following work, a scalable multi-agent autonomy solution for subterranean exploration developed by the University of Colorado's Team MARBLE for the SubT Challenge is presented, along with critical lessons learned and developments made along the way.

DARPA designed the SubT Challenge to simulate search-and-rescue scenarios in unknown subterranean environments, and consisted of three domain-specific circuit events, Tunnel Circuit, Urban Circuit, and Cave Circuit, followed by the Final Event, which was a combination of three subterranean domains. Teams were challenged with developing robot platforms to deploy in each of the events in search of sets of predefined artifacts, such as backpacks or a Bluetooth signal produced by a cell phone. Correct identifications, consisting of an artifact classification and location to within a 5m sphere of the ground truth location, resulted in a point scored. Placement in the competition was determined by the team which could score the most points over a series of one hour deployments. For the Final Event, teams were limited to a single ``human supervisor" who was able to interact with the systems and visualize any incoming data. Adding to the challenge, teams had a limited window of time in which five team members could set up and initialize robots at the entrance to the course.

Team MARBLE's initial approach to subterranean exploration for the Tunnel and Urban circuit events is presented in \cite{MARBLE_multi_agent}. Initially, a graph-based planning and exploration strategy was implemented, the details of which are presented in \cite{Ohradzansky2020}. This solution is suitable for tunnel-like mines that have mostly planar corridor-junction structures, because the environment can be easily represented by a graph of nodes and edges. A scanning lidar was used to center robots in corridors while navigating edge sections as well as avoid obstacles in the environment. However, this approach lacked multi-agent coordination, resulting in significant overlap of explored regions by different agents. For the Urban and Cave circuit events, a three-dimensional volumetric map representation of the environment was generated and used in a frontier-based exploration strategy \cite{ahmad20213d,ahmad_apf}. In this approach, the exploration rate of the robot is maximized using a frontier-based \cite{yamauchi1997frontier} sampling technique and a fast marching cost-to-go calculation \cite{sethian1999level} to select goal poses and plan paths to them in three dimensional space. An artificial potential function based obstacle avoidance algorithm enables the robot to path follow while avoiding small obstacles in the environment. Our initial approach also implemented limited forms of multi-agent coordination in the form of agents sharing goal points and paths.

Other teams developed impressive solutions to the initial Tunnel and Urban Circuit challenges. Team CSIRO, a collaboration between the Commonwealth Scientific and Industrial Research Organization (CSIRO), Emesent, and Georgia Tech, presents a unique homogeneous sensing solution \cite{CSIRO_tunnel_cave}. In this approach, heterogeneous teams of robots, including both ground and aerial platforms, share sensor information as a part of a decentralized multi-agent SLAM system. Initially, exploration was handled through manual waypoints commanded by the human supervisor, but eventually an autonomous exploration algorithm was implemented \cite{Williams_OnlineFrontier2020}. A common perception module, the CatPack, is used across all ground vehicles for easy reuse of the autonomy stack across different platforms. Similar to Team CSIRO, Team CERBERUS also used a heterogeneous team of ground and aerial platforms \cite{CERBERUS_tunnel_cave,papachristos2019,tranzatto_CERBERUS_finals}. In their approach, map information from different agents is fused into an optimized global map that is shared back to the agents \cite{khattak_complementary_2020}. Similar to the work presented in \cite{Ohradzansky2020}, other teams used graph-based planning approaches for global navigation \cite{dang2019explore,dang2020graph}. Other noteworthy teams and their approaches to autonomous subterranean navigation include Team CoSTAR \cite{santamaria-navarro_towards_2020,Ebadi2020,Agha2021NeBulaQF,otsu2020}, CTU-CRAS \cite{Roucek2020}, CMU (Carnegie-Mellon University) \cite{scherer2021}, and NCTU (National Chiao Tung University) \cite{Huang2019}. Additional discussions on the challenges, novel developments, and lessons learned from the Tunnel and Urban circuit events are included in the following works \cite{Miller2020,Lajoie2020}. One common theme common to many of these approaches is the use of heterogeneous teams of agents with multi-modal sensing solutions. By diversifying the sizes, types of locomotion, and sensor modalities of individual robots, teams can be more versatile when faced with varied environments, each with a unique set of challenges. This ability to be flexible and adapt to the needs of the mission is one of Team MARBLE's driving philosophies.

\begin{figure}[!htb]
		\centering
		\subfloat[]{{\includegraphics[width=.40\textwidth]{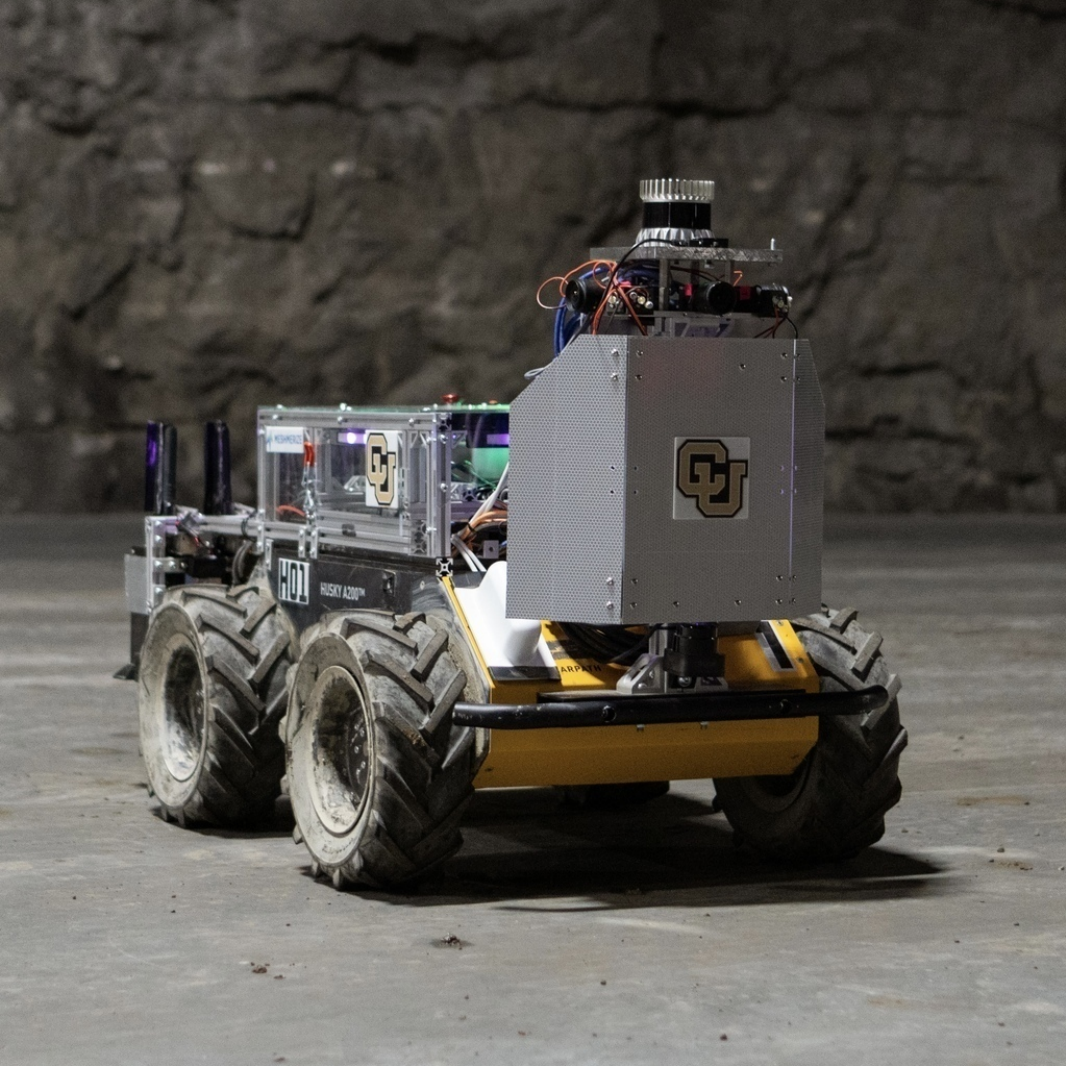}\label{fig:Husky} }}
		\subfloat[]{{\includegraphics[width=.40\textwidth]{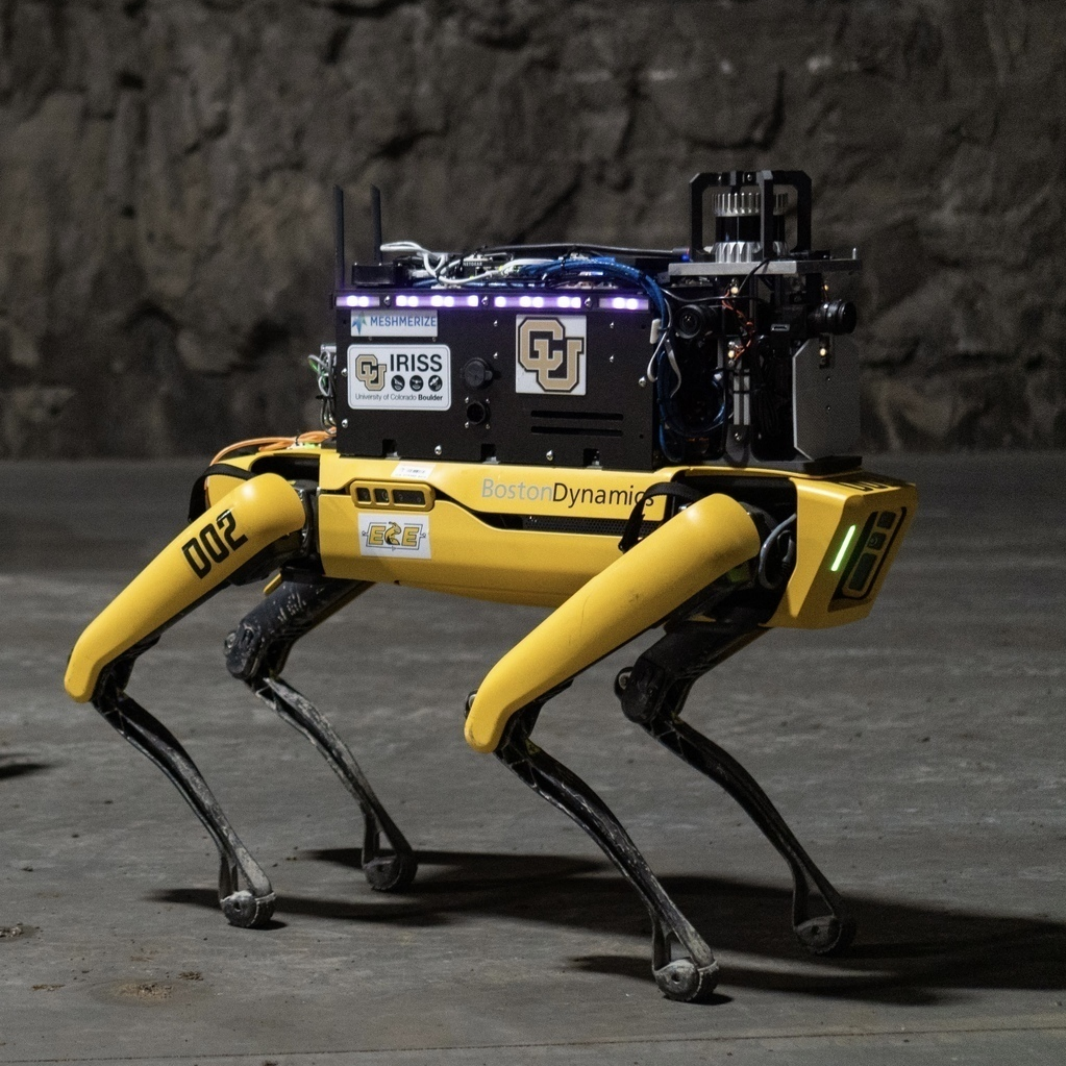}\label{fig:Spot} }}
		\caption{The fleet is composed of two classes of robotic agents: (a) Clearpath Husky A200, (b) Boston Dynamics Spot. Each platform carries a common sensor suite designed for exploration and object detection.}
		\label{fig:platforms}
\end{figure}

The format of the challenge necessitated advancements in platform design, robust communication networks, intelligent planners, and a balance between autonomy and human decision making for robot fleets. Team MARBLE's solution, which was developed over the course of three different circuit events and showcased at the Final Event, led to a third place finish. Specifically, we develop a heterogeneous fleet of autonomous robots, each capable of operating independently of human intervention. Our autonomy-first approach employs a lightweight graph-based planner that scales to large environments, and has been adapted to reason over dynamic environments, such as closing doors and passing agents, as well as take advantage of multi-agent coordination. Information sharing between agents is accomplished via a custom mesh networking solution with fast reconnect times and configurable message prioritization. The same network provides the human supervisor real-time visibility and control of the mission. Our autonomy-first philosophy inspired an autonomous mission management system that frees the human supervisor from agent-level micromanagement and deepens the opportunity to strategically accomplish mission goals. In this work we will present each of the components of our proposed autonomy system, as well as a detailed performance analysis of our solution and lessons learned along the way.

This paper is organized with the following structure. First, an overview of our system is provided in Section \ref{sec:system_overview}. The robot platforms developed for the Final Event are described in Section \ref{sec:platforms}, followed by a description of the localization system in Section \ref{sec:localization}, and the artifact detection system in Section \ref{sec:object_detection}. The multi-agent components of our system include the volumetric mapping pipeline in Section \ref{sec:mapping} and the graph-based path planning over those maps in Section \ref{sec:planning}. Information transmitted among agents, as well as to and from the human supervisor, is mediated by the wireless mesh network communications system described in Section \ref{sec:comm_systems}, and handled by the autonomous mission management system described in Section \ref{sec:mission_management}. Finally, we analyze how these systems performed in the DARPA SubT Challenge Final Event in Section \ref{sec:final_event} and discuss lessons learned in Section \ref{sec:lessons}.


\section{System Overview}\label{sec:system_overview}

In the following subsections we present a high-level description of Team MARBLE's approach to the DARPA Subterranean Challenge. First, the general concept of operation is described, followed by an overview of each of the major components of the developed autonomy solution. A full high level summary of the autonomy system can be seen in Figure \ref{f:block_diagram}.

\subsection{Concept of Operations}\label{ssec:concept_of_op}

Team MARBLE has emphasized development of multi-agent autonomy solutions that are able to operate without requiring intervention from a human operator. This aligns with the goals of the DARPA Subterranean Challenge where intermittent or unavailable communications with agents from a base station or between agents is expected. Therefore, our solution is centered around robust single-agent autonomy, where independent robots are able to explore unknown environments and report back to a base station with collected information about the environment including map data and detected artifact locations. In communication-limited environments where information sharing with other agents and the human supervisor may not be available, our agents persist and continue to execute the mission.

While it is important for single, independent agents to be able to explore autonomously, our solution incorporates several multi-agent components to improve exploration efficiency. Our fleet is also designed to be opportunistic, capitalizing on communication links when they are available to amplify fleet performance. Multi-agent coordination is an auxiliary capability that reduces redundant efforts when agents enter communication range with one another, and inform each other where they have been and where they plan to go next.  

Our system's performance can further be improved when communications are available which enables the human supervisor to have a holistic perspective of the specific search-and-rescue scenario. This holistic perspective empowers the human to make high-level contextual decisions through two types of intervention: directing the agent to a specific location by commanding a high-level waypoint, or teleoperating the agent by commanding low-level velocity signals. During the 60-minute final event prize run, Team MARBLE's robot fleet was completely autonomous, with the exception of five strategic low-level human supervisor interventions. The balance between human input and autonomy is further discussed in Section \ref{sec:mission_management}.

\scaledfig{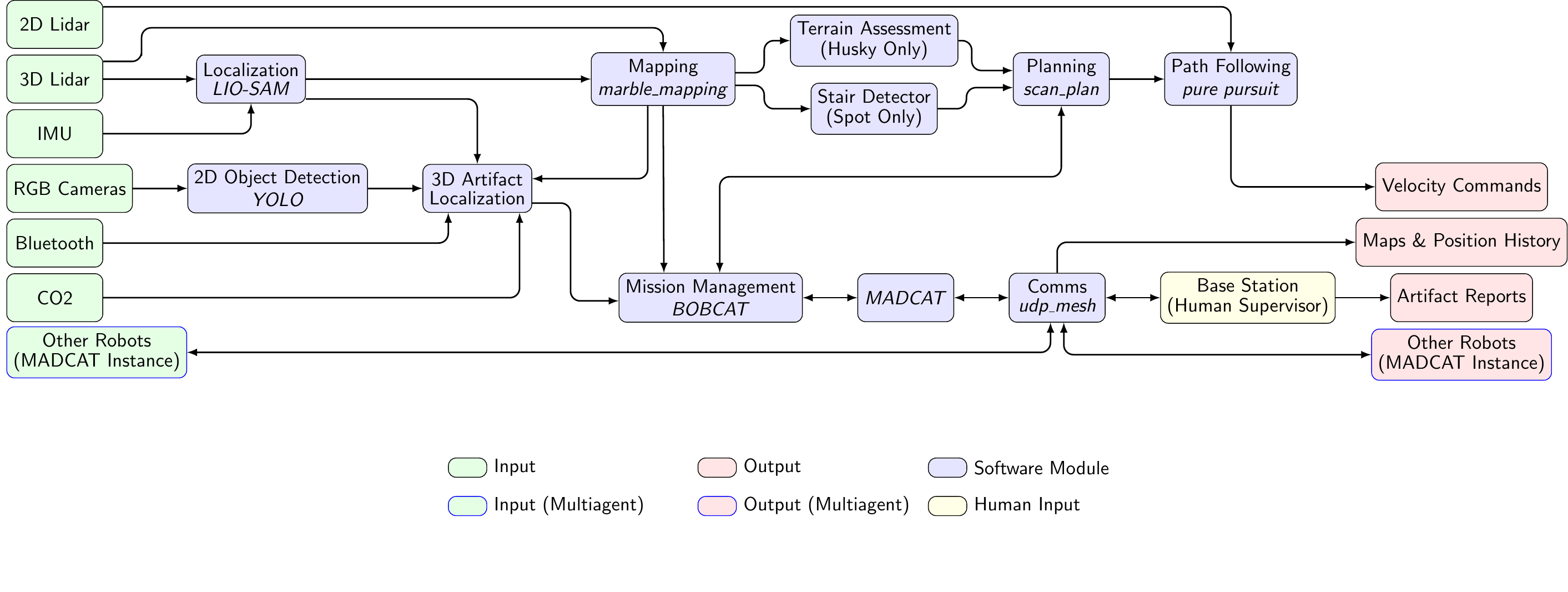}{block_diagram}{Overall block diagram showing the high level functionality of the autonomy stack. Inputs are shown in green and outputs are shown in red. Software package names are \emph{italicized} and inputs \& outputs which are shared between agents are outlined in blue. Terrain assessment and stair detection both add semantic information to the map but are only run on the Husky and Spot respectively.}{1.0}

\subsection{Perception}

A modular perception suite, shown in Figure \ref{fig:sensor_head}, was designed as the basis for the autonomy stack. The primary sensor is the Ouster OS1-64 lidar (Light Detection and Ranging), which provides 3D point clouds for mapping, localization, semantic mapping, and obstacle avoidance. A LORD Microstrain 3DM-GX5-15 IMU (Inertial Measurement Unit) is used to measure linear and angular acceleration of the sensor head for use in lidar-inertial state estimation. To identify visual artifacts, the systems are equipped with several FLIR Blackfly PGE-05S2C-CS cameras and an array of 5W dimmable LEDs for self illumination. The Husky platforms were equipped with four cameras facing forward, backward, to the left, and to the right. The Spot robots had a similar configuration, save omitting the rear camera due to occlusions caused by the custom-built compute and power management system.

\subsection{Localization}

Localization provides consistent pose information for many downstream autonomy processes including volumetric mapping, path planning, artifact detection, and multi-agent coordination. However, ensuring reliable localization is difficult in austere underground environments. Because conventional vision-based solutions can be unreliable due to dark, feature-poor settings, Team MARBLE utilized lidar-based methods and specifically tested and integrated LIO-SAM \cite{shan_liosam_2020}, which has fast online loop closures during long-duration missions. Several modifications were made to the system to improve localization accuracy and reliability, which are further discussed in Section \ref{sec:localization}. Methods used to align the robots with into a common reference frame are also presented in this section.

\subsection{Exploration}

The exploration algorithm generates safe and traversable paths that lead agents toward unexplored areas of previously unseen environments. The developed sampling based path planning algorithm is designed to be lightweight, so that it can operate on rapid exploration timescales, regardless of the extent of the environment. Computational efficiency is achieved by employing a bifurcated local-global graph for sampling unseen frontiers as well as a \emph{good enough} strategy for final selection. In such time-constrained search and rescue settings, even humans will often make rapid decisions rather than dwell for long periods of time to make globally optimal ones. The details of the planning algorithm is described in Section \ref{sec:planning}.

The planner is also constructed to be flexible, so that additional capabilities could be scaffolded on top of the core algorithm. Integration of the planning algorithm with semantic mapping was critical for rough terrain and stair traversal, as explained in Section \ref{sec:mapping}. Section \ref{ssec:planning_dynamic_replanning} explains how this planning algorithm re-plans in dynamic environments, whether due to doorways that are being opened or closed, or fellow agents that are passing by. This capability is crucial for robust operation in real-world environments, which cannot assumed to be static. Finally, the planner is able to run more efficiently with multiple robots using multi-agent coordination is covered in Section \ref{ssec:planning_multiagent_coordination}.

Agents follow paths via a modified pure pursuit controller \cite{purepursuit1992}. The yaw rate command is computed by comparing the current agent's heading against a lookahead point that is a fixed distance along the path. Forward speed is regulated based on local proximity to obstacles in the environment and the relative heading error to the lookahead point. This results in slower speeds when agents are in cluttered environments or experiencing large heading errors. A 2D (RP Lidar) was used for local obstacle avoidance on the Husky platform and the Spot platform had built in obstacle avoidance.

\subsection{Mapping}

Team MARBLE's mapping framework is based on the open source \emph{Octomap} package \cite{hornung2013octomap}, and has been customized with additional capabilities including map merging, transmission of difference maps, and encoding of semantic information. The core of the mapping framework is a log-odds based probability metric for occupied and unoccupied voxels or cells. These cells provide a 3D representation of the environment that is later used for navigation. This flexible framework enabled transmission of key environmental features such as rough terrain and the location of stairs efficiently through a bandwidth limited communication system. The details of our mapping system are provided in Section \ref{sec:mapping}. 

\subsection{Artifact Detection}

The artifact detection system precisely localizes visual artifacts using RGB sensors for visual classification and detection and lidar for the depth estimation. Non-visual artifacts such as cellphones and gas are localized based on the position of the robot. A weighted median filter fuses together all detections which are sent to the human supervisor for final validation as described in Section \ref{sec:object_detection}.

\subsection{Communication Systems}
A mesh networking solution transmits data between robots and back the base station for the human supervisor to review. Standard 2.4ghz 802.11 wireless radios based on the \textit{ath9k} chipset are used for the physical layer. The wireless radios are embedded in beacons that can be deployed from the back of the Husky platforms, allowing for ad-hoc mesh networks to be established. Meshing technology was provided by Meshmerize \cite{pandi2019meshmerize} and a custom UDP based transport layer (\textit{udp\_mesh}) was developed. Details of this innovative layer can be seen in Section \ref{sec:comm_systems}.

\subsection{Multi-Agent Coordination}
\label{ssec:multi_agent}
Exploring unknown environments with multiple agents can be made more efficient through coordination, especially when agents are not within communication range. Sharing information across agents, such as explored regions, discovered artifacts, and current behavioral states, allows for more intelligent management of multi-agent exploration. The framework called Multi-Agent Data Collaboration for Autonomous Teams (MADCAT), provides the multi-agent data sharing capabilities required for the Subterranean Challenge mission \cite{Riley2021}, including transmission of relevant coordination data and maps, as well as map merging functionality and decision making for each agent. Additionally, MADCAT implements Behaviors, Objectives and Binary states for Cooperative Autonomous Tasks (BOBCAT), originally presented in \cite{Riley2022}, for high-level autonomy, decision making, and interfacing with the human supervisor. The MADCAT algorithm is discussed in more detail in Section \ref{sec:mission_management}.

\subsection{Mission Management}
The human supervisor is able to monitor the fleet's progression through the unknown subterranean environment using a custom GUI operating on a computer at the entrance to the environment (base station). Current mission status of all agents in the field (`Reporting', `Exploring', `Home', etc.) as well as their location in the global map are displayed whenever robots are within communication range of the base station. Additionally, the goal point and goal path for each robot is visible, allowing the supervisor to see the intent of each robot. The human supervisor is able to take over control of a given agent by either sending manual goal points or tele-operating the vehicle using a joystick interface with the base station. Reported artifacts are displayed per robot, with the type (survivor, cell phone, backpack, helmet, rope), position, confidence, submission result, and an image. The GUI also enables the modification of artifact classes and locations prior to submission to the DARPA scoring server. An example of the GUI interface is shown in Figure \ref{f:marble_gui}.

\scaledfig{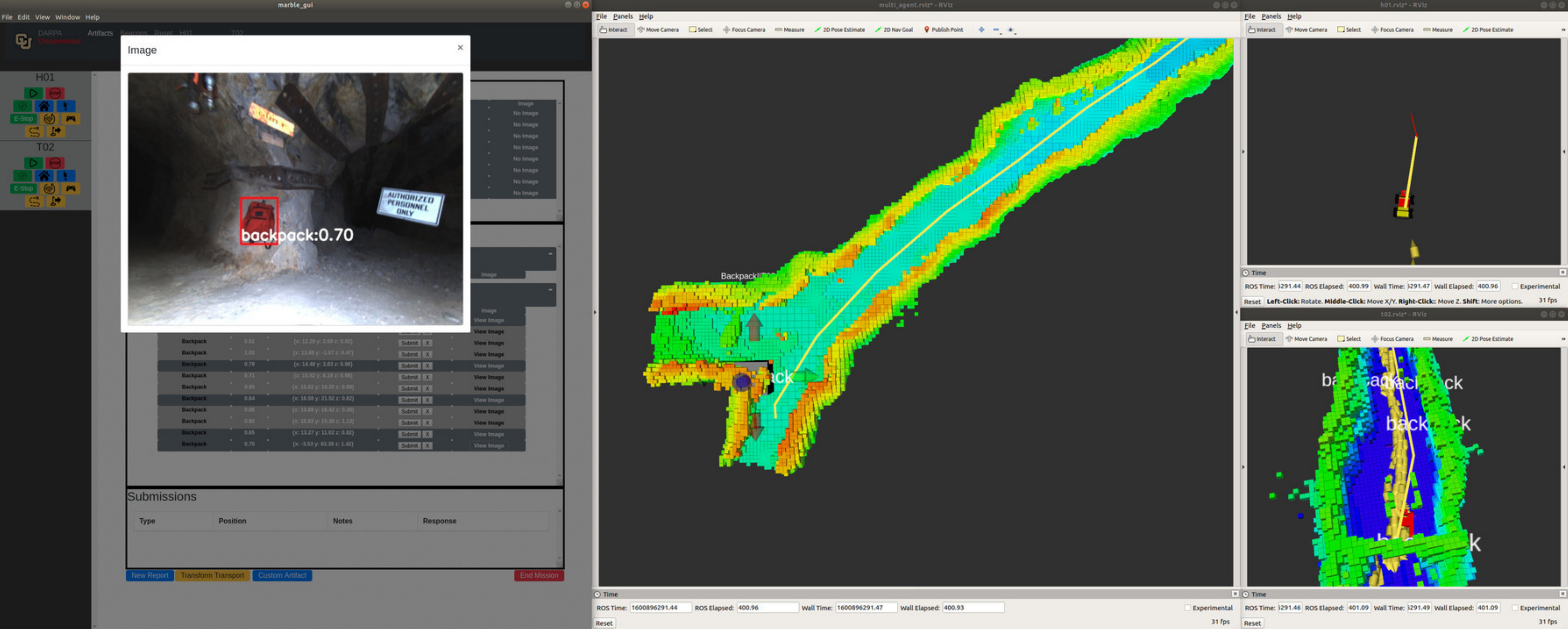}{marble_gui}{Example of the Human Supervisor's interface.  The custom GUI is on the left, showing a received artifact image.  The middle is the multi-agent RViz view, with all robots and the complete merged map.  The right is a third-person follower for each robot (two in this case) with that robot's original unmerged map.}{1.0}


\section{Platform Development} \label{sec:platforms}

For the final event of the SubT Challenge, Team MARBLE deployed a heterogeneous fleet consisting of two Clearpath Husky A200s (H01, H02) and two Boston Dynamics Spots (D01, D02). Examples of each platform are shown in Figure \ref{fig:platforms}. The Husky platforms are four-wheeled skid-steer ground vehicles capable of carrying heavy payloads, while the Spot quadrupedal ``dog" platforms are agile, capable of climbing staircases and traversing uneven terrain. The Husky platform is robust and stable, with its generous payload budget allowing it to carry six communication beacons. The intended deployment strategy is to first deploy the Spot platforms to maximize exploration, and then the Husky platforms to establish the mesh communication network.

\begin{figure}[!htb]
		\centering
		\subfloat[]{{\includegraphics[width=0.49\textwidth]{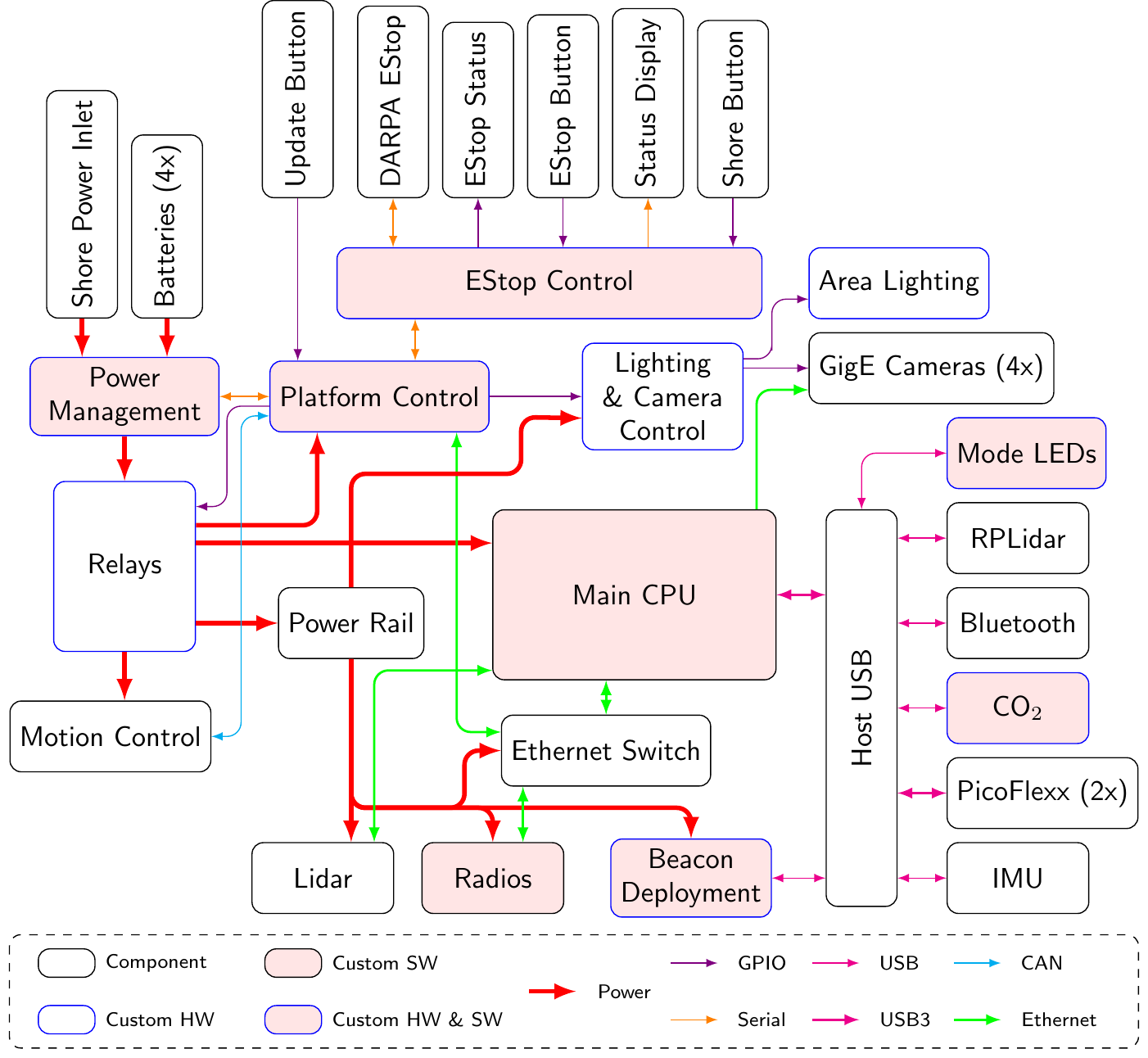}\label{fig:platform_ugv_routing} }}
		\subfloat[]{{\includegraphics[width=0.49\textwidth]{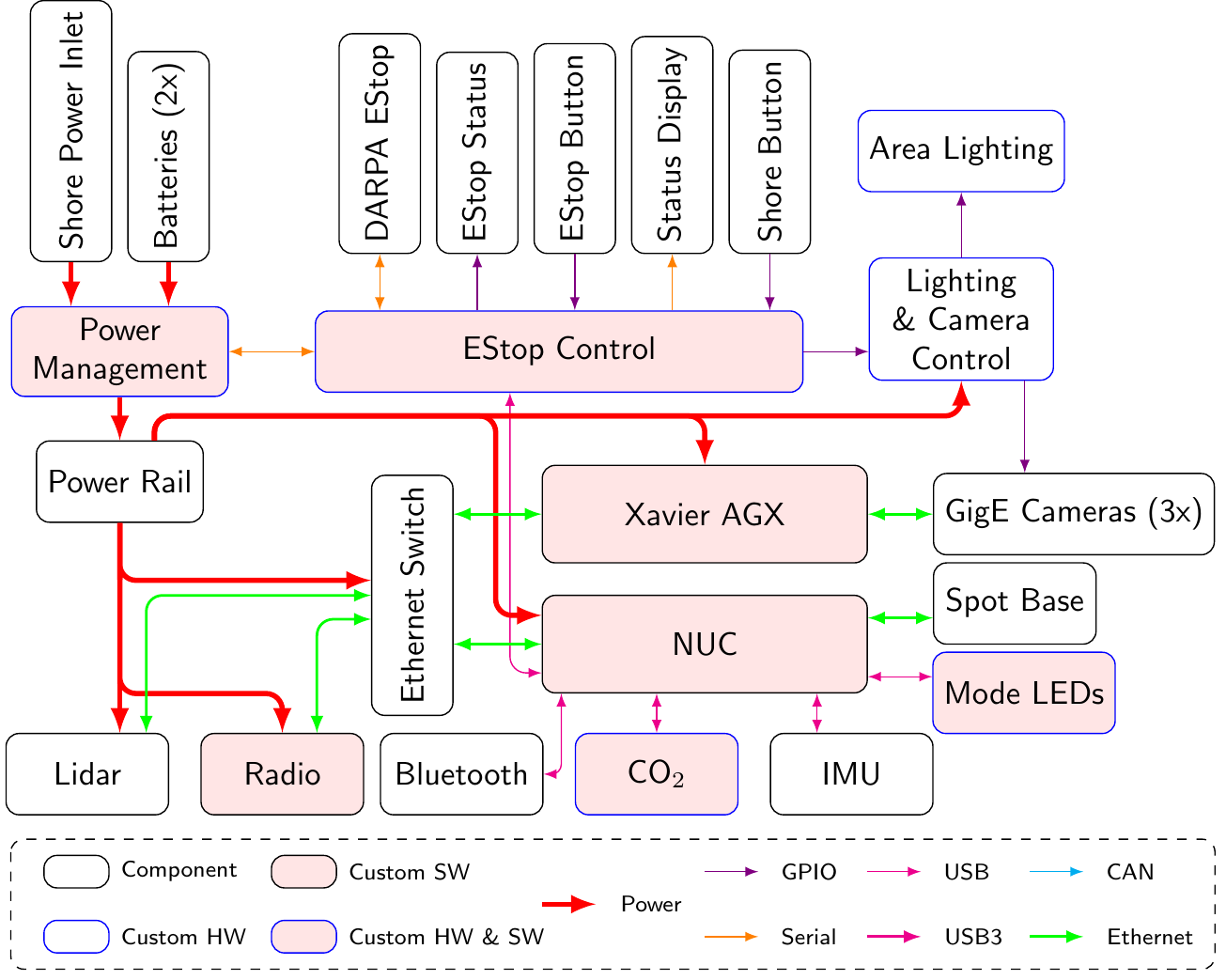}\label{fig:platform_spot_routing} }}\\
    \caption{Power and signal routing diagrams of customized (a) Husky and (b) Spot platforms.}
	\label{fig:platform_routing}
\end{figure}

For processing power, each Husky is equipped with a 32-core AMD Ryzen CPU equipped with 128GB of RAM and 4TB of SSD storage, integrated into a complete platform as shown in Figure \ref{fig:platform_ugv_routing}. Dual NVIDIA GTX 1650 GPUs were used to accelerate object detection inference speed. The primary computer on the Spot platforms is an AMD Ryzen 5800U with 64GB of RAM and 2TB of storage which is paired with a Jetson Xavier AGX to process the camera streams and perform artifact detection, following a similar integration as shown in Figure \ref{fig:platform_spot_routing}. Many purpose-built components are common between platforms to reduce field maintenance efforts and platform-specific code. Each system is outfitted with a custom power system, discussed in Section \ref{ssec:power_electronics}, which enables the ability to switch from a wired shore power supply to the onboard computer batteries. This leads to more efficient use of the onboard batteries, which are a limiting factor in the duration of field testing deployments. 

\subsection{Communication Beacons}

\begin{figure}[!htb]
		\centering
		\subfloat[]{{\includegraphics[width=0.49\textwidth]{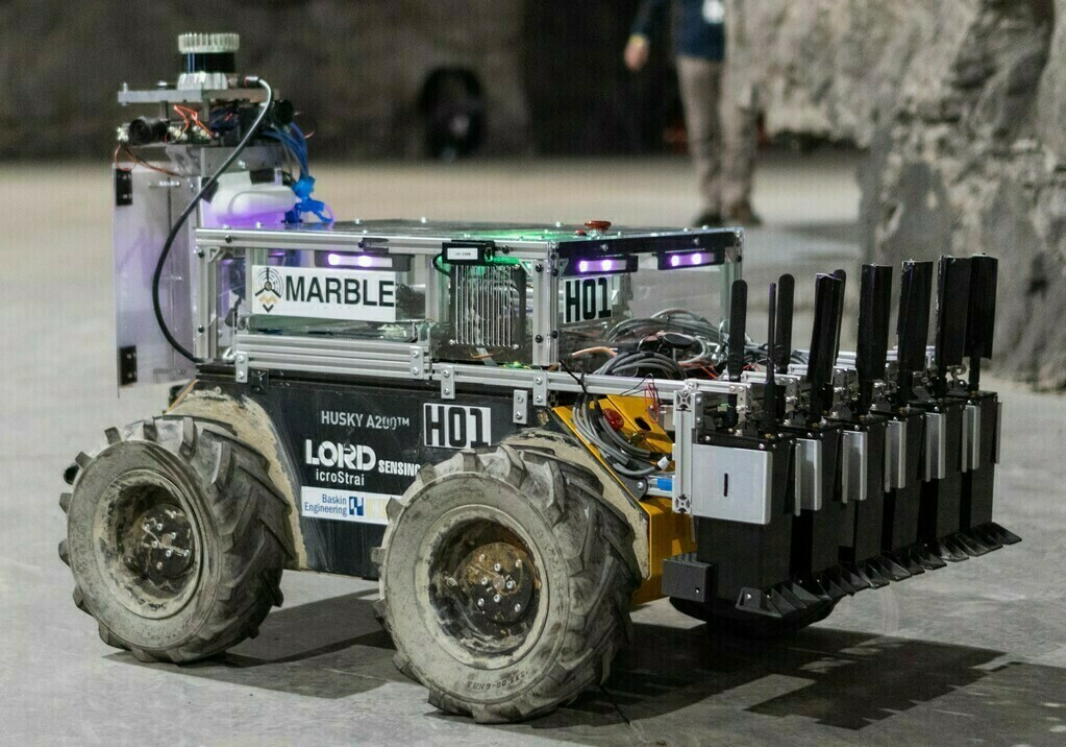}\label{fig:husky_with_beacons} }}
		\subfloat[]{{\includegraphics[width=0.49\textwidth]{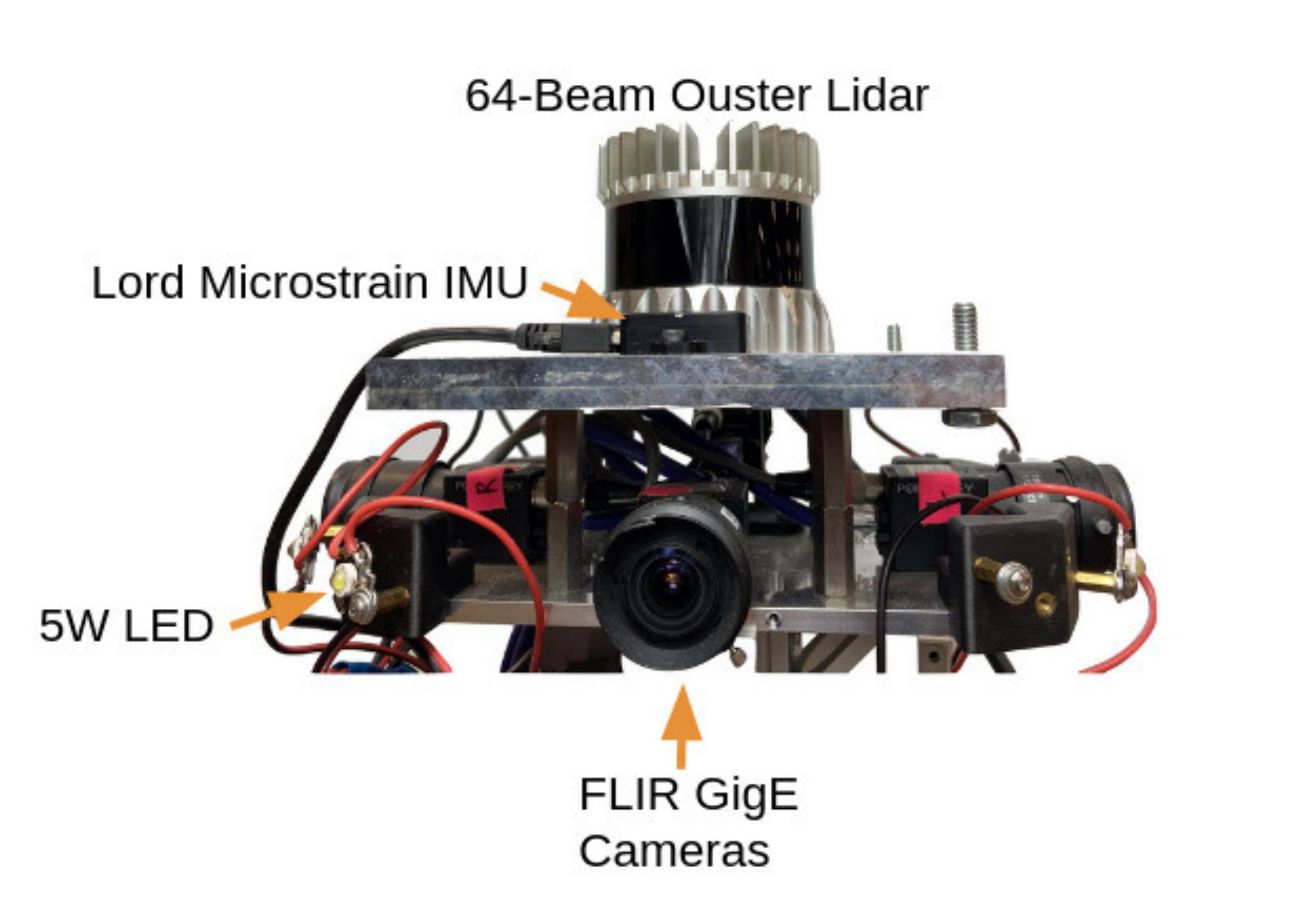}\label{fig:sensor_head} }}\\
    \caption{Final robot configurations, with (a) a deployment mechanism loaded with six communication beacons on Husky vehicles at Final Event and (b) a modular perception suite installed on both Huskies and Spots.}
	\label{fig:husky_final_sensor}
\end{figure}

Underground environments provide limited line-of-sight capabilities for wireless communications. As a result, Team MARBLE developed custom communication beacons to complement the custom multi-robot coordination solution. This allows for robots to share information with the base station and other robots in the field. Each Husky platform is capable of carrying six beacons, each containing a single 2W Doodlelabs 802.11n radio as seen in Figure \ref{fig:husky_with_beacons}. The autonomous beacon deployment mechanism relies on a latching solenoid release coupled with a novel passive system to gently lower each beacon to the ground to ensure maximum antenna height. Additional design details of the communication beacons can be found in Section \ref{ssec:sup_comm_beacon_design} of the Appendix.

\subsection{Power and Platform Control Systems} \label{ssec:power_electronics}

In order to support each platform's sensor and processing needs, as well as meet DARPA equipment requirements for emergency stop systems, the systems integration efforts relied on several custom hardware and software components. Where feasible, these components are shared between the Husky, shown in Figure \ref{fig:Husky}, and the Spot platforms, shown in Figure \ref{fig:Spot}, reducing the development and validation efforts, as well as team member operational training requirements. 

Of the custom capabilities developed, the power management subsystem deserves special mention. This component implements a hardware-interlocked, ideal diode system to permit downstream electronics to source power from either a wall-connected power supply, referred to as shore power, or onboard batteries. By switching to shore power, onboard systems can remain powered for development, testing, and analysis while the batteries are charged without carrying load. Further, the ideal diode component allows the battery packs to load share and charge independently. In contrast to a bus-tied battery system, this ideal diode design prevents high-energy charge equalization between packs and allowed each battery pack's onboard management board to function independently. The system also enables live monitoring of current consumption and battery voltage, as well as intelligent e-stop management which ensures the robot cannot exit the emergency stopped state while connected to shore power.

Emergency stop requirements dictated that each platform needed the ability to be stopped by a physical button, software, and via a DARPA deployed Xbee network. On the Husky control system the emergency stop system was integrated directly into the base controller. In contrast, the Spot platform's emergency stop tied into the available API to issue a the robot a ``sit'' command before terminating power to the motors which allowed the robot to be stopped gracefully.

Several important lessons learned emerged after three years of platform architecture development. We have highlighted the most critical lessons below and provide more details about the design and consequences of our platform compute systems in the Section \ref{ssec:sup_platform_compute_design} of the Appendix.

As part of our field testing campaign, we uncovered an issue where our USB-connected IMU was delayed in delivering measurements critical to localization performance. From an integration perspective, our IMU's USB interface was implemented using a standard USB Communications Data Class Abstract Control Model (CDC-ACM) interface. Using CDC-ACM for IMU measurements was particularly problematic due to the way in which CDC-ACM uses \textit{bulk} transport. USB has several methods of transferring data from device to host, including \textit{interrupt}, \textit{isochronous}, and \textit{bulk} transport. CDC-ACM uses \textit{bulk} transport, which does not include any guarantee for on-time delivery of data. As a consequence, during high CPU load, IMU measurements were occasionally delayed and resulted in localization error. In contrast, \textit{interrupt} and \textit{isochronous} transports are regularly serviced and can deliver on-time data. This problem could be solved in future deployments by either replacing the \textit{bulk} interface with an \textit{interrupt} interface or by using legacy serial interfaces such as RS-232 (not typically available on small form factor computing units). However, in practice we found adjusting the IMU timing as described in Section \ref{sec:localization}, was sufficient and did not require engineering new firmware. 


Another critical piece for reliable localization and consequently navigation is sensor synchronization. Synchronization is fundamentally necessary in order for our onboard sensors to communicate with their respective computers, and for those computers onboard individual agents to communicate with each other and the base station computer. The technical implementation of our solution is detailed in Section \ref{ssec:sup_sensor_sync_design} of the Appendix.

\section{Localization}
\label{sec:localization}

One of the major challenges in the DARPA Subterranean Challenge is ensuring reliable localization across a diverse set of austere environments. Localization is a critical process for an autonomous system, as it provides pose information to downstream autonomy processes including volumetric mapping, path planning, artifact detection, and multi-agent coordination. Section \ref{ssec:slam} details the simultaneous localization and mapping solution that was integrated into the autonomy stack, and Section \ref{ssec:gate_localization} describes the process used to align all robots to the common DARPA reference frame.

\subsection{Simultaneous Localization and Mapping}
\label{ssec:slam}

Simultaneous localization and mapping has relatively mature vision-based solutions  \cite{leutenegger_keyframe-based_2015,10.1007/978-3-319-50115-4_66,qin_vins-mono:_2017}, thanks to advances in feature extraction \cite{Cheung2009,Bay2008,Zhan_2018_CVPR}. However, in mission-critical applications such as underground search and rescue, visual-inertial solutions are not reliable enough when faced with irregular lighting, specular highlights, and feature-poor scenes. Recent work has illuminated the possibility of leveraging thermal-based odometry estimation in visually degraded environments \cite{khattak_keyframe-based_2019,wisth_unified_2021}.

Because underground environments are typically rich in geometric features, lidar-based localization solutions are a compelling alternative. Some spaces though, such as a smooth tunnels and corridors, contain relatively few longitudinal features, and therefore pose limits to lidar-based perception. Single-echo lidar also struggles in austere environments containing fog or smoke, though some recent work has focused on addressing these limitations \cite{shamsudin_fog_2016}.

For the Final Event, Team MARBLE transitioned from Google Cartographer \cite{hess2016real} to LIO-SAM \cite{shan_liosam_2020}, since its faster online loop closures during long-duration missions results in greater localization accuracy. Extensive testing was conducted in many different environments including parking garages, academic buildings, gold mines, and outdoor environments. The fast, lightweight loop closure performance can be attributed to performing scan-matching on a local level rather than a global level. LIO-SAM additionally performs IMU pre-integration to deskew point clouds, yielding better initialization for lidar odometry estimation. Because localization is the foundation to many autonomy modules, it was imperative to validate LIO-SAM's performance onboard the Spot and Husky platforms during large-scale, long-duration missions. Some examples of such validation efforts are shown in Section \ref{ssec:sup_localization_validation} of the Appendix.

Several modifications are made to the system to improve localization accuracy and reliability. First, the IMU and lidar sensors are fastened to a 6061 aluminum sensor plate, with a mounting configuration that is common between Huskies and Spots. By specifying the relative transform between the two sensors to a high precision, the need for extrinsics calibration is reduced. Specifically, the mounting configuration consists of a tight-tolerance, precision-ground plate with a flatness tolerance of 0.005", which greatly improves the roll and pitch alignment between the two sensors. By using a high-quality MEMS IMU and such precise sensor mounting, the LIO-SAM parameter specifying how much to weight IMU roll, pitch, and yaw measurements relative to lidar odometry was increased by a factor of 100. Taken together, these modifications greatly reduce accumulated rotation and translation drift, enabling smooth autonomous operation across long missions.

Secondly, LIO-SAM requires sensor timestamps to be aligned and sensor rates to be consistent. In particular, if IMU message rates fluctuate too greatly, the IMU pre-integration factors \cite{forster2015imu} can fail and lead to LIO-SAM instability. To reduce sensitivity to fluctuating IMU sensor rates caused by USB transmission delays (described in Section \ref{ssec:power_electronics}), the IMU timestamp assignment is adjusted when messages are not received within 15\% of the nominal rate. Additionally, the lidar sensor is synchronized with the onboard computer via PTP as discussed in Section \ref{ssec:sup_sensor_sync_design} of the Appendix. These two timing solutions reduce the probability of erroneous measurements and greatly improve the stability of LIO-SAM.

\subsection{Common Reference Frame Alignment}
\label{ssec:gate_localization}

\begin{figure}[!thb]
        \centering
		\includegraphics[width=0.45\textwidth]{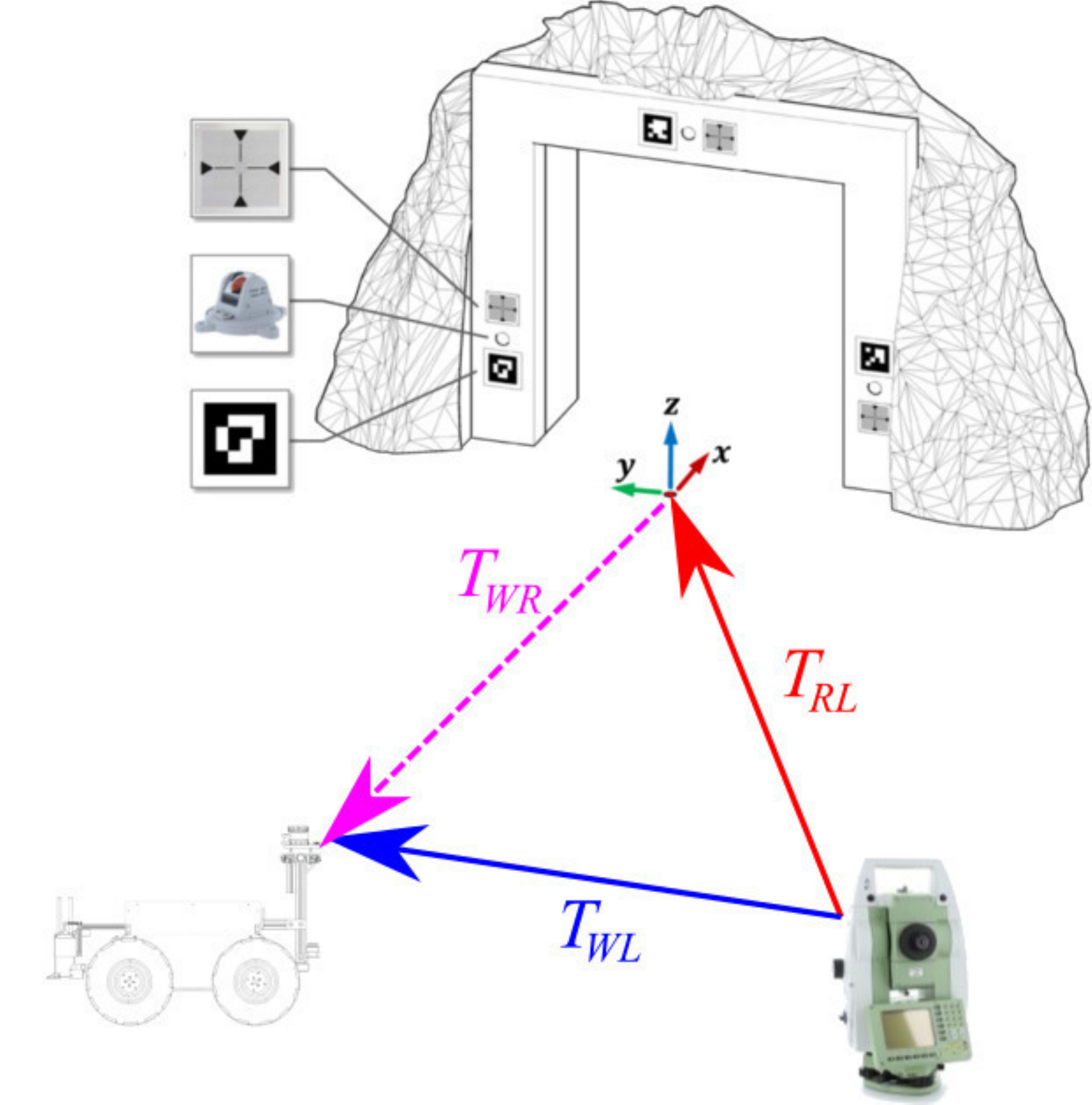}
		\caption{Figure of the MARBLE gate alignment setup including the Leica Total Station (LTS), gate, and an example robot. Transforms from the LTS to the robot $T_{RL}$, from the LTS to the world frame $T_{WL}$, and the resulting transform from the world frame into the $T_{WR}$}
		\label{f:gate_alignment}
\end{figure}

Accurate multi-robot alignment is a core design decision for the MARBLE localization, mapping, and planning systems. Robots are required to share globally aligned map data for planning and navigation. In addition, it allows the human supervisor and robots to share global coordinates for artifact locations relative to the DARPA-provided world frame. In order to align with the DARPA frame, Apriltags \cite{Malyuta2019,Brommer2018,Wang2016}, retro-reflective targets, and Leica Total Station (LTS) reflectors are attached to a gate with relative transforms to the DARPA origin frame. The global frame was assumed to be aligned with gravity, but each team was responsible for aligning yaw, and XYZ-translation from their robots to the common DARPA frame. In the context of the SubT Challenge, the DARPA frame was purely used to align robots into the measured ground truth frame for artifact scoring and map accuracy analysis. However, in practice, an accurate initial alignment between robots results in more reliable multi-agent coordination and global merging maps.

In order to align with the common reference frame, MARBLE primarily relied on the LTS reflectors. Based on conventional trigonometry and the assumption of needing to maintain less than 5m of error over the course of a 1km linear distance, it is determined that an initial alignment target required  less than $0.29^{\circ}$ of error.

To align the robot, 3 reflective prisms are attached to each robot, and their positions are scanned with an LTS. These points $V_{LR}$ are then compared with a ground-truth set $V_R$, determined by the relative locations of the prisms to the robots tracking frame via CAD. These two sets of points are used to estimate the transform between the LTS recorded positions and the assumed positions by minimizing across the pose $\mathcal{X}$ to solve:
\begin{equation}
    \text{argmin}_{\mathcal{X}}\sum ||V_{LR} - V_R T_{RL}||.
\end{equation}
The result is the robot's position in the LTS frame $T_{RL}$. An additional calculation is used between scanned points of the gate $V_{LG}$ and the provided coordinates $V_W$ are used to solve for the gate's position in the Leica frame $T_{WL}$ using the equation: 
\begin{equation}
     \text{argmin}_{\mathcal{X}}\sum ||V_{LG} - V_W T_{WL}||.
\end{equation}
Both minimization problems were solved using Horn's absolute orientation method \cite{horn1987closed}, a closed form solution to least squares alignment problems. Given these transforms, the robots position in the world frame was calculated by inverting the robot to LTS transform: 
\begin{equation}
T_{WR}=T_{WL}(T_{RL})^{-1}.    
\end{equation}
To further reduce the impact of minor errors in either prism localization or low observability, these transforms are altered slightly by each robot. The LTS-predicted pitch and roll is substituted with an estimated pitch and roll from the lidar-inertial localization system, largely based on the initial measurements of the IMU. 

After these adjustments, yaw estimates had the largest impact on our resulting transforms. Because yaw error has the potential to propagate to large translational discrepancies at far distances, it became imperative to modify our system. The solution involves increasing the lateral spacing of the prisms mounted on the robots, and is described in more detail in Section \ref{ssec:sup_ref_frame_optimization} of the Appendix.

\section{Artifact Detection}\label{sec:object_detection}

A core component of the SubT Challenge is the detection and localization of objects that could potentially indicate human presence. Each artifact needed to be reported within a 5m radius of the ground truth location. To achieve this requirement a lidar-inertial based state estimator as described in Section \ref{sec:localization} is used. Robots are put into a common reference frame based on survey-grade measurements from a Leica Total Station (LTS) and objects are projected using the mapping framework described in Section \ref{sec:mapping}. The available sensing modalities for various artifacts are discussed in Section \ref{ssec:sensing_modalities}, the visual detection system is described in Section \ref{ssec:visual_detection}, and the non-visual detection system is explained in Section \ref{ssec:non_visual_detection}. The resulting performance of the artifact detection system at the Final Event is detailed in Section \ref{ssec:results_artifact_detection}.

\subsection{Sensing Modalities}
\label{ssec:sensing_modalities}

Table \ref{tab:ad_sensing_modalities} shows the classes of artifacts present at the final event along with the types of sensing modalities capable of detecting each artifact. Each robot in the fleet is equipped with RGB cameras, Bluetooth modules, and CO$_2$ sensors which enable the detection of all classes of artifacts using a minimal sensor suite. The visual detection system is not trained to detect either the cell phone, due to its small form factor, or the cube artifact which was detectable using Bluetooth. The cube artifact had rotating colors which pose significant challenges for visual detection methods.

\begin{table}[htbp!]
\begin{center}
    \begin{tabular}{c c c c c c}
    \hline
    \textbf{} & \textbf{Artifact Class} & \textbf{Visual} & \textbf{Thermal} & \textbf{Wireless} & \textbf{CO\textsubscript{2}} \\ 
    \hline
    \rule{0pt}{4ex}
    \raisebox{-0.3\totalheight}{\includegraphics[height=0.04\textwidth]{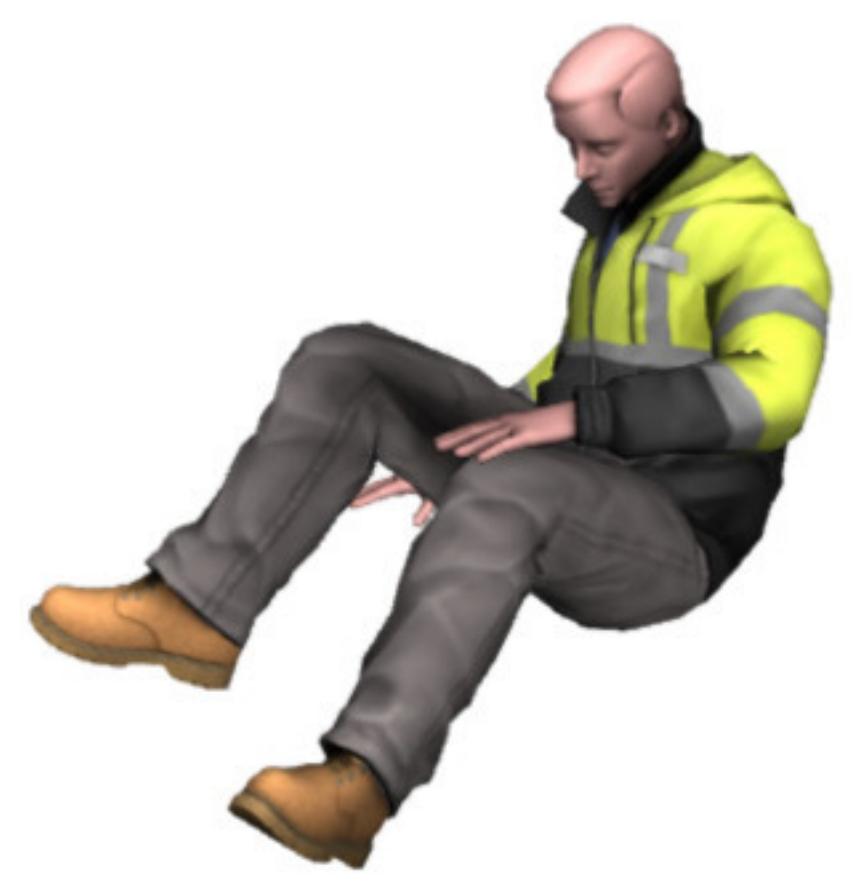}}
    & Survivor            & 	+   &   --   &       &       \\
    \raisebox{-0.3\totalheight}{\includegraphics[height=0.04\textwidth]{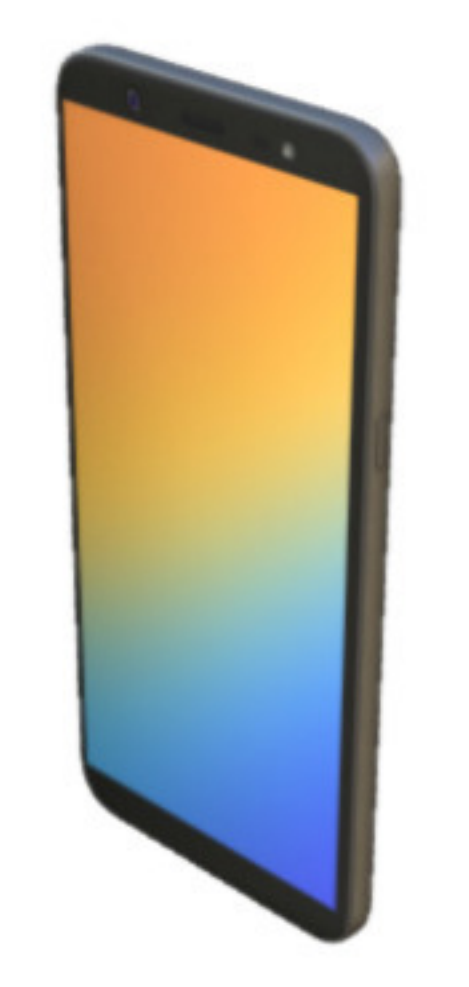}}
    & Cell Phone          &   --   &    	&   +   &       \\
    \raisebox{-0.3\totalheight}{\includegraphics[height=0.04\textwidth]{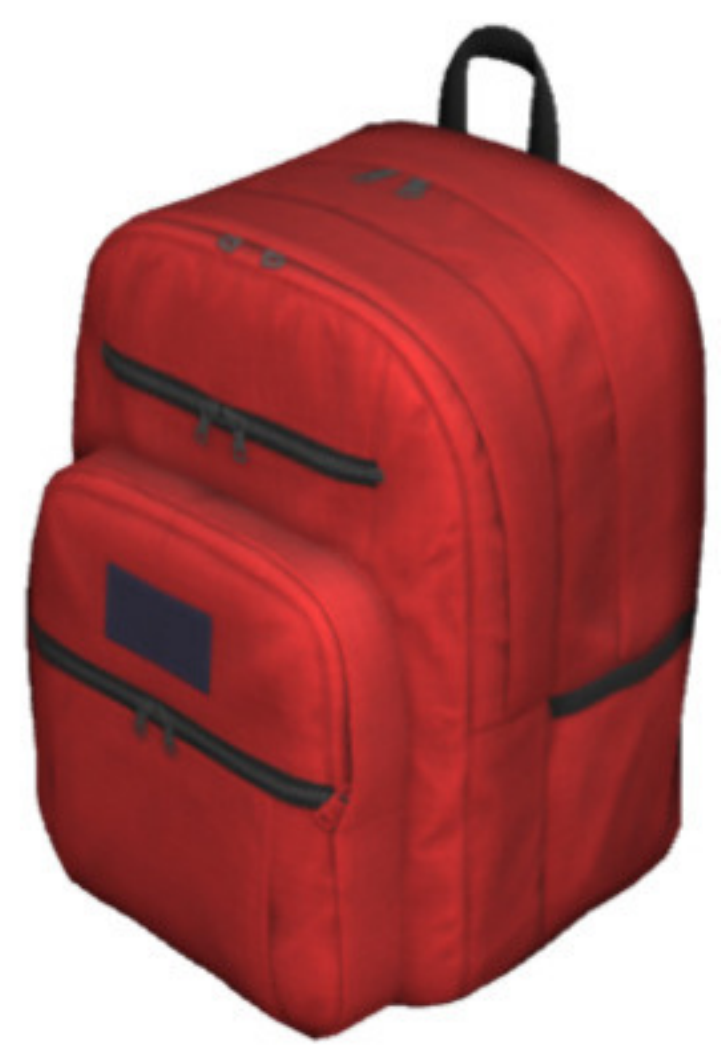}}
    & Backpack	        &   +   &    	&       &       \\
    \raisebox{-0.3\totalheight}{\includegraphics[height=0.04\textwidth]{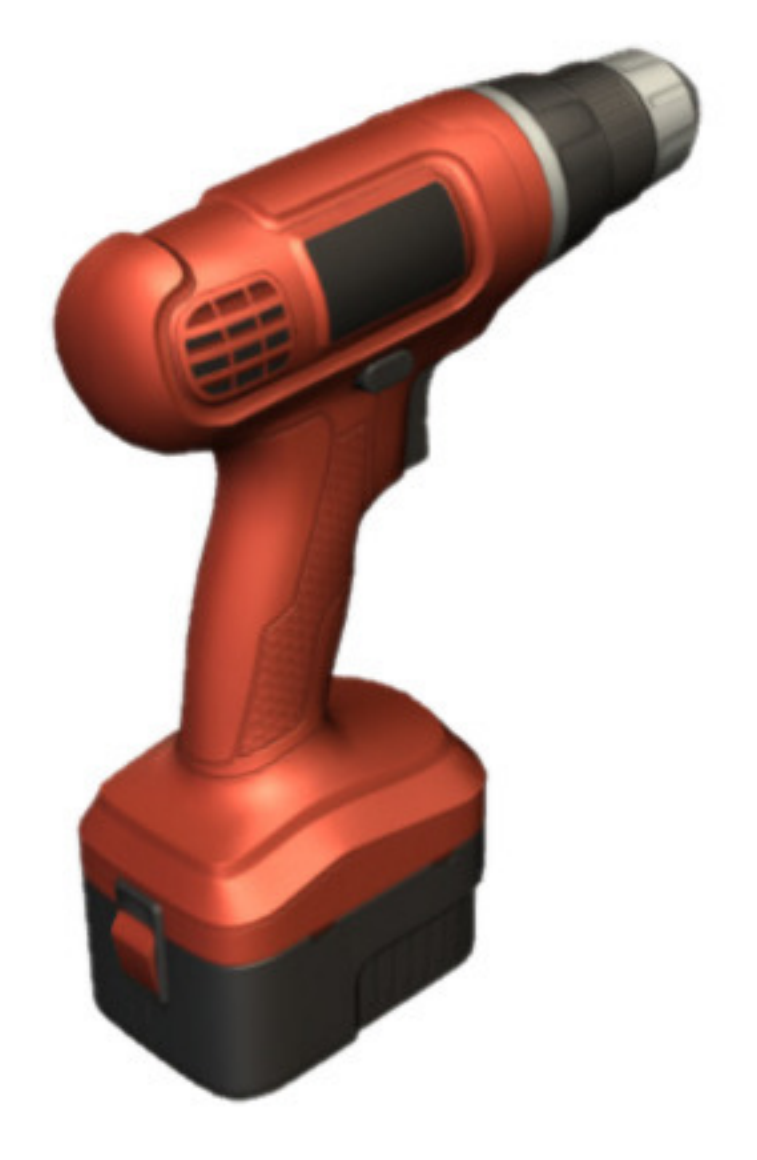}}
    & Drill	            &   +   &       &       &       \\
    \raisebox{-0.3\totalheight}{\includegraphics[height=0.04\textwidth]{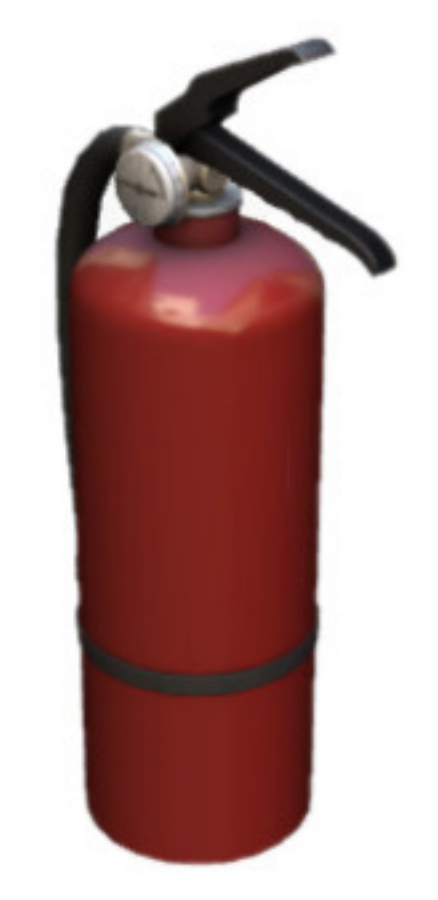}}
    & Fire Extinguisher   &	+   &       &       &       \\
    \raisebox{-0.3\totalheight}{\includegraphics[height=0.04\textwidth]{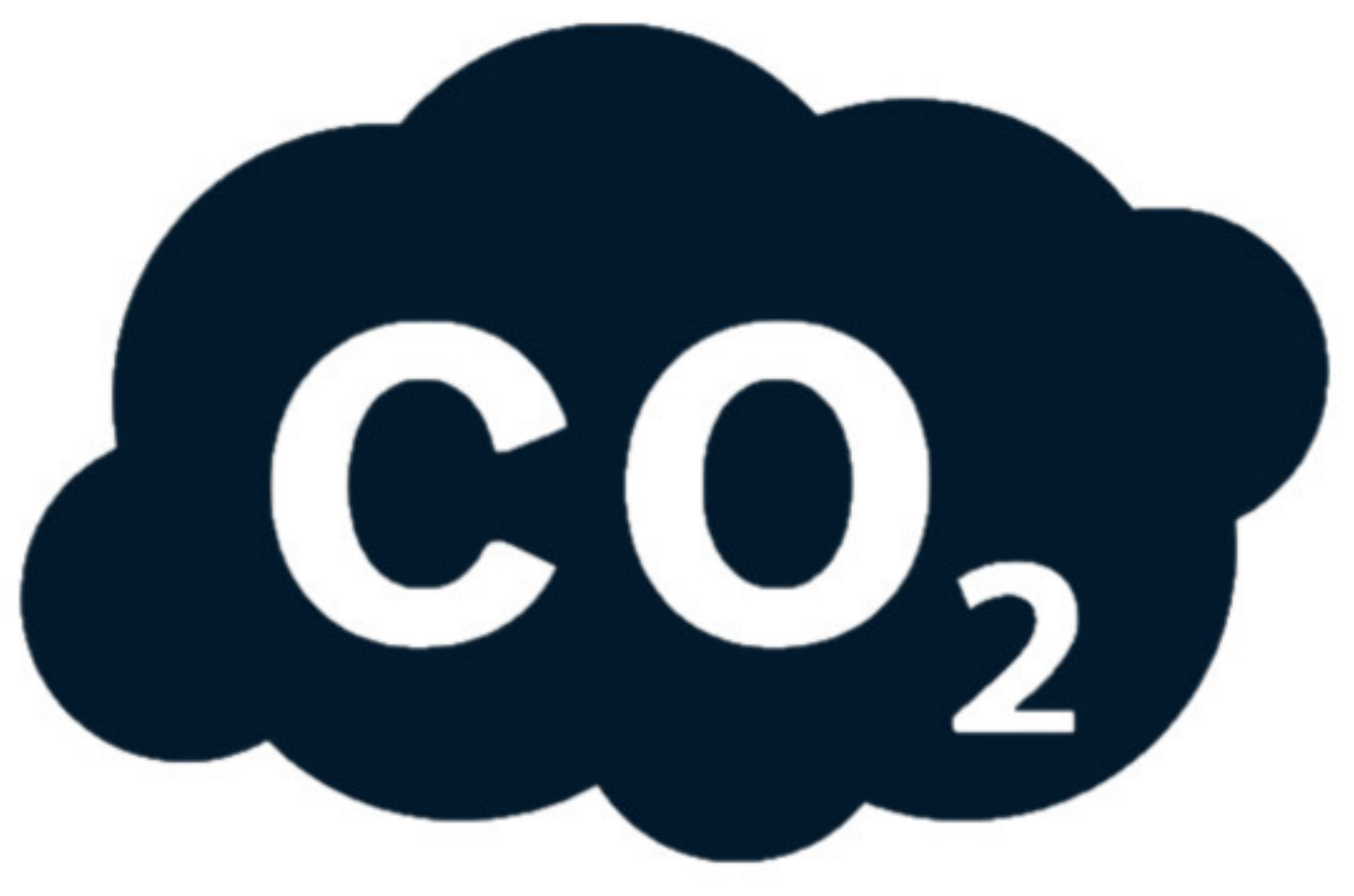}}
    & Gas	                &       &       &       &   +   \\
    \raisebox{-0.3\totalheight}{\includegraphics[height=0.04\textwidth]{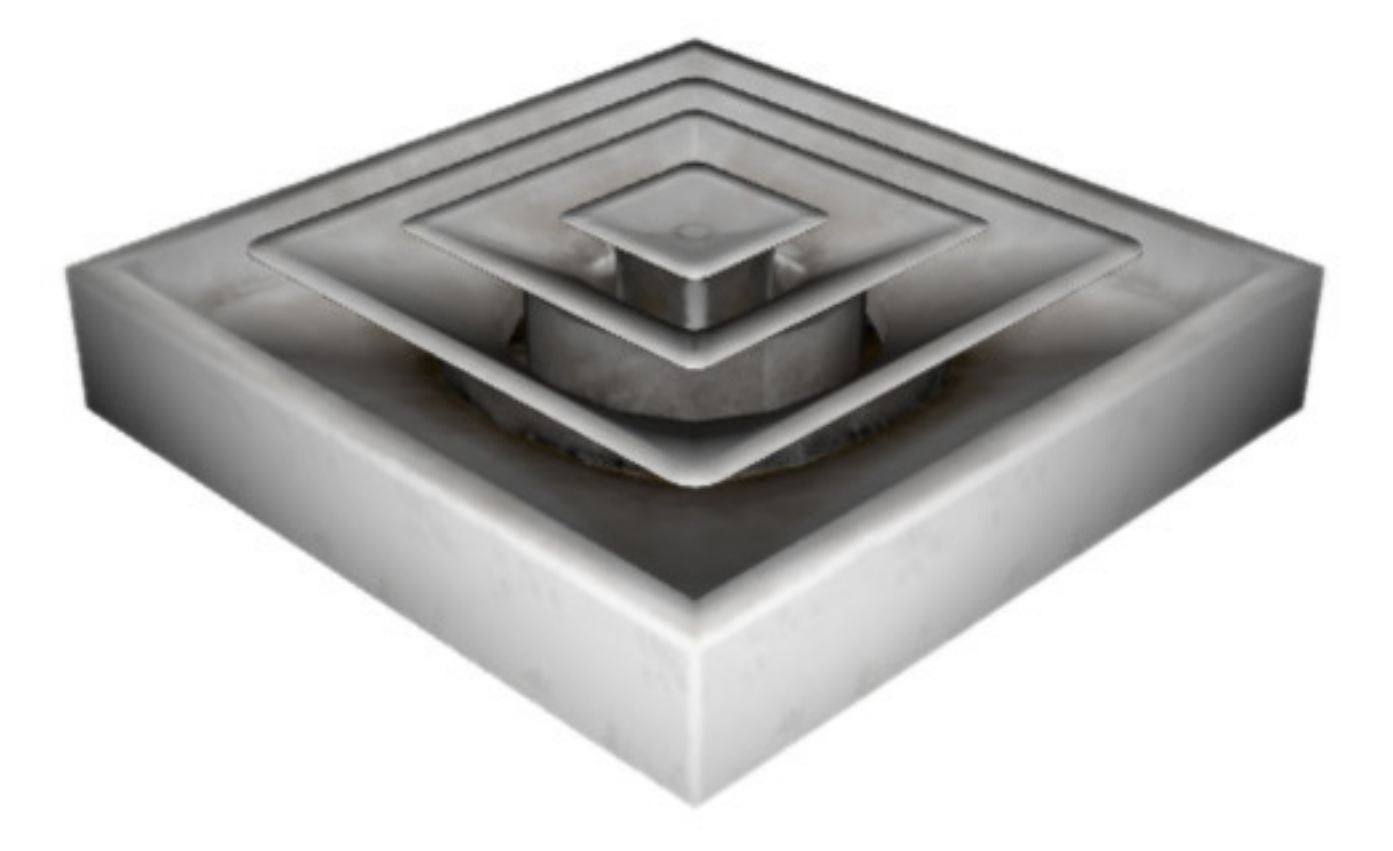}}
    & Vent                &   +   &   --   &       &       \\
    \raisebox{-0.3\totalheight}{\includegraphics[height=0.04\textwidth]{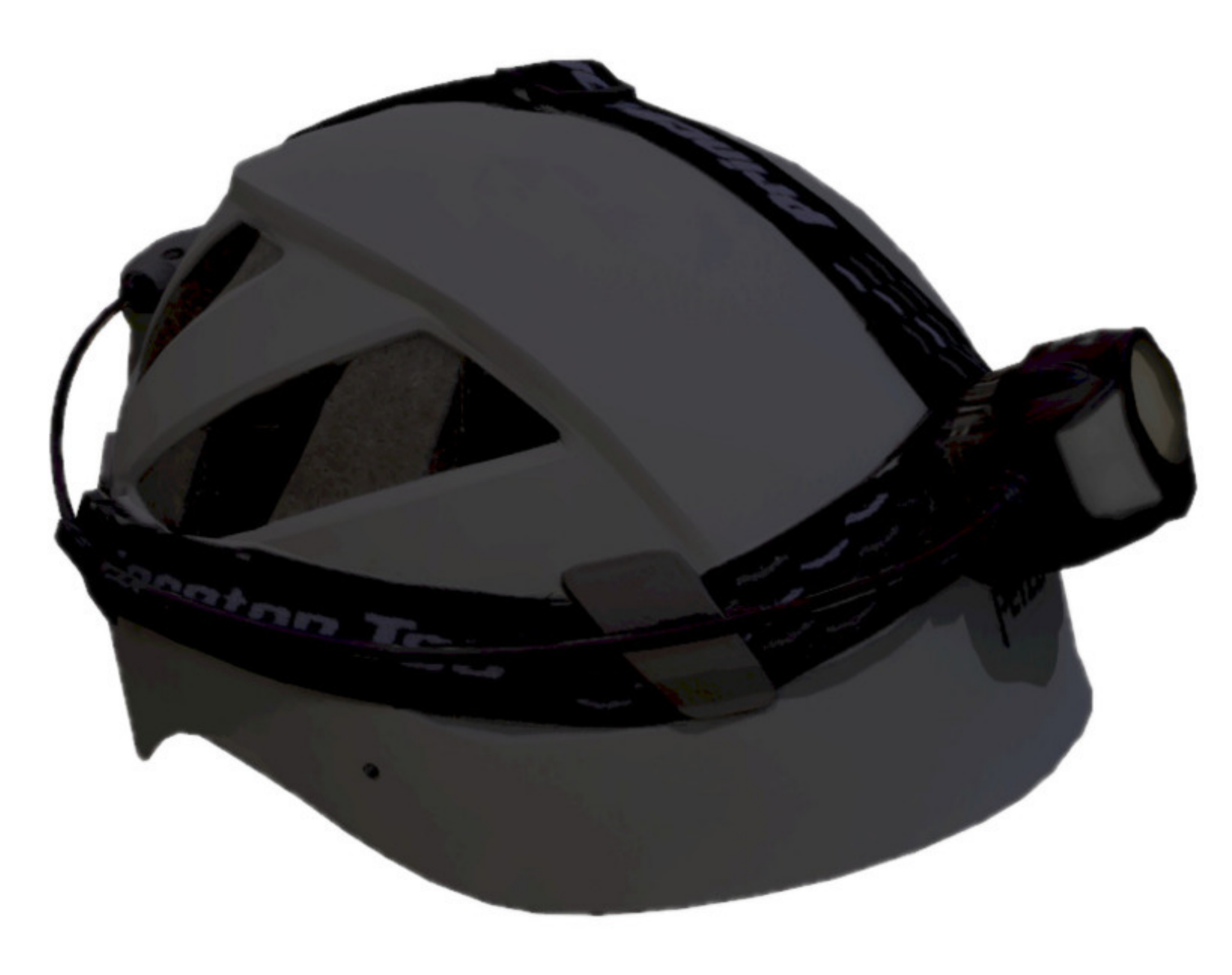}}
    & Helmet              &   +   &       &       &       \\
    \raisebox{-0.3\totalheight}{\includegraphics[height=0.04\textwidth]{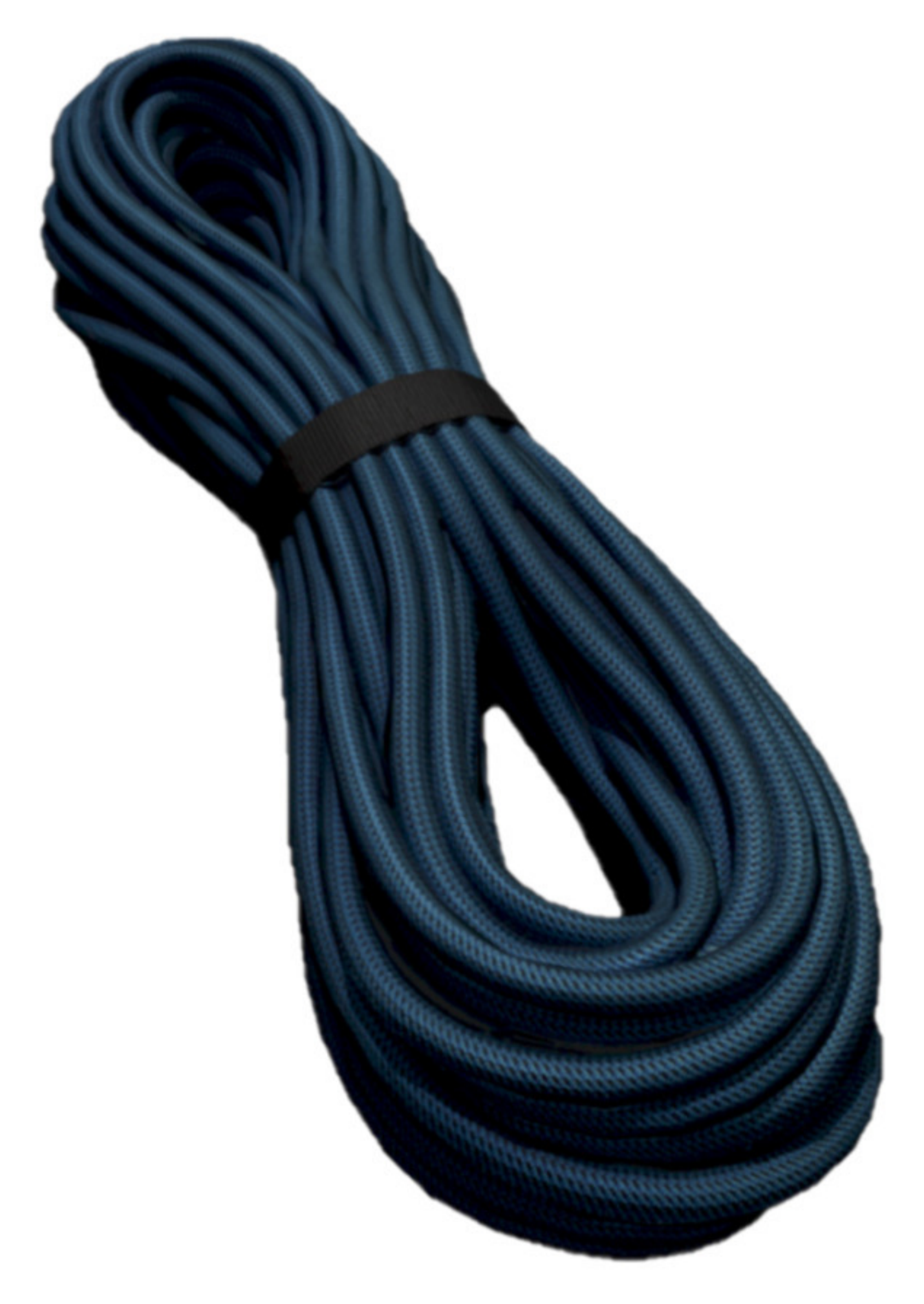}}
    & Rope	            &   +   &       &       &       \\
    \raisebox{-0.3\totalheight}{\includegraphics[height=0.04\textwidth]{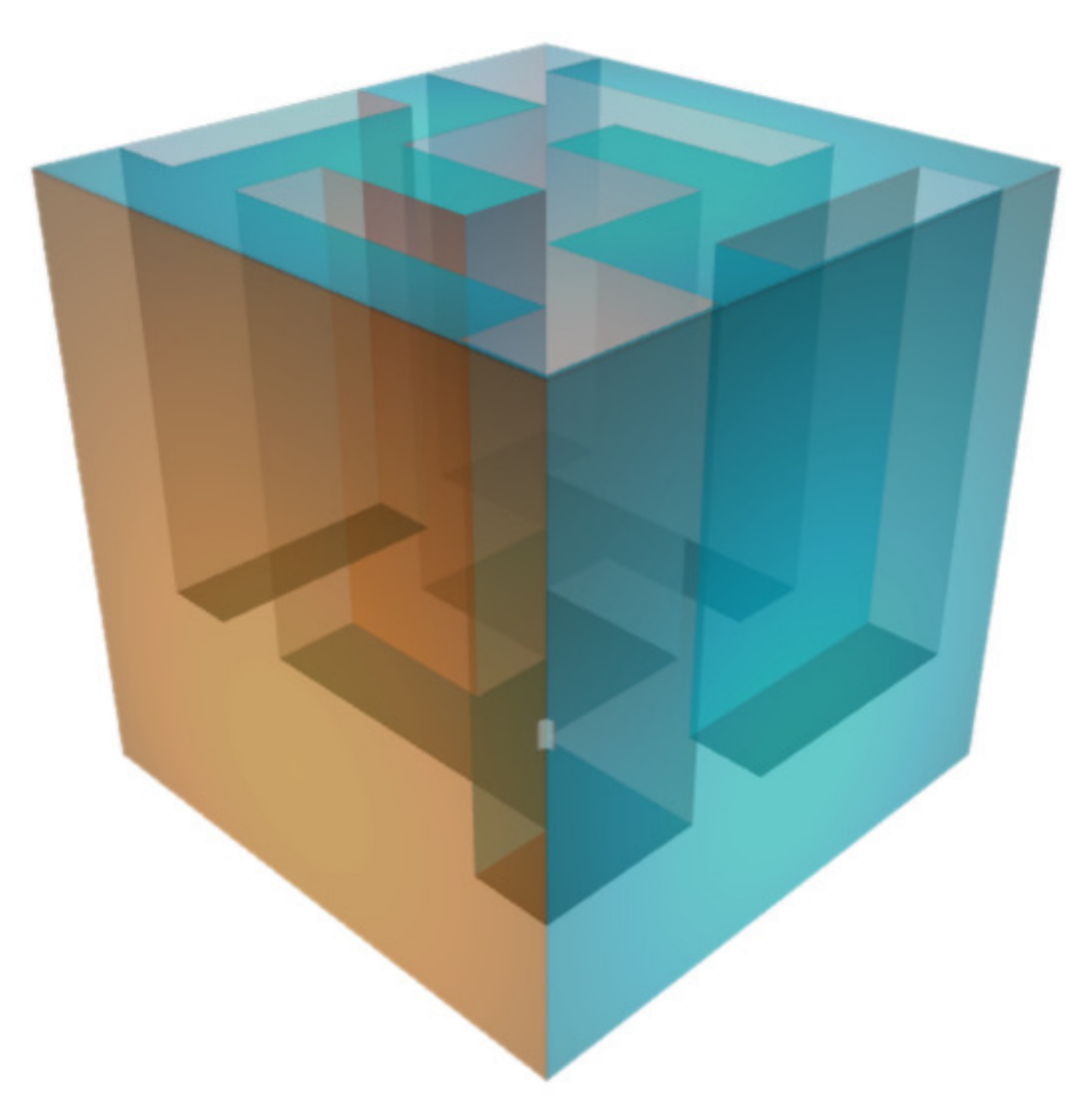}}
    & Cube                &   --   &       &   +   &       \\
    \hline
    \end{tabular}
\caption{\label{tab:ad_sensing_modalities}Sensing modalities, that Team MARBLE utilized (+) and did not utilize (--) for detecting the ten artifact classes. Blank entries indicate sensing modalities that are not useful for detecting specific artifact classes.}
\end{center}
\end{table}

\subsection{Visual Detection}
\label{ssec:visual_detection}

Visual object detection is a well-researched problem in computer vision and state of the art detectors are capable of identifying objects in both 2D and 3D. Common 2D detectors are typically based on Convolutional Neural Networks (CNN) \cite{zou2019object}, such as region proposal-based networks like Fast R-CNN \cite{girshick2015fast}. Typically these networks require multiple passes over an image to classify an object and then detect where the object is in the image. In contrast, YOLO \cite{redmon2016you} performs both classification and detection in a single regression making it a significantly faster detection: 0.5 FPS for Fast R-CNN and 45 FPS for YOLO.  Object detectors operating in 3D typically use point clouds obtained from a lidar and until recently were limited to classification rather than full detection \cite{maturana2015voxnet,qi2017pointnet}. Extensions to 3D classifiers such as Voxelnet \cite{zhou2018voxelnet} and PointRCNN \cite{shi2019pointrcnn} are capable of performing object detection on powerful GPUs. These GPUs are impractical from both a size and power consumption standpoint for mobile robots. We selected the Yolo V3 \cite{redmon2018yolov3} model due to the fast and accurate nature of the YOLO  \cite{redmon2016you} family of networks.

Specifically, for classification and detection, the visual pipeline utilizes a YOLO V3 Tiny \cite{redmon2016you} model with custom trained weights. The model is optimized for Nvidia TensorRT acceleration and we infer images at a resolution of 608x608. The Husky platforms are able to perform inference at 60FPS on a GTX 1650 based on Nvidia's Turing architecture with 896 CUDA cores and 112 RT cores. The Spot platform uses a Nvidia Jetson Xavier AGX based on the Volta architecture with 512 CUDA Cores and 64 Tensor cores to perform inference at 40 FPS. These GPUs were chosen to balance performance against size and power constraints for on-board compute. The TensorRT YOLO detector outputs a message containing the detected artifacts as well as the coordinates of their bounding boxes.

A systematic procedure targeted at low-light conditions is used to train the model. At each location, data was collected using three different brightness levels to minimize the impact of lighting conditions on the model's performance. Specifically, images were taken from past circuit events as well as separate field exercises. Images with excessive motion blur were subsequently filtered out and the data was later augmented with images that contained false positive defections. The full details of our training procedure can be found in Section \ref{ssec:sup_artifact_detection_training_procedure} of the Appendix.

Depth registration is performed using \emph{marble\_mapping} as described in Section \ref{sec:mapping} which is generated by the Ouster 64-beam lidar. Utilizing an Octomap based framework allowed us to avoid implementing any additional filtering due to the probabilistic nature of the map. Additionally, the Octomap structure aggregates scans into the map with temporal memory. This important feature resolves the inconsistency between the $33.2^\circ$ vertical field of view of the Ouster and the $68^\circ$ vertical field of view of the cameras. At further distances, the agent is able to incrementally build out regions near ceilings and floors, overcoming the vertical blind spots of the Ouster. Essentially, this temporal memory allows us to decouple the depth measurement from the visual artifact detection. The biggest drawback of this approach is the potential for an additional 0.15m of error on each detection due to the voxel resolution. However, this error figure still falls within the design constraints of localizing an object to within 5m of its desired location.

\tikzfigpos{artifact_diagram}{Overview of the artifact detection system. Sensor inputs are shown in red and outputs are shown in green.}
{

\usetikzlibrary{shapes.geometric,shapes.symbols,fit,positioning,shadows}
\makeatletter
\pgfdeclareshape{document}{
\inheritsavedanchors[from=rectangle] 
\inheritanchorborder[from=rectangle]
\inheritanchor[from=rectangle]{center}
\inheritanchor[from=rectangle]{north}
\inheritanchor[from=rectangle]{north east}
\inheritanchor[from=rectangle]{north west}
\inheritanchor[from=rectangle]{south}
\inheritanchor[from=rectangle]{south east}
\inheritanchor[from=rectangle]{south west}
\inheritanchor[from=rectangle]{west}
\inheritanchor[from=rectangle]{east}
\backgroundpath{%
\southwest \pgf@xa=\pgf@x \pgf@ya=\pgf@y
\northeast \pgf@xb=\pgf@x \pgf@yb=\pgf@y
\pgf@xc=\pgf@xb \advance\pgf@xc by-5pt 
\pgf@yc=\pgf@ya \advance\pgf@yc by5pt
\pgfpathmoveto{\pgfpoint{\pgf@xa}{\pgf@ya}}
\pgfpathlineto{\pgfpoint{\pgf@xa}{\pgf@yb}}
\pgfpathlineto{\pgfpoint{\pgf@xb}{\pgf@yb}}
\pgfpathlineto{\pgfpoint{\pgf@xb}{\pgf@yc}}
\pgfpathlineto{\pgfpoint{\pgf@xc}{\pgf@ya}}
\pgfpathclose
\pgfpathmoveto{\pgfpoint{\pgf@xc}{\pgf@ya}}
\pgfpathlineto{\pgfpoint{\pgf@xc}{\pgf@yc}}
\pgfpathlineto{\pgfpoint{\pgf@xb}{\pgf@yc}}
\pgfpathclose
}
}
\makeatother
\tikzset{doc/.style={document,fill=blue!10,draw,thin,minimum
height=1.2cm,align=center},
pics/.cd,
fusion/.style={code={%
\draw[fill=blue!50,opacity=0.2] (0,0) -- (0.5,-0.25) -- (0.5,0.25) -- (0,0.5) -- cycle;
\draw[fill=blue!50,opacity=0.2] (0,0) -- (-0.5,-0.25) -- (-0.5,0.25) -- (0,0.5) -- cycle;
\draw[fill=blue!60,opacity=0.2] (0,0) -- (-0.5,-0.25) -- (0,-0.5) -- (0.5,-0.25) -- cycle;
\draw[fill=blue!60] (0,0) -- (0.25,0.125) -- (0,0.25) -- (-0.25,0.125) -- cycle;
\draw[fill=blue!50] (0,0) -- (0.25,0.125) -- (0.25,-0.125) -- (0,-0.25) -- cycle;
\draw[fill=blue!50] (0,0) -- (-0.25,0.125) -- (-0.25,-0.125) -- (0,-0.25) -- cycle;
\draw[fill=blue!50,opacity=0.2] (0,-0.5) -- (0.5,-0.25) -- (0.5,0.25) -- (0,0) -- cycle;
 \draw[fill=blue!50,opacity=0.2] (0,-0.5) -- (-0.5,-0.25) -- (-0.5,0.25) -- (0,0) -- cycle;
\draw[fill=blue!60,opacity=0.2] (0,0.5) -- (-0.5,0.25) -- (0,0) -- (0.5,0.25) -- cycle;
}}}
\begin{tikzpicture}[font=\sffamily,every label/.append
style={font=\small\sffamily,align=center}]

\node[tape, draw,thin, tape bend top=none,fill=purple,
text=white,minimum width=2.2cm,double copy shadow,minimum height=1.5cm]
(Gige) {Gig-E Cameras};

\node[right=1cm and 1cm of Gige,doc,fill=blue!10] (Yolo)
{TensorRT YOLO};

\draw[-latex] (Gige) -- (Yolo);

\node[right=1cm and 1cm of Yolo,doc,fill=blue!10] (Localization)
{Artifact\\ Localization};

\draw[-latex] (Yolo) -- (Localization);

\node[draw,dashed,rounded corners,fit=(Gige) (Yolo),inner
sep=10pt,label={above:{Classification and Detection}}](box1){};

\pic[right=2cm of Localization,local bounding box=Fusion,scale=2] (Artifact Fusion) {fusion};
\node[below=1mm of Fusion,font=\small\sffamily,align=center]{Artifact\\ Fusion};

\draw[-latex] (Localization) -- (Fusion);

\node[above=1cm and 1cm of Localization,doc,fill=blue!10] (Mapping)
{Robot \\ Localization/Mapping};

\draw[-latex] (Mapping) -- (Localization);

\node[right=1cm of Fusion,doc,fill=green!10] (ArtifactMsg)
{Artifact Message};

\node[above=0.61cm and 2.25cm of Fusion,doc,fill=purple,text=white] (Bluetooth)
{Bluetooth \\ CO2};

\draw[-latex] (Fusion) -- (ArtifactMsg);
\draw[-latex] (Bluetooth) -- (Fusion);

\end{tikzpicture}
}

After 3D coordinates are obtained via the Artifact Localization node, we run a weighted median filter in the world coordinate frame to de-noise the projected location within the Artifact Fusion node in \fig{artifact_diagram}. Each localized artifact is considered to be part of the same measurement if it is the same class as a previous measurement and within 5m of that measurement. We then require five to 10 positive detection events and use the median position as the reported position to the human supervisor. The final detection is published in a custom ROS message which contains this position as well as a compressed version of a corresponding camera image and associated bounding box. The full overview of the artifact system can be seen in Figure \ref{f:artifact_diagram}.

\subsection{Non-Visual Detection}
\label{ssec:non_visual_detection}

Cell phone, cube, and gas reports are also fused using a weighted median filter. The Bluetooth and CO$_2$ detections are simply localized to the position of the robot at the time of detection. Bluetooth detections are also grouped together by unique SSIDs and gas detections within 10m of another detection are assumed to have originated from the same source. The final positioning of these non-visual artifacts relies on input from the human supervisor. Our human supervisor interface was designed to easily allow for movement of reported artifacts based on features observed in the map by the human operator. The details regarding the accuracy and success rate of these reports can be found in Section \ref{ssec:results_artifact_detection}.

\section{Mapping}
\label{sec:mapping}

Team MARBLE's custom mapping package, \emph{marble\_mapping} \cite{MARBLEmapping} is based on \emph{Octomap} \cite{hornung2013octomap} and is used to generate 3D occupancy grid representations of the world. The environment is sub-divided into voxels, or cells which are marked as either occupied, free, or unknown using a probabilistic log-odds based model operating on sensor returns. The output of \emph{marble\_mapping} is a direct input to the path planner and also provides depth measurements for visual artifact detection, as well as situational awareness for the human supervisor. The Octree \cite{meagher1982geometric} structure of Octomap's occupancy grids makes storing and transmitting maps more efficient than other representations such as point clouds; this efficiency is highly desirable when trying to transmit maps over low bandwidth mesh networks. \emph{marble\_mapping} extends \emph{Octomap} by enabling map differences for low bandwidth transmission, map merging between multiple robots, and the addition of semantic information.

\subsection{Difference-Based Map Merging}
\label{sec:map_merging}

\begin{figure}[!htb]
    \centering
    \includegraphics[width=0.65\linewidth]{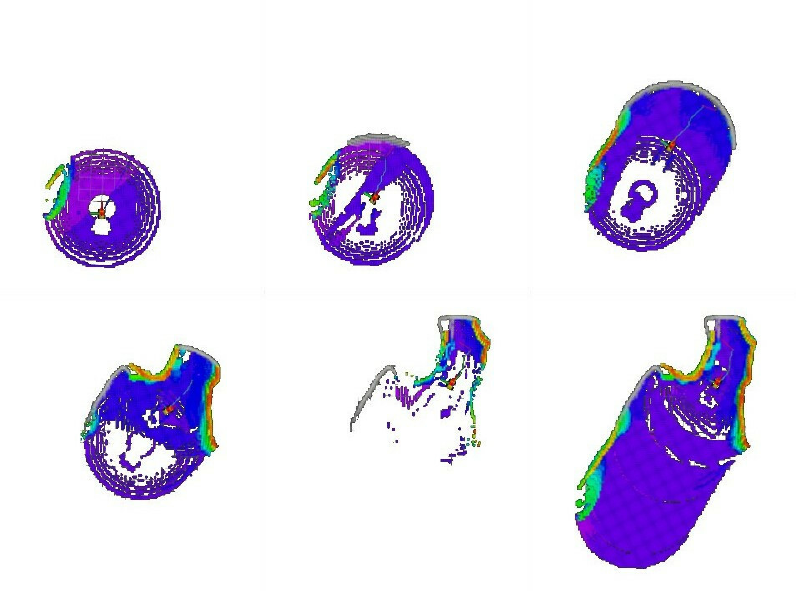}
    \caption{Sequential difference maps from top left to bottom right, with the final map on the far right constructed in real time for comparison. The diff maps can be merged to fully reconstruct the original map shown in the bottom right.}
    \label{fig:diff_maps_grid}
\end{figure}

Despite the efficient encoding of the Octree data structure, regularly transmitting full volumetric maps of the explored space is impractical in bandwidth-constrained subterranean environments. Map differences are both a natural solution to reduce bandwidth, and have been shown to facilitate efficient data transfers \cite{Sheng2004}. In the \emph{marble\_mapping} package, modifications to Octomap package were made to generate differences between different map sections, or ``diff maps" shown in Figure \ref{fig:diff_maps_grid}. The implementation allows for diff maps, or smaller Octree structures, to be created at a predetermined rate, and contains all the mapping data for that time interval. The sum of an agent's diff maps make up its ``self'' map and the differences can be transmitted to other agents. These differences are later merged into the robot's ``merged map'' in the the map merging process which is shown in Figure \ref{fig:diff_maps_grid}.

Merged maps generated from multiple agents are important both for a more complete view of the environment, and they also reduce redundant coverage in coordination strategies \cite{Ko2003,Simmons2000,Zlot2002}. The \textit{marble\_mapping} package enables map merging both for individual agents and on the base station which allows agents to intelligently act on the data and provides a holistic view for the human supervisor. The system does not re-align maps prior to merging, as it is assumes agents are already in a common reference frame as described in Section \ref{ssec:gate_localization}. The lack of a re-alignment feature has the potential to cause one agent's map to block pathways in a receiving agent's map. To mitigate this, each agent prioritizes its own map by only appending cells from other maps into ``unknown'' areas. Areas that have already been ``seen'' by the agent are left untouched which prevents misaligned data from blocking free space. In cases where this mitigation procedure is not enough, such as narrow hallways, or a complete loss of localization by an agent, ``bad'' map diffs can be removed by the human supervisor using the base station GUI described in Section \ref{sec:mission_management}.

\subsection{Semantic Mapping for Terrain-Aware Navigation}
\label{ssec:semantic_mapping}

While the volumetric-based mapping produced by the Octomap framework provides the high-level structure of the environment, its resolution, set to a voxel size of 0.15m, is too coarse to capture details needed for high fidelity motion planning. In order to augment the existing \emph{marble\_map} with terrain information, Team MARBLE evaluates the traversability of a given voxel using the normal and curvature values from raw point clouds. The planning solution is then able to utilize this semantic information to plan safe paths in Section \ref{sec:planning}. An additional label is attached to each voxel which enables the semantic labeling of staircases for the Spot platform.

Early approaches to evaluating the traversablity of an environment include elevation based maps based on a 2D lidar \cite{ye2003new} but are unable to take advantage of modern 3D sensors. The traversability classifier presented here is largely based on the Grid Map framework presented in \cite{fankhauser2016gridmap}, which evaluates the slope and roughness of point cloud regions to generate a multi-layer surface map but only creates a 2D grid rather than a 3D volumetric map. Other fielded approaches in subterranean environments include ``virtual surfaces'' on occupancy maps \cite{hines2021virtual} and Conditional-Value-at-Risk metrics, such as collision, step size, tip over, and slippage, which are incorporated into a dense 2.5D gridmap \cite{fan2021step}. These dense methods typically come in the form of high-resolution local maps, which enable more precise locomotion over varied terrain. An alternative approach presented in \cite{krusi2017driving} computes paths with continuous curvature over raw point clouds. However, by computing semantic traversability information, our planning approach only required a low-resolution global map, greatly simplifying both mapping and planning systems and allows for sharing of semantic information between agents.

\subsubsection{Traversability Classification \& Map Integration} \label{sec:trav_class}

To estimate the traversability of a voxel, we segment the 3D point cloud produced by the Ouster lidar, and evaluate the unit normal vector $\boldsymbol{\hat{n}}$ and curvature $K$ of each point $p$ at timestep $t$. All calculations are performed with the aid of the \emph{pcl} package \cite{rusu2011pcl} and a traversability value, $\tau_{p,t}$, is estimated for each point using Equation \ref{trav_est_eq} where $\boldsymbol{\hat{k}}$ is the gravity-aligned up vector, $(1-|\boldsymbol{\hat{n}} \cdot \boldsymbol{\hat{k}}|)^{3}$ is a measure of the slope of the terrain, and $c_{norm}$ and $c_{curv}$ are tunable parameters. The parameter values for the Final Event were set to $c_{norm} = 40.0$, and $c_{curv} = 4.0$, and $\tau_{p,t} \in [0,1]$.

\begin{equation} \label{trav_est_eq}
    \tau_{p,t} = c_{norm}(1-|\boldsymbol{\hat{n}_{p,t}} \cdot \boldsymbol{\hat{k}}|)^{3} + c_{curv} K_{p,t} \in [0,1]
\end{equation} 

\begin{figure}

\centering
\subfloat[]{
\begin{minipage}[b][2.55in][t]{0.5\textwidth}
\centering
\includegraphics[height=1.1in]{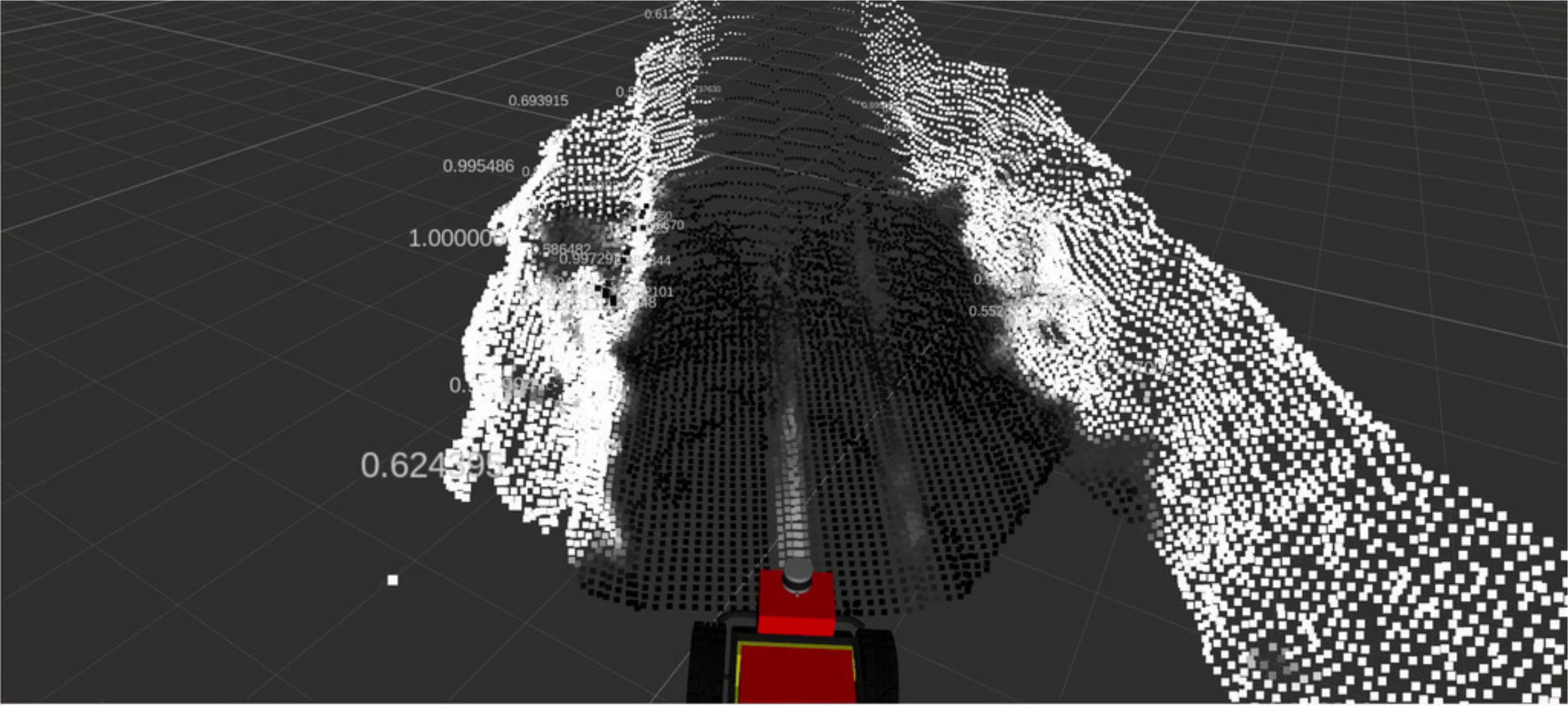}
\vfill
\includegraphics[height=1.1in]{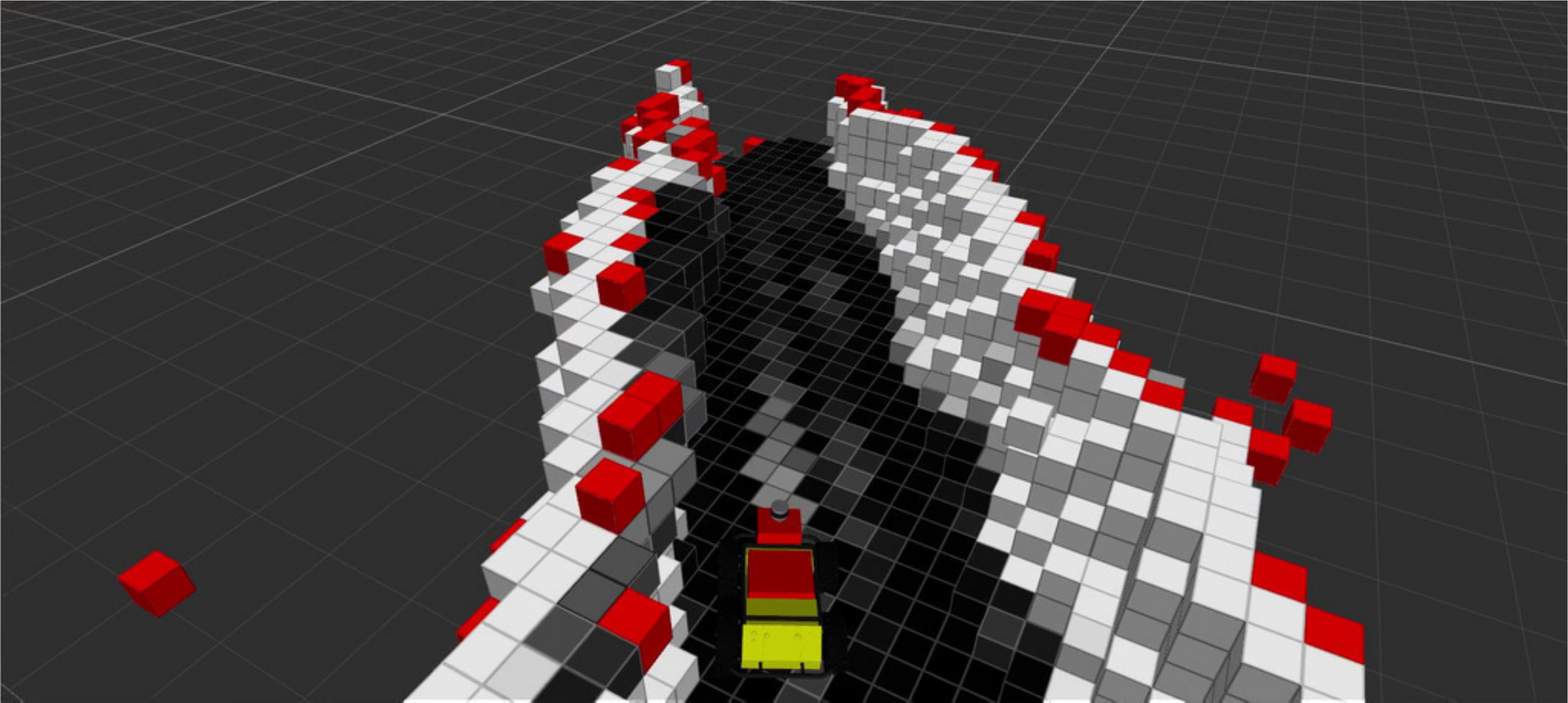}
\end{minipage}}%
\subfloat[]{\begin{minipage}[b][2.55in][t]{.5\textwidth}
\centering
\includegraphics[height=2.55in, width=0.9\textwidth]{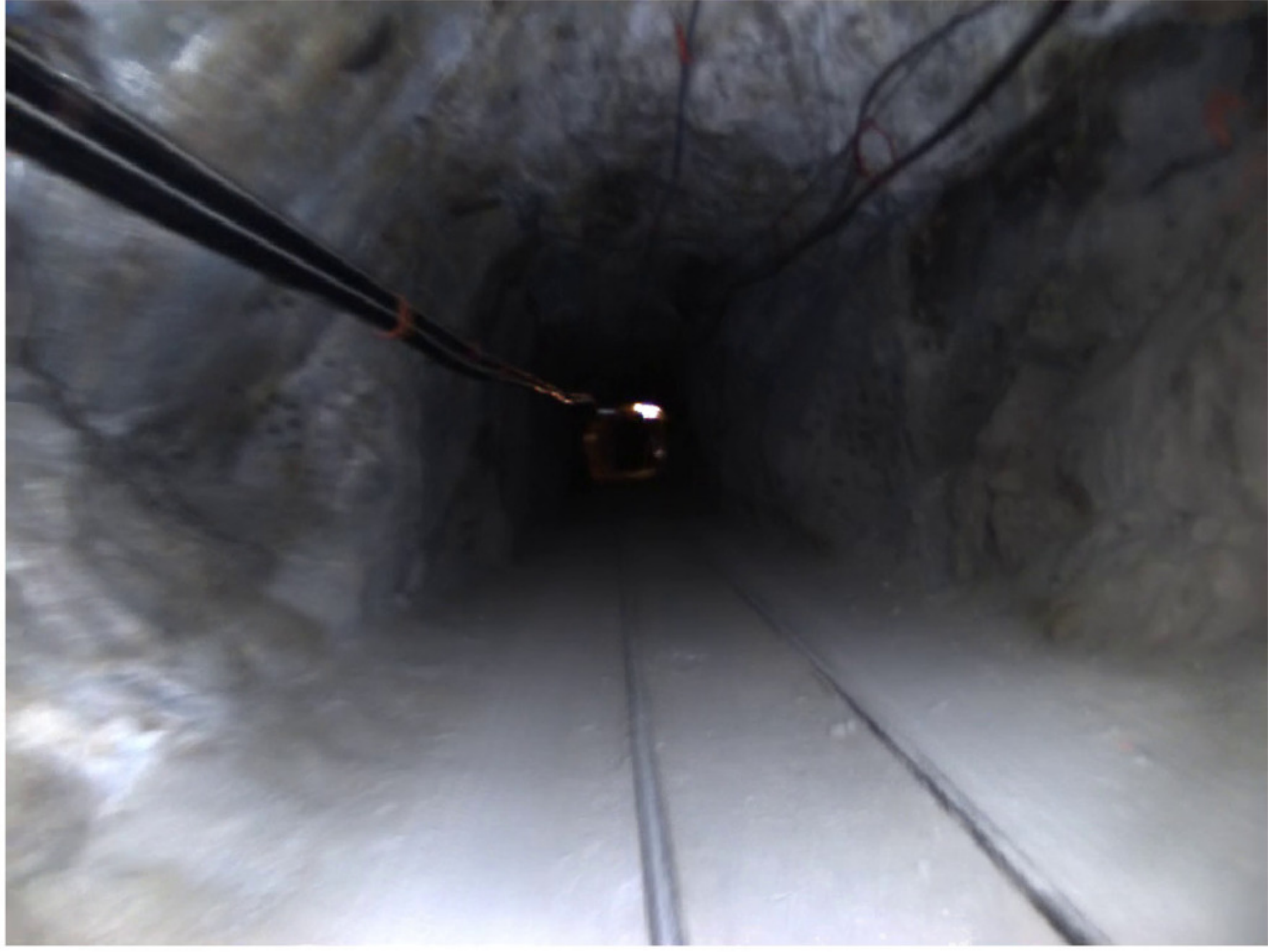}
\end{minipage}}

\caption{Traversability information (a) of section within the Edgar Experimental Mine in Idaho Springs, CO, USA, that contains railroad tracks. Raw traversability values of the lidar point clouds (top) are shown, where white is not traversable, and black is traversable. Resulting semantic map (bottom) illustrated non-traversable surfaces such as walls in white, traversable surfaces such as the ground in black, and semi-traversable surfaces such as the railroad tracks in grey. Note that red voxels do not contain traversability data. An accompanying photo (b) of the section of mine is shown for reference.}
\label{fig:traversability}
\end{figure}

Traversability is implemented in the Octomap framework using Equation \ref{eq:trav_fuse_eq} to estimate the traversability, $\tau_{v,t}$, of a given voxel, $v$, as a function of the voxel's occupancy probability, $P_{occ,v,t}$. The traversability estimate for the voxel is a linear combination its previous traversability estimate, $\tau_{v,t-1}$, and new estimate $\tau_{p,t}$ for the points in the voxel. An example of this process is shown in Figure \ref{fig:traversability}.

\begin{equation} \label{eq:trav_fuse_eq}
    \tau_{v,t} = \tau_{v,t-1} P_{occ,v,t} + \tau_{p \in \mathbb{V},t} (1-P_{occ,v,t}) \in [0,1]
\end{equation}

\subsubsection{Stair Classification \& Map Integration}
\label{sssec:stair_mapping}

Semantic information on stairs is fused into the mapping framework using the open source \emph{StairwayDetection} \cite{westfechtel2018robust} package and a binary Bayes filter \cite{thrun2005probabilistic}. Stair classification of point clouds via this approach consists of 4 major steps: (1) pre-analysis, in which the point cloud is downsampled and filtered, normal and curvature is estimated for each point, and floor separation is performed; (2) segmentation via a region growing algorithm, which segments the point cloud into smooth regions; (3) plane extraction, in which the surfaces that make up the riser and tread regions of each stair step are extracted; and (4) recognition, where the tread and riser regions are connected and analyzed via a graph to determine whether they make up a valid set of stairs.

\begin{figure}[!htb]
		\centering
		\subfloat[]{{\includegraphics[width=.45\textwidth]{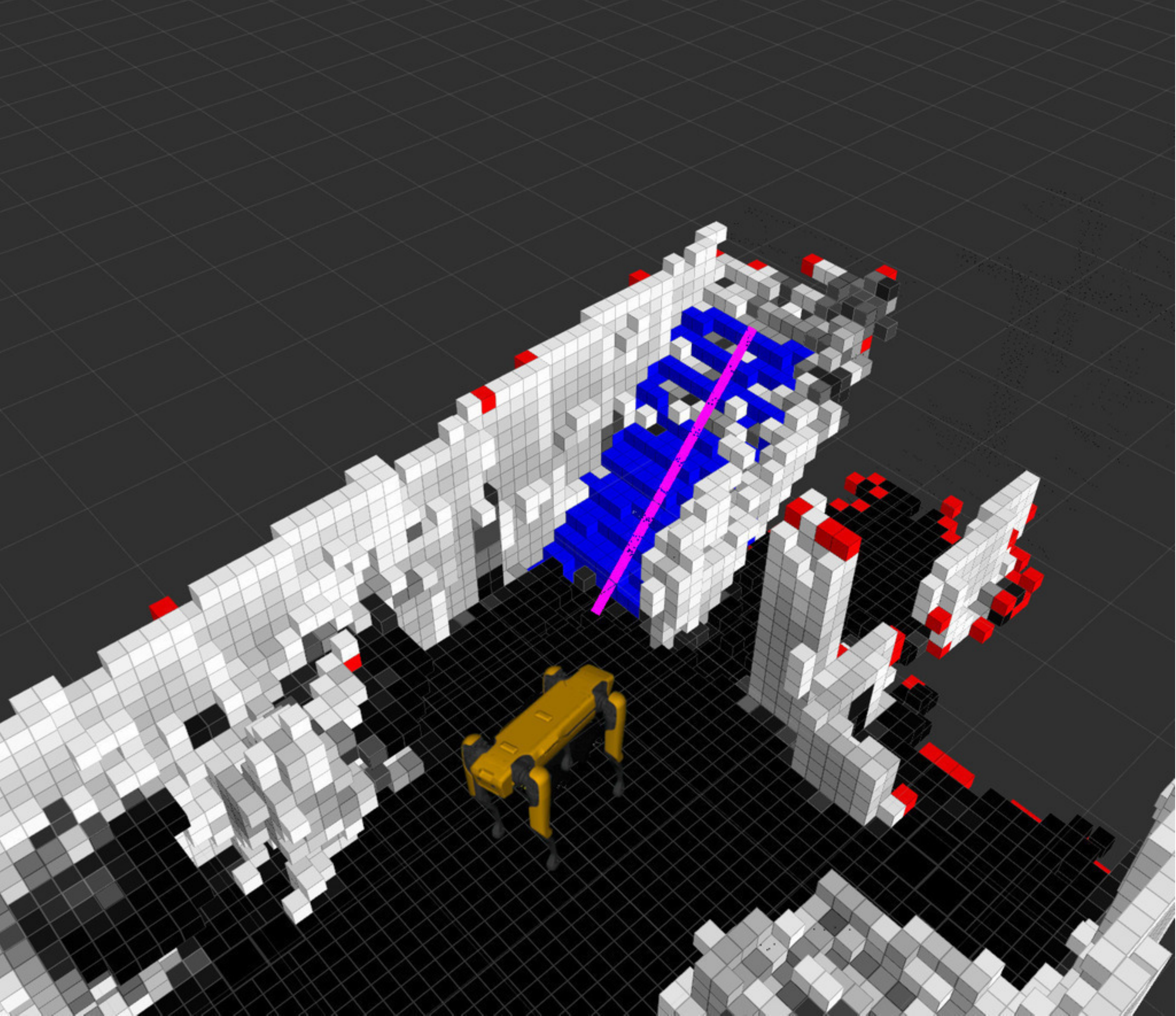}\label{fig:stairclass_map} }}
		\subfloat[]{{\includegraphics[width=.45\textwidth]{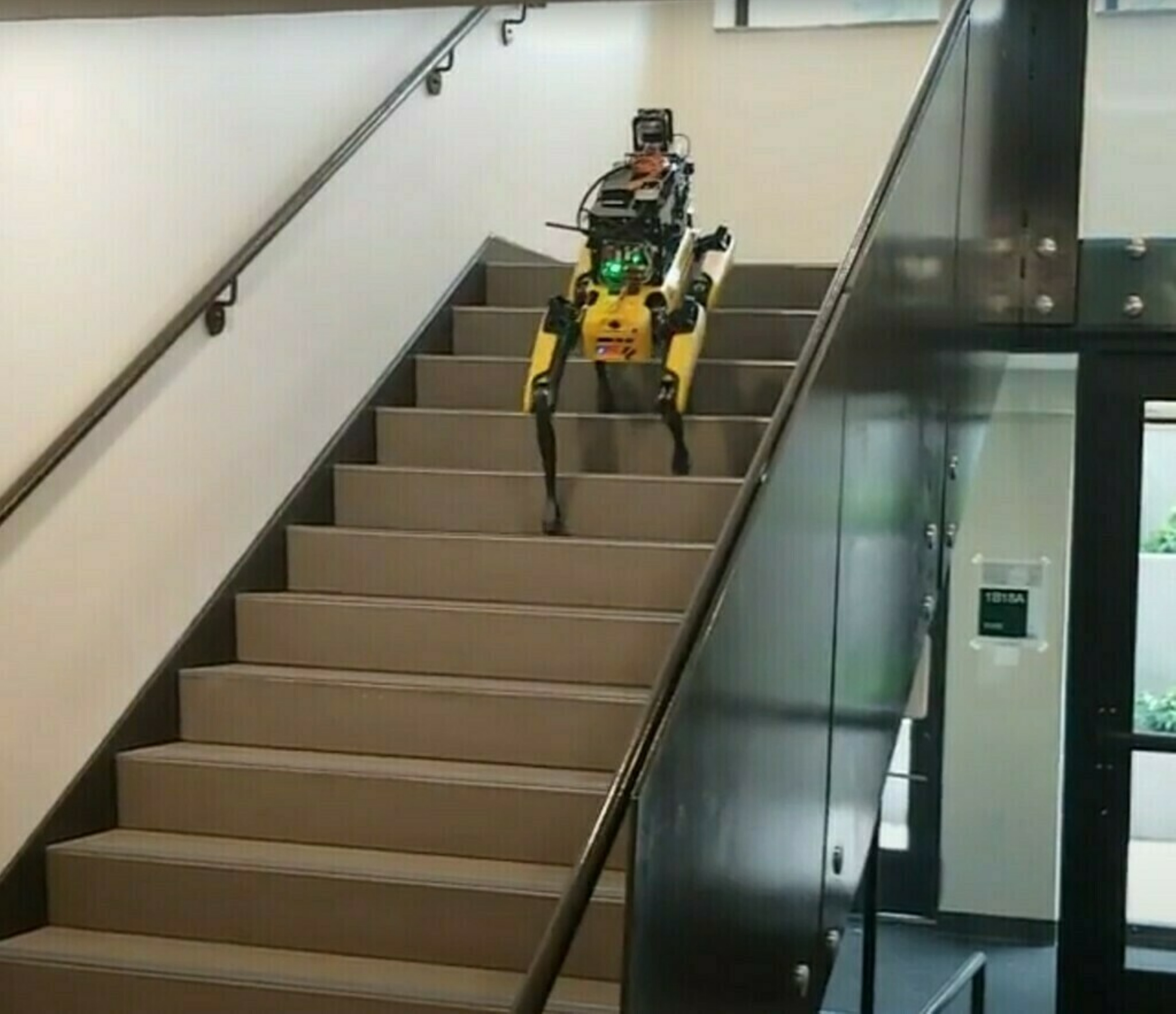}\label{fig:spot_on_stairs} }}
		\caption{Spot planning up a staircase using the estimated stair voxels in the Octomap shown in blue.}
		\label{stair_class_fig}
\end{figure}

Stair detections are integrated into the map using a similar mechanism to the log-odds probability which determines occupancy in octomap. A binary Bayes filter \cite{thrun2005probabilistic}, shown in Equation  \ref{eq:binary_bayes}, is used to estimate the probability that a given voxel is a part of a staircase. The extracted points from the stairway detector are modeled as measurements $z$ where $P_{stair}(n|z_{stair,t})$ is the probability that a voxel $n$ is part of a staircase. The measurement through time step $t$ is represented by  $z_{stair,1:t}$ as shown in Equation \ref{occ_prob_eq}. $\mathcal{L}_{stair}(n|z_{stair,1:t})$ as shown in Equation \ref{occ_prob_log_eq} are the corresponding log-odds probabilities which are used for fast updates updates to the probabilistic estimate of each voxel. More details of the log-odds formulation are provided in \cite{hornung2013octomap,thrun2005probabilistic}. Our filter is tuned to prioritize true positive detections with the following parameters: $P_{stair}(n) = 0.5$,  $\mathcal{L}_{stair,min} = -2.0$,  $\mathcal{L}_{stair,max} = 3.48$, $\mathcal{L}_{stair,hit} = 4.60$, $\mathcal{L}_{stair,miss} = -0.04$.

\begin{subequations} \label{eq:binary_bayes}
\begin{equation} \label{occ_prob_eq}
    P_{stair}(n|z_{stair,1:t}) = 
    \begin{bmatrix} 
        1 + \frac{1-P_{stair}(n|z_{stair,t})}{P_{stair}(n|z_{stair,t})} \frac{1-P_{stair}(n|z_{stair,1:t-1})}{P_{stair}(n|z_{stair,1:t-1})} \frac{P_{stair}(n)}{1-P_{stair}(n)}
    \end{bmatrix} 
    ^{-1}
     \in (0,1)
\end{equation} 

\begin{align} \label{occ_prob_log_eq}
    \mathcal{L}_{stair}(n|z_{stair,1:t})& = \mathcal{L}_{stair}(n|z_{stair,1:t-1}) + \mathcal{L}_{stair}(n|z_{stair,t}) \in [\mathcal{L}_{stair,min},\mathcal{L}_{stair,max}] \text{ ,} \\
    \text{where }
    & \mathcal{L}_{stair,min} \in (-\infty,0) \text{ , } \mathcal{L}_{stair,max} \in (0,\infty)  \text{ ,} \nonumber \\
    & \mathcal{L}_{stair}(n|z_{stair,t}) = 
    \begin{cases}
        \mathcal{L}_{stair,hit} > 0 \text{ on stairs } \\
        \mathcal{L}_{stair,miss} < 0 \text{ on non-stairs }
    \end{cases} 
    \nonumber 
\end{align} 
\end{subequations}

A raycast operation on the footprint of the vehicle is used to provide a binary signal indicating the robot is on stairs. Additionally, eigenvector decomposition is performed over each cluster of stair voxels to extract a straight path along the staircase. These triggers provide waypoints so that the local trajectory follower can navigate to the top of the staircase. It's important to note that since stairs would generally be classified as non-traversable, a stair label takes precedence over a traversability label for the Spot platform, which is capable of walking up stairs. Additionally, this method requires sufficient lidar scans of the staircase, which is generally available when located at the bottom of a staircase but is not when facing the stairs leading down. As a result, detecting and navigating a descending staircase is not feasible with the current configuration, but could be with a wider field-of-view sensor or programmed forward pitching behavior of the Spot.

The \textit{marble\_mapping} package enables the creation of difference-based Octomaps which allows for efficient transmission in underground environments. Furthermore the framework provides semantic and traversability information which the planner utilizes to ensure the robot is able to navigate safely. Details of the planner are described in Section \ref{sec:planning}.

\section{Path Planning}
\label{sec:planning}

Team MARBLE's heterogeneous fleet relies on autonomous path planning onboard each agent to reduce the workload of the human supervisor. The path planner running onboard each agent generates safe and traversable paths that lead to unexplored areas. Paths are planned on the Octomap-based \textit{marble\_mapping} framework described in Section \ref{sec:mapping}. Team MARBLE used the same planner on all robots with the only difference being the collision-function depending on vehicle's class. For instance, a wheeled robot cannot traverse stairs while a legged robot can. Existing methods discussed in Section \ref{ssec:planning_background} suffer computational costs that make it challenging to scale to large environments. Because the proposed planniner is computationally efficient and minimally dependent on tuning gains, it performs well in large-scale environments. Our planner makes several significant contributions, such as light on-demand terrain assessment, which is discussed in Section \ref{ssec:planning_sample_and_project}, hierarchical solution-search that also incorporates position history-based multi-agent coordination, which is discussed in Section \ref{ssec:planning_multiagent_coordination}, and handling of dynamic changes in the environment such as blocked passages, which is covered in Section \ref{ssec:planning_dynamic_replanning}.

\subsection{Background}
\label{ssec:planning_background}

One of the widely-known methods \cite{yamauchi1997frontier,ahmad20213d} for autonomous exploration relies on explicitly detecting potential frontiers on an explored map, followed by a path planned toward each cluster of frontiers. The method seeded significant developments in the area of autonomous robotic exploration since it was first proposed. However, this frontier-based method employs a computationally expensive optimization-based approach that plans paths to each frontier cluster, despite the fact that some may not be reachable.

In recent decades, the planning community has witnessed significant advancements in more computationally efficient sampling-based approaches for path planning and exploration. One instance of such development is an exploration planner that uses Rapidly Exploring Random Trees (RRT) \cite{lavalle2006planning} to sample an environment and chooses an \textit{optimal} path from the set of sampled ones. The method samples the environment as a single batch, and therefore is not scalable to large-scale environments. A rectification of this limitation is recently proposed by \cite{dang2020graph} where a bifurcation approach is introduced for sampling and exploration. This approach implies that the environment is sampled only in the local neighborhood of a robot while simultaneously building a sparse graph that scopes the entirety of the explored map. The latter is essential to deal with local minima such as dead-ends and also to plan a path back home. Our autonomous exploration solution for the SubT Final Event relies on the principle of bifurcation with additional contributions in the terrain assessment, solution-search, dynamic obstacle avoidance and coordination.

The graph-based planners based on sampling and bifurcation approach use high resolution depth images to compute a 2.5D grid-based elevation map using the technique presented in \cite{fankhauser2018probabilistic}. This elevation map is further filtered to segment terrain characteristics such as slope, roughness and step \cite{wermelinger2016navigation}. The authors of such graph planners mention the scalability challenges with such computationally expensive approaches, which limit their terrain awareness to regions local to the agent. This further leads to challenges such as the planned path and the underlying graph being generated with an over-optimistic view of the terrain, consequently needing the robot to be backed up if it encounters impassible terrain. 

\subsection{Sample-and-Project Strategy}
\label{ssec:planning_sample_and_project}

To rectify the terrain assessment challenges, a sample-and-project approach is followed, similar to the settling-based collision-check approach proposed in \cite{krusi2017driving}. We perform such checks on an Octomap with resolution 0.15m. At this resolution, all of team MARBLE's robots were at least three voxels wide, providing a decent amount of robot footprint to project a robot's pose on. SubT challenge rules highlight that the extremely narrow passages could be around 1m wide, with doorways as narrow as 36 inches. With this in mind, Octomap voxel length of 0.15m was small enough to navigate narrow passages and large enough to be able to keep up with computational complexity of generating such a map in a large-scale environment. In case of a wheeled robot, each voxel in the Octomap is labelled with a roughness value which is obtained using high resolution point clouds as described in Section \ref{ssec:semantic_mapping}. However, on Spot legged robots dense roughness information is not required because of their onboard terrain assessment. Additionally, for Spots, encode semantic information about stairways into the map which overrides the default height parameters of the planner. With explicit labels, the planner is able to plan paths over built up staircases despite elevation changes the robot would not normally traverse over. More formally, for the legged robots capable of traversing stairs, each Octomap voxel maps to a label from the set $\{ \text{`occupied'}, \text{`unknown'}, \text{`free'}, \text{`stair'} \}$, whereas in case of wheeled robots the label set is $\{ \text{`occupied'}, \text{`unknown'}, \text{`free'}, \text{`rough'} \}$.

\begin{figure}[!htb]
    \centering
    \includegraphics[width=0.65\linewidth]{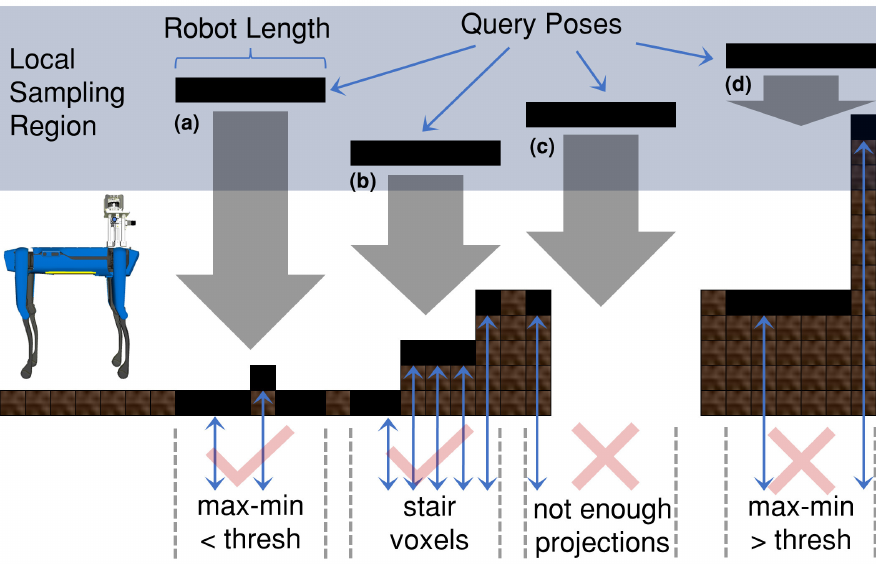}
    \caption{A depiction of sample-and-project strategy for terrain checks on an OctoMap. The figure shows four different types of query poses. (a) and (b) depict collision-free samples whereas (c) and (d) are marked under-collision or non-traversable.}
    \label{fig:planner_terrain_assessment}
\end{figure}

First, the environment is sampled in the local neighborhood of the robot using RRT$^*$, a variant of RRT with optimality considerations. Each tree sample is a robot position parameterized by the robot width and length. During sampling, the collision-checks are performed by vertically projecting a query sample to find the ground below it. Once the ground is found, the elevation change at the footprint of the sample is evaluated if there are enough projections on occupied voxels. In case of a wheeled robot, the average roughness information of the footprint voxels is also taken into consideration. For a legged robot, if a threshold amount of footprint voxels are labelled as `stairs', the sample is considered collision-free regardless of the elevation or roughness check. Expanding an RRT$^*$ requires checking path segments for collisions instead of isolated robot configurations. In order to check such a path segment, a set of robot configurations along the segment is checked for traversability. Fig. \ref{fig:planner_terrain_assessment} depicts the terrain assessment process.

\subsection{Solution Search and Multi-Robot Coordination}
\label{ssec:planning_multiagent_coordination}

At any replan iteration, a set of potential solutions include all of the local paths sampled using RRT$^*$ and all of the global paths ending at the graph frontiers. The former set of solutions is represented by $\mathcal{P}^L$ and latter solutions belong to the set $\mathcal{P}^G$. A set of all local paths that are leading the robot toward areas with greater than a defined threshold of volumetric gain and teammate separation are given as $\mathcal{P}^{LV}$ and $\mathcal{P}^{LS}$ respectively. Similarly, a set of all global paths that are leading the robot toward areas with greater than a defined threshold of volumetric gain and teammate separation are given as $\mathcal{P}^{GV}$ and $\mathcal{P}^{GS}$ respectively. These sets is highlighted in Figure \ref{f:planner_solution_search}. A typical approach to find an appropriate solution is to form an objective function with a combination of exploration objectives such as volumetric gain and exploration heading parameterized by the penalty gains. Volumetric gain calculation is, however, a computationally expensive operation and limits the amount of frontiers a robot can process in a reasonable amount of time. Our approach relies on finding a \textit{good enough} solution in terms of volumetric gain. To achieve multi-robot coordination, the position histories of teammate robots on the network are used. If a path is leading a robot to a point such that the minimum distance of the point from the position histories of the teammate robots is more than the mapping range, it is guaranteed that new areas are being explored. 

Following this intuition, the primary objective of Team MARBLE's solution search method is not to optimize for the volumetric gain and the teammate position histories separation, but to accept a solution that has a satisfactory amount of volumetric gain and distance from the teammate position histories. The formal objective of the path planner is to output a solution that belongs to $\mathcal{P}^{LV} \cup \mathcal{P}^{LS}$, $\mathcal{P}^{GV} \cup \mathcal{P}^{GS}$, $\mathcal{P}^{LV}$ or $\mathcal{P}^{GV}$ in the order of preference. The solution search process makes use of two different objective functions,

\begin{align}
    J^\alpha &= c_0 \mathcal{D}^P(p^\text{hist}, p^\text{cand}) - c_1 | \theta^\text{explore} - \theta^\text{cand}_\text{mean} | \nonumber  \\ 
    &- c_2 | h^\text{explore} - h^\text{cand}_\text{mean} |, \label{cost_fun_1} \\
    J^\beta &= -c_1 | \theta^\text{explore} - \theta^\text{cand}_\text{mean} | - c_2 | h^\text{explore} - h^\text{cand}_\text{mean} | \nonumber \\
    &+ c_3 \mathcal{G}^V(p^\text{cand}(1)) + c_4 \mathcal{D}^S(p^\text{cand}(1), p^\text{hist}_1,...,p^\text{hist}_1), \label{cost_fun_2}
\end{align}

where $J^\alpha$ is used to find a candidate path that aligns best with the current exploration heading of the robot and $J^\beta$ is leveraged to perform a thorough search if required. The sets of points $p^\text{cand}$ and $p^\text{hist}$ represent a list of candidate solutions and the position history of a robot respectively. The exploration heading $\theta^\text{explore}$ is calculated by averaging the most recent few points on $p^\text{hist}$ of the robot. The mean heading and mean height of a candidate path are denoted by $\theta^\text{cand}_\text{mean}$ and $h^\text{cand}_\text{mean}$ respectively. The function $\mathcal{D}^P$ accepts two paths as arguments and calculates the mean of minimum distance of all points along the first path with the second path. Furthermore, the function $\mathcal{D}^S$ calculates the minimum distance of a candidate path from the position histories of all other teammate robots.

Algorithm \ref{algo:planner_solution_search} provides a deeper insight into the solution search steps. As a first step, a collision-free local path is found that best aligns with the direction of travel of the robot. This path is then checked if it has a satisfactory amount of volumetric gain and distance from the teammate position histories. If a good-enough solution is found at this step, the solution is returned and only a single volumetric gain function call is required. Therefore, we save significant computation time during most replan iterations. If a solution is not found at this first step then a more thorough search is performed, first through the sampled local paths and then through the global paths leading toward graph frontiers. This search is highlighted in Algorithm \ref{algo:planner_solution_search}. The functions \textsc{Plan}\textsc{Locally}() shown in Algorithm \ref{algo:plan_locally},  and \textsc{Plan}\textsc{Globally}() shown in Algorithm \ref{algo:plan_globally}, are responsible for outputting solutions that satisfy both volumetric gain and teammate separation constraints if possible, otherwise they output solutions that only satisfy the volumetric gain constraint. In the worst case, when neither constraint can be satisfied, paths with maximum teammate separation are returned as a contingency.

This attempt of finding a solution by breaking the potential solution space down into subsets instead of having one objective function to optimize over the entire space, helped us avoid extensive gain tuning. During testing and final event runs, we found our approach to be scalable for environments of various sizes without a need for tuning gains for different environment types. The details of the sampling-based path planner can be found in \cite{ahmad2022efficient}. In this work, a simulation comparison of the proposed planner with an existing sampling-based planner \cite{dang2020graph} is presented, highlighting the improvement in scalability and computational efficiency.

\begin{algorithm}[!ht]
\caption{ScanPlan Solution Search.}
\label{algo:planner_solution_search}
    \begin{algorithmic}[1]
        \STATE $ p^l_\alpha \leftarrow$ $p^\text{cand} \in \mathcal{P}^L$ costing minimum $J^\alpha$
        
        \IF {$\mathcal{G}^V(p^l_\alpha) \geq v^g_\text{thresh}$ \AND $\mathcal{D}^S(p^l_\alpha) \geq s^g_\text{thresh}$ }
            \STATE return $p^l_\alpha$ 
        \ENDIF
        
        \STATE $ p^l_\beta \leftarrow$ \textsc{Plan}\textsc{Locally}()
        
        \IF {$p^l_\beta$ is non-empty \AND $\mathcal{D}^S(p^l_\beta) \geq s^g_\text{thresh}$ }
            \STATE return $p^l_\beta$ 
        \ENDIF
        \STATE $p^l \leftarrow p^l_\beta$
        
        \STATE $ p^g \leftarrow$ \textsc{Plan}\textsc{Globally}()
        
        \IF {$p^g$ is non-empty \AND $\mathcal{D}^S(p^g) \geq s^g_\text{thresh}$ }
            \STATE return $p^g$ 
        \ELSIF{$p^l$ is empty \AND $p^g$ is empty}
            \STATE return $p^l_\alpha$
        \ELSIF{$p^g$ is empty \OR \\ ($p^l$ is non-empty \AND $\mathcal{D}^S(p^l) \geq \mathcal{D}^S(p^g)$)}
            \STATE return $p^l$
        \ELSIF{$p^l$ is empty \OR \\ ($p^g$ is non-empty \AND $\mathcal{D}^S(p^g) \geq \mathcal{D}^S(p^l)$)}
            \STATE return $p^g$
        \ENDIF
    \end{algorithmic}
\end{algorithm}

\begin{algorithm}[!htb]
\caption{\textsc{Plan}\textsc{Locally}(). Returns a path in $\mathcal{P}^{LV} \cap \mathcal{P}^{LS}$ or $\mathcal{P}^{LV}$ in the order of preference.}
\label{algo:plan_locally}
    \begin{algorithmic}[1]
        \STATE $J^\beta_\text{min} \leftarrow +\inf$
        \STATE $p^\text{res} \leftarrow \text{none}$
        \STATE $\text{success} \leftarrow \text{false}$
        \FOR{$p_l \in \mathcal{P}^L$}
        
          \IF {$\mathcal{G}^V(p^l) \geq v^g_\text{thresh}$ \AND $\mathcal{D}^S(p^l) \geq s^d_\text{thresh}$ \AND $\sim \text{success}$ }
          \STATE $J^\beta_\text{min} \leftarrow J^\beta(p^l)$, $p^\text{res} \leftarrow p^l$, $\text{success} \leftarrow \text{true}$
          
          \ELSIF {$J^\beta(p^l) \geq J^\beta_\text{min}$}
            \STATE continue
          
          \ELSIF { ($\mathcal{G}^V(p^l) \geq v^g_\text{thresh}$ \AND $\mathcal{D}^S(p^l) \geq s^d_\text{thresh}$) \OR ($\mathcal{G}^V(p^l) \geq v^g_\text{thresh}$ \AND $\sim \text{success}$) }
            \STATE $J^\beta_\text{min} \leftarrow J^\beta(p^l)$, $p^\text{res} \leftarrow p^l$
          \ENDIF 
        
       \ENDFOR 
       \STATE return $p^\text{res}$ \\
    \end{algorithmic}
\end{algorithm}

\begin{algorithm}[!htb]
\caption{\textsc{Plan}\textsc{Globally}(). Returns a path in $\mathcal{P}^{GV} \cap \mathcal{P}^{GS}$ or $\mathcal{P}^{GV}$ in the order of preference.}
\label{algo:plan_globally}
    \begin{algorithmic}[1]
        \STATE $\mathcal{P}^{GV} \leftarrow \emptyset$, $\mathcal{P}^{GVS} \leftarrow \emptyset$
        \FOR{$p^g \in \mathcal{P}^G$}
            \IF{$\mathcal{G}^V(p^g) \geq v^g_\text{thresh}$ \AND $\mathcal{D}^S(p^g) \geq s^d_\text{thresh}$}
                \STATE $\mathcal{P}^{GVS} \leftarrow \mathcal{P}^{GVS} \cup \{p^g\}$
            \ELSIF{$\mathcal{G}^V(p^g) \geq v^g_\text{thresh}$}
                \STATE $\mathcal{P}^{GV} \leftarrow \mathcal{P}^{GV} \cup \{p^g\}$
            \ENDIF
        \ENDFOR
       
        \IF{ $\mathcal{P}^{GVS}$ is empty  }
            \STATE return path $p^g \in \mathcal{P}^{GV}$ with maximum $\mathcal{D}^S(p^g)$
        \ENDIF
        \STATE $p^g_r \leftarrow $ path from $\mathcal{P}^{GVS}$ leading to most recent frontier
        \STATE $p^g_c \leftarrow $ path from $\mathcal{P}^{GVS}$ leading to closest frontier
        
        \IF{\textsc{Path}\textsc{Length}($p^g_r$) $\geq$ \textsc{Path}\textsc{Length}($p^g_c$)}
            \STATE return $p^g_c$
        \ELSE
            \STATE return $p^g_r$
        \ENDIF
    \end{algorithmic}
\end{algorithm}

\threefer{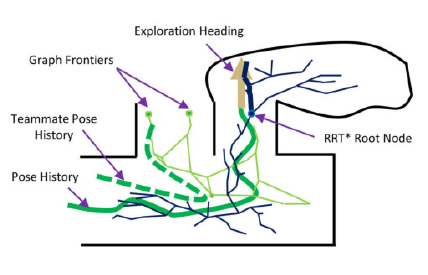}{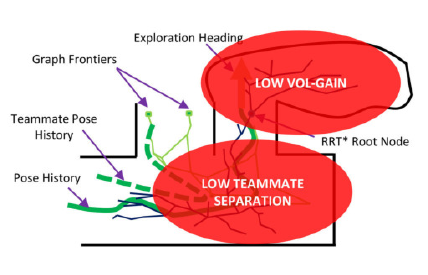}{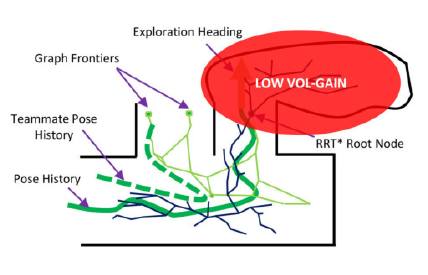}{planner_solution_search}{The figures highlight the planner solution search. The potential solutions for the planner include the paths leading toward the leaves of the RRT$^*$ tree (blue) and the frontiers of the graph (green). (a) The first preference of the solution is the path that aligns best with the robot's exploration heading (bold blue path). In case this path has both sufficient volumetric gain and teammate separation, it is returned as a solution. (b) As a second preference, a thorough search is performed to find a local or global path that satisfies both constraints. (c) If none of the paths is found at both of the steps above then a path that satisfies the volumetric gain constraint is accepted as a potential solution. }

\subsection{Dynamic Replanning}
\label{ssec:planning_dynamic_replanning}

Another challenge faced by the existing graph-based planners is that they rely on building a parallel graph representation of the environment. This representation does not naturally reflect changes in the environment, such as closed passages which were initially open at the time the graph is built. To handle this exception, the graph edges are labeled with a boolean representing its occupancy. During exploration, the planned paths are constantly checked for collisions. If a planned path is under-collision, all edges in the local neighborhood of the robot are validated for collision and marked accordingly. Moreover, all occupied edges are checked for occupancy all the time when the planner finds some idle time which mostly happens when the vehicle is following a path. This enables the planner to take into account the cases where an occupied area is free again. In the case where the graph search is performed to plan a global path, the occupied edges are ignored.

\section{Communication Systems}
\label{sec:comm_systems}

Effective communication with deployed systems from a fixed human operator is a crucial component of a complete robotic exploration system. While robots are capable of independent localization, mapping, and artifact detection, the addition of a communication infrastructure is a force multiplier to enable human supervisory control, inter-robot coordination, and timely artifact reporting. We developed a mesh network system to provide long-reach communications into underground environments which prioritizes reconnection times to maximize opportunities for data transfer.

\subsection{Background}

Previous work has developed several solutions to common problems encountered with deploying mesh networks, such as discovery and optimal routing. A wide variety of both closed-source and open-source solutions exist that include both hardware and software components.  Mesh networking can largely be subdivided into three layers: physical, logical, and transport. We will detail several prominent open-source or commercially available options for each layer before describing our final solution. 

From a logical layer standpoint, meshing layers lay between the physical transmission of frames over the medium and a higher-level protocol such as IP. For mobile robots operating in subterranean environments, a responsive mesh layer that minimizes lost link time is a major requirement due to the rapid movement of the robots. Further, to reduce integration effort, a mesh layer that operates at layer 2 of an OSI stack \cite{ISO74981} is desirable to allow transparent use of higher-level protocols such as ARP and IP.  Typically, meshing algorithms such as OLSR \cite{clausen2003optimized} and AODV \cite{perkins2003rfc3561} select a single best path for routing between nodes which hinders algorithmic performance in dynamic environments. A more recent example of a single-path logical meshing layer is Better Approach to Mobile Ad-hoc Networking-Advanced (\textit{batman-adv}) \cite{seither2011routing}, an open source implementation of a layer 2 mesh.  In contrast to \textit{batman-adv}, \textit{meshmerize} \cite{pandi2019meshmerize} provides multiple paths between nodes to ensure a reliable connection while still operating at layer 2; these multiple paths allow for a dramatic decrease in reconnect times when mesh topology changes. We relied on \textit{meshmerize} as our layer 2 meshing solution in cooperation with  Meshmerize GmBH. 

\begin{figure*}[!htb]
		\centering
		\subfloat[]{{\includegraphics[width=.24\textwidth]{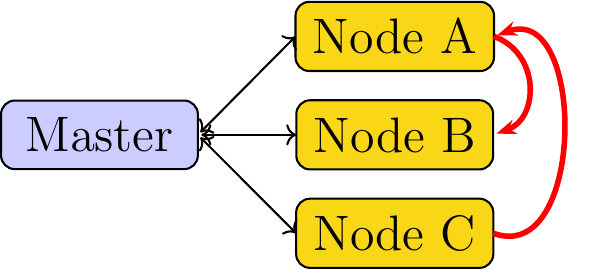}\label{f:comm_ros_l} }}%
		\subfloat[]{{\includegraphics[width=.34\textwidth]{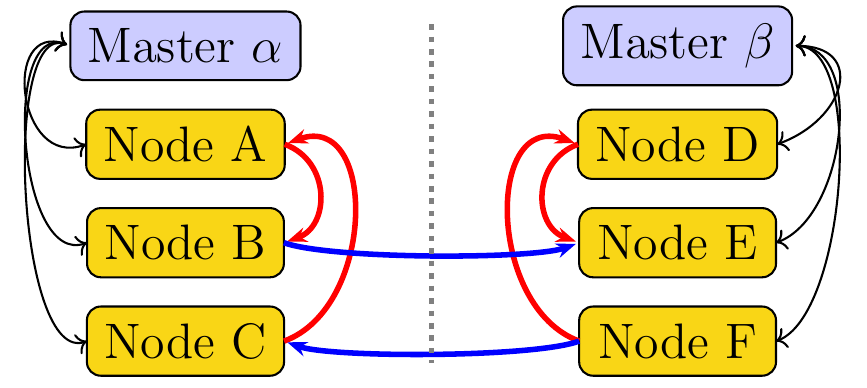}\label{f:comm_ros_c} }}%
		\subfloat[]{{\includegraphics[width=.39\textwidth]{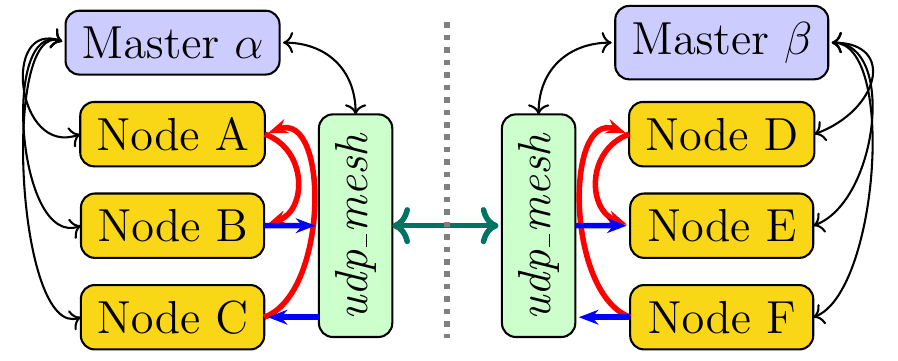}\label{f:comm_ros_r} }}%
		\caption{
		Several different types of ROS architectures. Red lines indicate data transfer, black lines indicate directory management, and blue lines are data paths that cross network segments. A basic, single-master ROS network node graph is shown in (a). 
		A \textit{fkie\_multimaster} multi-master ROS network node graph is shown in (b).
		In contrast to both (a) and (b), (c) shows how \textit{udp\_mesh} creates a single virtual channel between nodes, shown in green, to implement data prioritization.
		}
		\label{f:comm_ros}
\end{figure*}

Only transport layers designed for ROS were considered for ease of integration with the rest of the autonomy stack. In a traditional networked ROS architecture, a single computer runs a main node known as the \textit{rosmaster} that coordinates the publish-subscribe mechanisms. When a node wishes to exchange data with another node via named \textit{topics}, the master is consulted to determine the computer to connect to, as in \fig{comm_ros_l}. A single \textit{rosmaster} serves as a central directory of nodes and topics; when a subscription to a topic is requested, a list of publisher nodes is returned so that point-to-point TCP connections can be made directly between publisher and subscriber. These direct TCP connections break down when systems are linked over unreliable mesh networks which necessitates the need for an alternative transport mechanism. 

One open source transport layer \textit{multimaster\_fkie} \cite{juan2015multi} solves the discovery and advertisement problems using multi-cast packets and specialized nodes on each machine with an architecture shown in \fig{comm_ros_c}. However \textit{multimaster\_fkie} does nothing to establish prioritization of data flow. With the standard TCP transport provided by ROS, there is no centralized means of monitoring inter-node connections to arbitrate data priorities. Prioritization is crucial for monitoring the robot fleet in intermittent communication situations. Mission critical data such as artifacts needs to make it through to the human supervisor before other auxiliary data such as odometry and maps.  

One alternative to \textit{multimaster\_fkie},  \textit{Pound}\footnote{\url{https://github.com/dantard/unizar-pound-ros-pkg}} \cite{tardioli2019pound}, is specifically designed for use in unreliable mesh networks and implements many of the desired requirements. However, \textit{Pound} relies on hardcoded topic names and fixed addressing information, which limit the flexibility of the system. Alternatively, \textit{nimbro\_network} \cite{schwarz2016supervised} implements a similar set of functions with regards to transport over wireless networks, but omits prioritization. Crucially, \textit{nimbro\_network} still utilizes TCP for reliable inter-robot communication, preventing adaptation of core TCP behavior (particularly retransmits) to unreliable mesh networks; UDP links are only used for non-guaranteed data delivery.

\subsection{UDP Mesh}

The main innovation in our system is our transport layer, \textit{udp\_mesh} which allows for runtime reconfiguration, implements prioritization, and re-implements reliable communication over UDP to allow for more refined control over retransmits and fragmentation. Fundamentally, the \textit{udp\_mesh} layer uses only unicast and broadcast UDP datagrams to implement higher-level services without requiring multicast support. In principle, multicasting would offer a performance benefit by reducing broadcast traffic. However, in a wireless mesh environment, these potential gains are offset by multicast group membership management overhead.

\subsubsection{Discovery and Address Resolution}
\textit{Discovery} is the process of identifying nodes that are available for communication. We implement discovery through the use of a periodic heartbeat broadcast that advertises the node's availability and provides name resolution information. In concept, this service is similar to the \textit{mcast\_dns} service in Linux, where peers advertise their naming information to be able to address nodes by hostname instead of layer 2 MAC or layer 3 IP address. Nodes identified through discovery are added to the list of available nodes for communication as well as status reporting. This discovery heartbeat is also used as a lost-communications detector to prevent higher-level messages from queueing for unreachable nodes. 

\subsubsection{ROS Message Encapsulation}
In the ROS ecosystem, messages are translated from a message definition language specification into internal representations appropriate to the implementing language\footnote{\url{http://wiki.ros.org/msg}}. This same language specification is used to serialize and deserialize messages; that is, to transform a ROS message into a buffer of bytes suitable for transmission over an arbitrary channel. \textit{udp\_mesh} implements a generic message passing system such that the message to be transmitted is never deserialized, saving a significant amount of processing time in the case of complex, large message types such as images. Instead, a generic subscriber is used to acquire the serialized bytes for direct use to be transmitted to other nodes. On the receiver side, the transmitted byte stream is deserialized to instantiate the message in a format that other ROS ecosystem nodes can readily consume. These two functions abstract the transport of arbitrary messages over the \textit{udp\_mesh} layer and remove any requirement to define a list of acceptable message types.

\subsubsection{Point to Point Transport}
In the \textit{udp\_mesh} system, point-to-point transport is implemented via UDP datagrams. This envelope contains provisions for sequence tracking, fragmentation, and message reconstruction. When preparing a message for transmission, the byte buffer provided by the ROS encapsulation service is split into chunks that fit inside the underlying medium's maximum transmit unit (MTU). We use the standard 802.11 framing with an MTU size of 1500 bytes, out of which 100 bytes are reserved for overhead, leaving 1400 bytes for payload.

In the implementation of our system, a configurable number of message fragments are permitted to be `in flight' at any given time, similar to TCP congestion window control. In order for the next fragment to be transmitted, the receiver must send an acknowledgment. During unit testing to determine an appropriate value for the number of in-flight fragments permitted, an initial increase yields improved throughput. However, after a certain point, throughput decreases as multiple packets are queued for transmission on the medium and start to destructively interfere. As a compromise determined via empirical testing, three packets are permitted to be in-flight between any two nodes at a time. With this configuration, our transport-layer throughput is approximately 
20 Mbit/s of payload data, measured using raw images as representative high-density traffic over a wired gigabit Ethernet link.

Retransmits are automatically queued until either an acknowledgment is received or the host is marked offline due to non-reception of any heartbeat or acknowledgment messages. Once a host is marked offline, any attempts to send messages are discarded. Hosts may become online once again after receipt of a discovery message. On the receiver side, the message is kept in a temporary state while the fragments arrive. Should message fragments stop arriving, the partial message is purged after a timeout and the host is once again marked offline which indicates to higher levels that reliable transport is unavailable. In this case, the higher level is BOBCAT, which is discussed in Section \ref{sec:mission_management}.

\subsubsection{Quality of Service}
\textit{Quality of Service} (QoS) is the notion that some traffic should be prioritized over other traffic for use of a limited communications channel, e.g, artifact reports need to arrive before mapping data. Fundamentally, TCPROS (the default transport used in ROS v1) is not capable of implementing a QoS scheme where a limited channel is shared between different topics (\fig{comm_ros_r}), as every node subscribing to a topic uses an individual TCP point-to-point link with no information about other links. This need to prioritize traffic was the driving rationale behind the development of the \textit{udp\_mesh} layer. As part of the configuration of the layer, each topic to be transported includes a priority number. Internally, this priority number is used as a sorting key to order message fragments for transmission.

\subsubsection{Point-to-Multipoint Transport}
Although \textit{udp\_mesh} is based around point-to-point message transfer, mission requirements sometimes necessitate system-wide messaging. For example, broadcast methods are used within the \textit{udp\_mesh} layer to manage name resolution. To facilitate these type of messages originated at higher levels, a broadcast mechanism is provided by the transport layer. For messages that fit within a single MTU, a single, unacknowledged UDP broadcast is used to distribute the message. For larger messages, individual links to each node are used to send the broadcast as a series of unicast fragments using the same accounting and acknowledgments as the point-to-point mechanism.

\subsection{Final Solution}
The final communication solution used meshmerize as the logical layer with \textit{udp\_mesh} as the transport layer. Both robots and beacons acted as nodes in the mesh with robots carrying 1W radios and beacons carrying 2W radios. Beacon drops are controlled by the methodology described in Section \ref{sec:relay_deployment}. Table \ref{tab:comm_system_dev} shows the evolution of our final networking solution from the Tunnel event through the Final Event. Our meshing solution, including the \textit{meshmerize} layer 2 software stack, was implemented on \textit{ath9k}-compatible 802.11 hardware, while \textit{udp\_mesh} was implemented on high-level compute units. Because of this split and radio hardware commonality, all of our radios ran essentially the same firmware image built off of an OpenWRT\footnote{\url{http://www.openwrt.org}} base. Our beacons only participated in the mesh at layer 2, and as such did not contribute to any broadcast traffic associated with \textit{udp\_mesh} services. By providing a reliable ROS-compatible mesh networking layer, higher-level autonomy and human interface via BOBCAT could be provisioned without knowledge of the underlying infrastructure. 

\begin{table}[hbt!]
\begin{center}
    \begin{tabular}{c c c c c}
    \hline
    \textbf{Event} & \textbf{Physical} & \textbf{Data Link} & \textbf{Transport} & \textbf{Application} \\ 
    \hline
    Tunnel  &   ath9k   &   B.A.T.M.A.N.    &   fkie\_multimaster   &   marble\_multi\_agent   \\
    Urban   &   ath9k   &   meshmerize      &   fkie\_multimaster   &   marble\_multi\_agent   \\ 
    Final   &   ath9k   &   meshmerize      &   udp\_mesh           &   BOBCAT                 \\
    \hline
    \end{tabular}
\caption{\label{tab:comm_system_dev} Evolution of Team MARBLE's communication system. As a note, ath9k and meshmerize are commercially available, B.A.T.M.A.N. and fkie\_multimaster are open-source software, and upd\_mesh, marble\_multi\_agent, and BOBCAT are custom packages developed for the SubT Challenge.}
\end{center}
\end{table}


\section{Mission Management}\label{sec:mission_management}

While the combination of Team MARBLE's large scale positioning system, mapping, and planning solutions provide a solid foundation for autonomy, higher level cognition and reasoning is required to take full advantage of the system. For Team MARBLE, this higher level reasoning consists of a flexible mission management solution which keeps the robots on task and allows for higher level instructions from a human supervisor. The core of the mission management solution is Behaviors, Objectives and Binary States for Coordinated Autonomous Tasks (BOBCAT) \cite{Riley2021}. BOBCAT controls the decision-making process for each individual agent while a separate process known as Multi-Agent Data Collaboration for Autonomous Teams (MADCAT) controls the data sharing and waypoint deconfliction between robots. In this section we highlight the design decisions, and algorithm details behind BOBCAT and MADCAT.

\subsection{BOBCAT}

BOBCAT simplifies the robot and environment states using \textit{Monitors} such as communication status. The \textit{Monitors} are combined with weighted goals which \textit{Objective} such as finding artifacts or extending communications can be fulfilled. BOBCAT then selects the best \textit{Behavior} such as exploring or deploying a beacon to fulfill and the most important \textit{Objectives} to execute. A full list of implemented \textit{Monitors}, \textit{Objectives}, and \textit{Behaviors} can be seen in Section \ref{sec:sup_bobcat} of the Appendix.

Formally, a BOBCAT is defined by the tuple $\{{\bf x}, {\bf y}, {\bf w}, M, O, B, \pi_B\}$ where 
\begin{itemize}
    \item ${\bf x} \in X$ is the system state with state space $X$.
    \item ${\bf y} \in Y$ are the sensor measurements with measurement space $Y$.
    \item ${\bf w} \in W = \mathbb{R}^{|O|}_+$ is a vector of \textit{input weights}.  These weights are used by the respective \textit{Objective} functions and represent the relative importance of the \textit{Objective} to the overall mission
    \item $M$ is the set of \textit{Monitor} functions of the form $M_i: X \times Y \rightarrow \{0,1\} \; \forall M_i \in M$. \textit{Monitor} functions $M_i$ return a binary value based on the robot state and measurements.
    \item $O$ is the set of \textit{Objective} functions of the form $O_j: W \times \{0,1\}^{|M|} \rightarrow \{0, W_j\} \; \forall O_j \in O$. \textit{Objective} functions $O_j$ use the input weight $W_j$ and a logical combination of \textit{Monitor} outputs to return either the input weight or a 0, which indicates the current preference of the objective to be fulfilled.
    \item $B$ is the set of \textit{Behavior} functions of the form $B_k: \{0,1\}^{|M|} \times \{0, W_j \}^{|O|} \rightarrow \mathbb{R}_{\geq 0} \times F_k \; \forall B_k \in B$. \textit{Behavior} functions $B_k$ sum the outputs of the \textit{Objectives} associated to that \textit{Behavior}.  \textit{Monitor} outputs may be used to selectively inhibit specific \textit{Objective} weights during evaluation steps. The \textit{Behavior} function returns a real value that indicates the current utility score of the actions associated with that \textit{Behavior}, and a pointer to an execution function.  A \textit{Behavior} may have a null execution function.
    \item $\pi_B$ is the policy for selecting the \textit{Execution Behavior} $B_E$ based on each of the \textit{Behavior} utility scores.
\end{itemize}

\begin{figure}[htbp]
    \centering
    \includegraphics[width=\textwidth]{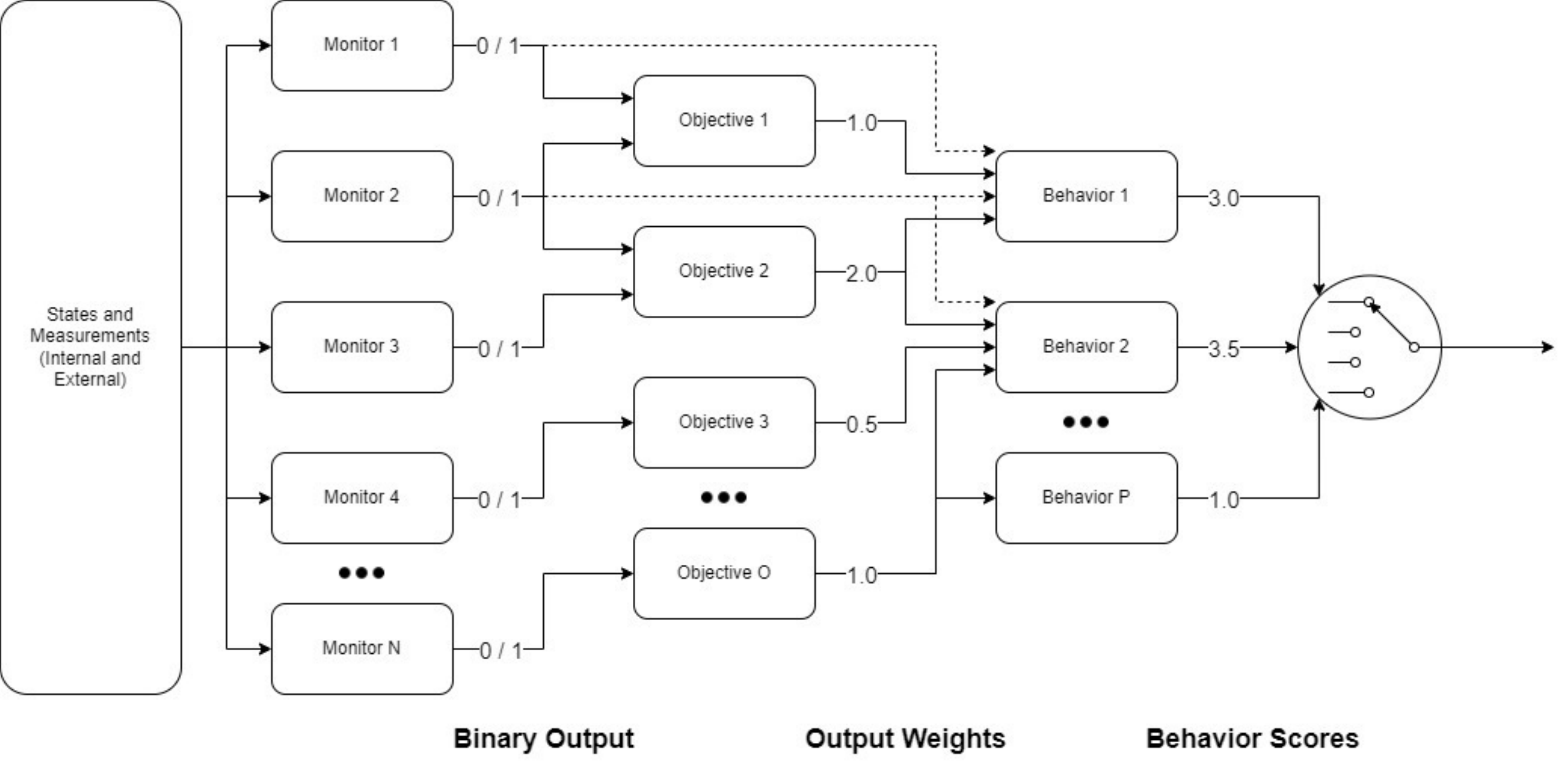}
    \caption{Graphical overview of \textit{BOBCAT}.  Numbers represent binary outputs, output weights, and behavior scores, respectively. A full list of monitors, objectives and behaviors is provided in the Appendix in Tables  \ref{tab:monitor_table}, \ref{tab:objective_table}, \ref{tab:behavior_table} respectively.}
    \label{fig:bobcat_overview}
\end{figure}

\textit{BOBCAT} can be represented graphically as in Figure \ref{fig:bobcat_overview}.  States and measurements from both the robot itself and external agents in a multi-agent scenario feed the various \textit{Monitors}.  This represents what the robot ``knows", and provides a binary output to the rest of the system. The \textit{Monitor} output lines in Figure \ref{fig:bobcat_overview} and other figures represent the cases where the \textit{Monitor} is associated with the respective \textit{Objective} or \textit{Behavior}.

\subsection{MADCAT}
\label{madcat}
\begin{figure}[!htbp]
    \centering
    \includegraphics[width=0.90\textwidth]{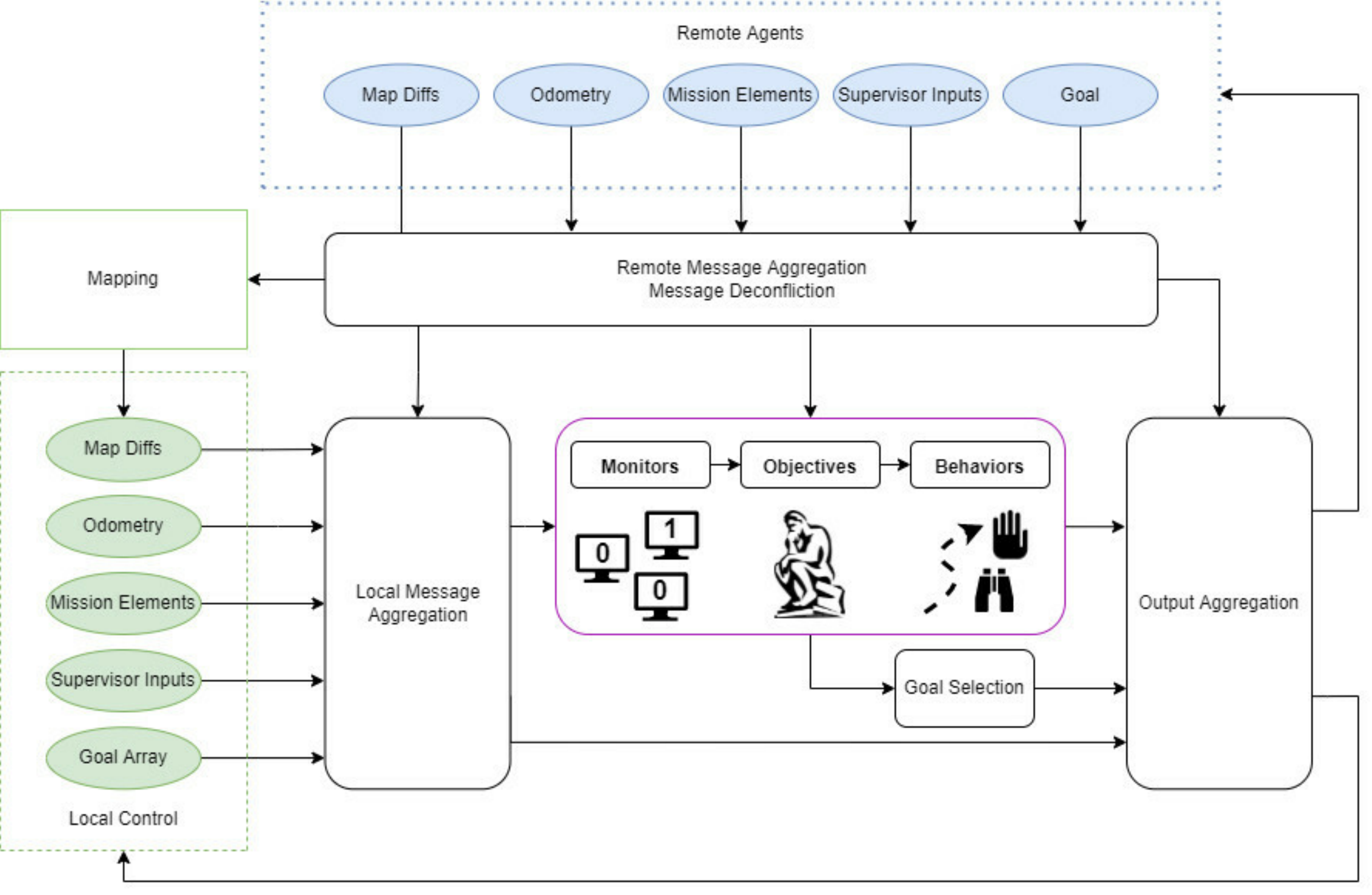}
    \caption{Overview of the Multi-Agent Data Collaboration for Autonomous Teams (\textit{MADCAT}) framework.}
    \label{fig:mad_cat_framework}
\end{figure}

The MADCAT framework depicted in Figure \ref{fig:mad_cat_framework} provides the multi-agent data sharing capabilities required for the mission.  The framework includes transmission of relevant coordination data and maps, as well as map merging functionality and decision making for each agent. \textit{MADCAT} uses \textit{BOBCAT} to accomplish the high-level mission management for individual agents with additional higher-level direction provided by the human supervisor. 

\subsubsection{Messages}

MADCAT sends most messages by broadcast with no acknowledgement required, and therefore does not require the sender to needlessly wait. This allows any agent who receives the message to act accordingly without a requirement to respond. This is helpful in the event the sender leaves communications range shortly after the broadcast.  An exception to this policy is made for high bandwidth data such as maps and images, because the receiver can not act on incomplete data. Bandwidth is not strictly managed, but instead uses a `best-available' strategy consistent with the prioritizations assigned to differing message classes, e.g. telemetry, supervisor commands, maps, and FPV.

An agent's pertinent local messages are concatenated into a single message in order to limit the number of messages broadcast over the communications channels.  Messages are re-built and broadcast every second.  Only the most recent message is needed for time-varying data such as odometry or the current goalpoint and any older messages are discareded. Other data such as artifact reports and relay locations are appended to a growing list, so any message a remote agent received has all of this type of data.  Larger data that could grow to become impractical to transmit repeatedly, such as maps and images, use a point-to-point handshake transmission. Messages are deconflicted using sequence numbers, to allow agents to share the messages of other agents but ensure only the latest data is used by the receiver.

Each agent's broadcast message contains not only its own local data, but that of any neighbor agents as well.  This allows downstream agents who can communicate with agent A but not agent B to still receive relatively current information from agent B.

\subsubsection{Artifact Report Management}
Agents keep track of both their own detected artifacts using the procedure described in Section \ref{sec:object_detection} as well any artifacts they have received from other agents. The framework aggregates all of the artifact reports and the \textit{Artifact} monitor determines if the agent needs to return to communications to report the new information to the base station.  The base station further parses these messages for display, selection, and transmission to the scoring system. More details of this display can be found in Section \ref{sec:hs_interface}. Images are sent using a point-to-point request system over a low priority channel to reduce bandwidth requirements.

The \textit{BOBCAT} \textit{Artifact} monitor which is triggered by 3 unreported artifacts or 5 minutes of exploration with a pending artifact, ultimately determines whether the robot should return to communications for unreported artifacts.  The raw artifact reports are always used in this determination, but transmission of images is configurable.  By default, and as configured during the Final Event Prize Run, artifact images not received by the base station are considered unreported artifacts, and will force the robot to return to communications until they are fully transmitted.

\subsection{Beacon Deployment}\label{sec:relay_deployment}
The framework is responsible for identifying locations to deploy communications relays to extend the communications reach into the environment. It uses a combination of communications status, distance and turn detection to identify potential locations. The human supervisor is also able to command drops based on a robot's location on the map.

\subsection{Goal Selection}
Some behaviors, particularly \textit{Explore}, require a goal selection step once that behavior has been chosen to execute. The goals either come from the global planner described in Section \ref{sec:planning} or from human supervisor input. If two agents goals are found to be conflicting, BOBCAT requests a new goal from the planner which provides a path to the next closet goal point.

\subsection{Human Supervisor Interface}\label{sec:hs_interface}

\begin{figure}[!htb]
    \centering
    \includegraphics[width=0.8\textwidth]{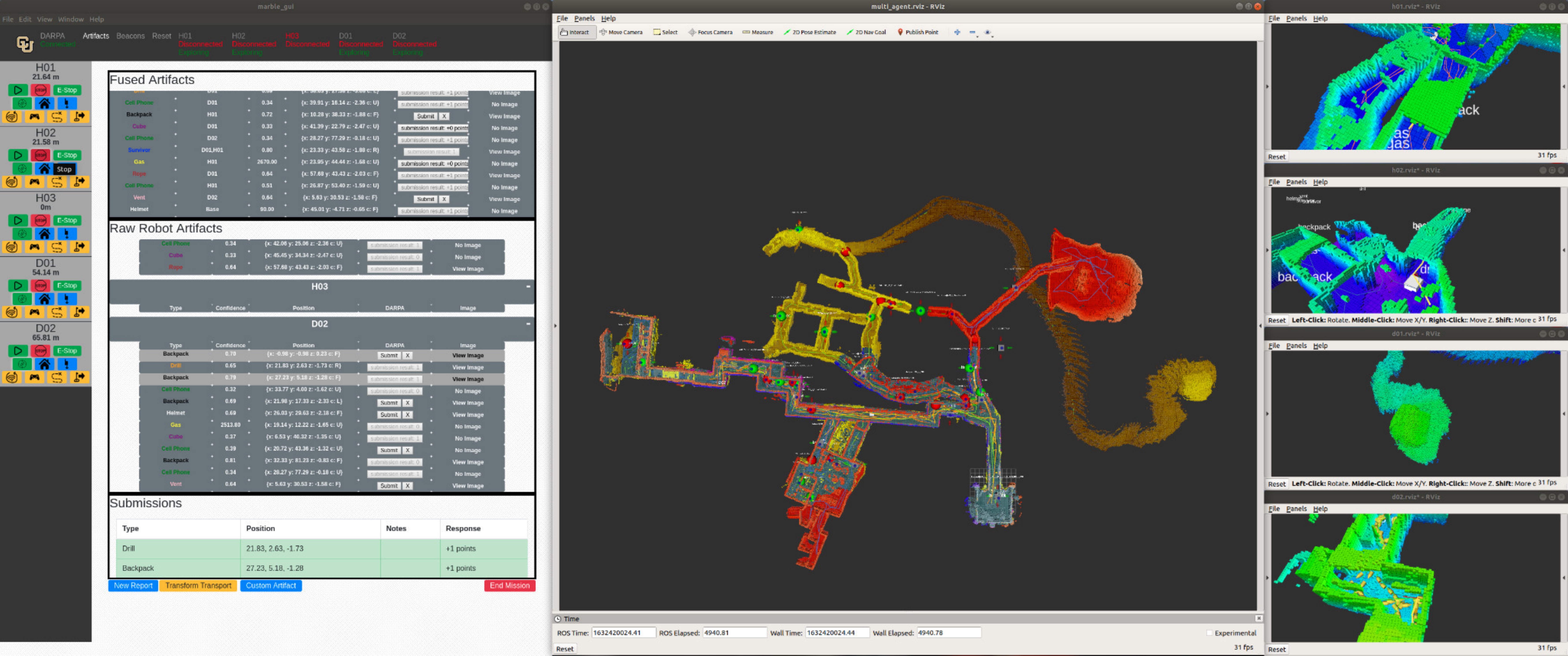}
    \caption{Example of the human supervisor interface showing the end of Team MARBLE's final run}
    \label{fig:hs_interface}
\end{figure}

The human supervisor interacts with BOBCAT using a custom GUI shown in Figure \ref{fig:hs_interface}. This interface allows the human supervisor to set a goal point for the robot using an Interactive Marker. MADCAT then passes this goal to the robot through the communications network if a connection is available. The human supervisor can also remotely control robots using an Xbox controller when communication systems allow.

First-person view (FPV) allows the human supervisor to see the environment from the robot's perspective in semi-real-time, instead of just through the map representation and infrequent artifact images.  In addition to increased situational awareness it allows the human supervisor to identify visual artifacts that may be missed by the on-board artifact detection system, or identify them more rapidly.  Finally, FPV helps the human supervisor during manual teleoperation in the event it is needed to direct the robot.

Flexibility of our \textit{udp\_mesh} communication architecture made it possible to rapidly adapt to mission specific constraints. During the Final Event, we observed extra bandwidth in the communication system and decided to add FPV to our Spot robots to enhance their exploration potential. Compressed images from the Spot forward facing cameras were transmitted as 1 Hz low-priority messages. On the base station computer, these images were displayed live in RViz, and saved locally so the human supervisor can review that at any time. The additional scoring potential of FPV is highlighted in Section \ref{ssec:results_artifact_detection}.

Several features of the human supervisor GUI were designed to help reduce operator workload. First, another artifact fusion process runs on the Base Station computer, to aid the operator in tracking artifacts reported by multiple vehicles.  If reports of the same type are within 3m of prior reports, they are fused to the mean position. Redundant artifact reports already been seen by another robot appear in a light gray color. In contrast, new reports flash with large white text to bring attention to the human supervisor. When submitting artifact reports, the human supervisor can select from individual or fused reports. If an artifact is successfully scored, the submission is locked out to prevent re-submitting. If it does not result in a score, the operator can utilize additional map, trajectory, and FPV information to improve the estimated position. New reports can be submitted by shifting fused artifacts icons in the map or specifying a manual position.

\section{Final Event Results}\label{sec:final_event}

The SubT Final Event Prize Run on September 23, 2021 provided Team MARBLE an excellent opportunity to evaluate the performance of our complete supervised autonomy solution, and we share our results in this section. First, an overview is provided in Section \ref{ssec:results_overview}, which includes an outline of the mission objectives, a description of the previously unknown course in Section, as well as a high-level summary of the results. Localization and mapping results are presented in Section \ref{ssec:results_localization_and_mapping}. Further analysis of the planner and resulting exploration effort is described in Section \ref{ssec:results_planning}. Artifact detection results are thoroughly analyzed in Section \ref{ssec:results_artifact_detection}. The communication environment was friendlier than expected, and in Section \ref{ssec:results_comms}, we discuss how we capitalized on that opportunity. Mission management results are detailed in Section \ref{ssec:results_mission_management}, including the five instances where the human supervisor manual intervened, as well as the seven artifacts that were scored via FPV imagery. Together, this section elucidates how our systems worked together to score 18 artifacts, while also discussing the areas that limited even higher performance. Data from Team MARBLE's deployment during the Final Event Prize Run is publicly available and discussed in Section \ref{ssec:results_opensource_data}.

\subsection{Overview}
\label{ssec:results_overview}

The mission objectives were to accurately report as many of the 40 artifacts in the course as possible during the mission. There are three hard constraints: the mission is 60 minutes long, there are a total of 40 attempts to report artifacts, and only one human, the human supervisor, is permitted to supervise the mission, manipulate robotic agents, and submit artifact reports.

\begin{figure*}[hbt!]
    \centering
    \includegraphics[width=0.95\textwidth]{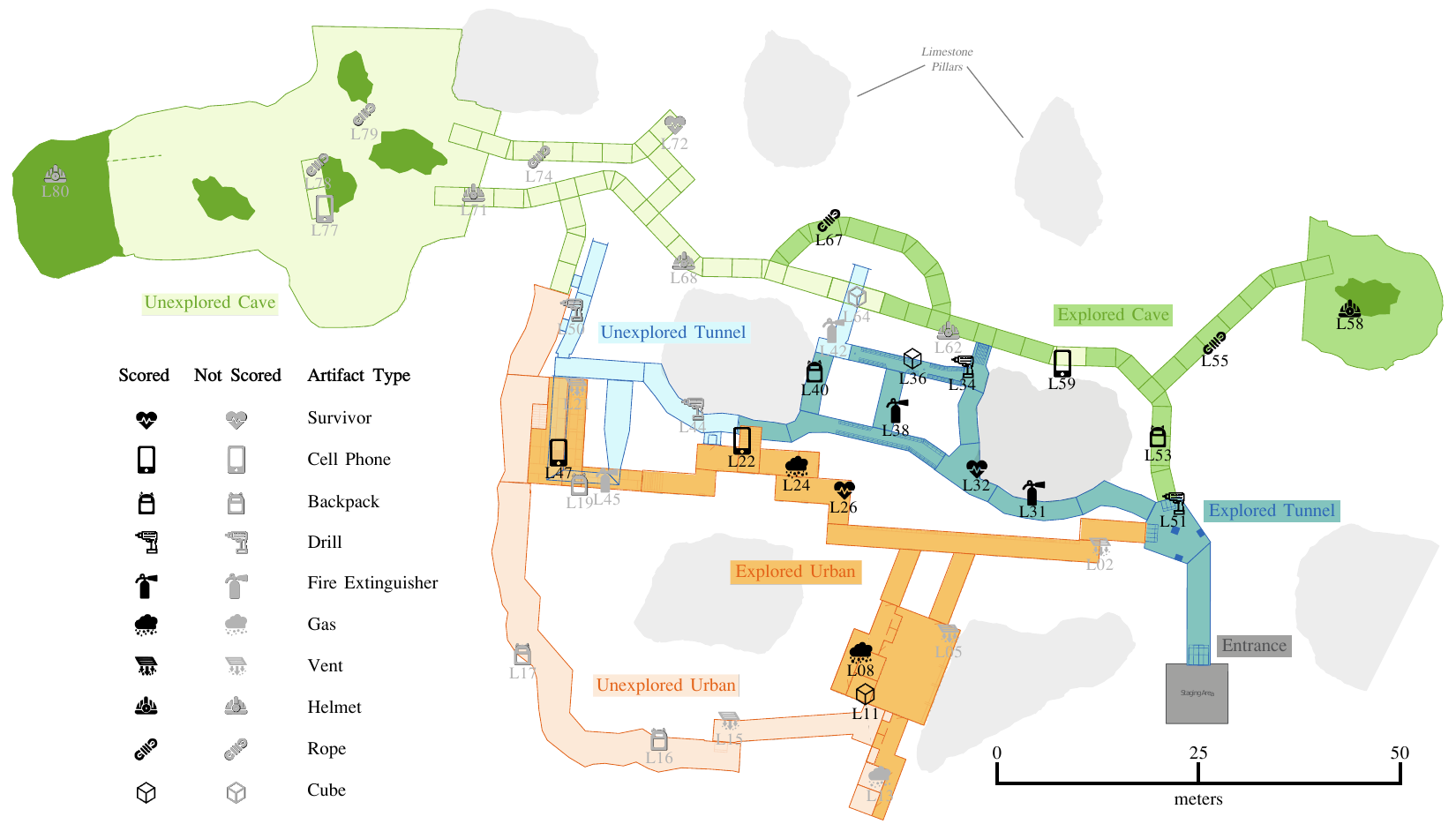}
    \caption{Course map of the 60-minute Final Event Prize Run designed by DARPA, highlighting the 18 of 40 artifacts scored by Team MARBLE, along with the areas of the tunnel, urban, and cave sections were explored by agents.}
    \label{fig:darpa_course_map_marble_results_overview}
\end{figure*}

The final course was custom constructed as illustrated in Figure \ref{fig:darpa_course_map_marble_results_overview}, which breaks the course out into distinct tunnel, urban, and cave environments. The course contains numerous hazards and challenges: rough terrain, railroad tracks, slippery surfaces, ramps, stairs, large drop-offs, rocky cliffs, narrow hallways, low-to-the-ground corridors, wide-open caverns, fog, standing water, dynamic obstacles, trap doors, and a degraded communications environment. These challenges are discussed further in Section \ref{ssec:results_planning}.

Here, we provide a brief high-level summary of the artifacts scored and the extent of the environment explored. Team MARBLE scored 18 of the 40 artifacts and explored roughly half of the environment. For reference, the top-scoring team scored 23 artifacts, and the performance for all teams is listed in Section \ref{ssec:sup_competition_results} of the Appendix. The location and class of all 40 artifacts can be visualized in the context of the course map shown in Figure \ref{fig:darpa_course_map_marble_results_overview}. This map also indicates which regions of the course that Team MARBLE explored as well as the 18 scored artifacts. The artifacts that Team MARBLE scored are also listed in Table \ref{tab:scored_points}, ordered chronologically from mission start to mission end. Each of the 18 artifacts in Table \ref{tab:scored_points} correspond by ID to the scored artifacts in Figure \ref{fig:darpa_course_map_marble_results_overview}. Further analysis of artifact detection results are detailed in Section \ref{ssec:results_artifact_detection}.

\begin{table}[hbt!]
\begin{center}
    \begin{tabular}{c c c c c c}
    \hline
    \textbf{Score} & \textbf{Time [mm:ss]} & \textbf{Type} & \textbf{ID} & \textbf{Error [m]}\\ 
    \hline
    1   &   01:08   &   Drill               &   L51 &   0.57 \\
    2   &   01:23   &   Backpack            &   L53 &   2.23 \\
    3   &   06:23   &   Rope                &   L55 &   0.84 \\
    4   &   12:03   &   Survivor            &   L26 &   0.62 \\
    5   &   16:35   &   Survivor            &   L32 &   1.40 \\
    6   &   17:23   &   Gas                 &   L08 &   1.80 \\
    7   &   28:08   &   Fire Extinguisher   &   L31 &   1.31 \\
    8   &   35:51   &   Drill               &   L34 &   1.43 \\
    9   &   36:53   &   Fire Extinguisher   &   L38 &   2.82 \\
    10  &   37:08   &   Cube                &   L36 &   3.94 \\
    11  &   37:58   &   Backpack            &   L40 &   1.40 \\
    12  &   38:47   &   Rope                &   L67 &   2.87 \\
    13  &   47:53   &   Cube                &   L11 &   1.83 \\
    14  &   50:33   &   Cell Phone          &   L22 &   4.06 \\
    15  &   50:45   &   Cell Phone          &   L47 &   4.00 \\
    16  &   51:48   &   Cell Phone          &   L59 &   2.15 \\
    17  &   52:45   &   Gas                 &   L24 &   2.55 \\
    18  &   56:33   &   Helmet              &   L58 &   1.74 \\
    \hline
    \end{tabular}
    \caption{\label{tab:scored_points} List of all artifacts scored by Team MARBLE during the 60-minute Final Event Prize Run, along with the corresponding mission time when artifacts were reported and scored, artifact type, unique DARPA-assigned ID numbers, and Euclidean distance error between the reported and ground truth location of the artifact.}
\end{center}
\end{table}

\subsection{Localization \& Mapping}
\label{ssec:results_localization_and_mapping}

\begin{figure*}[hbt!]
    \centering
    \includegraphics[width=0.95\textwidth]{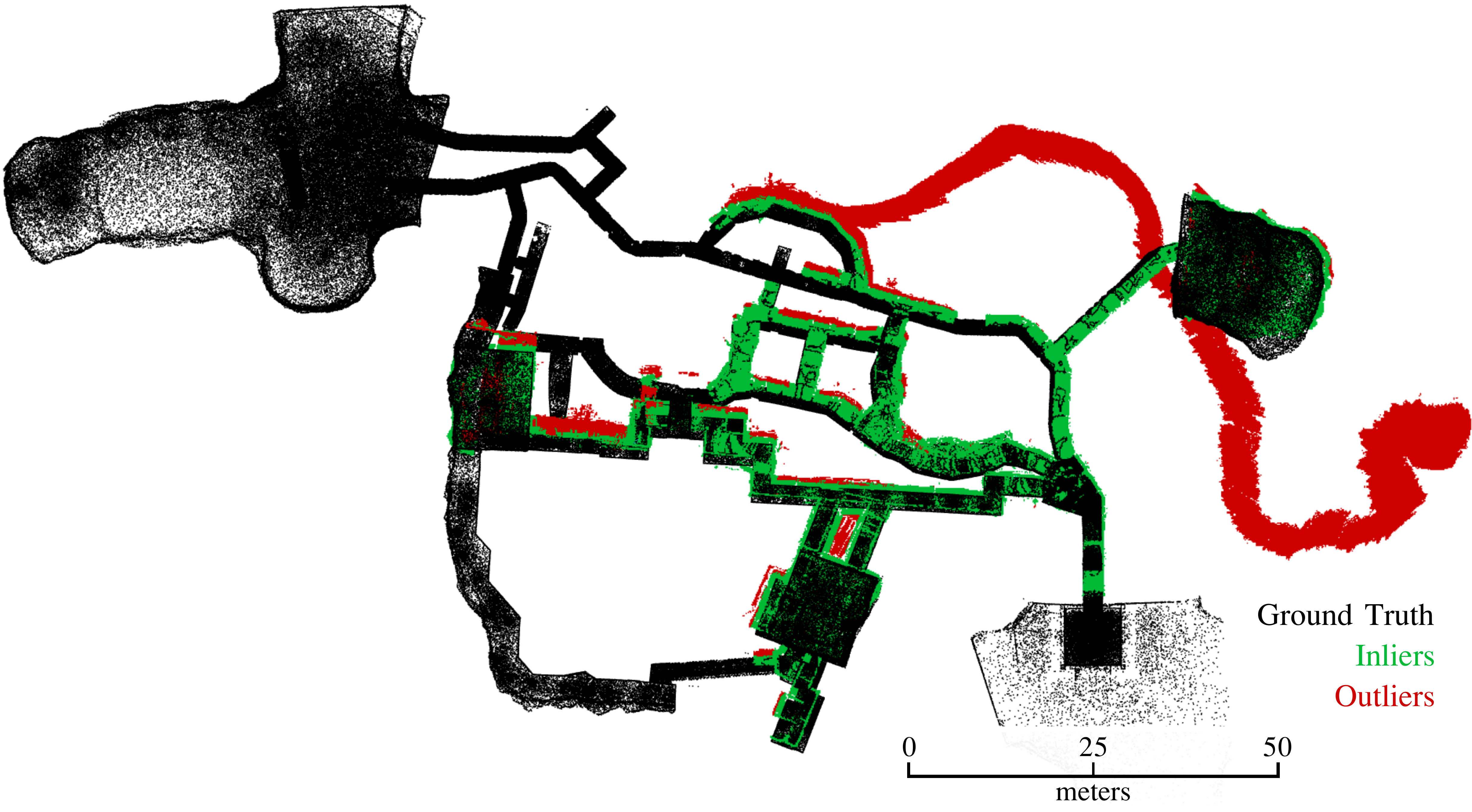}
    \caption{Final point cloud map submitted by Team MARBLE to DARPA staff during the Final Event Prize Run.}
    \label{fig:darpa_final_pcl_map}
\end{figure*}

A secondary objective of the 60-minute Final Event Prize Run was to rapidly map the environment and transmit the real-time map back to DARPA every ten seconds. The map takes the form of a point cloud, or a collection of three-dimensional points that represent occupied space in the environment, e.g. floors, walls, ceilings. Figure \ref{fig:darpa_final_pcl_map} provides a comparison of the DARPA-generated ground truth map in black against the final map submitted by Team MARBLE, split into inliers in green and outliers in red.

DARPA has generously collected, processed, and shared this map data with participating teams \cite{darpa_map_analysis_2021}. Inliers are defined as points within 1m of ground truth map points, and outliers are defined as points outside 1m. Map coverage is a metric representing the ratio of the environment explored, and is defined as

\begin{equation} \label{eq:coverage}
    \textrm{map coverage} = \frac{\textrm{ground truth points within 1m of an inlier point}}{\textrm{total ground truth points}}.
\end{equation} 

Map error or deviation is a metric representing the ratio of the submitted map that is inaccurate relative to the ground truth map, and is defined as

\begin{equation} \label{eq:deviation}
    \textrm{map deviation} = \frac{\textrm{outlier points}}{\textrm{total submitted points}}.
\end{equation}

\begin{figure*}[hbt!]
    \centering
    \includegraphics[width=0.85\textwidth]{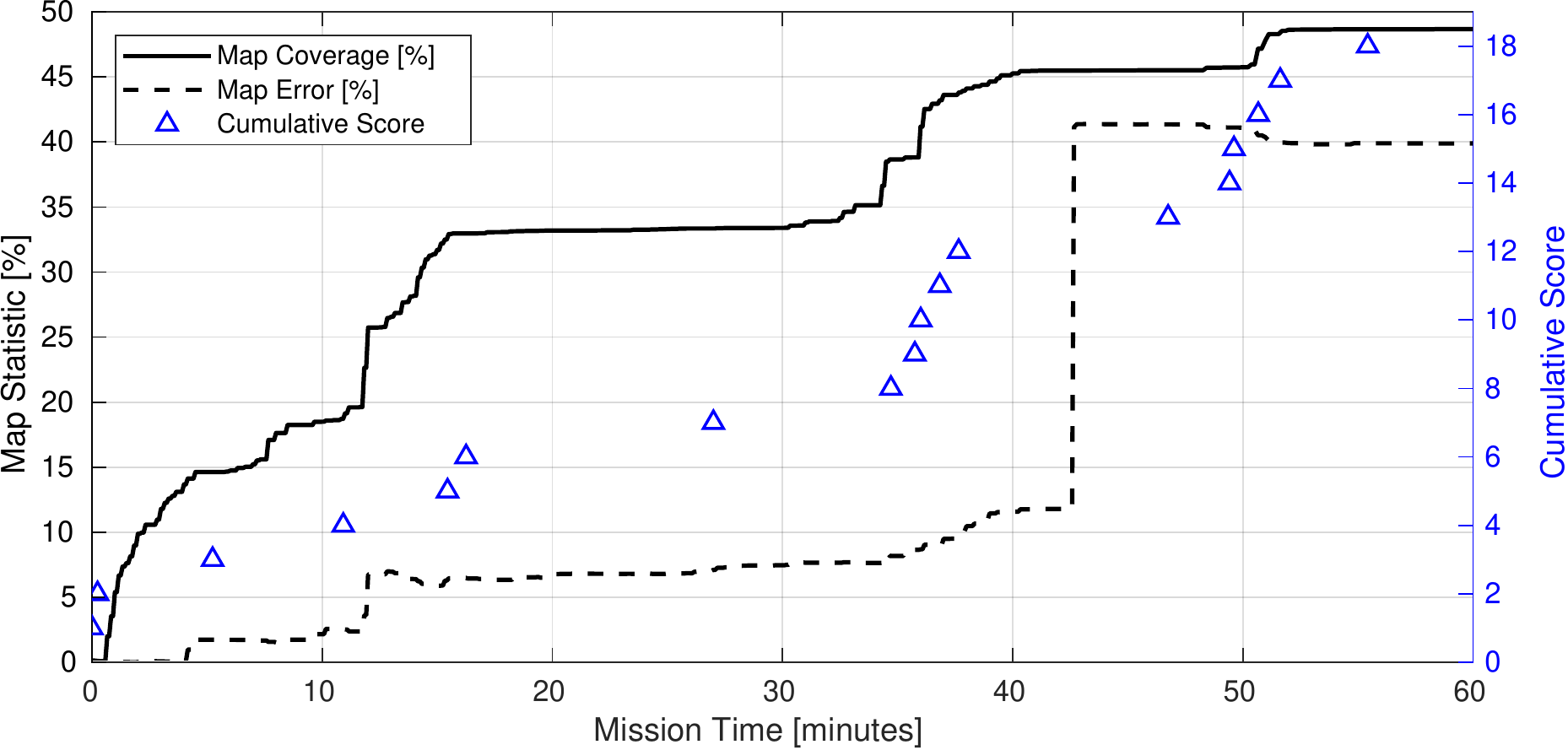}
    \caption{Map coverage, error, and cumulative score throughout Team MARBLE's deployment during the DARPA SubT Final Event Prize Run.}
    \label{fig:map_statistic_score_figure}
\end{figure*}

Figure \ref{fig:map_statistic_score_figure} shows that map coverage steadily increases throughout the mission, with some periods of rapid exploration, and by the end of the mission, nearly 50\% of the environment has been mapped. Map error on the other hand, increases modestly throughout the mission due to localization drift. However, it increases significantly due to a localization failure on D01, which is discussed further in Section \ref{sssec:results_planning_limitations}. This failure generated erroneous sections of map which are mistakenly appear as a long winding corridor in Figure \ref{fig:darpa_final_pcl_map}.

\begin{figure}[hbt!]
    \centering
    \subfloat[]{{\includegraphics[width=.45\textwidth]{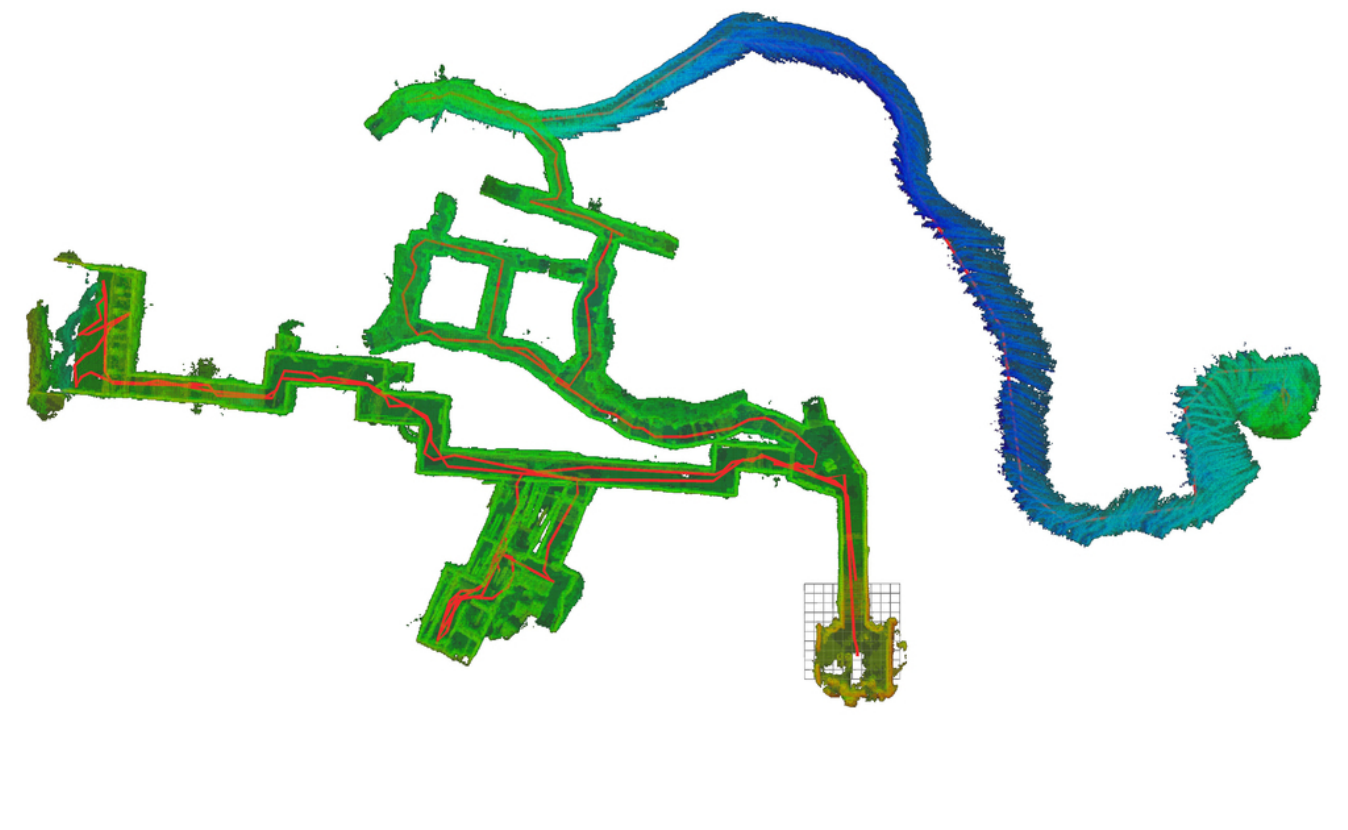}\label{fig:D01_final_map} }} \hspace{2.0pt}
    \subfloat[]{{\includegraphics[width=.45\textwidth]{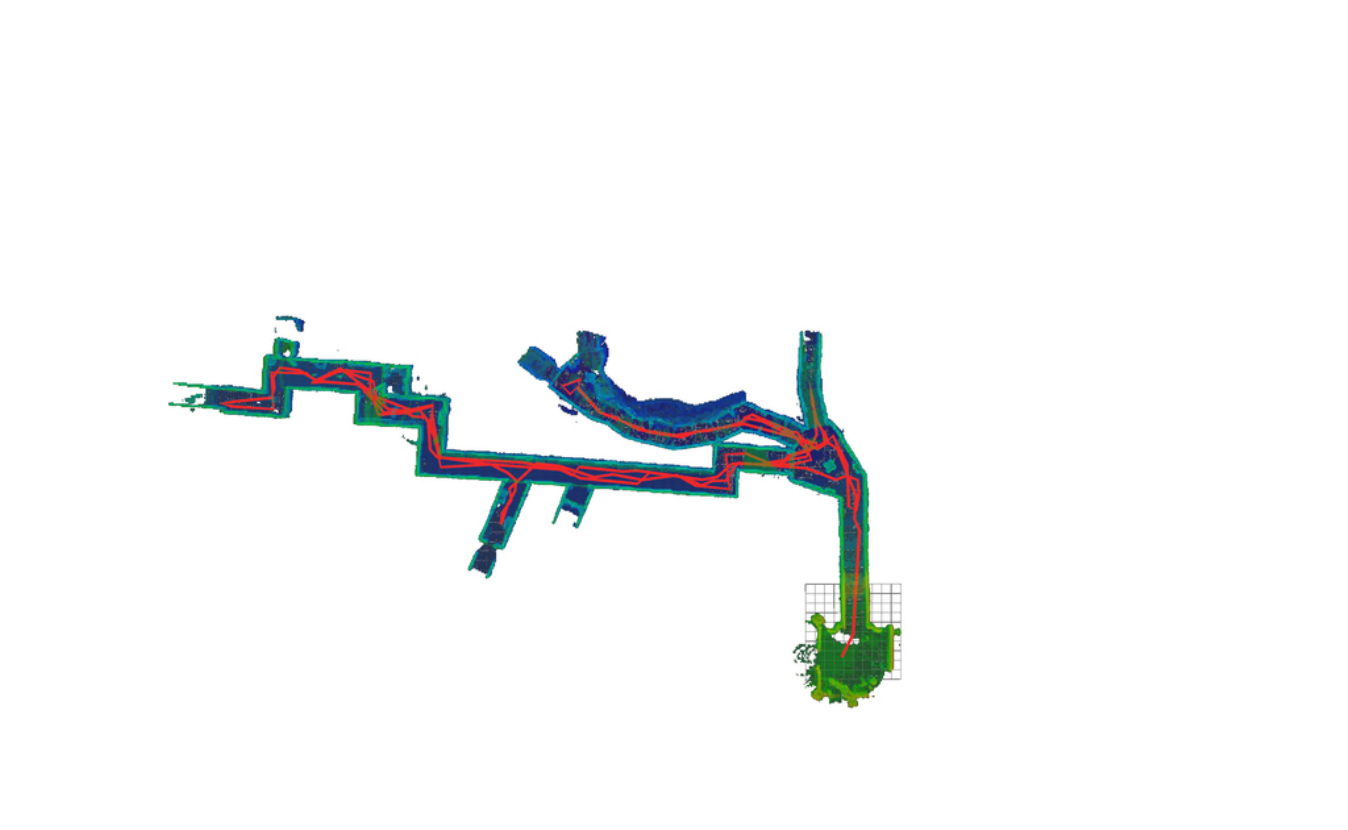}\label{fig:H01_final_map} }} \\
    \subfloat[]{{\includegraphics[width=.45\textwidth]{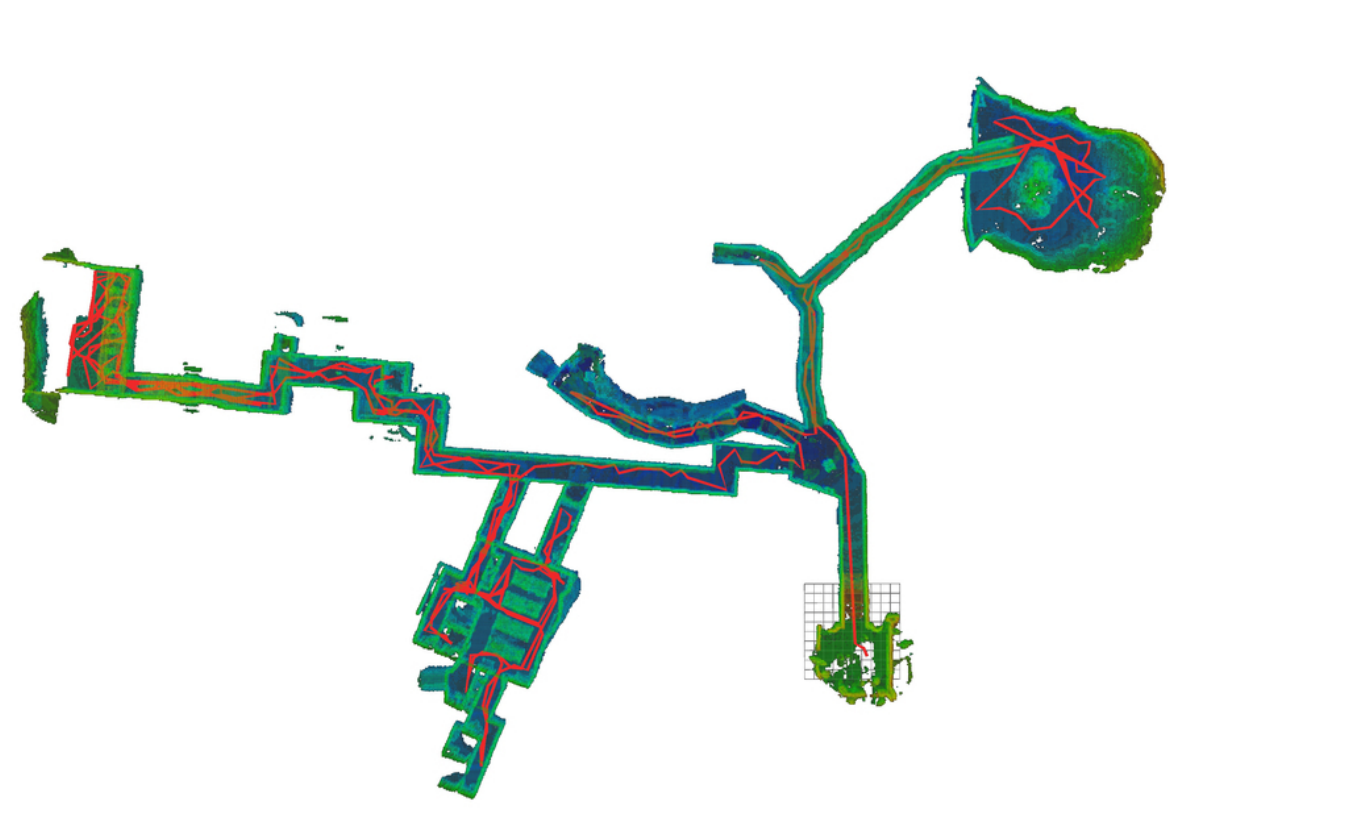}\label{fig:D02_final_map} }} \hspace{2.0pt}
    \subfloat[]{{\includegraphics[width=.45\textwidth]{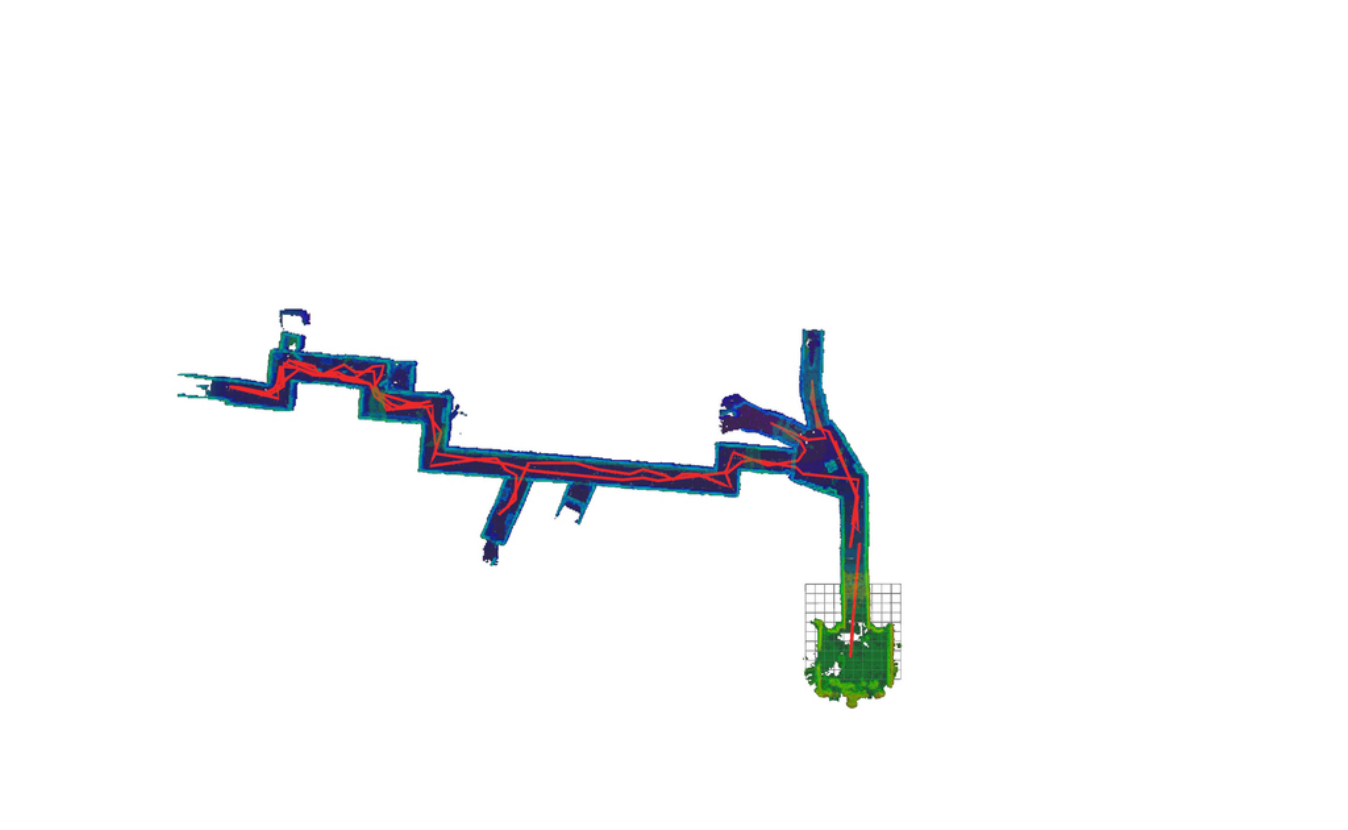}\label{fig:H02_final_map} }}\\
    \caption{The final maps from (a) D01, (b) H01, (c) D02, and (d) H02.}
    \label{fig:final_maps_event}
\end{figure}

\subsection{Planning}
\label{ssec:results_planning}

During the entire 60-minute mission and across a diversity of environments, none of the robots were teleoperated due to a planner failure or to improve the volumetric gain, which is seen as a successful demonstration of the planning reliability and flexibility. The human supervisor only intervened to complement autonomy with human-level cognition and intelligence, as discussed further in Section \ref{ssec:results_mission_management}. 

As an overview, Figure \ref{f:subt_finals_pose_graph_coordination_scan_plan} presents three scenarios from the Final Event Prize Run, which demonstrate negative obstacle avoidance, multi-agent coordination based on teammate position histories, and teleoperation initiation by the human supervisor. Below, Section \ref{sssec:results_planning_exploration} highlights the exploration performance, Section \ref{sssec:results_planning_treacherousterrain} shares examples of agents avoiding treacherous terrain, and Section \ref{sssec:results_planning_dynamic_environment} demonstrates the planner adapting to dynamic changes in the environment. Section \ref{sssec:results_planning_limitations} discusses system limitations that prevented the agents from exploring the entire course.

\centerfig{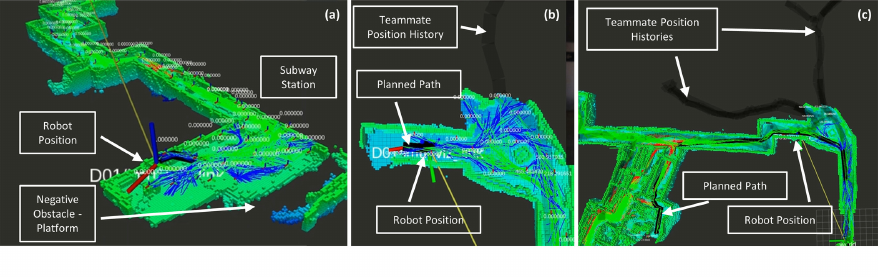}{subt_finals_pose_graph_coordination_scan_plan}
{ The figure shows three snapshots from the SubT Final Event Prize Round run. The blue lines in the figure represents the locally sampled RRT$^*$ tree and the green lines represent the global graph. (a) The robot can be seen to avoid sampling over the negative obstacles i.e., the edge of the subway platform, owing to the settling-based collision-checks. (b) The agent planned away from the position history of a teammate robot that was launched before it, demonstrating effective multi-agent coordination. (c)  This snapshot shows an instance where teleoperation was initialized on one of the robots. The robot can be seen to plan a path that had sufficient volumetric gain leading the robot toward a frontier in the urban area that had not been seen by any other robots. However, the human supervisor decided to teleoperate the robot through the initial section of the tunnel area and then let it autonomously explore the tunnel.}

\subsubsection{Exploration}
\label{sssec:results_planning_exploration}

The volume of unseen area explored by each agent across the mission is illustrated in \fig{subt_finals_vol_gain_scan_plan}. Unlike the statistics of the global map submitted to DARPA in Figure \ref{fig:darpa_final_pcl_map}, \fig{subt_finals_vol_gain_scan_plan} presents agent-specific exploration information stored onboard each robot. D02 was the largest contributor to overall exploration, partly because it was launched first and had the most time to explore. D01 played a complementary role by exploring most of the tunnel environment, which remained unexplored by D02. Because the subset of the environment safely traversable by the wheel robots was relatively more constrained, it naturally led to less exploration from the Huskies across the mission. 

Besides showing the volume explored by each of the robots, \fig{subt_finals_vol_gain_scan_plan} also shows time periods during which the default planner exploration behavior was paused for higher-level mission management and teleoperation commands. These interjections by the autonomous mission management system were primarily triggered when agents approached each other, and resulted in lower-priority agents pausing and higher-priority agents resuming their task. During these encounters, agents appear as fast-approaching dynamic objects, and this simple procedure was employed rather than incorporating reactive obstacle avoidance into the planner.

\centerfig{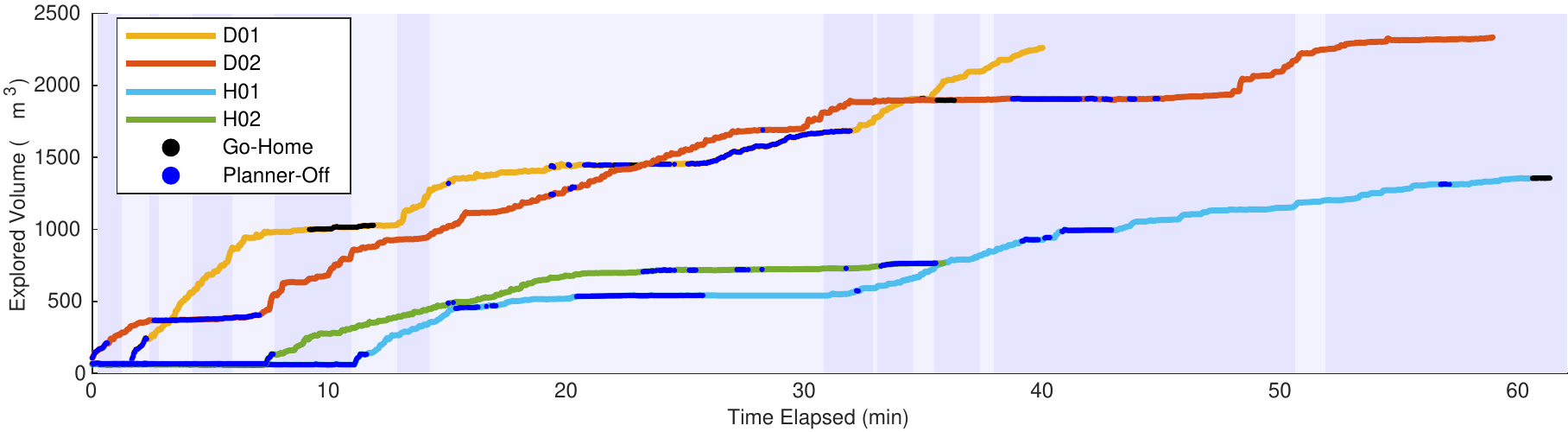}{subt_finals_vol_gain_scan_plan}{Volumetric gain explored by each robot across the 60-minute Final Event Prize Run. Planner-Off, denoted by blue lines, represents instances when the planner was paused to allow the autonomous mission-management system to take over when robots are in close proximity, as well as the several interventions when the human supervisor manually teleoperated robots. Go-Home, denoted by black lines, represents instances when the planner began returning home to reconnect to the network and report new artifact reports and map data.}

\subsubsection{Treacherous Terrain}
\label{sssec:results_planning_treacherousterrain}

The path planning solution successfully kept each agent safe from collision throughout the entire mission. The Spot robots, which explored more challenging features in the environment, fully avoided negative obstacles, such as shear drop-offs and rocky slopes, and traversed up and down stairs. 

\begin{figure}[hbt!]
		\centering
		\subfloat[]{{\includegraphics[width=.4\textwidth]{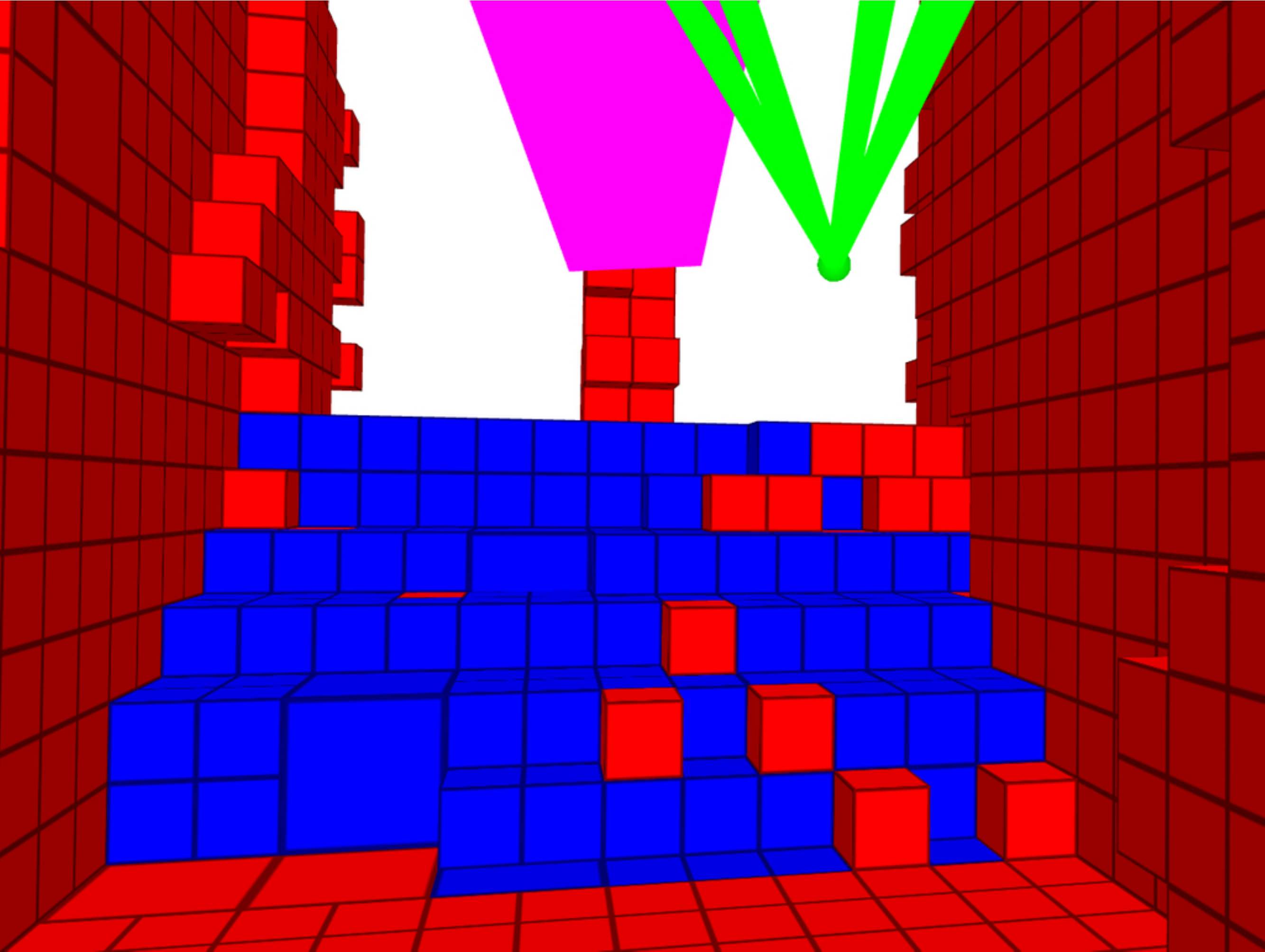}\label{fig:D01_0212s_stairsmap} }}
		\subfloat[]{{\includegraphics[width=.40\textwidth]{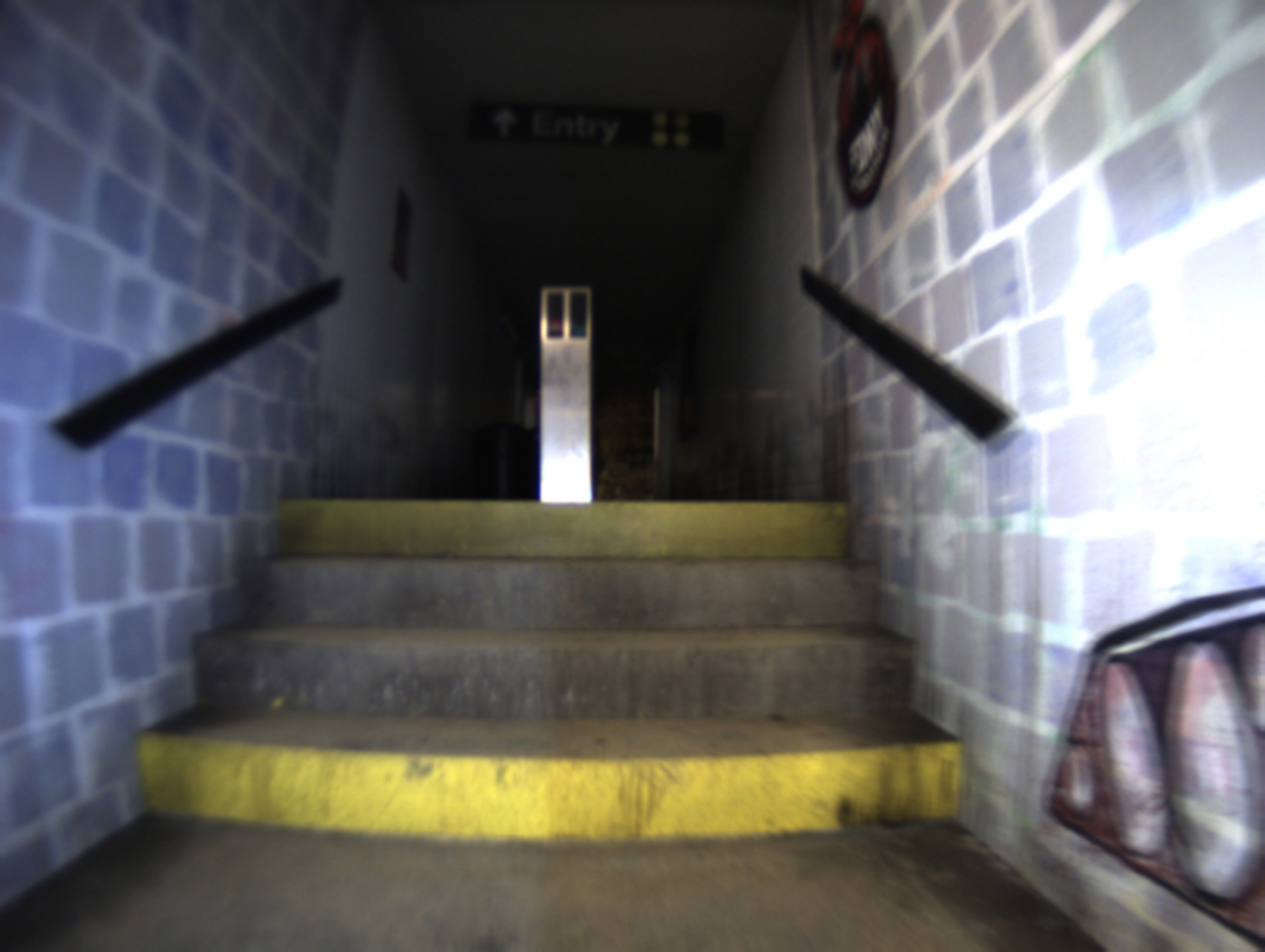}\label{fig:D01_0209s_stairs} }}
		\caption{Instance of (a) D01 planning up stairs with the green edges of the graph, pink planned path, and associated semantic map with blue voxels representing stairs, along with (b) FPV imagery.}
		\label{fig:planning_stairs}
\end{figure}

\begin{figure}[hbt!]
		\centering
		\subfloat[]{{\includegraphics[width=.40\textwidth]{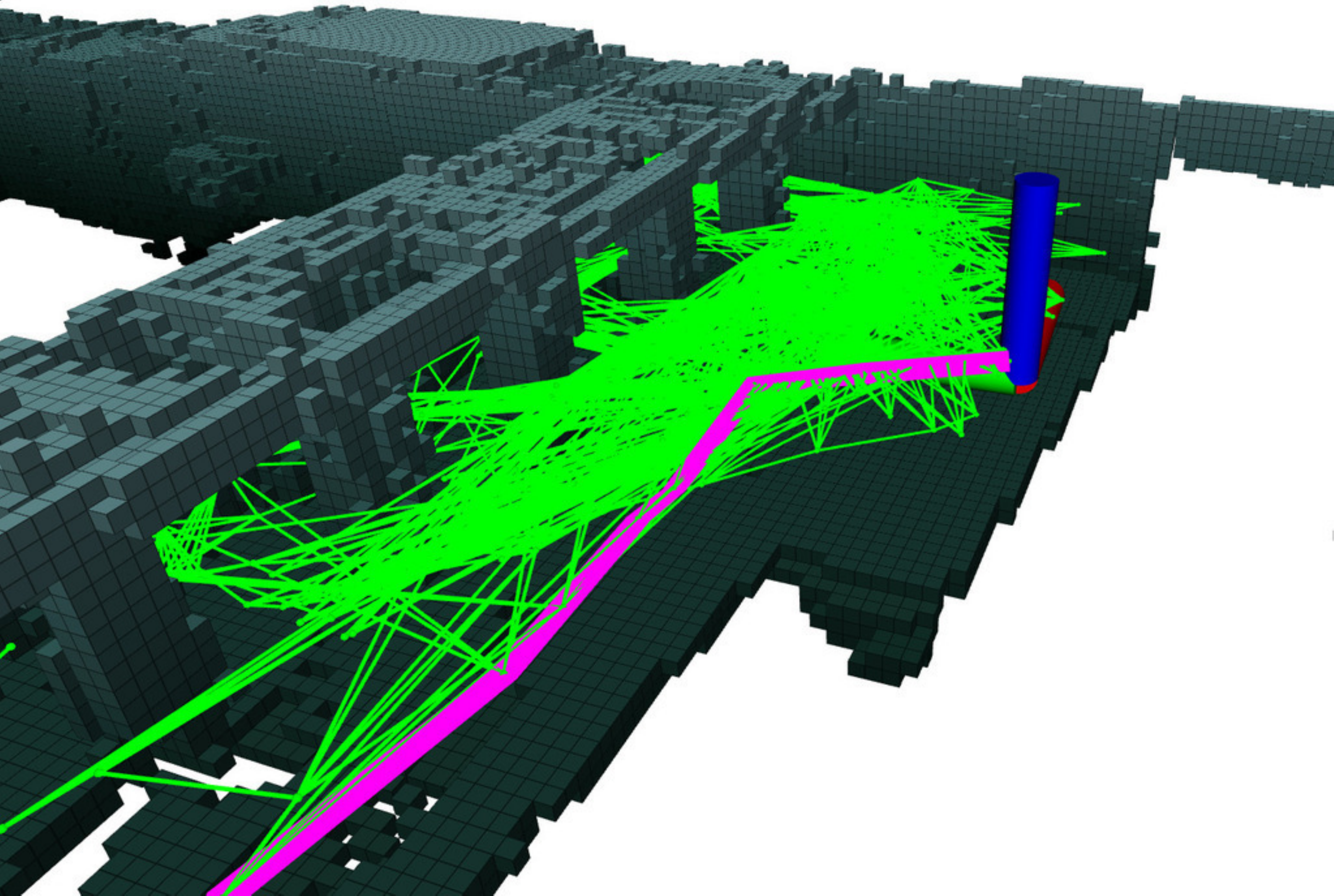}\label{fig:D02_1884s_subwaymap} }}
		\subfloat[]{{\includegraphics[width=.40\textwidth]{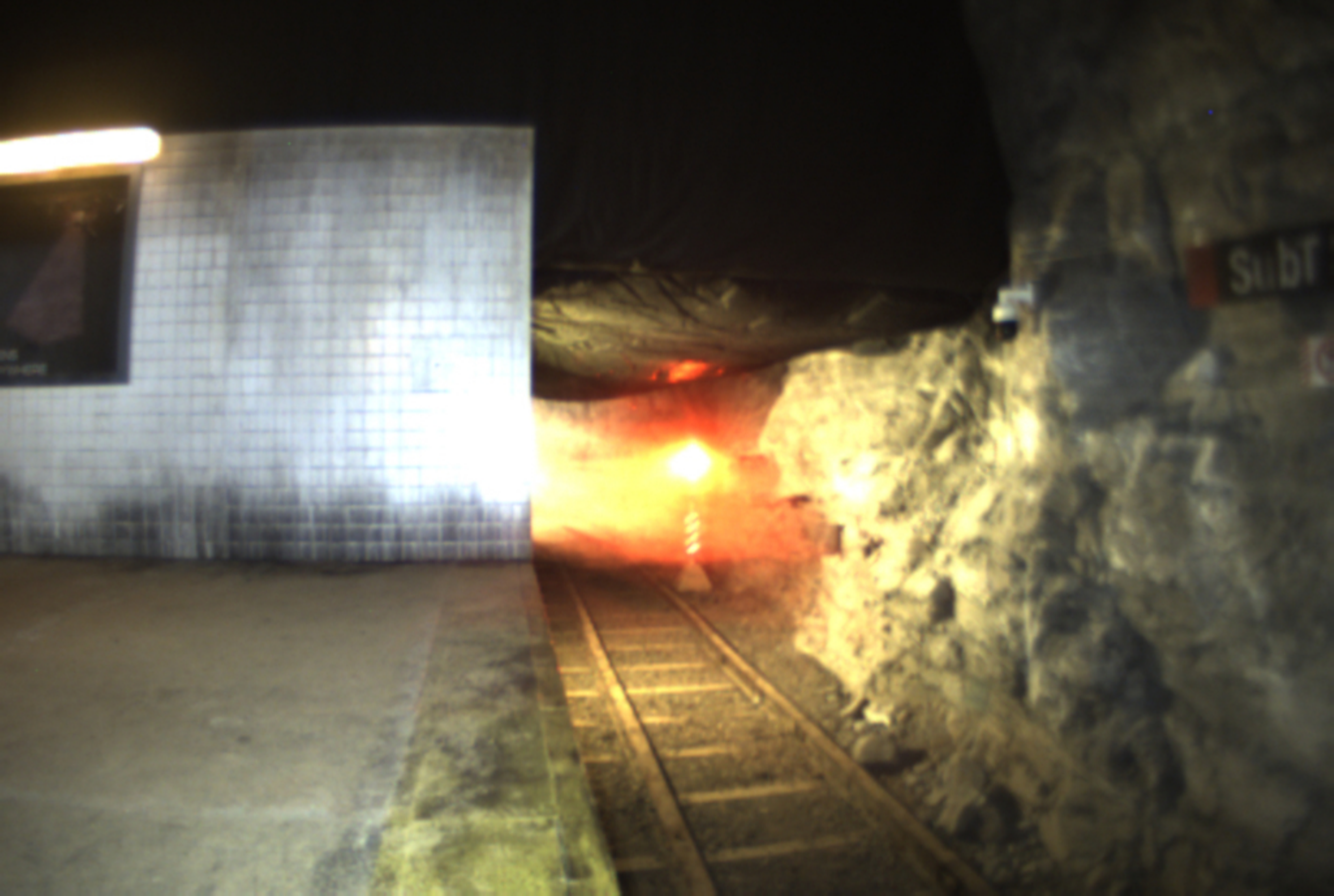}\label{fig:D02_1884s_subway} }}
		\caption{Instance of (a) D02 thoroughly exploring the subway platform without planning over the edge, with green edges of the graph, pink planned path, and associated map, along with (b) FPV imagery.}
		\label{fig:planning_subway}
\end{figure}

The Spot robots successfully traversed up and down the small set of stairs leading up to the subway platform, as shown in Figure \ref{fig:planning_stairs}. However, when stairs where first encountered from above, agents did not plan down them due to the limited ${\pm}16.5^{\circ}$ vertical field of view of onboard Ouster OS-1-64 lidar sensors, as shown in Figure \ref{fig:D02_1884s_subwaymap}. In addition, the Spots can only safely walk down stairs backwards, and therefore additional logic would be be required to autonomously traverse those stairs.

\begin{figure}[hbt!]
		\centering
		\subfloat[]{{\includegraphics[width=.40\textwidth]{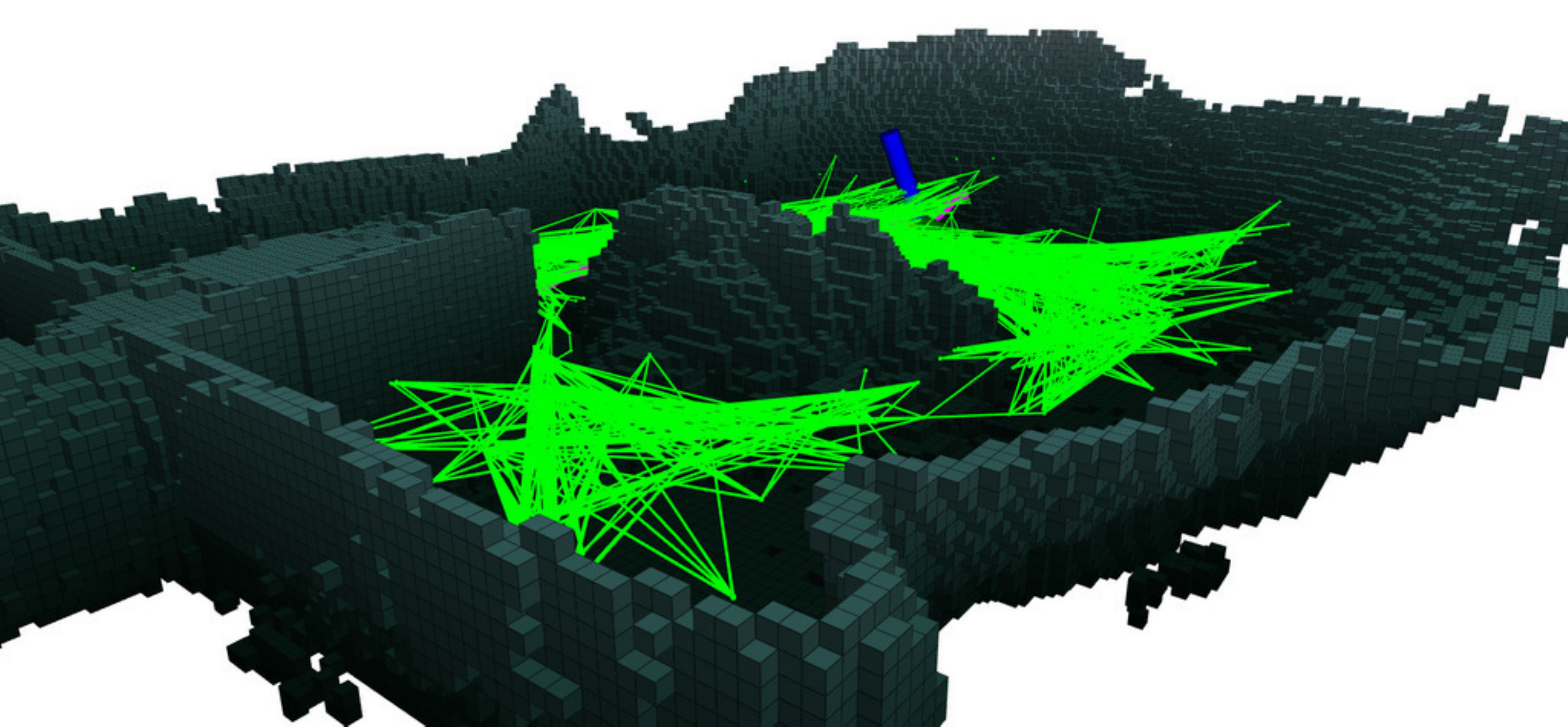}\label{fig:D02_0731s_cavern} }}
		\subfloat[]{{\includegraphics[width=.40\textwidth]{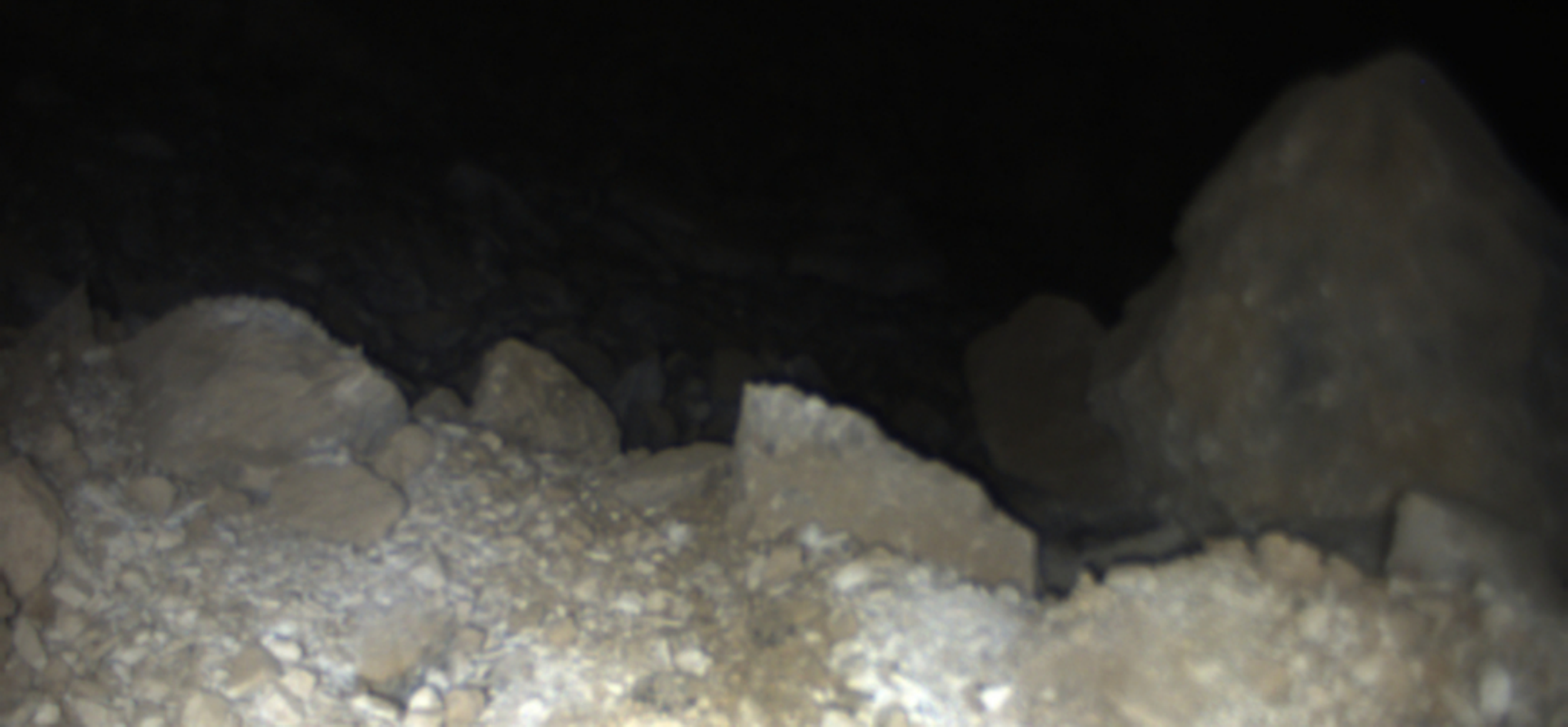}\label{fig:D02_0533s_rubble} }}
		\caption{Instance of (a) D02 autonomously exploring the entire cavern, traversing the safer surfaces and avoiding treacherous areas, such as the one shown in (b).}
		\label{fig:mobility_cavern}
\end{figure}

Both Spot robots thoroughly explored the subway platform, approaching the edge, but never stepping and falling over, as demonstrated by D02 in Figure \ref{fig:planning_subway}. Additionally, the same Spot robot, D02, explored the entire cavern autonomously without entering treacherous terrain, as shown in Figure \ref{fig:mobility_cavern}.

\subsubsection{Dynamic Environment}
\label{sssec:results_planning_dynamic_environment}

The planner has the ability to adapt to dynamic environments, such as closing or opening of doors, falling rubble, as well as other situations also lead to dynamic changes in the map, including other nearby mobile agents, and localization and mapping error.

\begin{figure}[hbt!]
		\centering
		\subfloat[]{{\includegraphics[width=.24\textwidth]{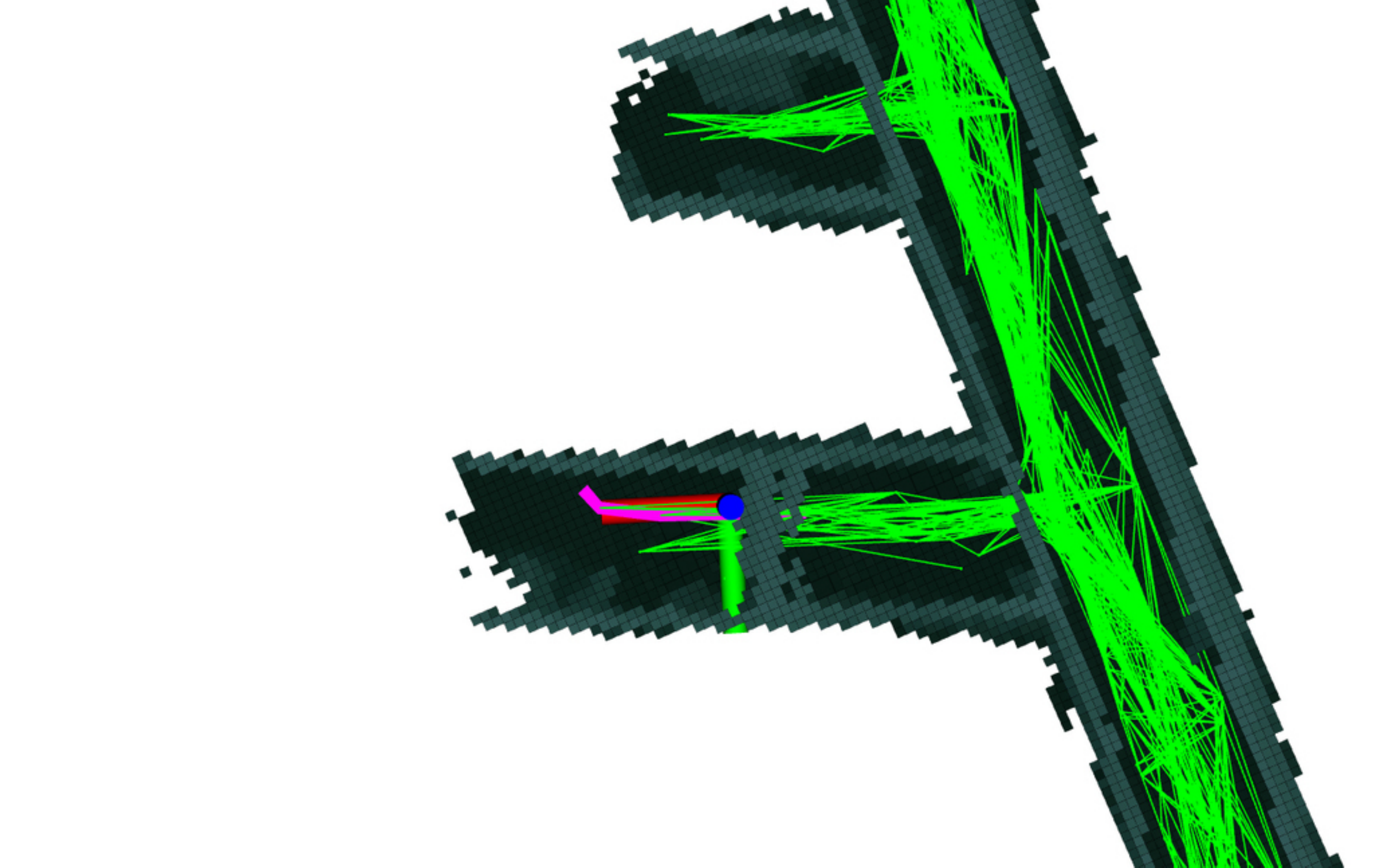}\label{fig:D01_0107s_beforetrapdoor} }}
		\subfloat[]{{\includegraphics[width=.24\textwidth]{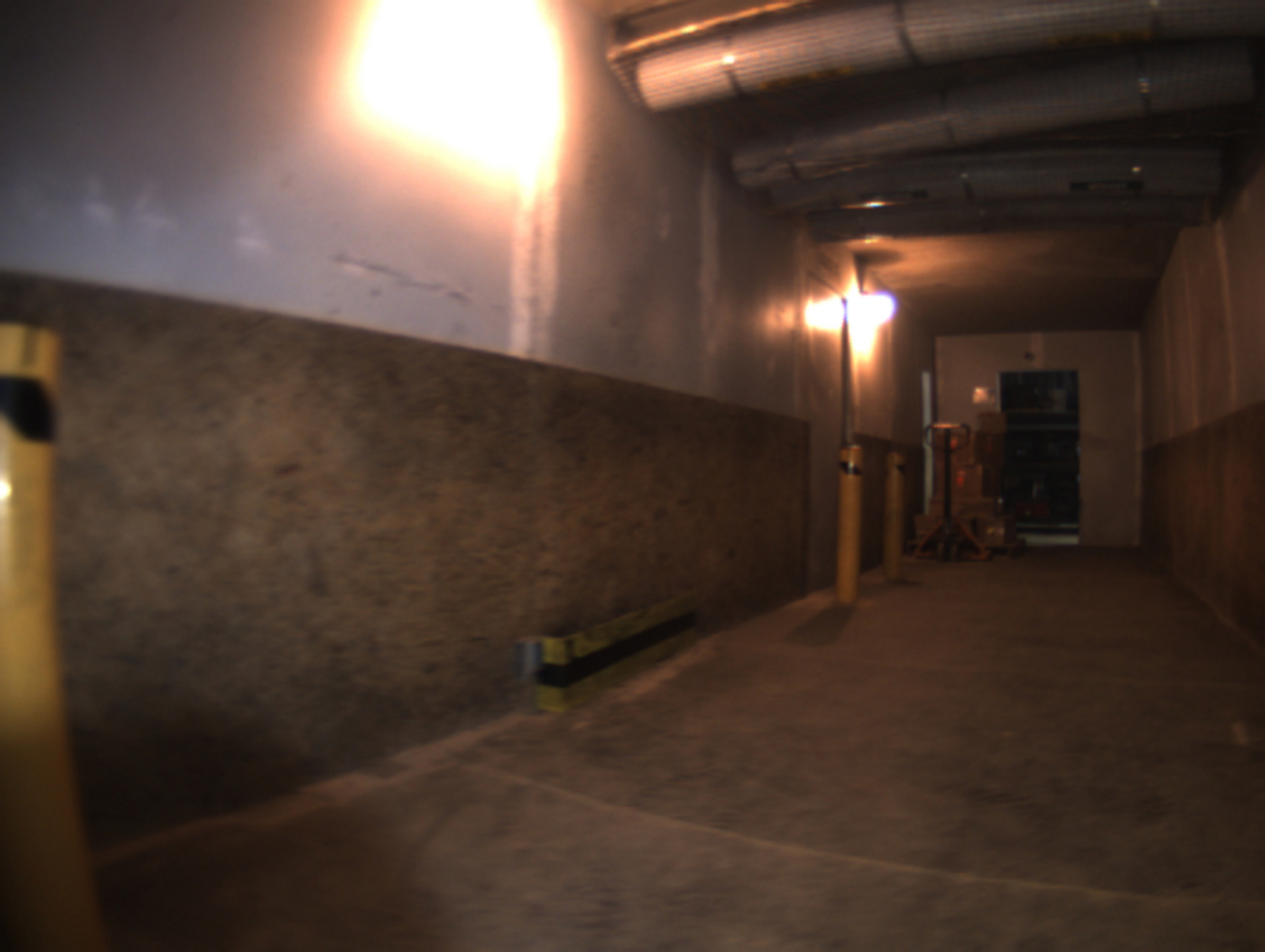}\label{fig:D01_0634s_beforetrapdoormap} }} 
		\subfloat[]{{\includegraphics[width=.24\textwidth]{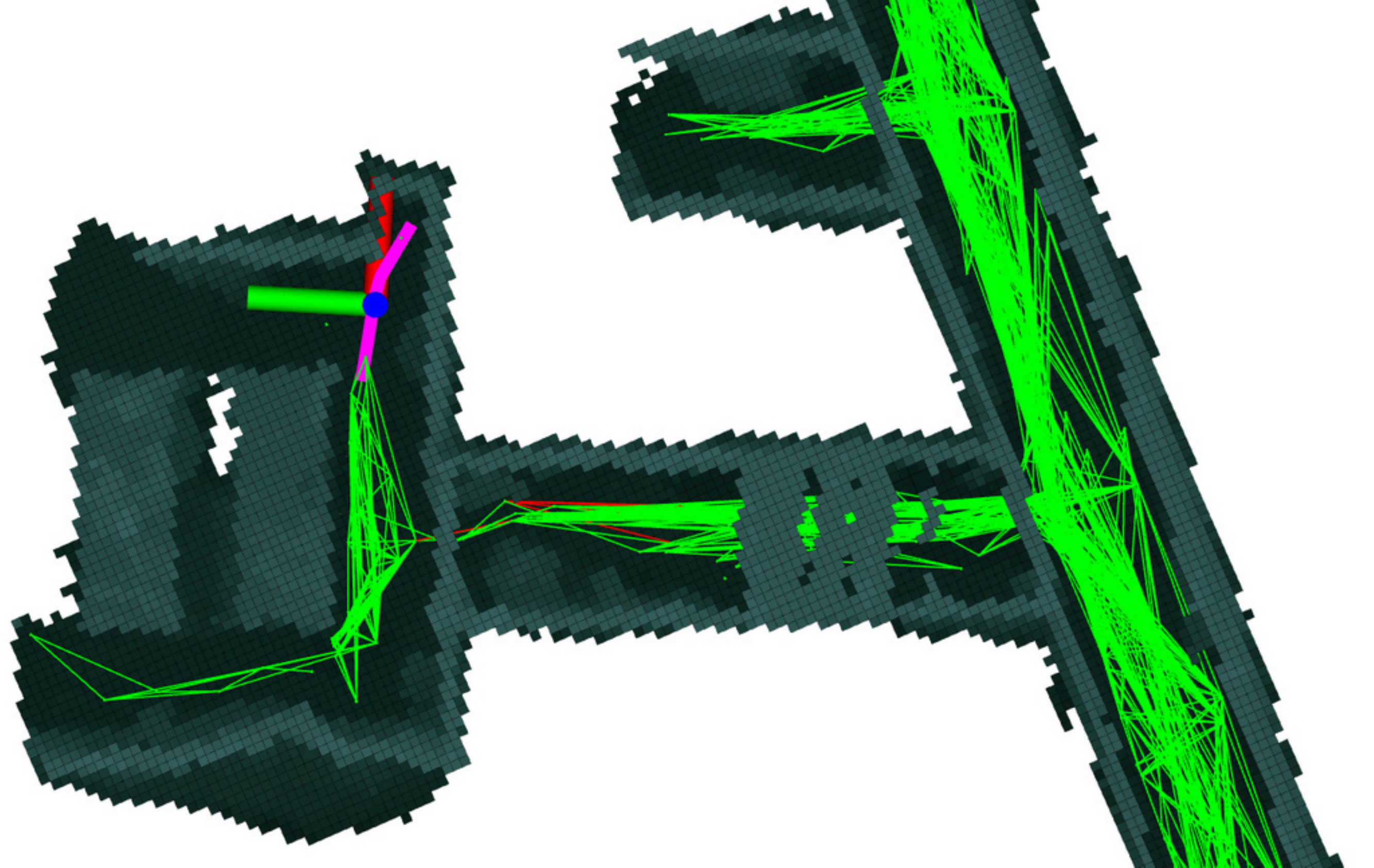}\label{fig:D01_0661s_aftertrapdoor} }}
		\subfloat[]{{\includegraphics[width=.24\textwidth]{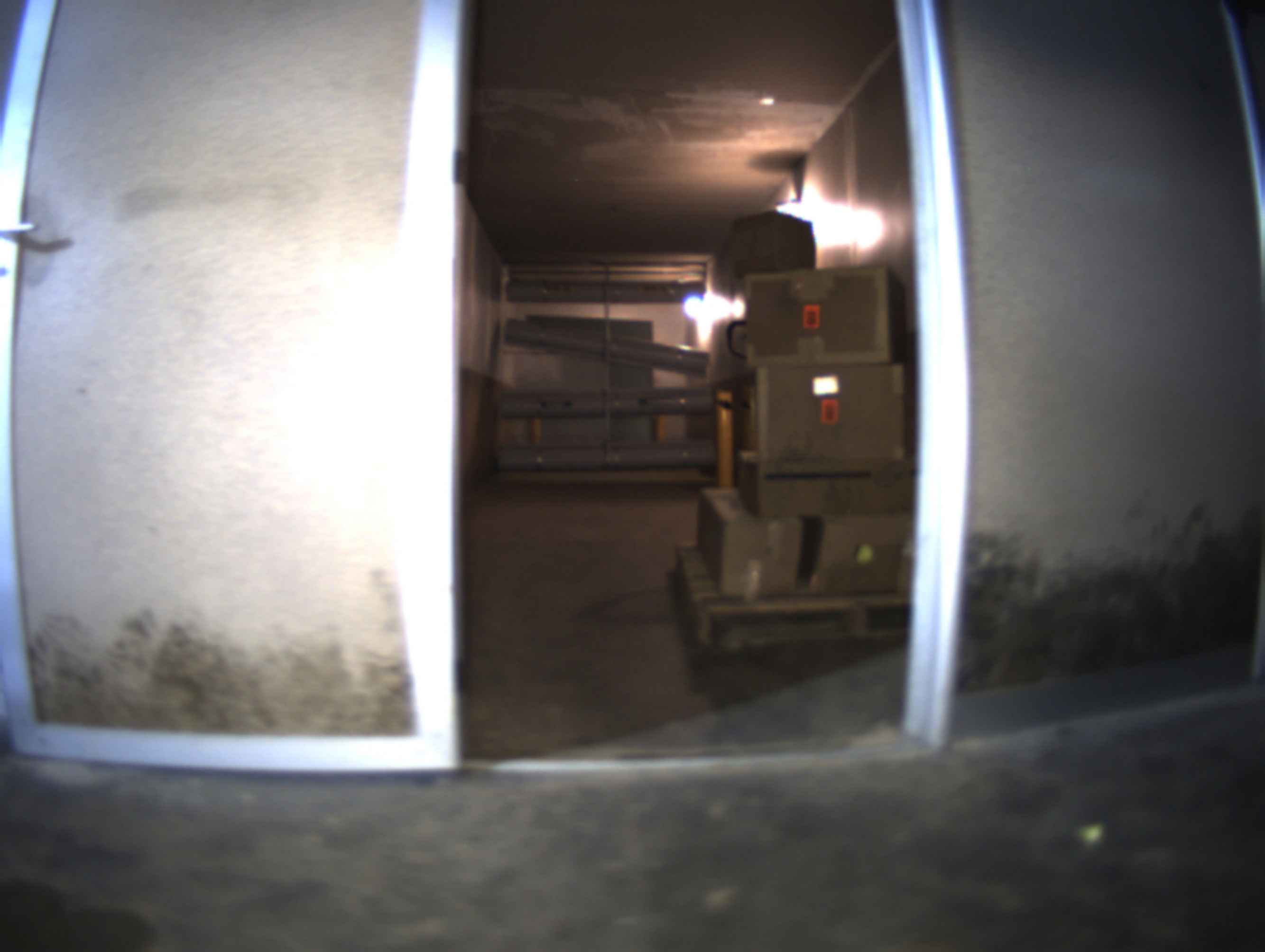}\label{fig:D01_0671s_aftertrapdoormap} }}
		\caption{Early in the mission, D01 walked by the corridor with the trap door, as shown by (b) the left camera (1:47). The agent later to returns to the corridor,  (a) walks under the trap door (10:34), and soon after sees it has closed, as shown by (d) the right camera (11:01). After seeing the trap door close, (c) the updated map and graph show (red) edges as untraversable (11:11). Had the agent moved closer to the trap door, and fully been within mapping range, all edges would have updated as untraversable.}
		\label{fig:trap_door}
\end{figure}

During the Final Event Prize Run, D01 traveled through a side branch of the urban environment, triggering a trap door. Figure \ref{fig:trap_door} shows the planner adapting to the dynamic environment by re-assigning previously traversable edges as untraversable.

In addition, there were several instances of temporary localization and mapping error, which caused erroneous new map data to change previously traversable edges of the planning graph to untraversable. In each case, the planner adapted to the new scenario, and continuously operated throughout the localization drift and loop closure correction. An example of this is included in Section \ref{ssec:sup_planner_dynamic_environment} of the Appendix.

When agents pass each other, they appear as fast-moving dynamic obstacles, and cannot re-plan around one another fast enough. Therefore, an agent-based prioritization scheme prevents both collision and deadlock, by enforcing one agent to wait while the other passes. Examples can be found in Section \ref{ssec:sup_planner_dynamic_environment} of the Appendix.

\subsubsection{Limitations}
\label{sssec:results_planning_limitations}

However, several limitations did prevent agents from exploring roughly half of the course. In total, there were three types of bottlenecks: constrained corridors, slippery surfaces, and downward sets of stairs, each exposing unique limitations within the autonomy system. None of these are limitations of the planner itself, but rather limitations of mapping, mobility, and perception.

The planner did not plan through all constrained spaces because the selected planning parameters for agent width and height did not allow the graph to propagate through especially short and narrow spaces. These parameters were intentionally chosen to be conservative to prevent the agent from moving along an unsafe trajectory. Utilizing a higher-resolution local map and planner could result in a more agile robot that could safely traverse those spaces. Additionally, autonomously transitioning into a crouching gait could improve the Spots ability to traverse spaces with low ceilings. Examples are included in Section \ref{ssec:sup_planner_constrained_spaces} of the Appendix.

The second limitation is slippery surfaces, and led to D01 slipping and falling in the cave section. Some of these rocky surfaces were intentionally designed to be slippery, and plenty of humans walking through the course after the event also slipped and fell. After D01 fell, it also experienced a localization failure. Methods to recovery the system from an event such as this one would involve implementing a fall detection algorithm, as well as autonomous self-righting and localization reset logic.

\subsection{Artifact Detection}
\label{ssec:results_artifact_detection}

\begin{figure*}[hbt!]
    \centering
    \includegraphics[width=0.95\textwidth]{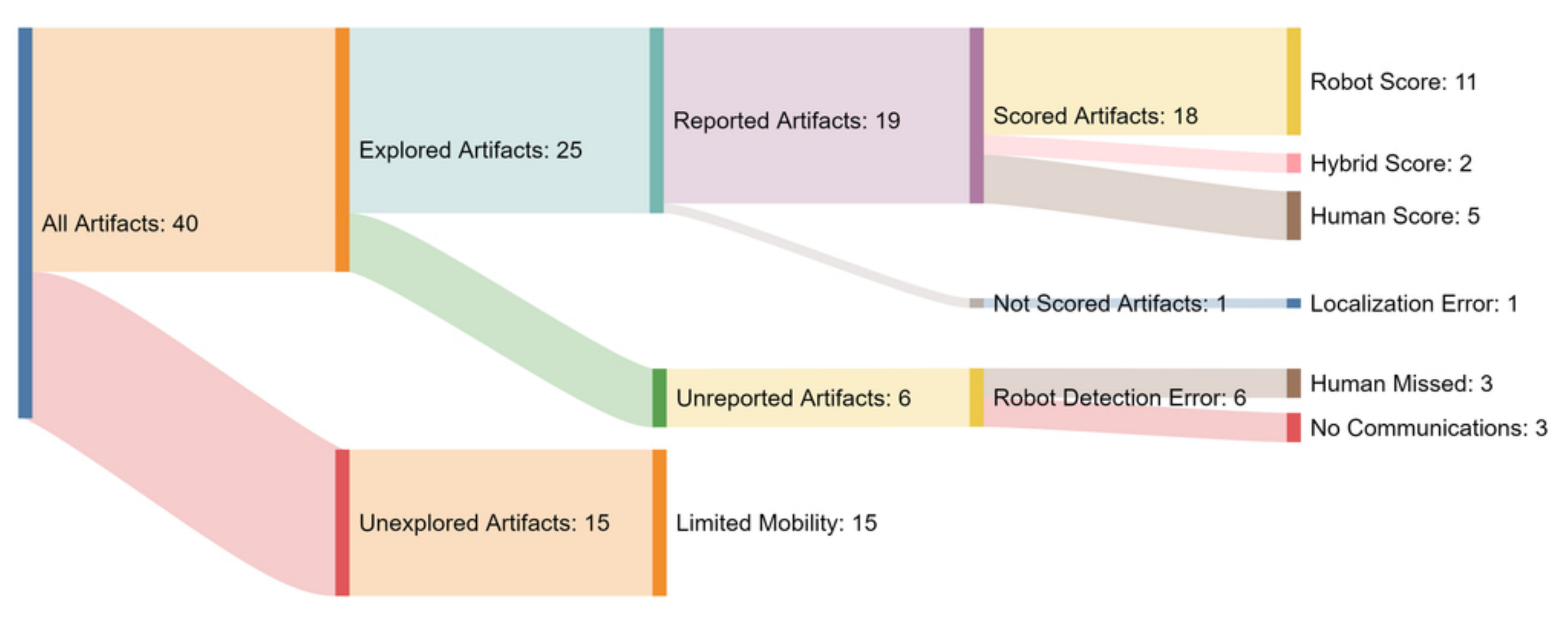}
    \caption{Flow diagram illustrating how Team MARBLE scored 18 of the 40 artifacts in the course, and the limitations preventing the remaining 22 artifacts from being scored.}
    \label{fig:sankey}
\end{figure*}

\begin{figure}[!htb]
    \centering
    \includegraphics[width=0.95\textwidth]{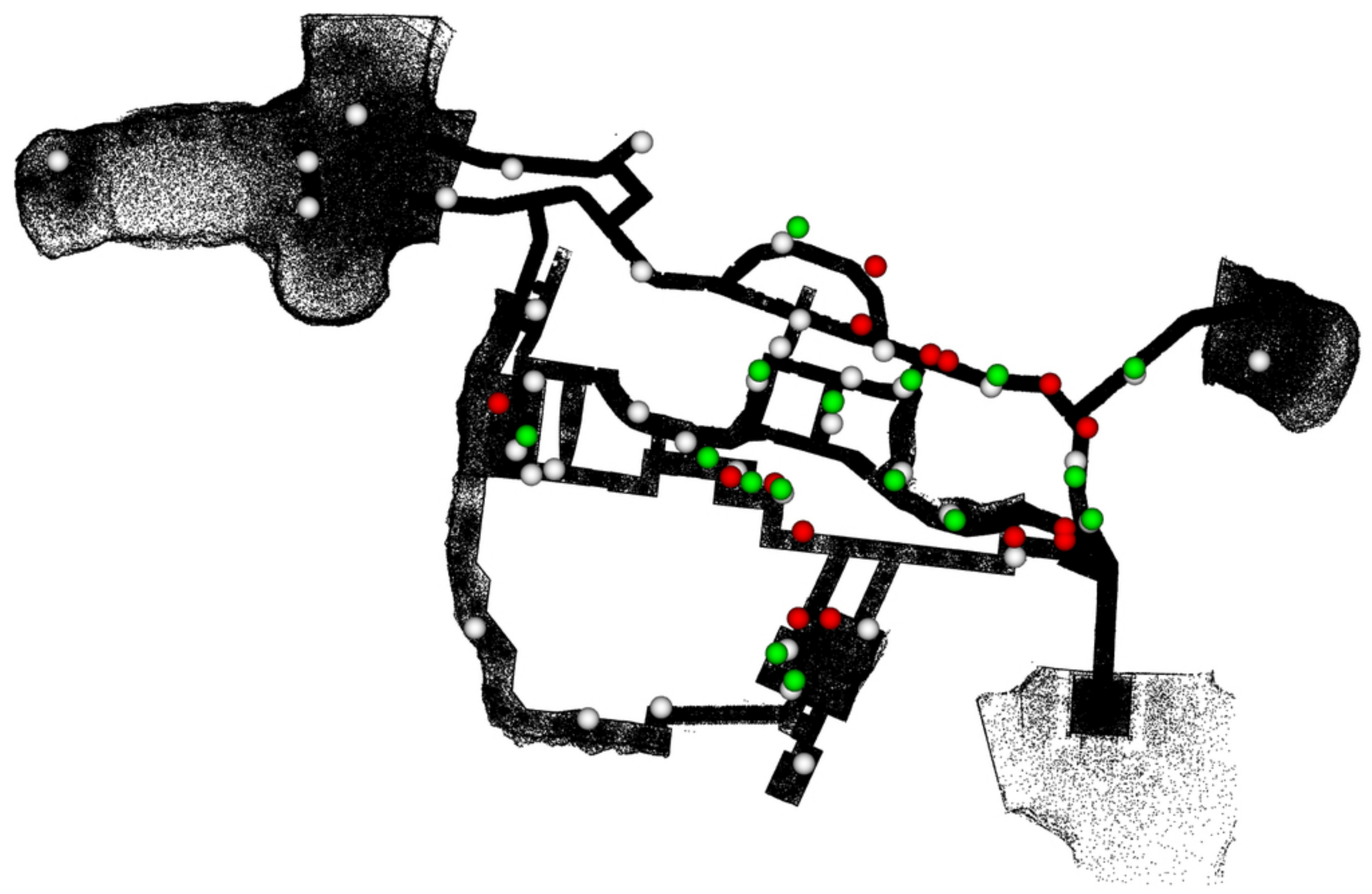}
    \caption{Locations of ground truth artifacts in white, reports that scored in green, and reports that did not score in red, overlaid on ground truth map of the course.}
    \label{fig:merged_map_with_scored_not_scored_reports}
\end{figure}

In this section, we present performance results of the artifact detection and reporting system. During the 60-minute Final Event Prize Run, Team MARBLE scored a total of 18 artifacts out of the 40 artifacts in the environment. Figure \ref{fig:sankey} presents a flow diagram that summarizes how our team scored 18 artifacts and the limitations that resulted in the remaining 22 from being scored. Of all 40 artifacts, our agents explored enough of the environment that they were in the vicinity of 25 artifacts, leaving 15 unexplored due to mobility challenges discussed in Section \ref{ssec:results_planning}. Team MARBLE reported 19 of the 25 artifacts that were explored, and successfully scored 18 of those 19 reported. A map of the area explored and scored artifacts is shown in Figure \ref{fig:merged_map_with_scored_not_scored_reports}. Details of these 25 explored artifacts, of which 18 were scored, one was missed, and six were unreported, are shared in Section \ref{ssec:sup_explored_artifacts} of the Appendix. 

Of the 18 artifacts that Team MARBLE successfully scored, 11 were scored by autonomous robot reports, two were scored by the human supervisor modifying the position of autonomous reports, and five were scored by the human supervisor via robot FPV imagery. One artifact was reported but did not score due to localization error in excess of 5m. There were six artifacts that agents saw, but did not report due to errors in the autonomous artifact detection system. The human supervisor received information regarding three of these six artifacts but missed them due to high workload demands during the mission. The other three artifacts were located in areas that prevented agents from communicating back to the human supervisor. 

\subsubsection{Visual Detection}

The focus of this section is to quantify the performance of the visual artifact detection system. Of the 18 artifacts that Team MARBLE scored, 11 of them were visual artifacts, as shown by Table \ref{tab:artifact_scored_not_scored}. First, we focus on the six artifacts (L51, L53, L26, L34, L40, L67) that were successfully reported by the agents' autonomous artifact detection systems in the course. All six artifacts were accurately localized to within 5m. In fact, the largest error for a visual artifact was 2.87m (L67). This eliminated the need for the human supervisor to spend time trying to correctly localize artifacts.

The autonomous artifact detection system filters raw frame-to-frame detections onboard the agent, with the aim to reduce the number of false positive and redundant artifact reports. Because the human supervisor has limited bandwidth, unnecessary distractions detract from the completing other mission-related tasks. In the process of scoring these six visual artifacts, the human supervisor had to process 21 artifact reports from the automated artifact detection systems onboard agents. Of these, 11 were true positives, six were approved by the human supervisor and successfully reported, and five were ignored because they were redundant reports that were previously scored. The other 10 reports were also ignored by the human supervisor because they were false positives.

In total, there were only four false reports that the human supervisor submitted. One was cause by human error, the other three were caused by erroneous CO$_2$ reports, as detailed in Section \ref{ssec:sup_false_artifacts} of the Appendix.

Agents in the course failed to autonomously report the other five artifacts (L55, L32, L31, L38, L58), but did transmit FPV imagery back to the human supervisor, who manually reported and scored them. The fact that the autonomous artifact detection system did not detect five of the 11 visual artifacts it saw, indicates that certain reliability limitations exist. Team MARBLE acknowledged this limitation and relied on the the human supervisor and FPV system to fill in that void, which is further in Section \ref{ssec:results_mission_management}.

\subsubsection{Non-Visual Detection}

Team MARBLE scored seven non-visual artifacts, i.e. cell phone, cube, and gas, but as shown in Table \ref{tab:artifact_reports}, required submitting more reports due to difficulty around accurately localizing the source. The main limitation is that the detection scheme relies on threshold-based logic for RF and CO$_2$ levels, and when triggered, simply reports the current location of the agent. The thresholds were intentionally set low to increase the probability of detection when agents pass by the vicinity of these non-visual artifacts. Of the seven non-visual artifacts scored, five of them (L08, L36, L22, L47, L24) were scored via the autonomous robot reports, with an average error of 3.27m. The remaining two artifacts (L11, L59) were scored by the human supervisor manually adjusting the reported artifact locations.

\begin{table}[hbt!]
\begin{center}
    \begin{tabular}{c c c c c c}
    \hline
    \textbf{Artifact Type} & \textbf{Scored} & \textbf{Not Scored} & \textbf{Unreported} & \textbf{Unexplored} & \textbf{Total} \\ 
    \hline
    Survivor            & 	2   &   0   &   0   &   1   &   \textbf{3}   \\
    Cell Phone          &   3   &   0   &   0   &   1   &   \textbf{4}   \\
    Backpack	        &   2   &   0   &   0   &   3   &   \textbf{5}   \\
    Drill	            &   2   &   0   &   0   &   2   &   \textbf{4}   \\
    Fire Extinguisher   &	2   &   0   &   1   &   1   &   \textbf{4}   \\
    Gas	                &   2   &   0   &   1   &   0   &   \textbf{3}   \\
    Vent                &   0   &   0   &   3   &   1   &   \textbf{4}   \\
    Helmet              &   1   &   0   &   1   &   3   &   \textbf{5}   \\
    Rope	            &   2   &   0   &   0   &   3   &   \textbf{5}   \\
    Cube                &   2   &   1   &   0   &   0   &   \textbf{3}   \\
    \hline
    \textbf{Total}      &   \textbf{18} &   \textbf{1} & \textbf{6} &   \textbf{15} & \textbf{40} \\
    \hline
    \end{tabular}
\caption{\label{tab:artifact_scored_not_scored} Artifact statistics for Team MARBLE during the Final Event Prize Run. A total of 18 artifacts were \textit{scored}. Only one artifact was reported but \textit{not scored}, which was due to localization error greater than 5m. Agents were in the vicinity of six artifacts, but they went \textit{unreported} due to autonomous artifact detection failure and in some cases, also missed by the human supervisor due to excess workload. The remaining 15 \textit{unexplored} artifacts were never seen by agents because they were located in parts of the course that were never reached.}
\end{center}
\end{table}

\begin{table}[hbt!]
\begin{center}
    \begin{tabular}{c c c c c}
    \hline
    \textbf{Attempt Type} & \textbf{Scored} & \textbf{Missed} & \textbf{False} & \textbf{Total} \\ 
    \hline
    Survivor            & 	2   &   0   &   0   &   \textbf{2}   \\
    Cell Phone          &   3   &   5	&   0   &   \textbf{8}   \\
    Backpack	        &   2   &   0	&   1   &   \textbf{3}   \\
    Drill	            &   2   &   0   &   0   &   \textbf{2}   \\
    Fire Extinguisher   &	2   &   0   &   1   &   \textbf{3}   \\
    Gas	                &   2   &   1   &   3   &   \textbf{6}   \\
    Vent                &   0   &   0   &   0   &   \textbf{0}   \\
    Helmet              &   1   &   0   &   0   &   \textbf{1}   \\
    Rope	            &   2   &   0   &   0   &   \textbf{2}   \\
    Cube                &   2   &   5   &   0   &   \textbf{7}   \\
    \hline
    \textbf{Total}      &   \textbf{18} &   \textbf{11} &   \textbf{5} & \textbf{34} \\
    \hline
    \end{tabular}
\caption{\label{tab:artifact_reports} Artifact report statistics for Team MARBLE during the Final Event Prize Run. A total of 34 reports, or attempts, were made throughout the mission. A total of 18 attempts resulted in \textit{scores}, 11 attempts were \textit{misses} and did not result in a score due to localization error greater than 5m, and five attempts were \textit{false} attempts in that they were false positives and no artifact of that class was in the vicinity.}
\end{center}
\end{table}

\subsection{Communications}
\label{ssec:results_comms}

\begin{figure}[!htb]
		\centering
		\subfloat[]{{\includegraphics[width=0.57\textwidth]{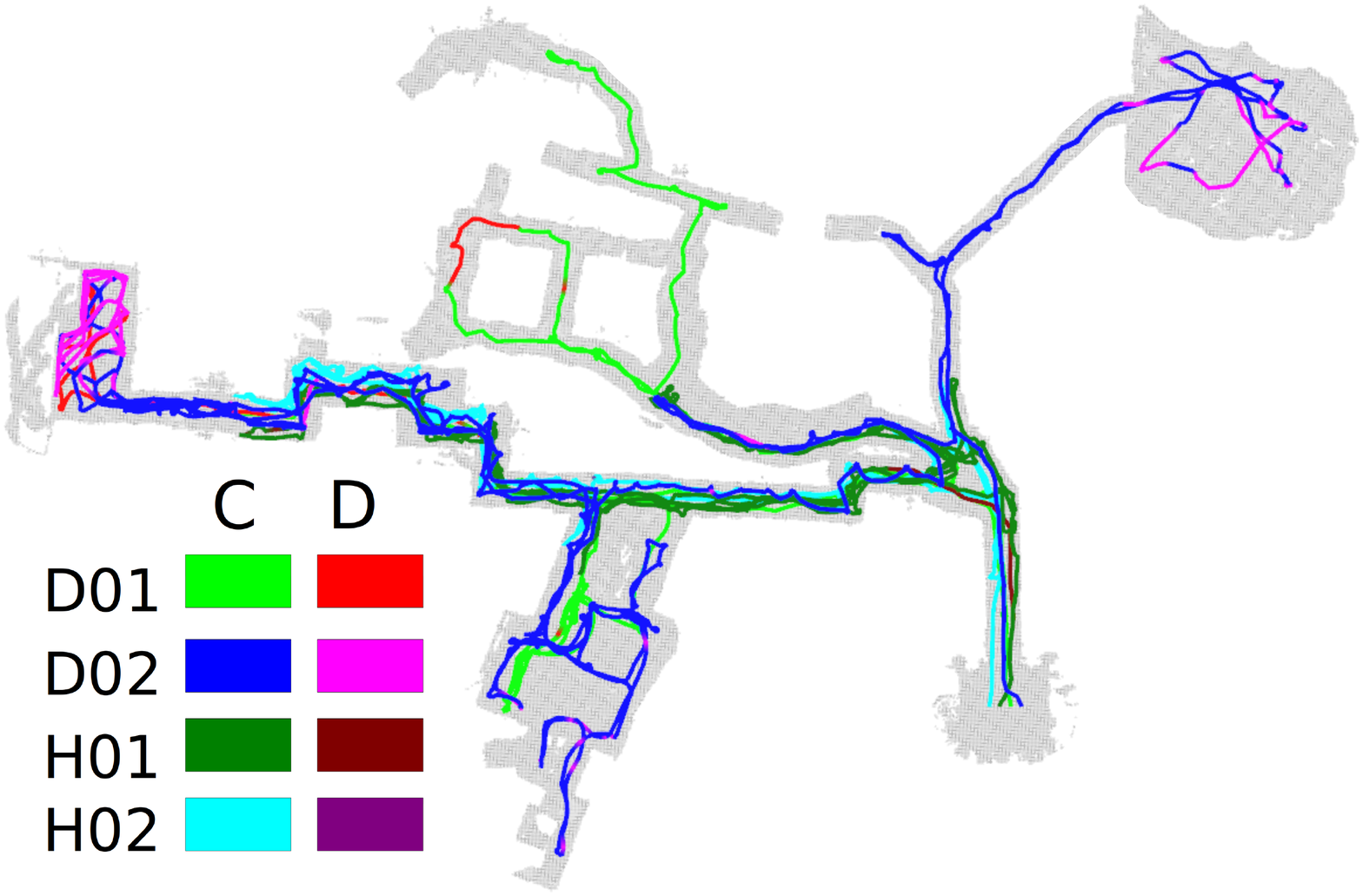}\label{fig:comm_results_map} }}
		\subfloat[]{{\includegraphics[width=0.41\textwidth]{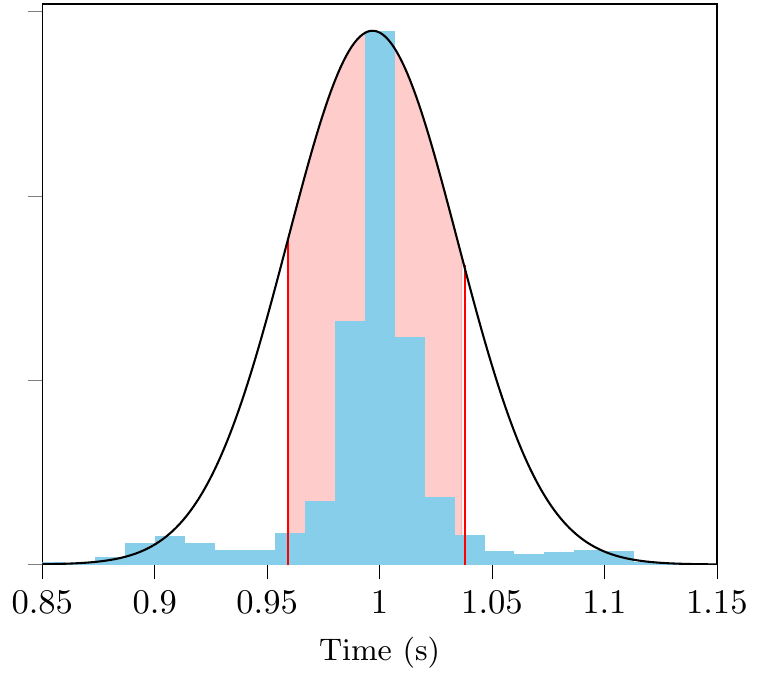}\label{fig:comm_results_latency} }}\\
    \caption{Performance results of the communication systems, including (a) the map of the Team MARBLE's deployment during the Final Event Prize Run, overlaid with locations of robot connection (C) to the network in blue-green, and locations of robot disconnection (D) from the network in red-magenta, as well as (b) the distribution of inter-message arrival times for D01 with a nominal publishing rate of 1 Hz, overlaid with $\mathcal{N}(1.00,0.04)$. The highlighted red region represents the first $\sigma$ value which contains 68\% of the message times.}
	\label{fig:comm_results}
\end{figure}

The performance of the communication system was evaluated in the final run using both qualitative and quantitative measures. Subjectively, the human operator was able to employ live FPV video from the robots, a capability that directly contributed to team's third place finish. The robots were in communication with the base station over the majority of the explored regions of the course, as shown by the blue-green hues in Figure \ref{fig:comm_results_map}, allowing the human supervisor to monitor and intervene as needed. Overall, 125.2 MB of data was transferred through the communications network, including all map segments, telemetry, artifact reports, and other data products. Of that, FPV video comprised 51.6 MB. The latency of the mesh networking solution was evaluated using inter-message arrival times of a heartbeat message sent from a long-ranging robot to the base station. This mission management message, transmitted regularly from the robot, was part of our protocol scheme and is analyzed as a message of convenience. A distribution of the inter-message arrival times of these heartbeat messages from D01 to the base station is shown in Figure \ref{fig:comm_results_latency}.
 
Since these messages originate from the robot at 1 Hz, an ideal system would observe all inter-message arrival times to be one second in duration. As our mission management system on the robot does not run with hard realtime constraints, the 1 Hz publish rate is an estimate that includes noise due to process load, etc. On the base station, there was an approximately normal ($\mathcal{N}(1.00,0.04)$) distribution of arrival times. Since messages may experience delays, the immediately following message may exhibit an inter-message time of less than one second, leading to the symmetry apparent in Figure \ref{fig:comm_results_latency}. Our key observation of this plot is that the bulk of messages arrive within five percent of their expected times across a distance of hundreds of meters and multiple mesh hops, validating the performance of the entire mesh networking solution.

\subsection{Mission Management}
\label{ssec:results_mission_management}

\begin{figure*}[htbp]
    \centering
    \includegraphics[width=1.00\textwidth]{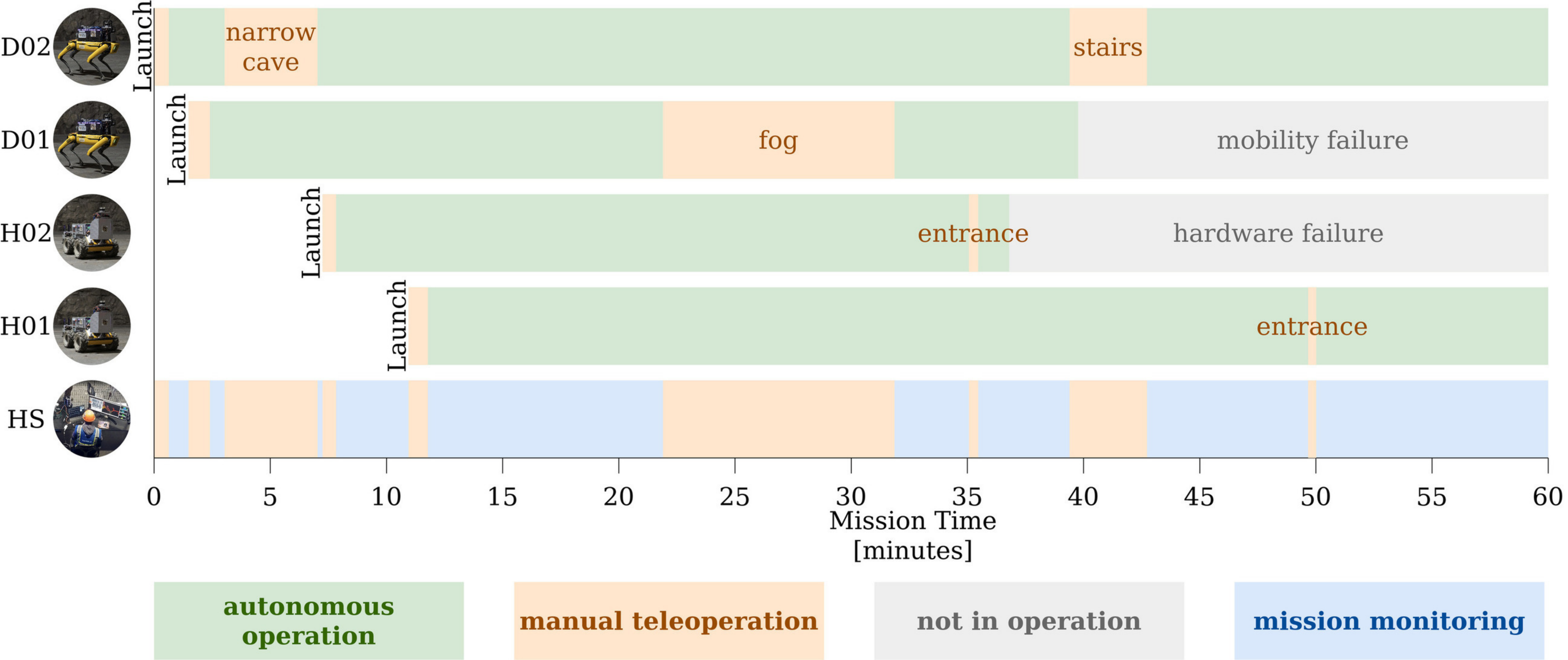}
    \caption{Mission management timeline for all robots and human supervisor during the 60-minute Final Event Prize Run. When not manually teleoperating a robot, the human supervisor was monitoring the mission, which includes watching live FPV streams from D01 and D02, reviewing incoming artifact reports from agents, reporting artifacts to DARPA, and in the last five minutes of the mission, checking archived FPV images from the Spots for previously missed artifacts.}
    \label{fig:teleoperation_timeline}
\end{figure*}

Overall, the mission management system was able to keep the robots on task with minimal human supervisor intervention. However, when needed, the interventions were crucial towards both the exploration capabilities of the system and the final event performance. Figure \ref{fig:teleoperation_timeline} presents a detailed timeline of the four agents in the field, as well as the human supervisor. The four robot launches and five robot interventions were the only times when the human supervisor used teleoperation. There were only two other types of instructions agents received from the human supervisor. One was commanding H02 to drop two communication beacons. The other was commanding D01 to return home three times, each occurring while the human supervisor was also teloperating D01 through the fog.

\subsubsection{Robot Launches}

The human supervisor was under immense pressure to optimally balance many competing tasks during the 60-minute Final Event Prize Run. With such limited time, the primary objective at the beginning of the mission is to launch robots into the environment as fast as possible. Through extensive practice and full-scale comprehensive field deployments, discussed further in Section \ref{ssec:lessons_learned_deployments}, Team MARBLE launched all four robots, with a mean launch time of 41 seconds, as shown in Table \ref{tab:launch_times}.

Two agents experienced failures late in the mission, reducing overall fleet utilization rate from 92\% to 73\%. H02 experienced a hardware failure (36:48), and post-event inspection revealed better vibration isolation of the computing system would reduce the likelihood of such a failure in the future. D01 experienced a mobility failure (39:46), in which the Spot slipped on a slick rock and fell over. The agent then experienced a localization instability due to the large induced velocity. To recover from such an incident in the future, Team MARBLE could implement an autonomous self-righting maneuver and localization reset logic.

\begin{table}[hbt!]
\centering
\begin{tabular}{c c c c c}
    \hline
    \textbf{Launch}   &   \textbf{Agent}  & \textbf{Duration} & \textbf{Teleoperation Window}      & \textbf{Course Entry}  \\
    \textbf{}               &   \textbf{}       & \textbf{[s]}      & \textbf{[mm:ss - mm:ss]}  & \textbf{[mm:ss]}          \\
    \hline
    RL1   &   D02 &   75  &   -00:37 - 00:38  &   00:02      \\
    RL2   &   D01 &   55  &   01:29 - 02:24   &   01:44      \\
    RL3   &   H02 &   35  &   07:15 - 07:50   &   07:25      \\
    RL4   &   H01 &   50  &   10:57 - 11:32   &   11:08      \\
    \hline
        &   \textbf{Mean} & \textbf{41} & & \\
    \hline
\end{tabular}
\caption{List of the four robot launch (RL) sequences executed by the human supervisor during the 60-minute Final Event Prize Run. Extensive process streamlining and repeated practice deployments resulted in quick and repeatable launch sequences. The course entry column represents the mission time at which the agent crossed into the course.}\label{tab:launch_times}
\end{table}

Despite the fact that some artifacts were not detected and reported by the autonomous board artifact detection system, the human supervisor filled in the void. The human supervisor reported and scored five artifacts that were seen via robot FPV imagery, but not autonomously detected. Of these, the human supervisor saw four (L55, L32, L31, L38) from live FPV streams, while one (L58) was found while reviewing archived FPV images near the end of the mission. More details of these artifact reports are presented in Section \ref{ssec:sup_all_reports} of the Appendix.

\subsubsection{Robot Interventions}

\begin{figure}[hbt!]
		\centering
		\subfloat[]{{\includegraphics[width=.4\textwidth]{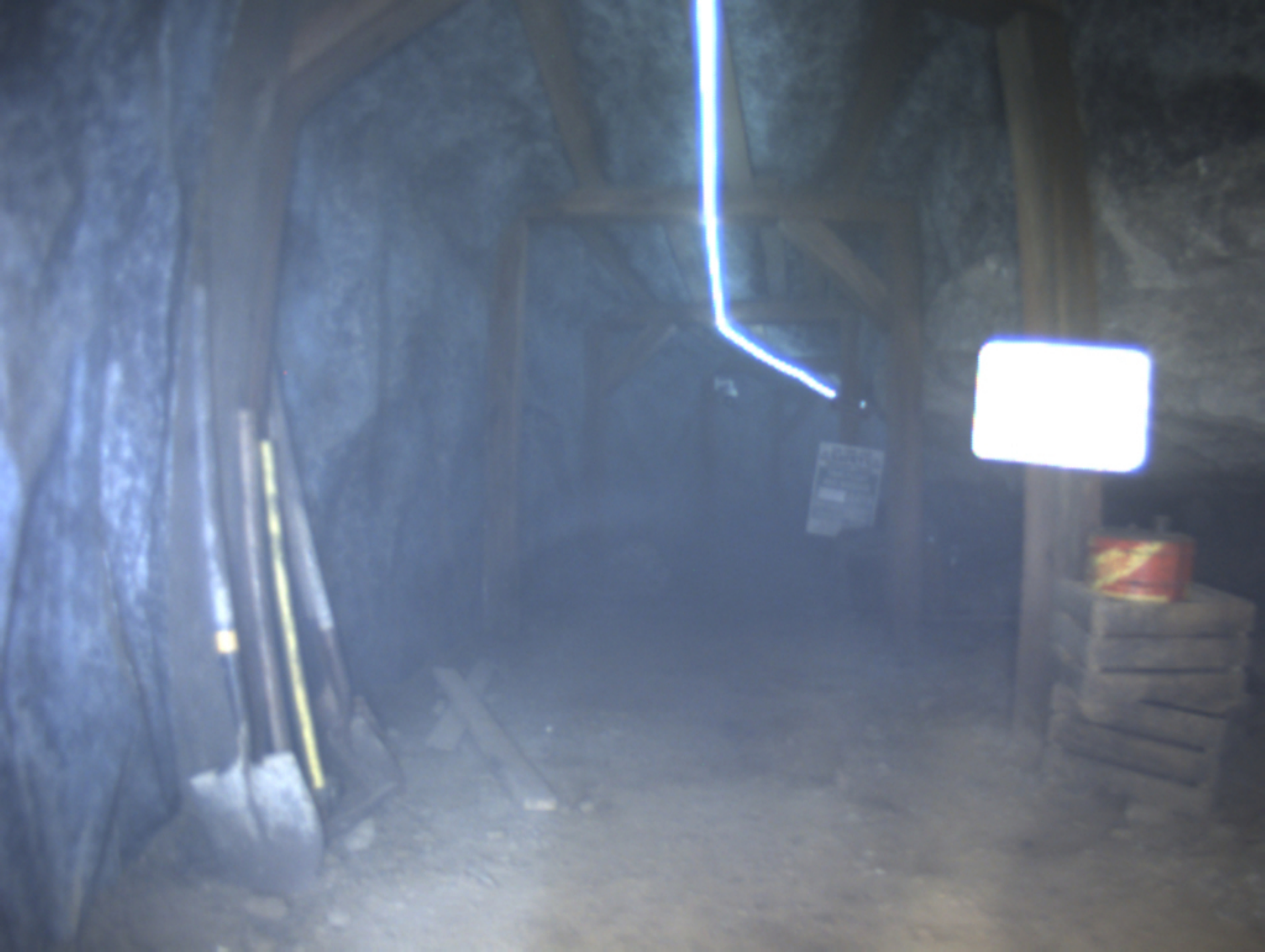}\label{fig:fpv_imagery_comparison_fullres} }}
		\subfloat[]{{\includegraphics[width=.4\textwidth]{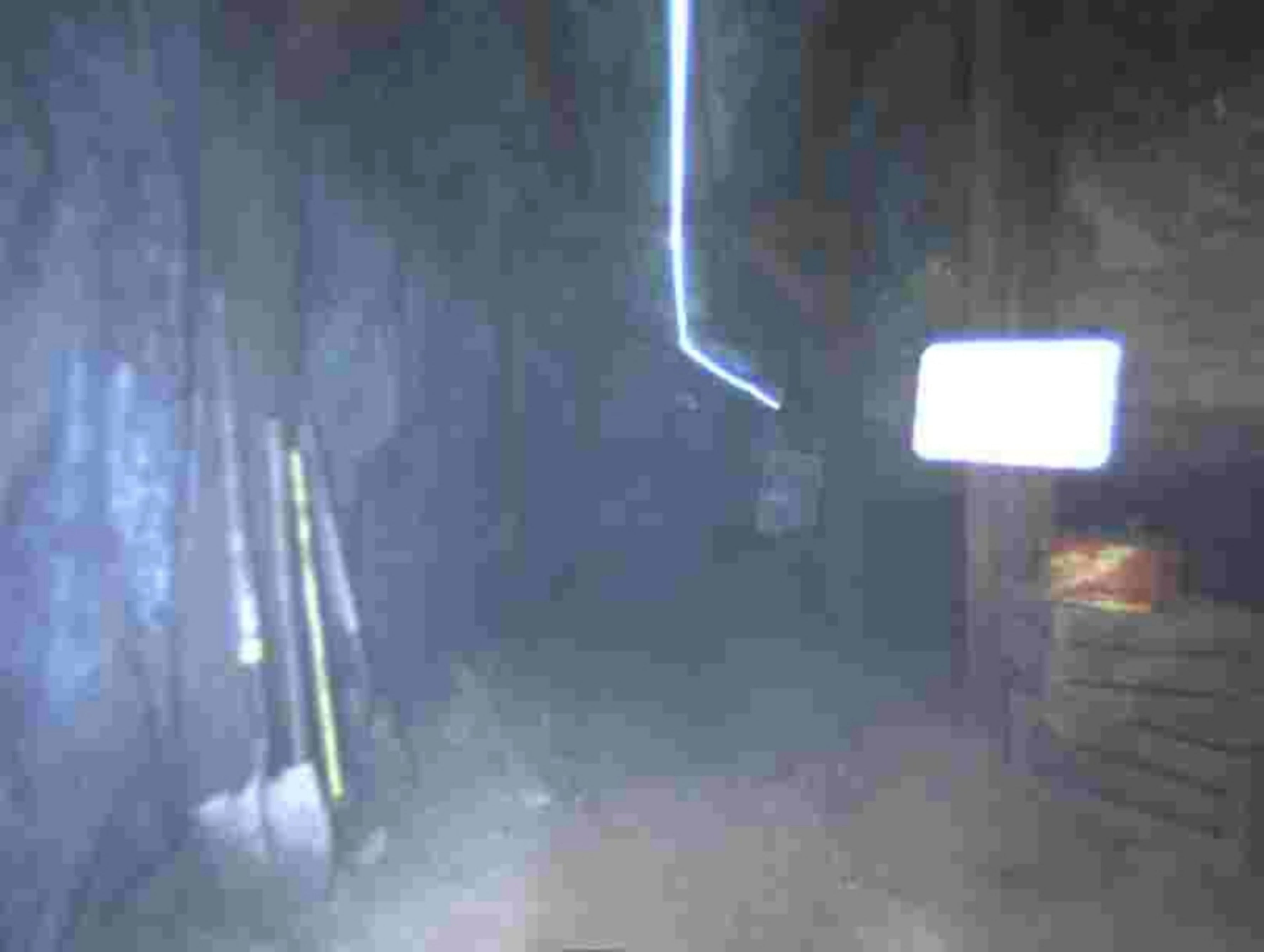}\label{fig:fpv_imagery_comparison_lowros} }}
		\caption{Imagery during robot intervention 2 (RI2). when human supervisor was pushing D01 through the foggy area in the tunnel environment, eventually leading to six additional points. Shown is a comparison of (a) full-resolution FPV imagery onboard the agent and (b) compressed low-resolution FPV imagery transmitted to the human supervisor.}
		\label{fig:fpv_imagery_comparison}
\end{figure}

In total, the human supervisor intervened in the autonomous fleet five times during the 60-minute mission. Table \ref{t:teleop} presents the duration of each intervention, with a mean length of 217 seconds, as well as the reason for intervening. Each of these interventions was in the form of manual teleoperation, in which FPV streams were available at the base station, and velocity commands from the human supervisor was transmitted to the remote agent in the field. Of the five manual teleoperation interventions, only two had significant impact on the mission. The first (RI1) was navigating D02 through the narrow cave corridor early in the mission (3:02 - 7:02), which led the agent to the small cavern and two artifact scores. The second (RI2) was navigating D01 through fog in the tunnel section midway through the mission (21:54 - 31:52), leading to six artifact scores. Figure \ref{fig:fpv_imagery_comparison} provides a side-by-side comparison of the full-resolution FPV imagery processed onboard by the autonomous artifact detection system and the compressed FPV imagery transmitted to the human supervisor.

Two interventions (RI3, RI5) commanding agents back into the course, was out of a shear abundance of caution. The planning algorithm is configured so that the staging area is treated as explored, so agents should not attempt to explore it. One intervention failed (RI4), in which the human supervisor attempted to teleoperate D02 down the stairs by the subway platform. This attempt failed because communication to and from the robot was intermittent.

The main takeaways from these results is that our agents are highly autonomous, leaving the human supervisor to focus on mission monitoring and targeting strategic, high-value intervention opportunities. The mission management system enabled convenient transition between autonomy and manual operation, while the communication system enabled visibility and control over the agents in the field.

\begin{table}[hbt!]
\centering
\begin{tabular}{c c c c c c c}
    \hline
    \textbf{Intervention}   &   \textbf{Agent}  &   \textbf{Duration}   &   \textbf{Window}   &  \textbf{Goal} &  \textbf{Success} &  \textbf{Points} \\
    \textbf{}               &   \textbf{}       &   \textbf{[s]}        &   \textbf{[mm:ss]} &  & & \\
    \hline
    RI1   &   D02 &   240 &   03:02 - 07:02   &   Enter narrow cave corridor  &   +   &   6   \\
    RI2   &   D01 &   598 &   21:54 - 31:52   &   Enter foggy tunnel area     &   +   &   2   \\
    RI3   &   H02 &   24  &   35:04 - 35:28   &   Avoid course exit           &   +   &   0   \\
    RI4   &   D02 &   200 &   39:24 - 42:44   &   Walk down stairs            &     --  &   0   \\
    RI5   &   H01 &   21  &   49:40 - 50:01   &   Avoid course exit           &   +   &   0   \\
    \hline
        &   \textbf{Mean} & \textbf{217} & & & & \textbf{1.6} \\
    \hline
\end{tabular}
\caption{List of the five robot interventions (RI) executed by the human supervisor during the 60-minute Final Event Prize Run. Our concept of operations relies on autonomous multi-agent exploration, and does not necessitate manual waypoints or teleoperation from the human supervisor. Therefore, agents were completely autonomous, except for the human supervisor input during these five instances of teleoperation. The interventions goals varied, but were all specific scenarios where human intervention would augment autonomous agent capabilities in a mission-relevant manner.}\label{t:teleop}
\end{table}

\subsection{Open-Source Data}
\label{ssec:results_opensource_data}

During the final run at the final event, our team collected a significant amount of data related to autonomous subterranean exploration in the form of ROS ``rosbags". Our datasets are split up by agent, and each set contains a rosbag of the system inputs, mostly consisting of raw sensor data, and another rosbag of the outputs used for visualization and performance monitoring. The complete collection of data recorded at the final event can be found at \href{https://arpg.github.io/marble}{https://arpg.github.io/marble}.

\section{Lessons Learned}
\label{sec:lessons}

Underground exploration of previously unknown environments, especially in a time and resource-constrained search-and-rescue context, requires a highly adaptable human-robot team. The lessons learned presented in the following sections enhance our proposed system's flexibility across mobility, communications, human-robot teaming, and multi-agent coordination.

\subsection{Platform Mobility}
\label{ssec:lessons_learned_platform_mobility}

Systems with heterogeneous platforms allow for specialization by each platform for both specific environments, and roles which benefit the entire mission. Specifically, in Team MARBLE's case, the addition of Spot platforms into an exploration role enabled rapid multi-story expeditions. This capability was further augmented with the utility of higher-payload, wheeled Huskies, carrying communication beacons. When deployed, these beacons allowed the Spot platforms to report artifacts without the need to traverse ``home,'' leading to much more efficient exploration. No single, platform is as performant as the combination of platforms operating in different roles based on each one of their strengths.

\subsection{Testing and Validation} 
\label{ssec:lessons_learned_deployments}
\begin{figure}[htb!]
		\centering
		\subfloat[]{{\includegraphics[width=.33\textwidth, height=1.2in]{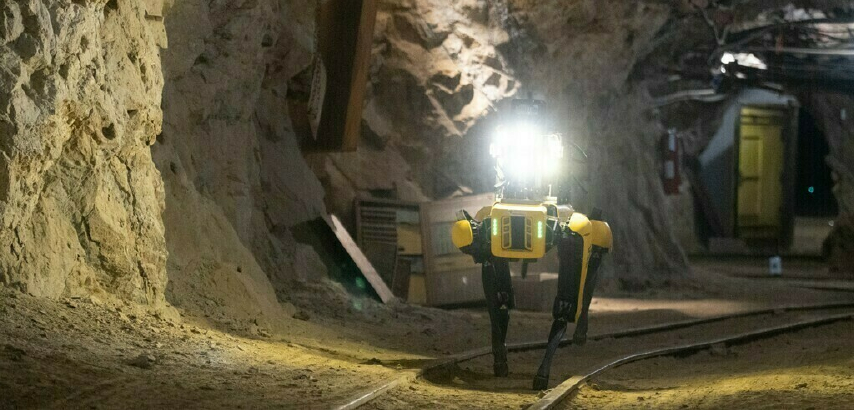}\label{fig:edgar1_cropped} }}
		\subfloat[]{{\includegraphics[width=.33\textwidth, height=1.2in]{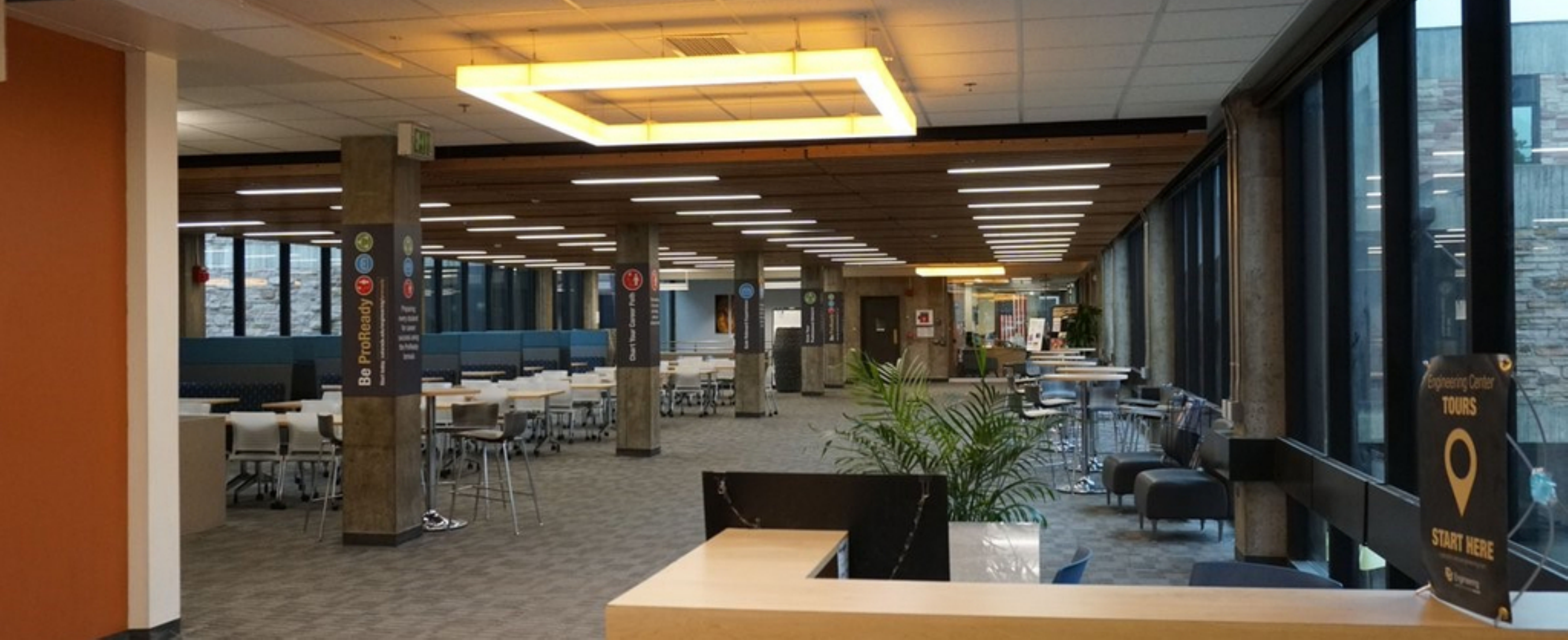}\label{fig:eng_center_first_floor} }} 
		\subfloat[]{{\includegraphics[width=.33\textwidth, height=1.2in]{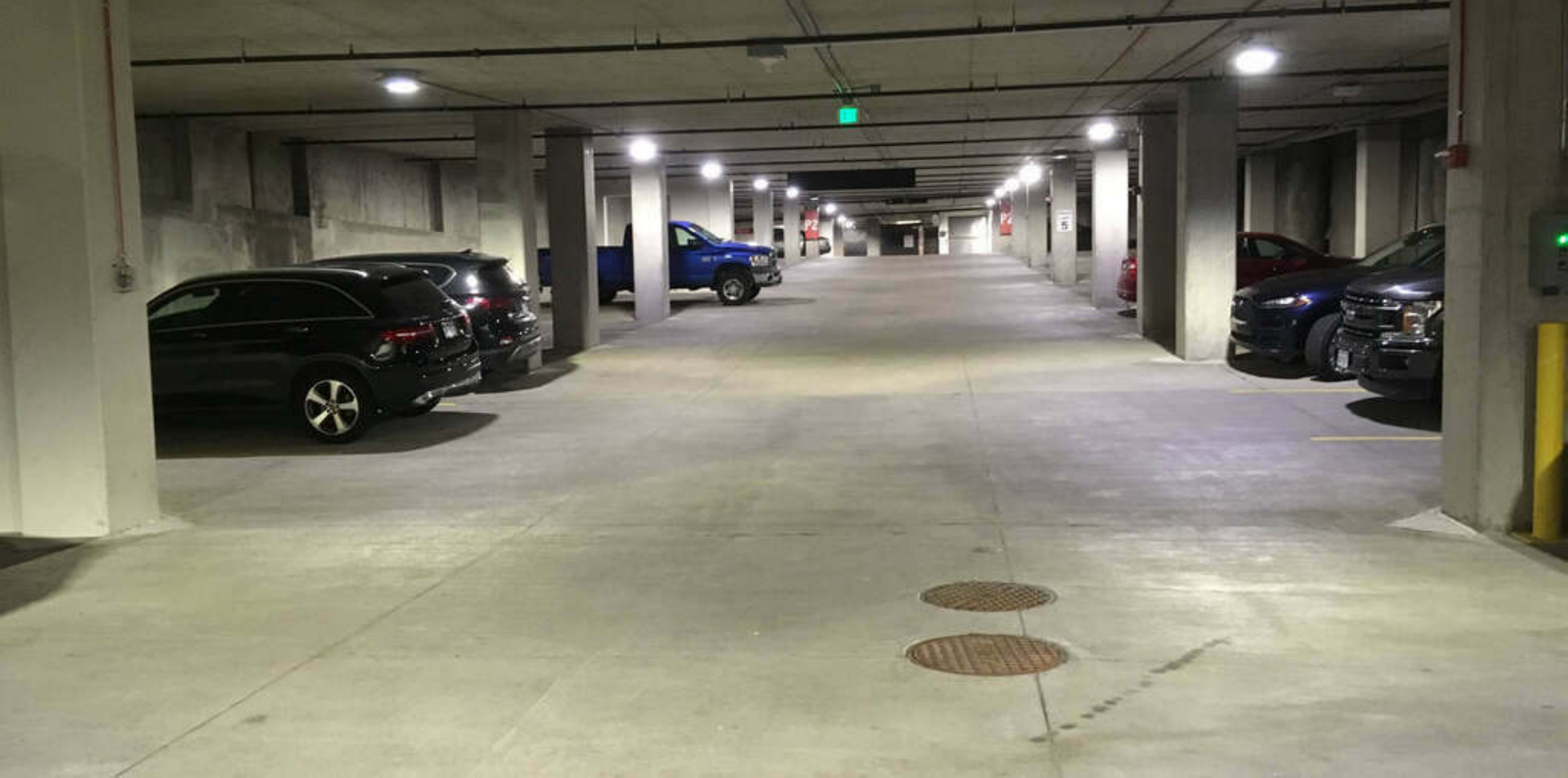}\label{fig:folsom_garage} }} 
		\caption{Team MARBLE conducted comprehensive field deployments at various sites including (a) the Edgar Experimental Mine operated by Colorado School of Mines, located in Idaho Springs, CO. which tested terrain, distance travelled, and communications; (b) the Engineering Center complex also located on University of Colorado Boulder Main Campus which tested multi-story navigation and repeated features from urban environments; and (c) the Folsom Parking Garage located on University of Colorado Boulder Main Campus in Boulder, CO. which tested planning in open spaces and vertical localization across multiple stories.}
		\label{fig:various_urban_environments_result}
\end{figure}

Significant testing and validation was conducted for every element of the Team MARBLE system. These tests taken over a variety of environments, shown in Figure \ref{fig:various_urban_environments_result}, reduced mission-to-mission variability, and increased system-wide adaptability. Through these tests, Team MARBLE locked in well tested solutions with minimal unexplained errors. This reduced changes on the system as we approached the final competition, given that new solutions had to be verified against similarly stringent tests. To our knowledge our team made the fewest hardware and software based adjustments from the preliminary runs to the final run to achieve our score, instead relying on our testing having eliminated all significant errors outside of a so-called "Poisson distributed fatal error." These types of errors were mission ending to any individual robot but difficult to predict or adapt to in a meaningful sense. These can be seen in our final competition run where D01 fell over on rough steep terrain, and where H02 suffered a hardware error. 

A secondary goal of these full-scale deployments was to reduce human-based variability in performance. They helped both the human supervisor and pit crew prepare for the stressors of operating and responding quickly to complex interactions between robots, the environment, and potential failure modes. At no point was a single test considered sufficient for validating a solution as sufficient; instead all solutions released demonstrated repeatability across the same and varied environments. For reference, all full scale deployments are tabulated in Section \ref{ssec:sup_testing_and_validation} of the Appendix.

\subsection{System Adaptability}
The challenges posed by operating a system in an unknown environment necessitate a high level of system adaptability. Predicting every capability that a system will need for a given mission, such as search and rescue, is impractical. Instead, having a highly flexible system capable of adapting to unknown situations is crucial. Having already addressed the mobility considerations in Section \ref{ssec:lessons_learned_platform_mobility}, we also found software adaptability key to our success. For instance, the flexible communication network enabled the ability to pass on FPV to the human supervisor with a simple configuration change. While our system was designed with autonomy in mind and the capability was not previously planned for, the adaptation proved invaluable. Minimal human supervision directly impacted the final score and exploration capabilities of the system as a whole. 

The notion of adaptability extends to other software architecture as well. The artifact detection system was adjusted during the final competition to include SSID information for Bluetooth artifacts. These artifacts could then be more accurately merged between agents and by the human supervisor, despite inaccurate positioning from wireless signals. 

In some cases, adaptability is explicitly accounted for in our design, most notably in the mission management system. BOBCAT has many parameters which allowed the human supervisor to adjust exploration activity, including how long a robot can explore before reporting detected artifacts, whether a robot should find multiple artifacts before reporting, and how the system should adjust to time constraints in the mission. 

\subsection{Autonomy and Human Robot Interaction}
Team MARBLE emphasized autonomous system design as our primary goal. Since only one person, the human supervisor, was permitted to interact with agents while they were deployed, having robots controlling their own paths and decision making was key to reducing the cognitive load required to manage mission objectives. Despite the focus on autonomy, the human supervisor is still necessary to maximize system performance. As Team MARBLE approached the final competition, we found targeted areas where direct human supervisor control improved results. This influenced decisions regarding the number of deployed platforms, how artifacts were passed from agents to the human supervisor, and how we recovered from anomalous robot behavior. 

To maximize the human supervisor's ability to track the data streams present across the fleet, only four robots were deployed. While a larger fleet might have enabled a rapid exploration of the environment, it could reduce the human supervisor's ability to meaningfully address problems arising from any specific robot. 

The artifact detection system was designed to filter against false positives before passing information to the human supervisor. Artifacts had to consistently detected in multiple frames, and be a sufficient distance from existing artifact estimates. Where possible, artifacts were returned with additional information including SSID for Bluetooth artifacts and images for visual artifacts. This data allowed the human supervisor to sort remaining false positives quickly without being distracting from the core mission objectives. 

Specifically, the human supervisor spent most of their operational load solving problems the robots were incapable of correcting through their autonomy stack. For example, during the Final Event Prize Run, the human supervisor used the first person vision to navigate through dense fog, which the robot was incapable of planning through autonomously. After this intervention, the human supervisor allowed the robot to return to autonomous operations, where it explored several new areas, and reported a total of six new artifacts. The human supervisor was able to opportunistically intervene like this, only because the other agents were operating autonomously without supervision.

\subsection{Retrospective}
It's important, after a three-year effort of this size and scope, to examine some of the bigger picture questions. What would we do differently next time? What did we wish we knew at the start, that we know now? The answer to the first two questions is that we would have spent more time at the onset to scope out short-term and long-term development goals. We excelled at the Tunnel Event, placing fourth amongst a large group of competitors. This was mostly due to meeting well-scoped short-term goals within the one year cycle. However, during the six-month development period for the Urban Event, we focused on long-term goals that we ultimately only partially validated, leading to a disappointing performance. In hindsight, we should have focused on making our already capable platforms more capable, rather than spreading our resources thin across many thrusts. However, what our team excelled at after this experience, was pivoting and adopting two new strategies. First, we leaned into student-led project management, which led to greater team coordination, a key component of rapid and effective system development. Secondly, we focused on long-term development goals, and given a year and a half, had the time to adequately develop, test, and validate each subsystem, each fully autonomous robot, as well as our entire fleet in numerous search and rescue missions. Together, these two powerful changes allowed a small, lean team, produce a highly functional autonomy solution that could perform under pressure.


\section{Conclusions}
\label{sec:conclusions}

In this paper, we showcase our flexible autonomy solution for exploring unknown subterranean environments. The highly performant autonomy solution directly lead to a third place finish at the DARPA SubT Final Event which was focused on search and rescue. Moreover, the specific innovations presented in graph planning, flexible communications, and mission management are directly applicable to other multi-robot teaming applications under limited human supervision.

Specifically, the combination of legged and wheeled robots allowed for heterogeneous teaming enabling both rapid exploration and a robust communication network. The deployed mesh network which is described in Section \ref{sec:comm_systems} enables flexible configuration and prioritization of data sent both to other robots, and a human supervisor for review. A powerful graph planning framework, as described in Section \ref{sec:planning}, paired with semantically encoded Octomaps enabled safe, rapid exploration. The final system was able to explore a variety of underground environments, including gold mines and subway stations, with minimal human input.

Human input was reserved for specific situations where higher-level reasoning had the potential to improve the mission. The flexible mission management system described in Section \ref{sec:mission_management} enables safe transitions between human input and the underlying autonomy system. One of the most important lessons learned from the developed system is that focusing on autonomy is core for human robot teaming. Robots need to be able to make decisions on their own, enabling the limited human resource to only act in critical situations.

\clearpage
\newpage

\subsection*{Acknowledgments}
This work was supported through the DARPA Subterranean Challenge, cooperative agreement number HR0011-18-2-0043. Results presented in this paper were obtained using the Chameleon testbed supported by the National Science Foundation. A special thanks to  Andrew Beauthard, Nikolaas Bender, Cesar Galan, Nicole Gunderson, Davis Landry, Greg Lund, Cole Radetich, Ben Rautio, Zoe Turin for assisting with the design, testing, and deployments of our platforms and algorithms. Thank you to the Colorado School of Mines and the Edgar Experimental Mine for allowing us to conduct mock deployments in the mine. Special thanks also to Simon Wunderlich and his team at Meshmerize GmBH for their mesh networking support. Finally, thank you to all of the DARPA staff who have planned and executed absolutely incredible Subterranean Challenge system track circuit events.

\clearpage
\newpage
\bibliographystyle{apalike}
\bibliography{refs}

\clearpage
\newpage


\section{Appendix}
\label{sec:supplementary_material}

\subsection{Communication Beacon Design}
\label{ssec:sup_comm_beacon_design} 

\begin{figure}[!htb]
		\centering
		\subfloat[]{{\includegraphics[width=0.475\textwidth]{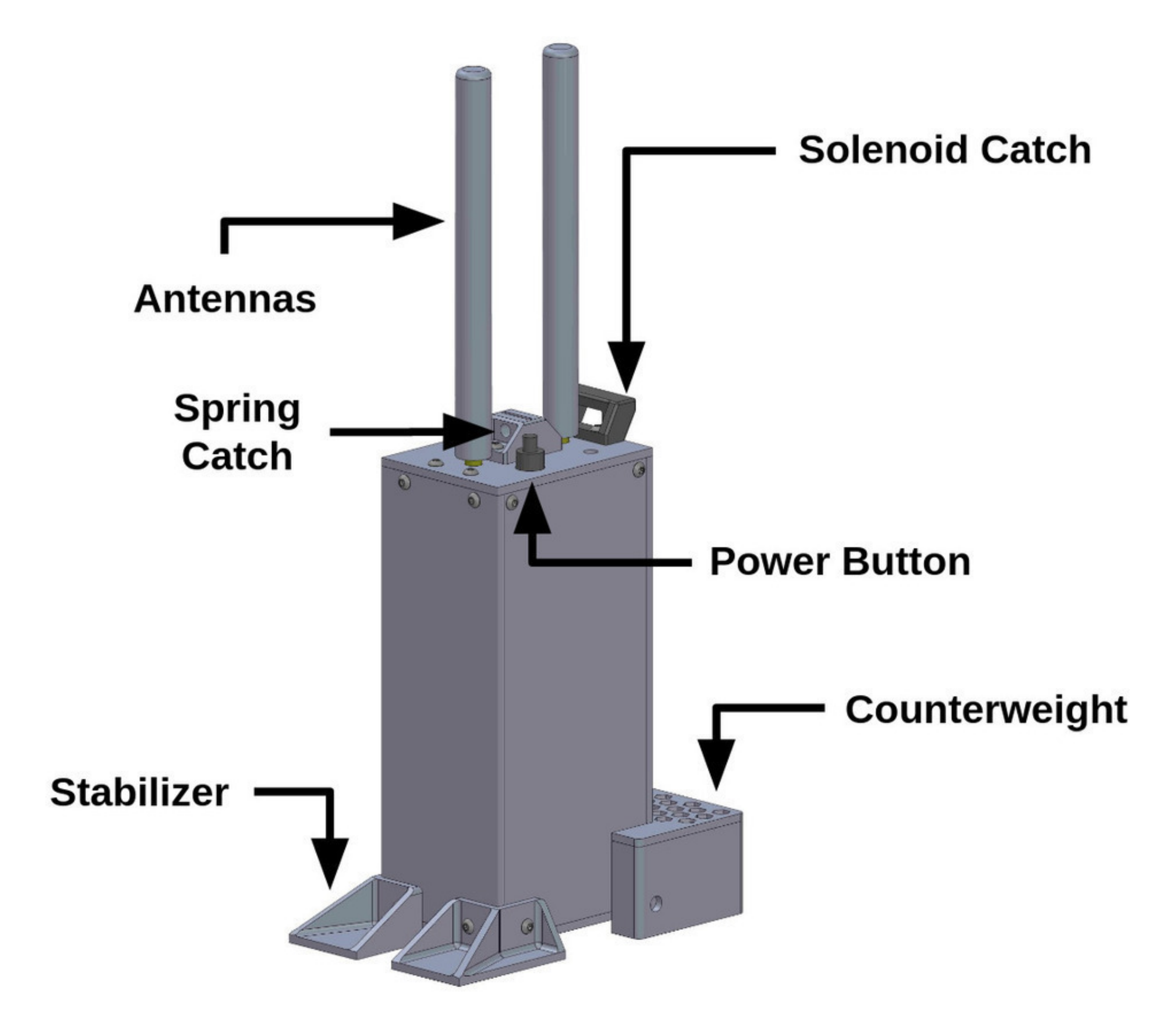}\label{fig:beacon_diagram} }}
		\subfloat[]{{\includegraphics[width=0.475\textwidth]{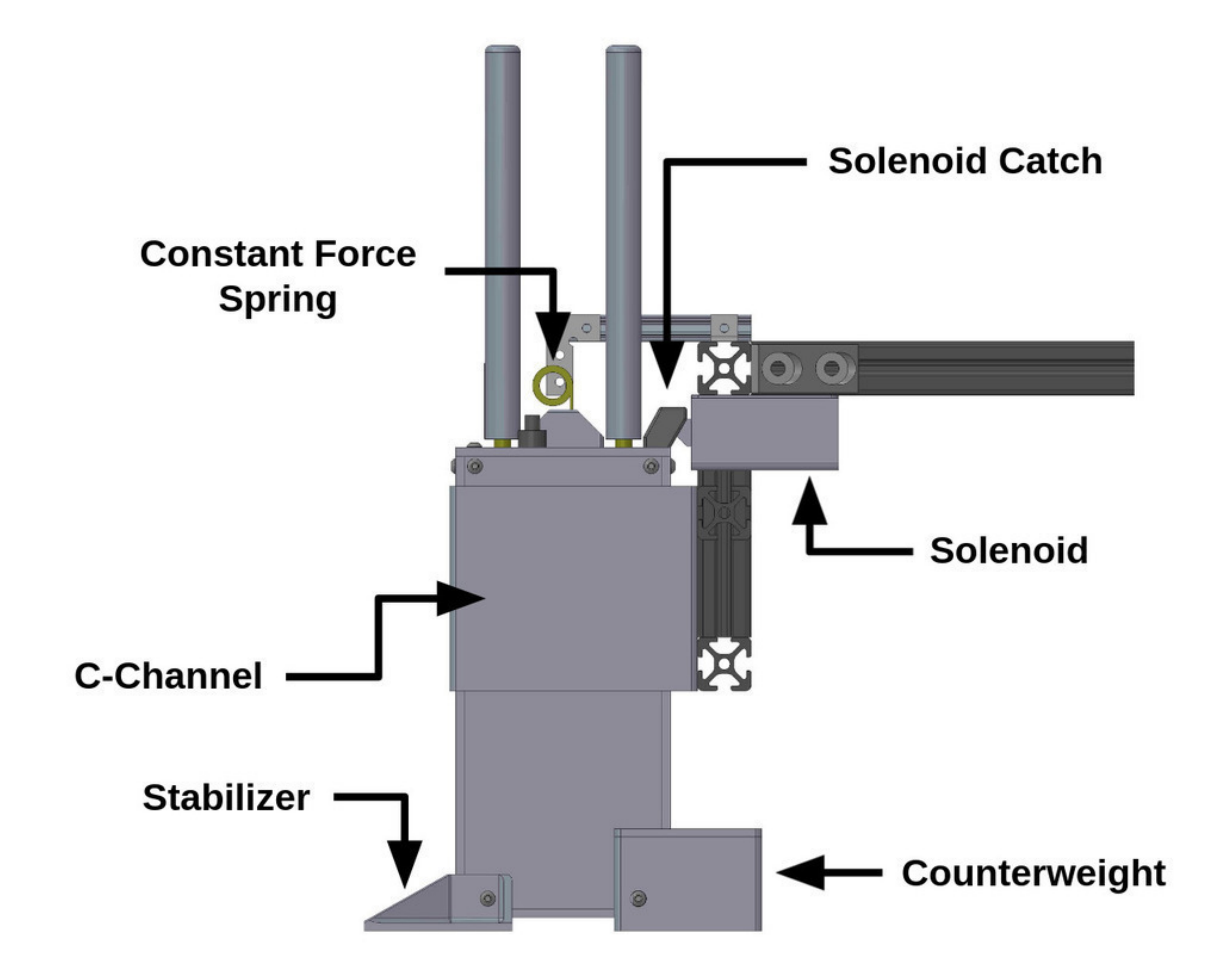}\label{fig:deployment_mech_cross_section} }}\\
    \caption{On the left (a) a perspective view of the beacon design as well as (b) a side view of the beacon attached to the deployment mechanism.}
		\label{fig:beacon_cad_drawings}
\end{figure}

Each beacon consists of two 3D printed nylon-carbon fiber infused internal brackets serving as structural support components. Figure \ref{fig:beacon_cad_drawings}a shows the top of each beacon that contains a charging port, power button and LED power status indicator between the antennas. The two stabilizers seen towards the front provide longitudinal stability for the beacon once deployed on the ground. These complement the steel counterweights on the rear of the beacon. The deployment mechanism shown in Figure \ref{fig:beacon_cad_drawings}b uses solenoids to hold the beacons in place. When the solenoid is released, the beacon falls at a consistent rate using a constant force spring.

\subsection{Platform Compute Design}
\label{ssec:sup_platform_compute_design}

When designing our system, we sought to balance the ease of development with a single, monolithic compute unit versus potential integration challenges of a distributed system. Early on, we decided that, given uncertain compute requirements, we should attempt to pack as much compute as possible into the Husky platforms. This decision had a wide array of collateral consequences, including power system requirements, cooling requirements, and mechanical considerations. Further, the decision to not utilize an industrial-style motherboard, but rather a consumer gaming motherboard, created integration difficulties that might have otherwise been avoided. For example, the power requirements of our Ryzen Threadripper-based system were roughly 425W at peak consumption of both CPU and GPUs, at the limit of potential commercially available DC/DC ATX power supplies. Using discrete GPUs mounted in PCI Express slots presented a mechanical challenge, particularly for shock and vibration mounting, which was discovered as a failure mode late in the design process. In contrast, the distributed architecture developed for our Spot platforms utilized significantly less power and space while delivering similar performance. Future system designs are more likely to follow a distributed approach, rather than a monolithic approach, to ease mechanical and electrical integration efforts at the expense of only minimal added software effort. 

At a deeper level of inter-robot module communications, we underestimated the challenges of differing ground potentials. After shorting out several serial links between components and having unreliable USB communication, we realized that several ground loops were responsible. By adding serial optocouplers and converting to Ethernet-based platform control, these ground loops were eliminated, resulting in highly reliable platforms. Future development would rely exclusively on differential signalling such as controller-area network or Ethernet for inter-module communication.

\subsection{Sensor Synchronization Design}
\label{ssec:sup_sensor_sync_design}

To effectively share sensor data between robots, sensor and system timing has to be considered. Our lidar solution could utilize IEEE 1588v2 timing (also known as the Precision Time Protocol v2 or PTP), but our wireless mesh networking solution could not support IEEE-1588v2. Therefore, we implement Network Time Protocol (NTP) between robots and PTP within the same robot. On startup, each robot attempts to synchronize with the Base Station using NTP over the mesh network. This synchronization step is critical for multi-robot operations and coordination to provide a consistent time basis across all nodes. In testing, the relative time drift (a few milliseconds) over the course of a run (1 hour) was not significant enough to cause problems. If the Base Station is unreachable (say for single-robot testing), the robot falls back to its own battery-backed realtime clock as a time source. In either case, after attempted synchronization, no further attempts are made to match times with any other robot or the Base Station. In testing, we observed that clock slews resulting from attempted time synchronizations as robots entered and left communications range had a negative impact on localization performance. 

In the Threadripper monolithic architecture used on the Husky platforms, PTP on the secondary Ethernet interface functioned perfectly, allowing the Threadripper to become the grandmaster and the lidar to follow along, as shown in \fig{ptp_diagram_l}. However, in the distributed Xavier+NUC architecture (shown in \fig{ptp_diagram_r}), designing an effective PTP interface encountered significant challenges. 

\twofer{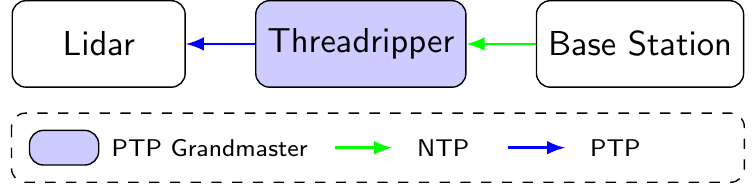}{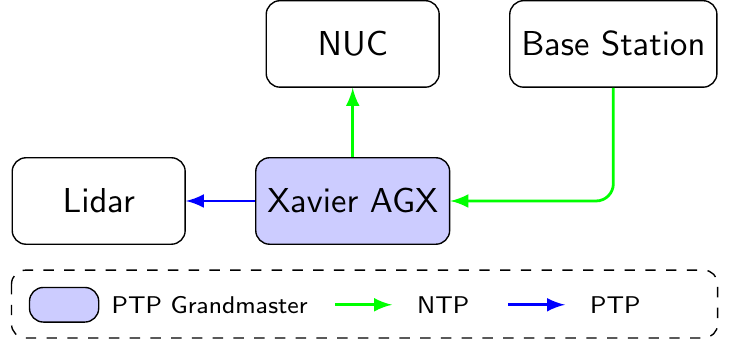}{ptp_diagram}{Block diagrams showing Precision Time Protocol (PTP) distribution between system components on the Threadripper (a) and Xavier+NUC (b).}{}{}

Fundamentally, PTP requires hardware support in order to function by performing sensitive timing operations as close to transport medium as possible. The network hardware options on the Spot included a Realtek r8125, an Nvidia platform SoC module, and a quad-port Intel i210 card. Realtek r8125 support for PTP was not functional, as verified by the \textit{phc\_ctl} utility; this relatively new chipset relies on an out-of-tree kernel driver for Linux at the time of our development. The NVidia platform module appeared to support PTP when interrogated by \textit{phc\_ctl}, but on further investigation through network traffic inspection, the platform module was not inserting the correct information into outbound Ethernet traffic. Our final solution is based on using a spare port on the Intel i210 card, which had robust, verifiable PTP support. As the NUC lacks a port with viable PTP support, we fall back on NTP as a synchronization method, relying on the Xavier as the robot's grandmaster time source.

As an aid to the community, we offer the following verification steps to assist in debugging PTP issues. First, verify that there is hardware support via \textit{ethtool -T $<$iface$>$} to verify kernel-level hardware PTP support. Second, use \textit{phc\_ctl $<$iface$>$ cmp} to verify that the Ethernet hardware clock is synchronized to the Linux system clock. Finally, ensure that the hardware timestamps encoded in the PTP network traffic match the local system by capturing network traffic from a PTP-enabled grandmaster Ethernet interface. These steps can verify that the PTP software stack is broadcasting the system time via the Ethernet hardware to downstream consumers. 

\subsection{Large-Scale Localization Validation}
\label{ssec:sup_localization_validation} 

\begin{figure}[!htb]
		\centering
		\subfloat[]{{\includegraphics[height=0.475\textwidth]{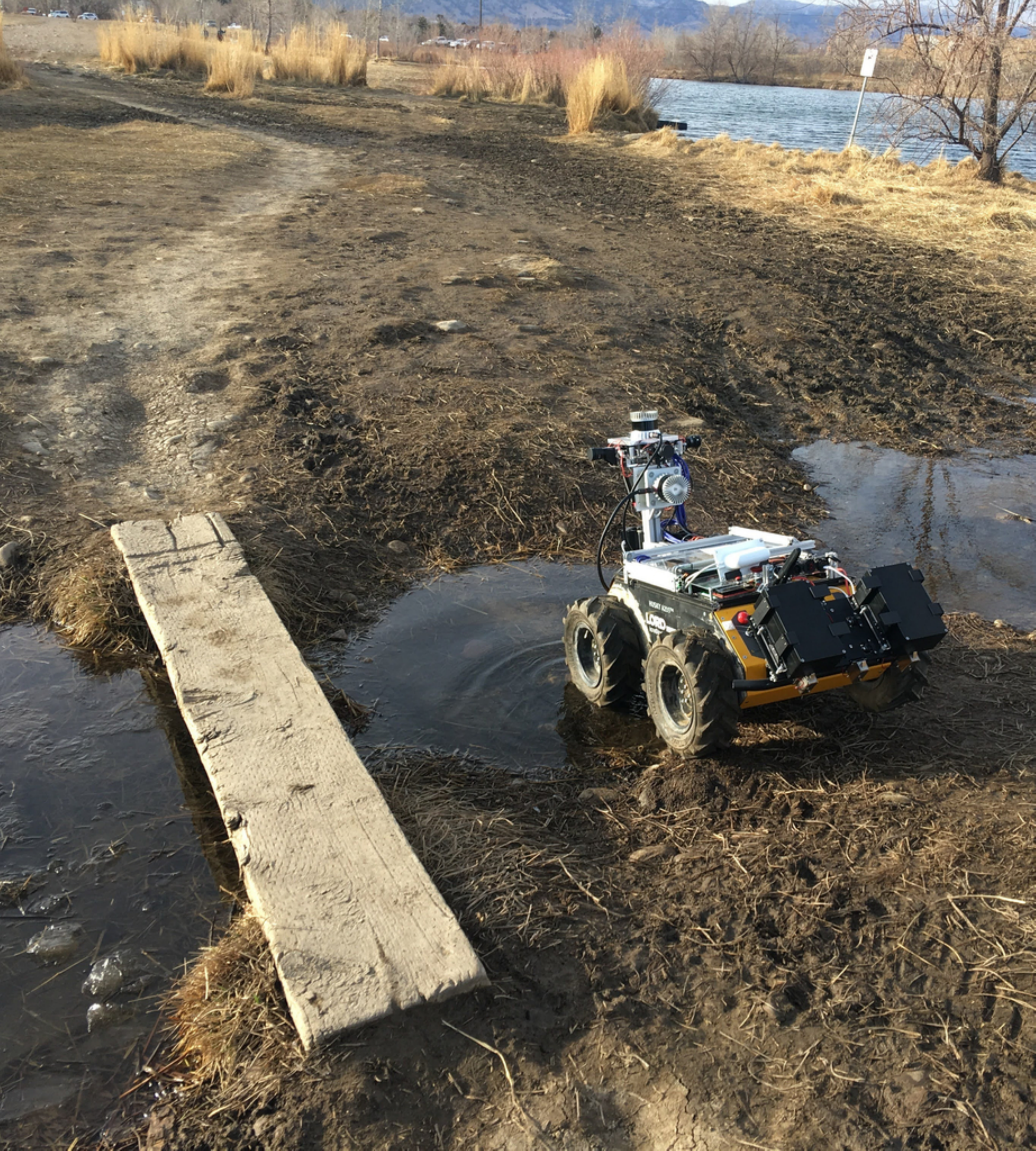}\label{fig:cu_south_photo} }}
		\subfloat[]{{\includegraphics[height=0.475\textwidth]{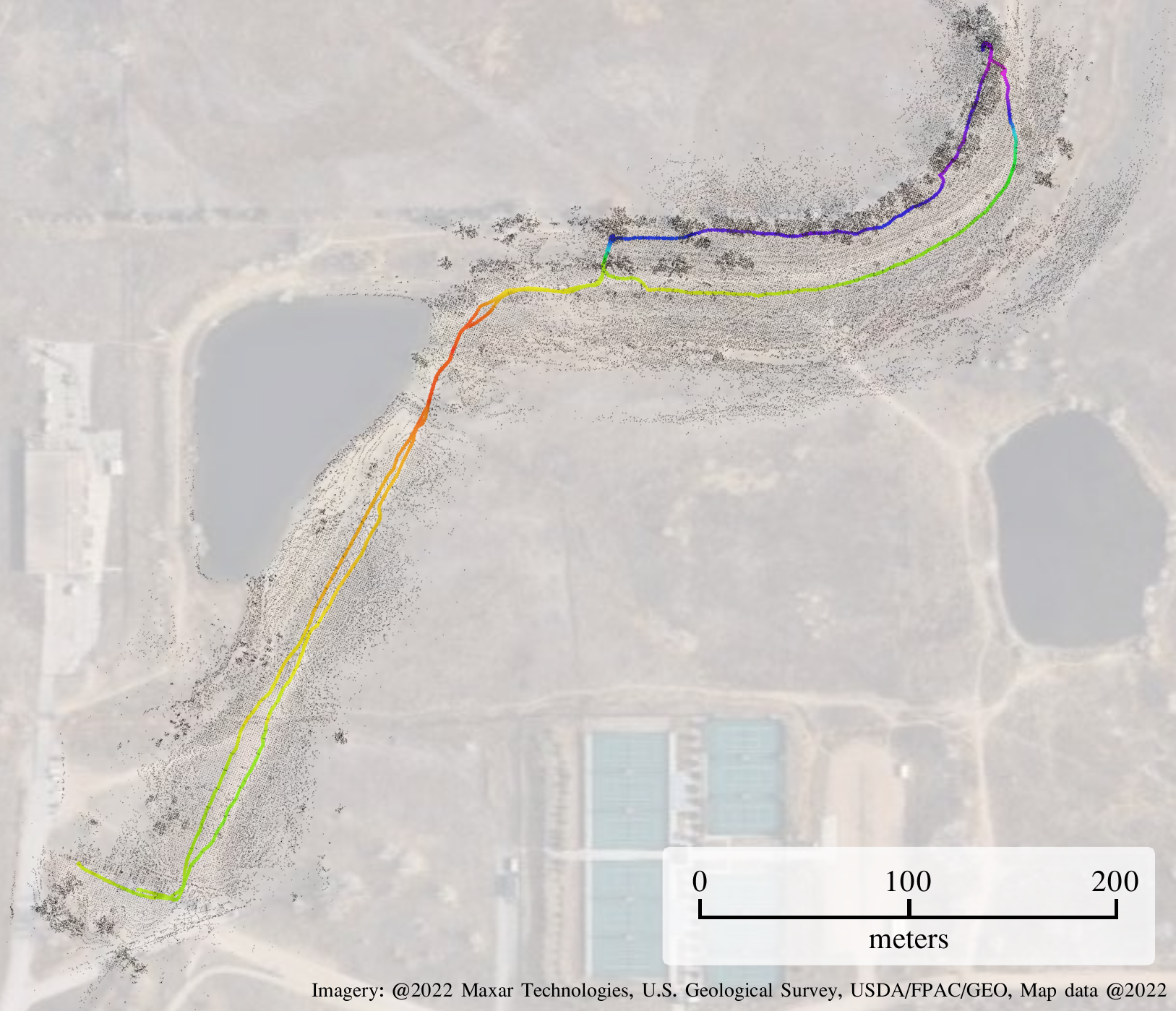}\label{fig:cu_south_map} }}\\
    \caption{On the left (a) photo taken at CU South Campus of a Husky robot during a large-scale, long-duration localization test. On the right (b), is an LIO-SAM point cloud map, denoted by small black dots, and LIO-SAM robot trajectory denoted by larger dots colored by elevation, overlaid with Google Maps satellite imagery of the area. The Husky was manually controlled, beginning in the CU South parking lot, continuing along a dirt path and up a hill, looping back, and ending back at the parking lot.}
		\label{fig:cu_south_test}
\end{figure}

\begin{figure*}[!htb]
    \centering
    \includegraphics[width=0.99\textwidth]{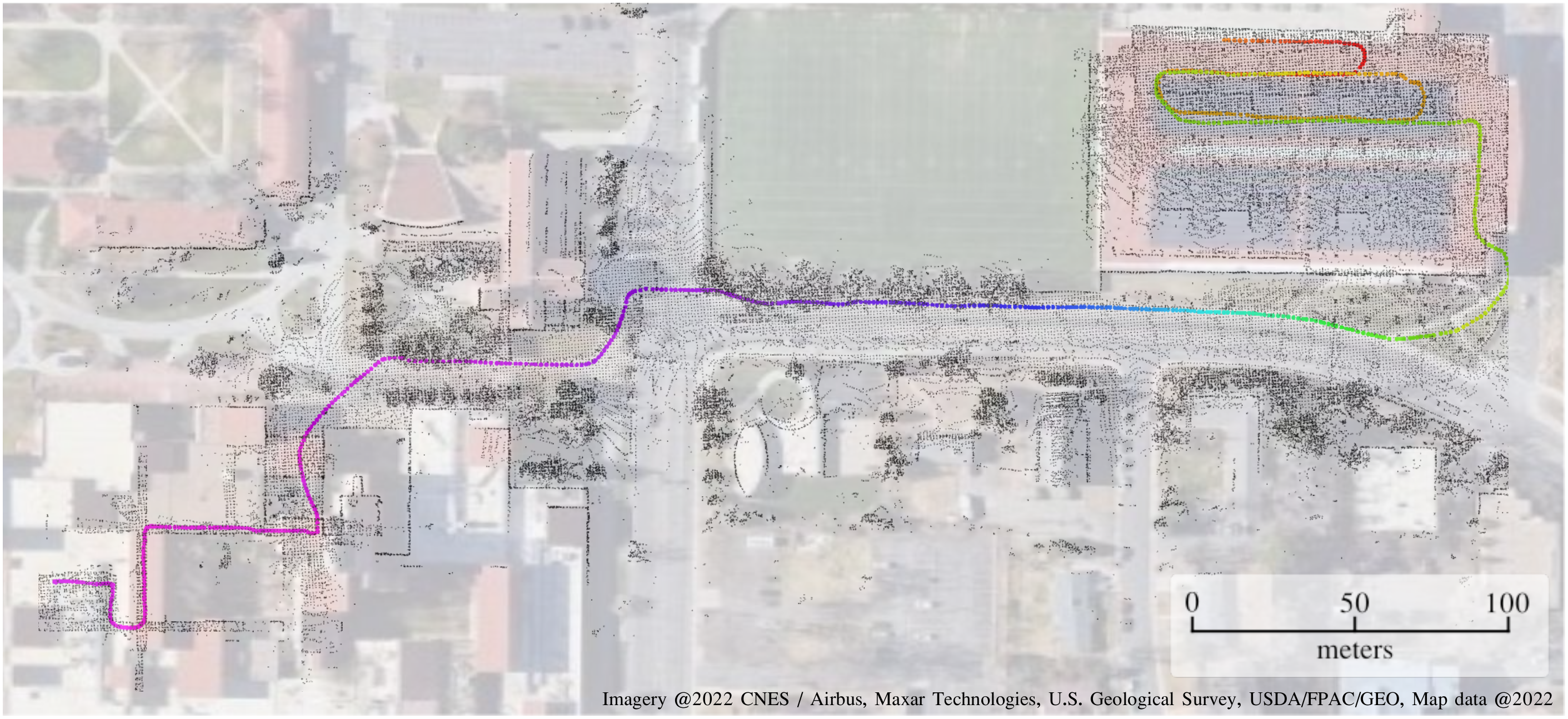}
    \caption{Overlay of LIO-SAM point cloud map, denoted by small black dots, and LIO-SAM robot trajectory denoted by larger dots colored by elevation, with Google Maps during a large-scale, long-duration localization test at CU Main Campus. This test was conducted on a Spot robot, that was manually controlled, beginning in the CU Engineering Center building on the bottom left corner of the map, continuing across campus, ending at the bottom of the three-story underground Folsom Parking Garage on the top right corner of the map.}
    \label{fig:folsom_liosam_google_map}
\end{figure*}

Validation testing of LIO-SAM onboard Spot and Husky platforms was imperative for ensuring sufficient accuracy, speed, and stability for long-duration and large-scale missions. Some examples include an outdoor test at CU South Campus, as shown in Figure \ref{fig:cu_south_test}, and as well as test that begins at the CU Engineering Center building and treks across campus to the bottom of a three-story underground parking garage, as shown in Figure \ref{fig:folsom_liosam_google_map}.

\subsection{Common Reference Frame Alignment Optimization}
\label{ssec:sup_ref_frame_optimization}

To further reduce the yaw error of the common reference frame alignment, the lateral spacing of the robot prisms was increased.

In our initial testing, we found that the roughly 120mm plates attached to the robots along with the prisms 1.5mm centering error lead to consistent variance in the resulting yaw of approximately 0.7$^{\circ}$. This aligned with a calculated value of $arcsin(1.5/120)=0.72^{\circ}$. In order to reduce the impact of this error, a mechanical bar holding two of the prisms at a distance of 655mm, is added to each robot. The resulting angle error after adding the bar was $arcsin(1.5/655)=0.13^{\circ}$. This was consistent with external testing. For a rough comparison see Figure \ref{f:prism_tests_l}, where a test was conducted using a mock robot plate and the prism separating bar. The difference from the average of each test shows a higher precision for the tests conducted with the prism bar in place. An example setup for these test is shown in Figure \ref{f:prism_tests_r}.

\twoferheight{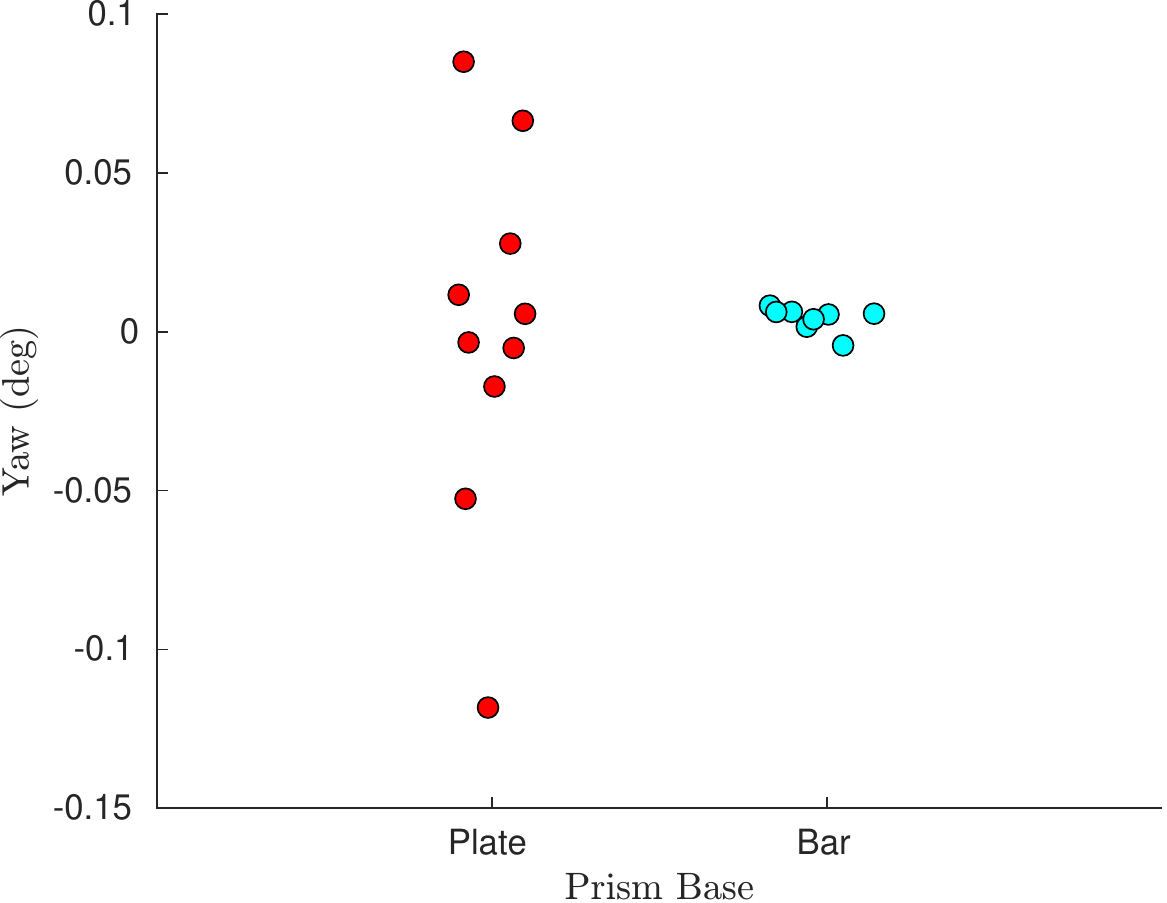}{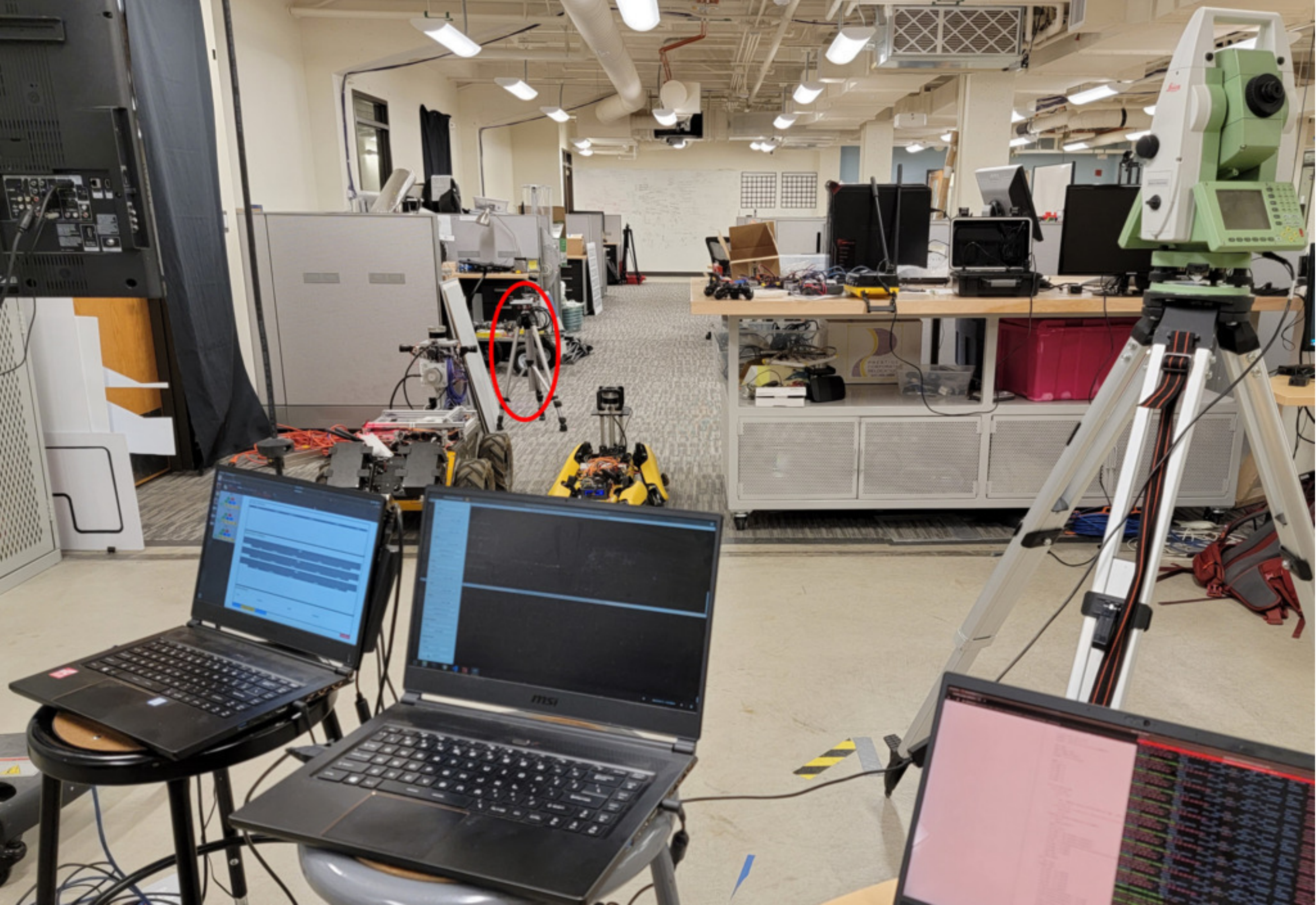}{prism_tests}{A comparison of transforms generated by LTS using the standard robot sensor plate, and after attaching prisms to a bar placed on the plate instead (a). The difference from the average, and precision of the bar is higher than without the bar. Setup for a prism test with a mock gate highlighted in red (b). The mock gate was designed to have the same dimension as a robot sensor plate}{}{}{2.2in}

\subsection{Artifact Detection Training Procedure}
\label{ssec:sup_artifact_detection_training_procedure}

A systematic procedure targeted at low-light conditions is used to train the model. At each location, data was collected using three different brightness levels to minimize the impact of lighting conditions on the model's performance. Specifically, images were taken from past circuit events as well as separate field exercises. Remote data collection sessions took place inside the dark and rocky Edgar Experimental Mine in Idaho Springs, Colorado. Local data collection took place on University of Colorado Boulder campus, primarily within the outdoor courtyard of our Engineering Center building, and during evening hours when there was no natural illumination. The data was collected with three onboard illumination levels: 0\%, 50\%, and 100\%. The cameras, FLIR Blackfly PGE-05S2C-CS GigEVision cameras, were mounted in cardinal directions on the robots as shown in Figure \ref{fig:sensor_head}.

Photos in which the artifacts suffered excessive motion blur and occlusions, determined by the ability of the human reviewer to detect the artifact, were removed from the data set. After the AD pipeline was trained on this initial data collection effort, we found that it did not generalize well to new environments. Therefore, we augmented the dataset with additional imagery collected from a greater diversity of backgrounds, including a nearby university loading dock. The datasets used for training are summarized in the Appendix. The dataset was later augmented with data from areas where false positives were frequently identified in order to reduce the identification of these false positives.

\subsection{BOBCAT Components}
\label{sec:sup_bobcat}

\textit{BOBCAT} calculates \textit{Objective} weights and \textit{Behavior} scores to select an \textit{Execution Behavior} whenever one or more \textit{Monitor} outputs or \textit{Objective} input weights change.  \textit{Behavior} execution functions may be either blocking or non-blocking.  They should be non-blocking to the maximum extent possible, to increase reactivity and allow \textit{Behaviors} to change at any time.  \textit{Behaviors} that need to block may be required for certain actions that must be completed before the robot can do something different.  If a \textit{Behavior} is blocking, \textit{BOBCAT} will delay evaluation until the actions have completed. Tables \ref{tab:monitor_table}, \ref{tab:objective_table}, and \ref{tab:behavior_table} provide an exhaustive list of the \textit{Monitors}, \textit{Objective}, and \textit{Behaviors}, respectively.

\begin{table}[!htb]
\begin{center}
\begingroup
\renewcommand{\arraystretch}{1.75}
    \begin{tabular}{l l}
    \hline
    \textbf{Monitor}    &   \textbf{Criteria} \\
    \hline
    ExploreToGoal       & \makecell[l]{Received command to explore to a specific goalpoint, either submitted by the \\ human supervisor or generated by a node other than the global planner.} \\
    \makecell[l]{iExplore \\ iGoToGoal \\iStop \\ iDeployBeacon \\ iGoHome}       & \makecell[l]{The associated input command has been sent by the human supervisor from the GUI  \\ or joystick to execute a specific behavior.} \\
    NearbyRobot      & \makecell[l]{This and any other robot's paths are within 2m of each other.} \\
    Beacon           & \makecell[l]{A communications beacon is available and other criteria has been met to deploy it.} \\
    ReverseDrop      & \makecell[l]{A communications beacon is available and communications have been lost with the \\ base station for 10 seconds.} \\
    Comms            & \makecell[l]{Any message has been received from the base station in the last 3 seconds.} \\
    Artifact         & \makecell[l]{There are at least 3 unreported artifacts, or it has been at least 5 minutes since the \\ first unreported artifact was detected.} \\
    \hline
    \end{tabular}
\endgroup
\caption{MARBLE \textit{Monitors}, with a description of robot state requirements for an output of 1.}
\label{tab:monitor_table}
\end{center}
\end{table}

\begin{table*}[htbp]
\begingroup
\renewcommand{\arraystretch}{1.75}
\centering
\begin{tabular}{>{\raggedright\arraybackslash}m{2.5cm} >{\raggedright\arraybackslash}m{5cm} >{\raggedright\arraybackslash}m{7.65cm}}
\hline
\multicolumn{1}{l}{\textbf{Objective}} & \multicolumn{1}{l}{\textbf{Evaluation Function}} & \multicolumn{1}{l}{\textbf{Description}}                                                        \\ \hline
FindArtifacts                           & $iwFindArtifacts$                         & Find, identify and localize artifacts.  Always active.               \\
Input                                   & $iwInput\ *\ $OR(Input \textit{Monitors})                        &  Allow supervisor to override autonomy when necessary.                           \\
BeSafe                                  & $iwBeSafe * NearbyRobot$                                                      &  Safety of the robot, particularly collision with other robots. \\
ExtendComms                             & $iwExtendComms * (Beacon\ ||\ ReverseDrop)$                    & Extend communications as far into the environment as possible to reduce any delay in reports and minimize robot travel back and forth. \\
MaintainComms                           & $iwMaintainComms\ *\ !Comms$                              & Communicate with the base station, either by staying in or returning to communications.                                \\
ReportArtifacts                         & $iwReportArtifacts * Artifact$                                                      & Report artifact types and locations to base station. \\ \hline
\end{tabular}
\endgroup
\caption{MARBLE \textit{Objectives}.  Evaluation functions calculate the weight for each objective.  Components prefixed by ``$iw$" represent the input weight of the objective, while monitor names represent the binary output of the monitor.}
\label{tab:objective_table}
\end{table*}

\begin{table*}[htbp]
\begingroup
\renewcommand{\arraystretch}{1.75}
\centering
\begin{tabular}{>{\raggedright\arraybackslash}m{2cm}>{\raggedright\arraybackslash}m{6.5cm}>{\raggedright\arraybackslash}m{6.5cm}}
\hline
\multicolumn{1}{l}{\textbf{Behavior}} & \multicolumn{1}{l}{\textbf{Evaluation Function}} & \multicolumn{1}{l}{\textbf{Actions}}                                                        \\ \hline
Explore                                 & $owFindArtifacts\ *\ !ExploreToGoal + owInput * iExplore$                         & Request global planner to plan to unexplored areas.                                       \\
GoToGoal                                & $owFindArtifacts * ExploreToGoal + owInput * iGoToGoal$                        & Request global planner to plan a path to a specific goalpoint.                           \\
Stop                                    & $owBeSafe + owInput * iStop$                                                      & Controller stops using plan generated by global planner.  Causes robot to stop autonomous movement, but will still accept manual movement by human supervisor. \\
DeployBeacon                            & $owExtendComms\ *\ !ReverseDrop + owInput * iDeployBeacon$                    & Initiate beacon deployment maneuver, which positions robot, stops, and deploys a beacon. \\
GoHome                                  & $owReportArtifacts + owMaintainComms + owExtendComms * ReverseDrop + owInput * iGoHome$                              & Request global planner to plan a path to the starting point.                                \\ \hline
\end{tabular}
\endgroup
\caption{MARBLE \textit{Behaviors}.  Evaluation functions calculate the score for each behavior for use by the policy $\pi_B$.  Components prefixed by ``$ow$" represent the output weight of the objective, while monitor names represent the binary output of the monitor.}
\label{tab:behavior_table}
\end{table*}

\subsection{Artifact Detection Training Data}
\label{ssec:sup_artifact_detection_training_data}

Table \ref{tab:artifact_training_data} presents a summary of the training data Team MARBLE collected and annotated. Our main data collection effort was conducted in the Engineering Center Courtyard at University of Colorado Boulder, during late evening hours when there was no natural illumination. This location was chosen because we believed it be more representative of a subterranean environment, with the large amount of concrete and low illumination levels. Another smaller dataset was collected at the Edgar Mine in order to introduce data from tunnel-level environments. Real-world imagery from the Tunnel Event at the NIOSH Mine and the Urban Event at the Satsop Nuclear Power Plant was incorporated as well. To round out training, a final data set at the loading dock at the Engineering Center was added.

\begin{table}[!htb]
\begin{center}
    \begin{tabular}{c c c c c c c c c c c}
    \hline
    \raisebox{-0.3\totalheight}{\textbf{Dataset Name}} & \raisebox{-0.3\totalheight}{\textbf{FPS}} & \raisebox{-0.2\totalheight}{\textbf{Images}} & \raisebox{-0.5\totalheight}{\includegraphics[height=0.0325\textwidth]{figures/artifact_thumbnails/survivor}} & \raisebox{-0.5\totalheight}{\includegraphics[height=0.0325\textwidth]{figures/artifact_thumbnails/backpack}} & \raisebox{-0.5\totalheight}{\includegraphics[height=0.0325\textwidth]{figures/artifact_thumbnails/drill}} & \raisebox{-0.5\totalheight}{\includegraphics[height=0.0325\textwidth]{figures/artifact_thumbnails/fire_extinguisher}} & \raisebox{-0.5\totalheight}{\includegraphics[height=0.0325\textwidth]{figures/artifact_thumbnails/vent}} & \raisebox{-0.5\totalheight}{\includegraphics[height=0.0325\textwidth]{figures/artifact_thumbnails/rope}} & \raisebox{-0.5\totalheight}{\includegraphics[height=0.0325\textwidth]{figures/artifact_thumbnails/helmet}} & \raisebox{-0.35\totalheight}{\textbf{Labels}} \\
    \hline
    \textbf{Courtyard} & 10 & 54523 & 4837 & 8299 & 2682 & 3190 & 6219 & 3928 & 3654 & 32809 \\
    \textbf{Edgar Mine} & 10 & 3603 & 254 & 159 & 148 & 142 & 122 & 187 & 84 & 1096 \\
    \textbf{NIOSH Mine} & 1 & 13437 & 118 & 45 & 106 & 173 & 0 & 0 & 0 & 442 \\
    \textbf{Satsop Nuclear} & 15 & 33953 & 0 & 630 & 0 & 0 & 302 & 0 & 0 & 932 \\
    \textbf{Loading Dock} & 10 & 6923 & 184 & 234 & 143 & 143 & 76 & 329 & 116 & 1225 \\
    \hline
    \textbf{Total} &  & \textbf{112439} & \textbf{5393} & \textbf{9367} & \textbf{3079} & \textbf{3648} & \textbf{6719} & \textbf{4444} & \textbf{3854} & \textbf{36504} \\
    \hline
    \end{tabular}
    \caption{\label{tab:artifact_training_data} List of data sets utilized to train the artifact detection system. Also listed is the frames per second (FPS) of the imagery, the total number of images, the number of labels for each artifact, and the total number of artifact labels.}
\end{center}
\end{table}

\begin{figure*}[htbp]
    \centering
    \includegraphics[width=0.33\textwidth]{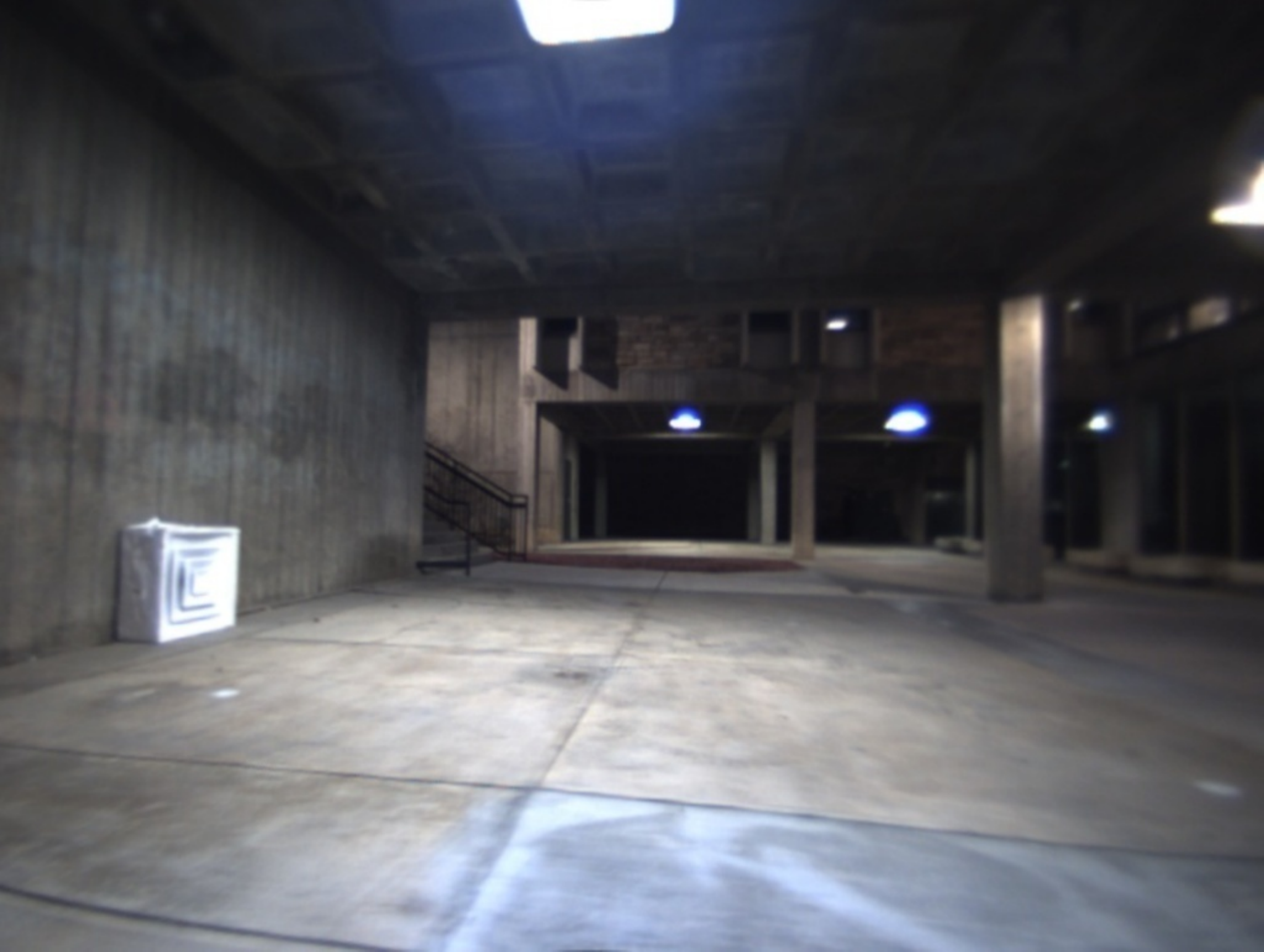}
    \caption{Team MARBLE trained on a vent with white-walled sides.}
    \label{fig:vent_box}
\end{figure*}

At the Final Event, our artifact detection system did not correctly detect any vents. It also falsely detected many white walls as vents. One reason for this poor performance is likely because the vent we trained on was different than the vent at the Final Event. To support the vent, white sides were added, making it appear as a box, as shown in Figure \ref{fig:vent_box}.

\subsection{Competition Results}
\label{ssec:sup_competition_results}

Table \ref{tab:list_of_team_scores} lists the eight teams that competed in the DARPA Subterranean Challenge Final Event, and the number of artifacts each team found during the 60-minute Prize Run, out of a maximum possible score of 40.

\begin{table}[!htb]
\begin{center}
    \begin{tabular}{c c c}
    \hline
    \textbf{Score} & \textbf{Team} & \textbf{Funding}\\ 
    \hline
    23  &   CERBERUS                &   DARPA   \\
    23  &   CSIRO Data61            &   DARPA   \\
    18  &   MARBLE                  &   DARPA   \\
    17  &   Explorer                &   DARPA   \\
    13  &   CoSTAR                  &   DARPA   \\
    7   &   CTU-CRAS-NORLAB         &   DARPA   \\
    2   &   Coordinated Robotics    &   Self    \\
    2   &   Robotika                &   Self    \\
    \hline
    \end{tabular}
    \caption{\label{tab:list_of_team_scores} List of final scores of all teams that participated in the 60-minute Final Event Prize Run.}
\end{center}
\end{table}

\subsection{Planning in a Dynamic Environment}
\label{ssec:sup_planner_dynamic_environment}

The planning algorithm presented in this paper has the capability of adapting to dynamic environments, i.e. closing or opening of doors, falling rubble, etc. Other situations also lead to dynamic changes in the map, including other nearly mobile agents, as well as localization and mapping error. The paper presents the planner response to a trap door, and here in the Appendix, additionally present examples of planner responses to robot-robot encounters as well as localization and mapping error.

\begin{figure}[hbt!]
		\centering
		\subfloat[]{{\includegraphics[width=.24\textwidth]{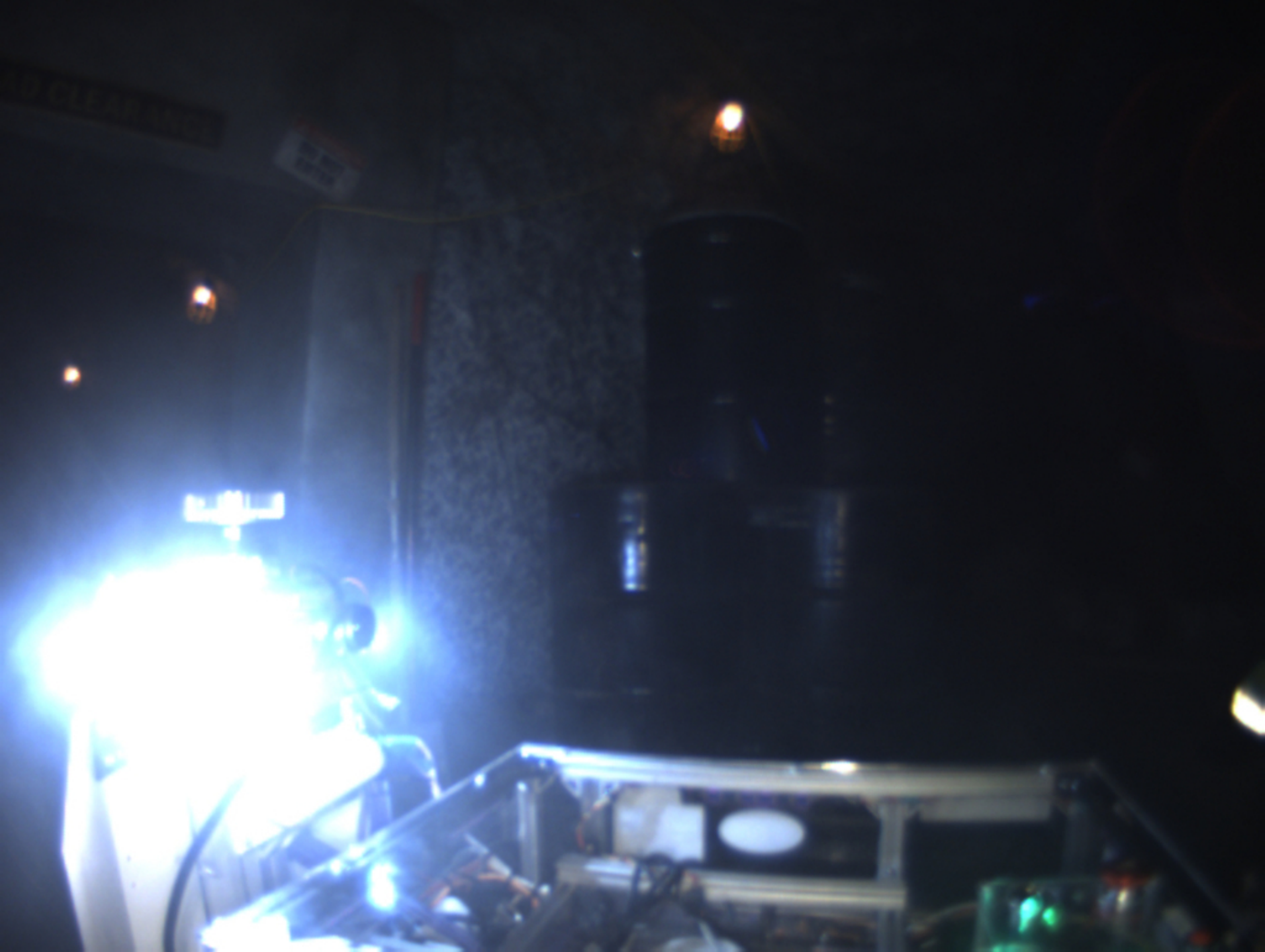}\label{fig:D02_0937s_H01} }}
		\subfloat[]{{\includegraphics[width=.24\textwidth]{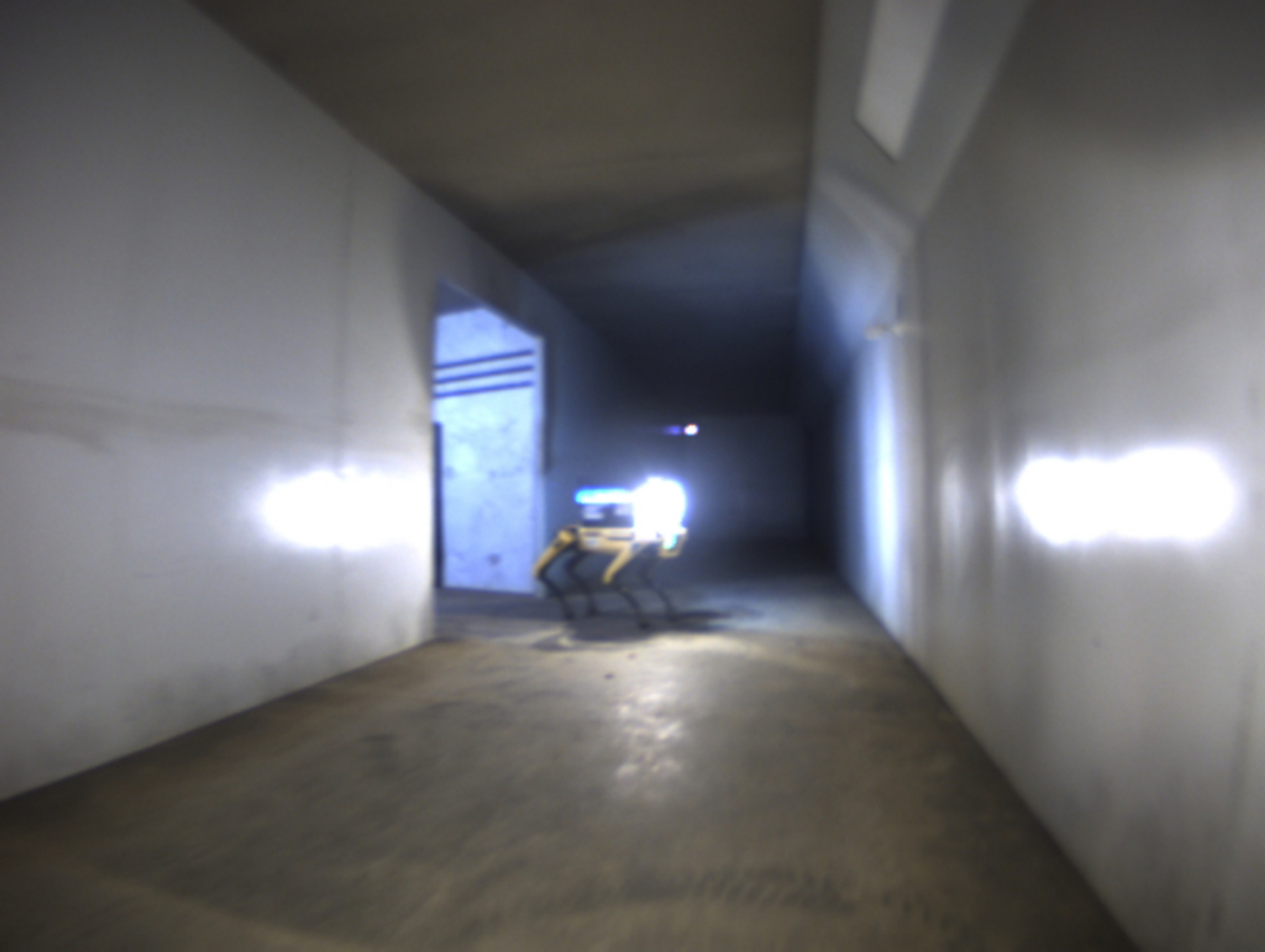}\label{fig:D02_1198s_D01} }} 
		\subfloat[]{{\includegraphics[width=.24\textwidth]{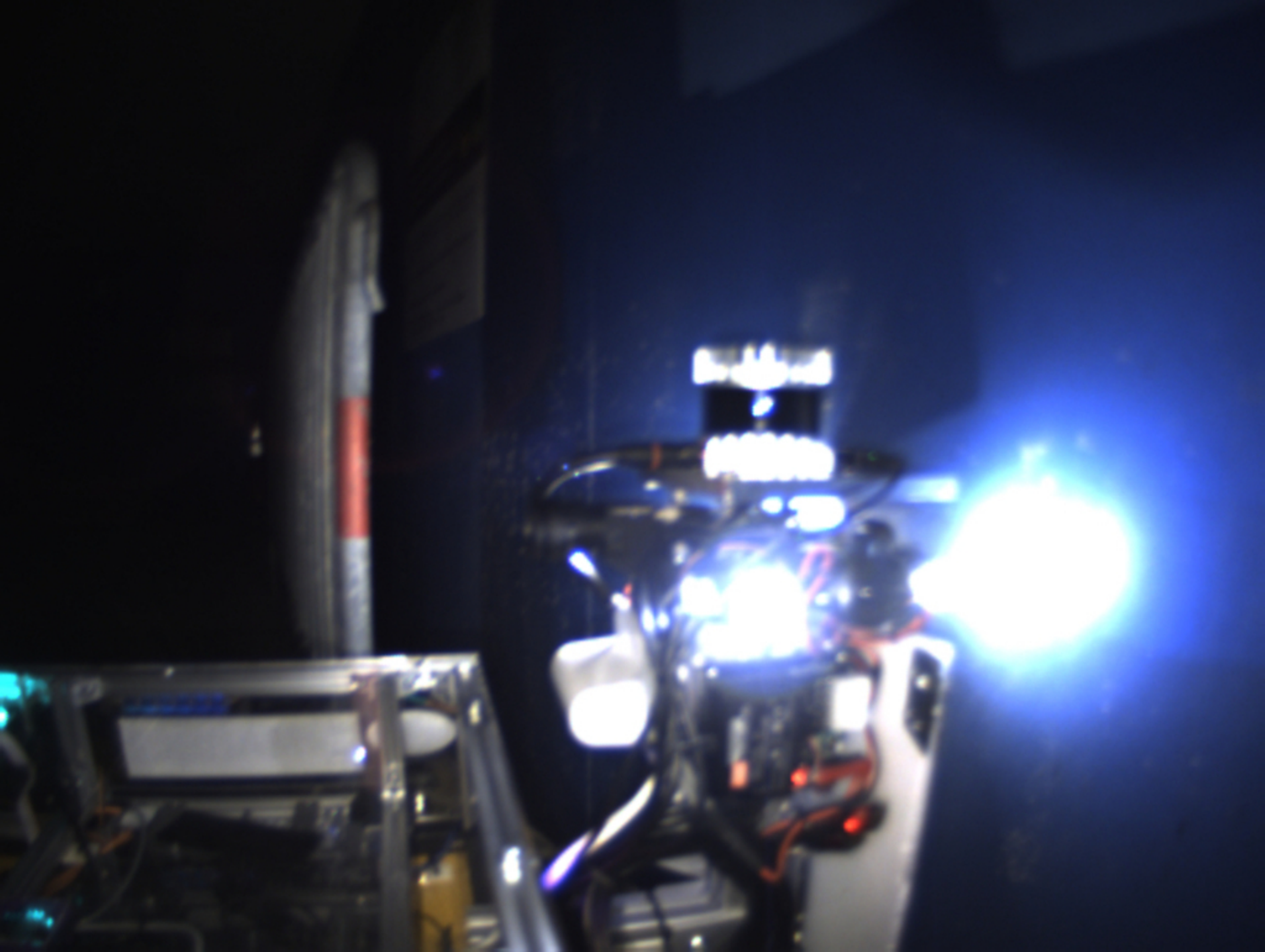}\label{fig:D02_1428s_H02} }} 
		\subfloat[]{{\includegraphics[width=.24\textwidth]{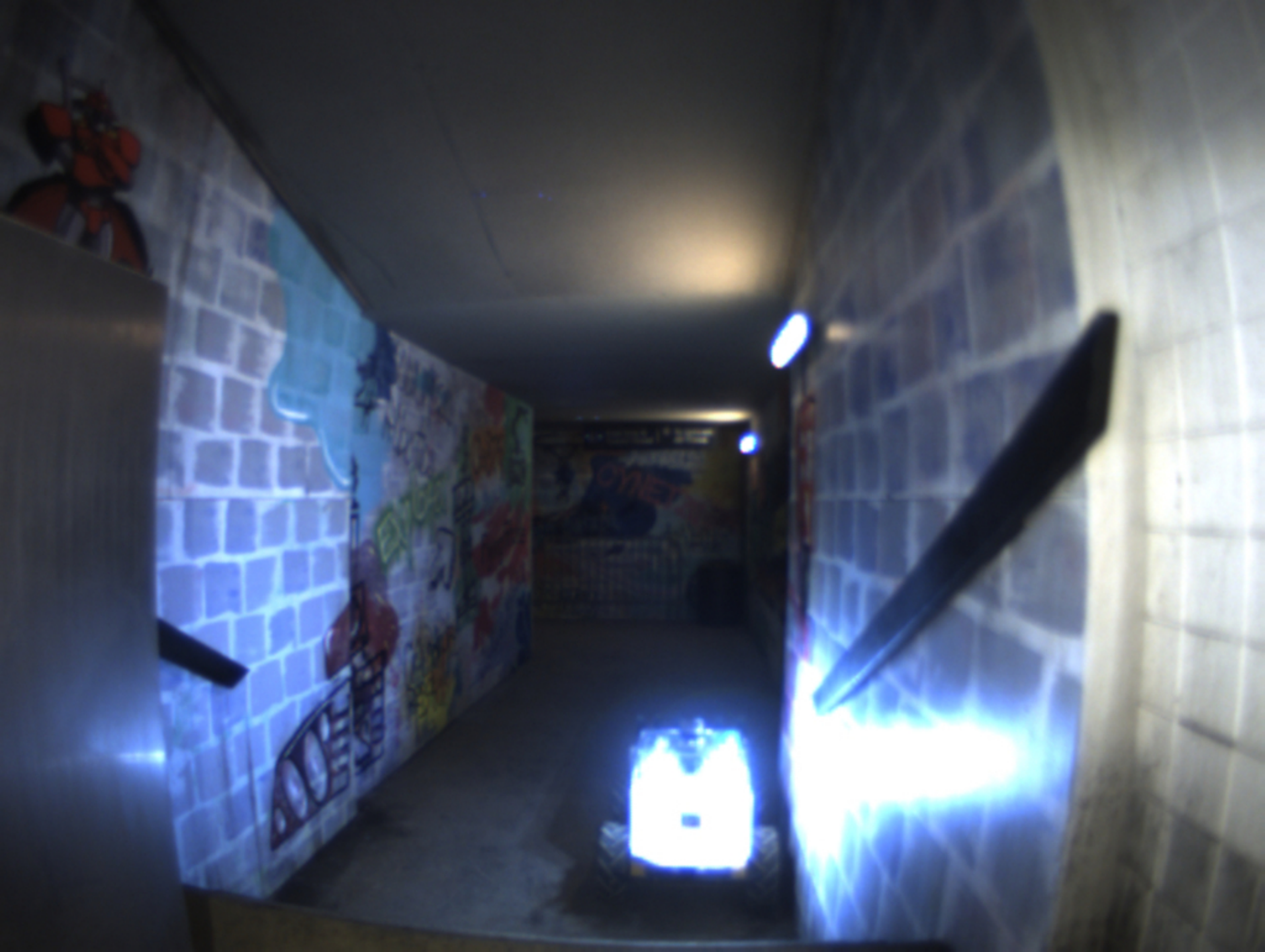}\label{fig:D02_2619s_H02} }}
		\caption{Robot-robot interactions}
		\label{fig:agent_deconfliction}
\end{figure}

Robots often came withing close proximity of other robots during the mission. Some examples of this are shown in Figure \ref{fig:agent_deconfliction}. Rather than resolve these interactions through the planner, which operates on a slower timescale that fast-approaching robots, the autonomous mission management system specifies the higher-priority agent continue while the lower-priority agent wait. The planner marks previously traversable edges as untraversable, and later updates them as traversable again once the other agents leaves the vicinity.

\begin{figure}[hbt!]
		\centering
		\subfloat[]{{\includegraphics[width=.45\textwidth]{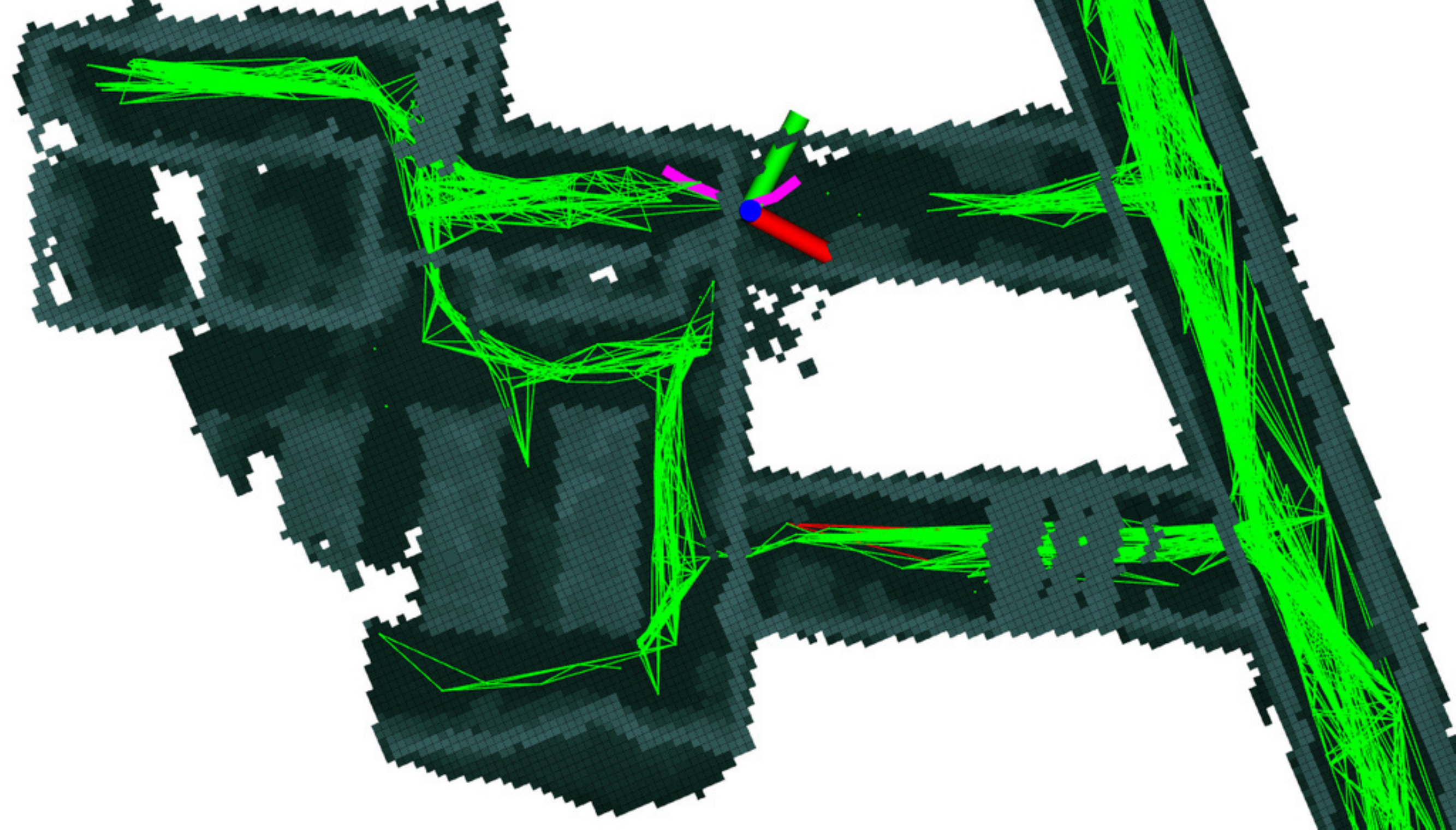}\label{fig:D01_0783s_beforedrift_cropped} }}
		\subfloat[]{{\includegraphics[width=.45\textwidth]{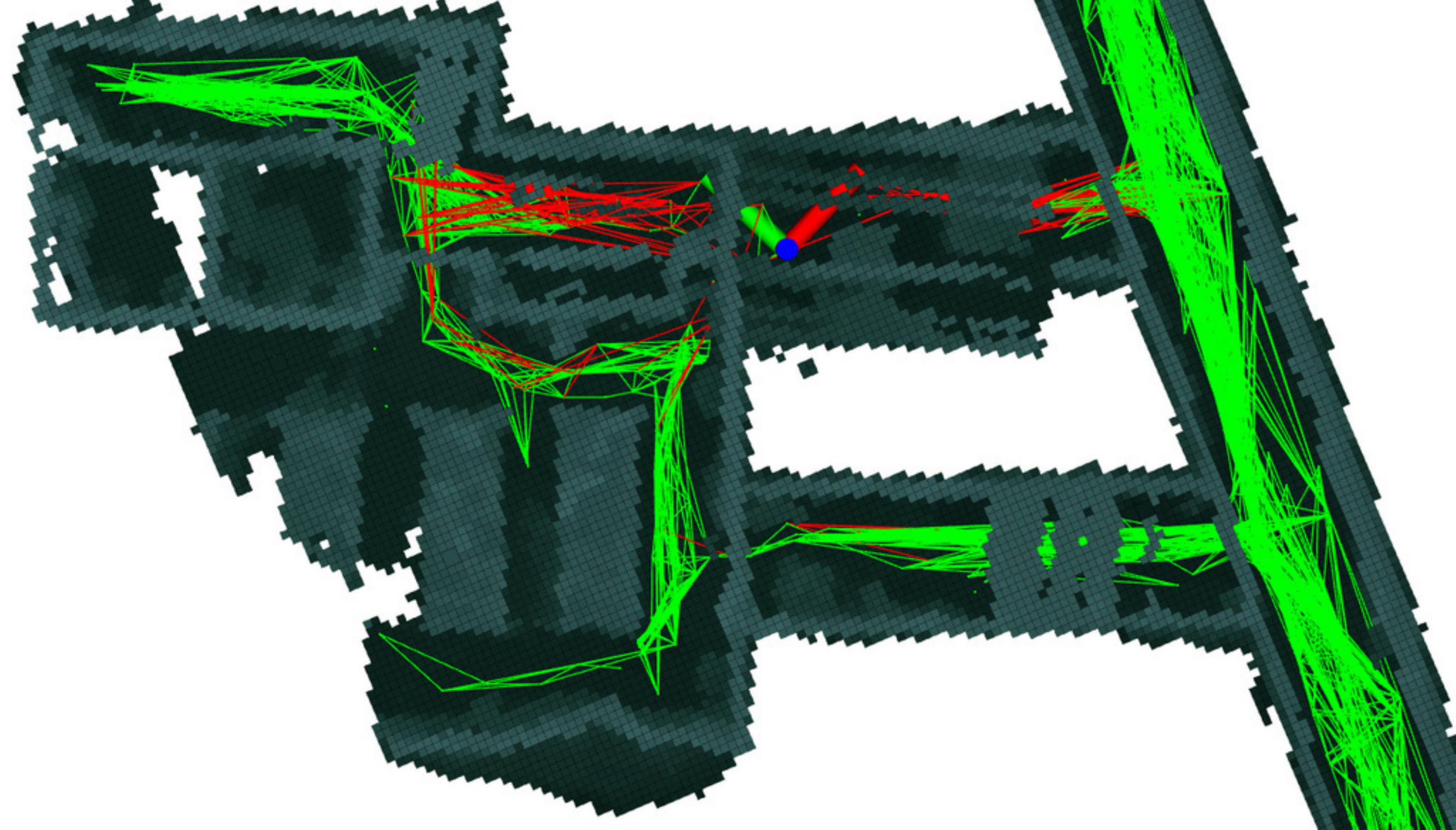}\label{fig:D01_0792s_afterdrift_cropped} }} \\
		\subfloat[]{{\includegraphics[width=.45\textwidth]{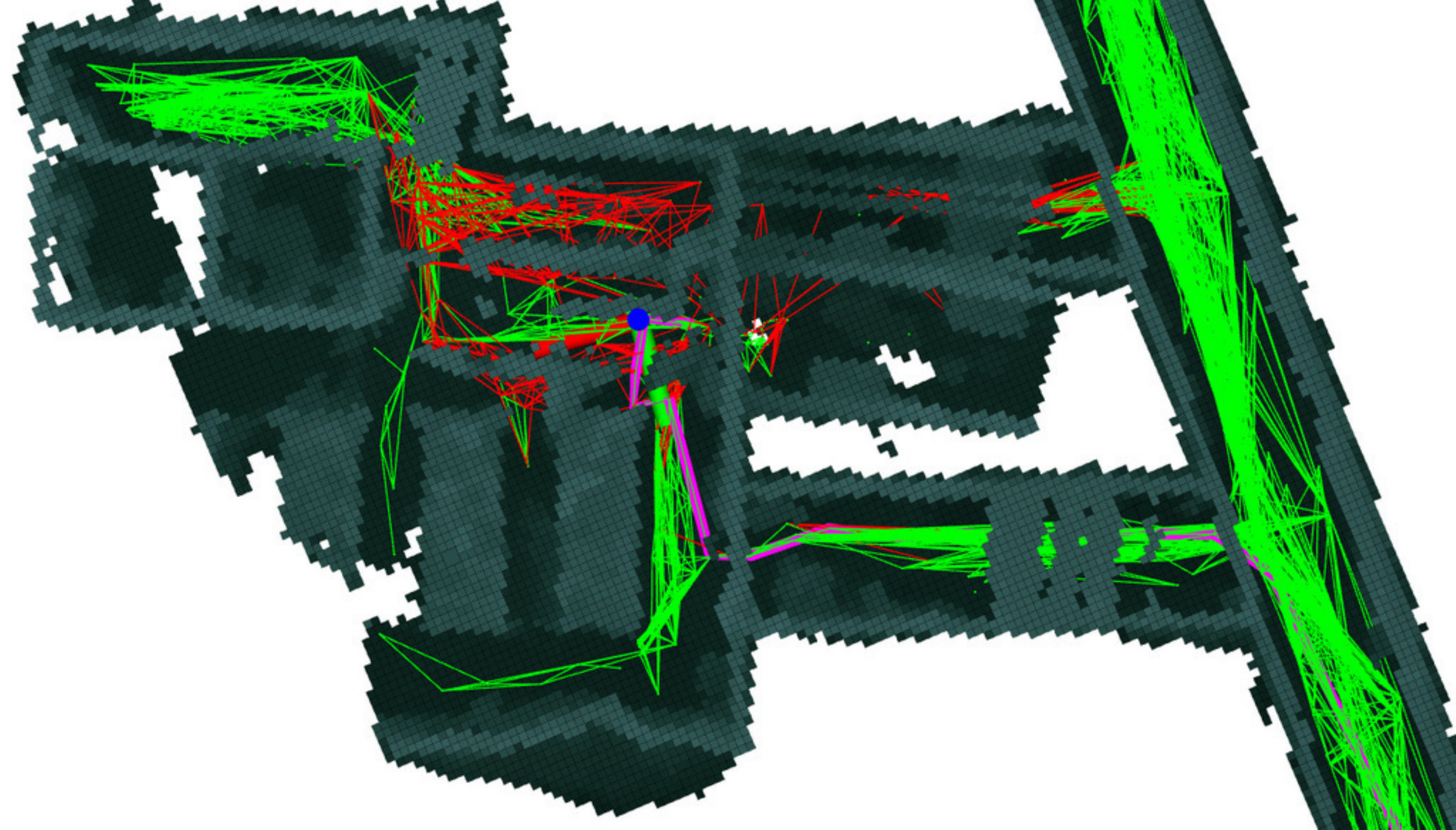}\label{fig:D01_0833s_afterdrift_cropped} }}
		\subfloat[]{{\includegraphics[width=.45\textwidth]{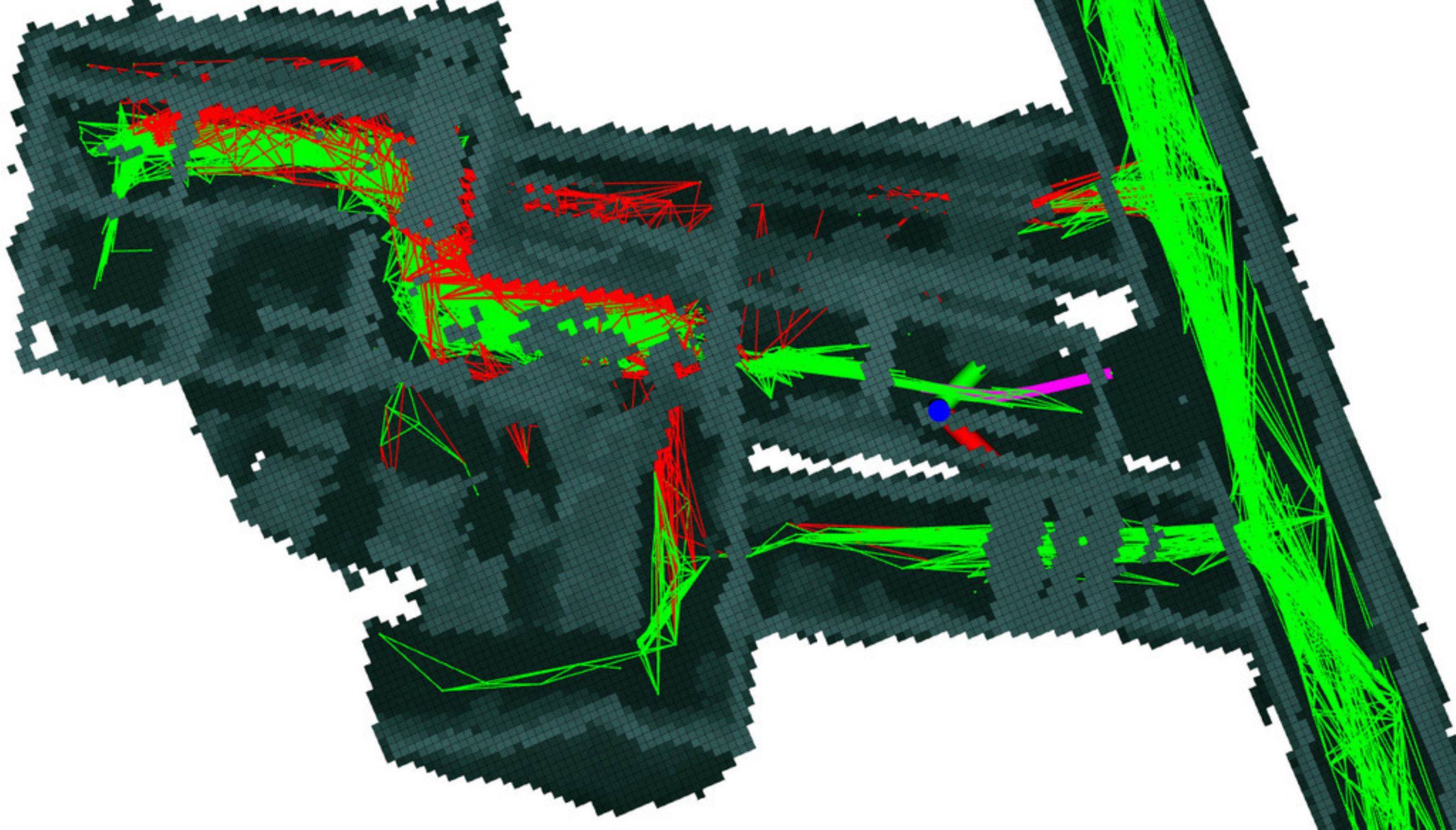}\label{fig:D01_1036s_afterdrift_cropped} }} \\
		\subfloat[]{{\includegraphics[width=.45\textwidth]{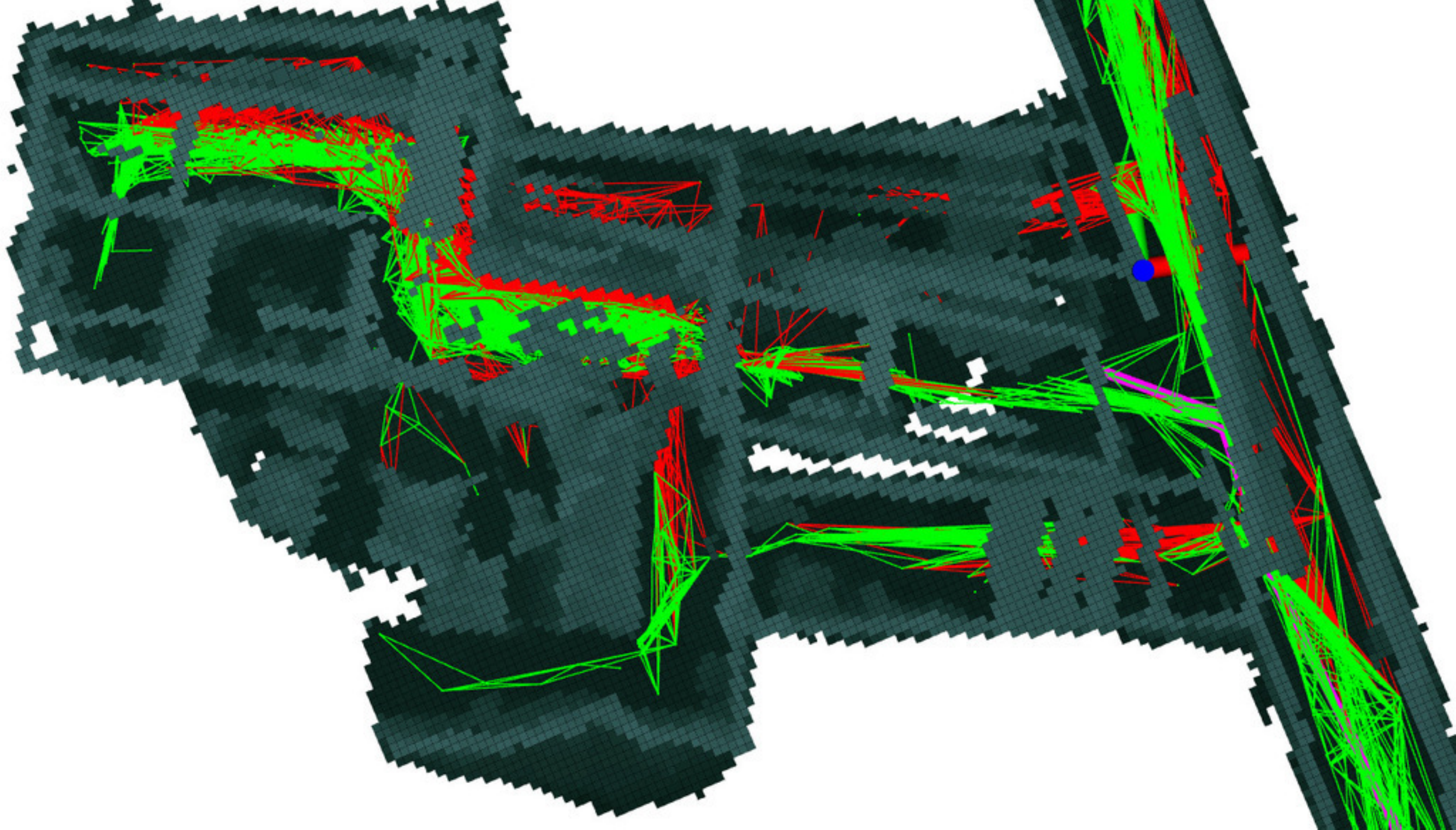}\label{fig:D01_1054s_afterdrift_cropped} }}
		\subfloat[]{{\includegraphics[width=.45\textwidth]{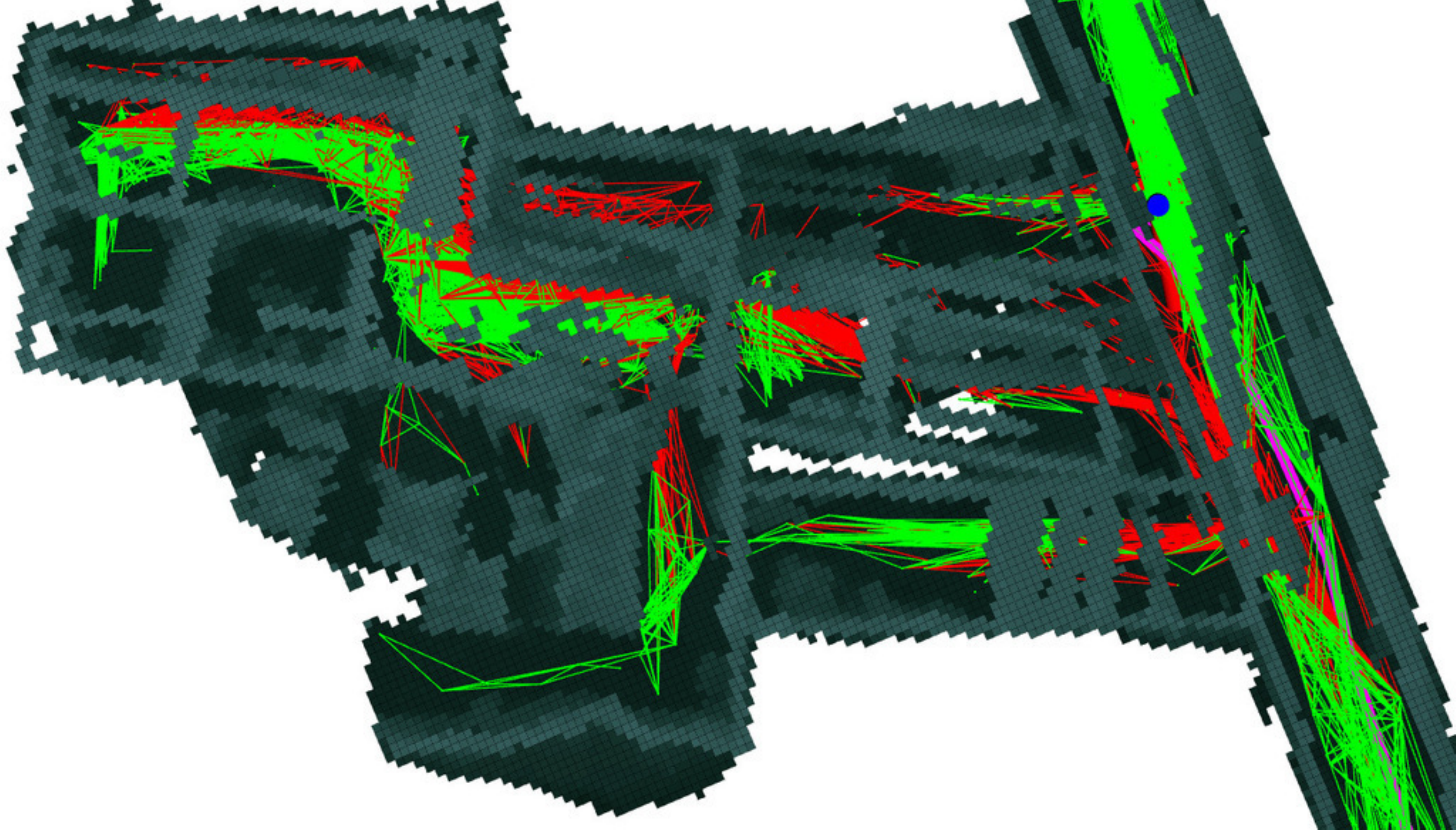}\label{fig:D01_1103s_afterdrift_cropped} }}
		\caption{Instance of D01 experiencing localization drift and erroneous mapping, causing planner to mark traversable (green) edges in the graph as no longer traversable (red). The progression of events is first (a) no localization error (13:03), (b) initial localization and mapping error (17:12), (c, d) continued localization and mapping error (13:53, 17:16), (e) loop closure correcting the robot pose (17:34), teleporting the agent away from its planned path (pink), and (f) continuing on the mission (18:23).}
		\label{fig:D01_localization_drift}
\end{figure}

Figure \ref{fig:D01_localization_drift} presents an instance when the planner on D01 adapted to localization and mapping drift. The erroneous new map data caused many of the previously traversable edges of the graph to change to untraversable. The planner adapted to the new scenario, helping the agent return to the main corridor, at which point a loop closure occurred, correcting the agent's pose. The planner operated continuously throughout this localization drift and loop closure correction.

\subsection{Planning in Constrained Spaces}
\label{ssec:sup_planner_constrained_spaces}

\begin{figure}[hbt!]
		\centering
		\subfloat[]{{\includegraphics[width=.40\textwidth]{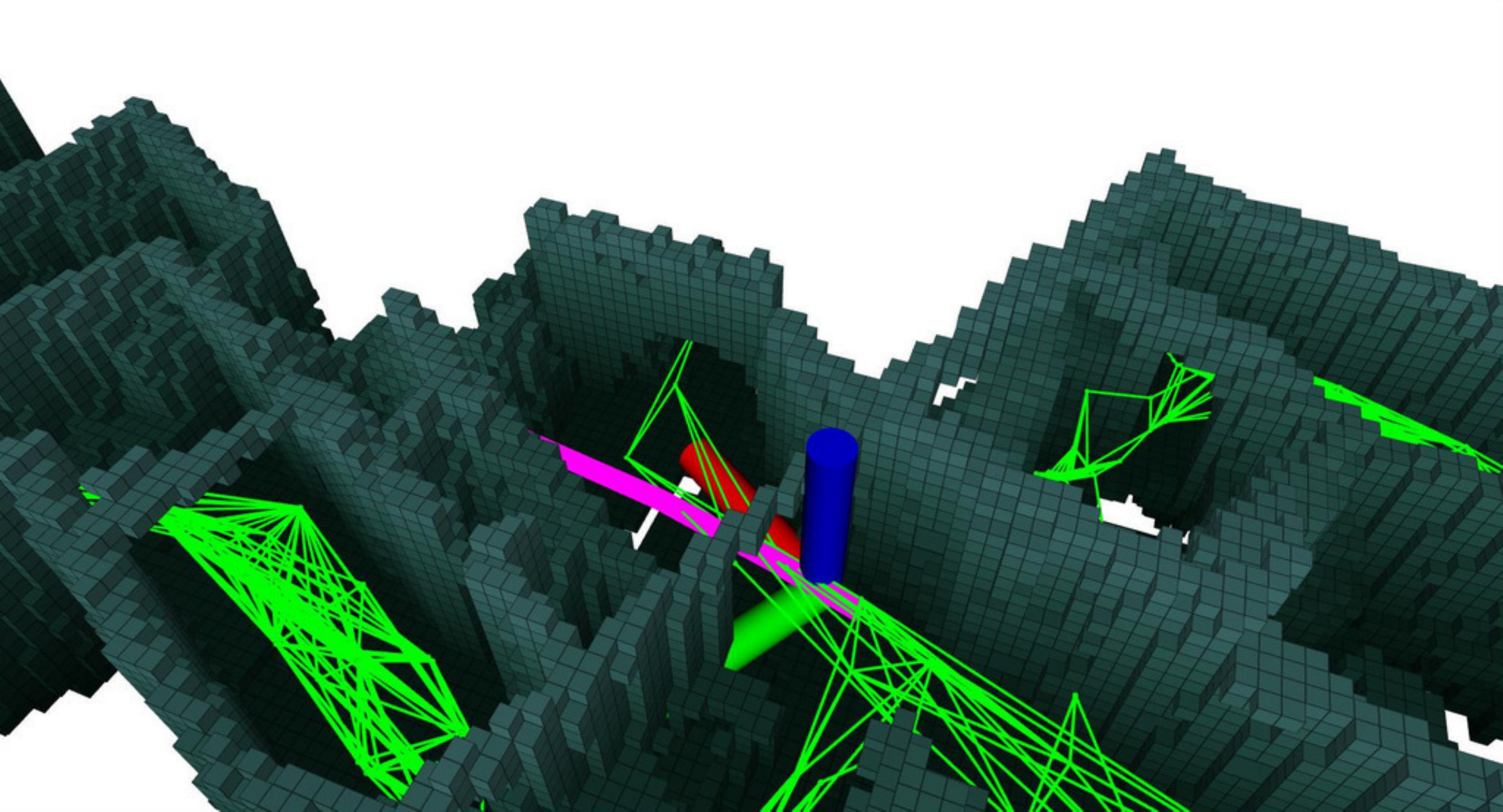}\label{fig:D02_3081s_short_corridor_map} }}
		\subfloat[]{{\includegraphics[width=.40\textwidth]{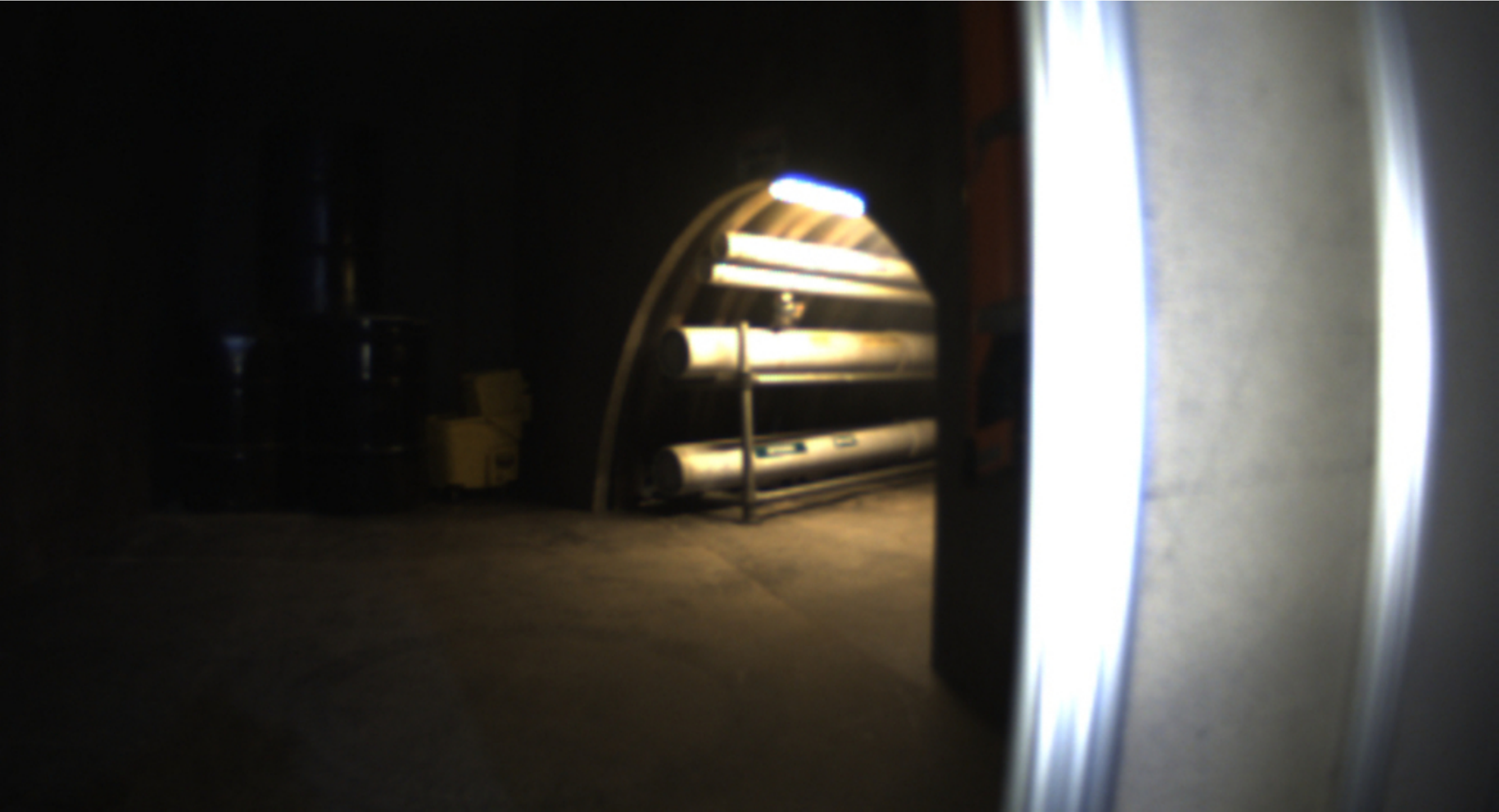}\label{fig:D02_3081s_short_corridor} }}
		\caption{Mobility limitations.}
		\label{fig:mobility_short_corridor}
\end{figure}

\begin{figure}[hbt!]
		\centering
		\subfloat[]{{\includegraphics[width=.45\textwidth]{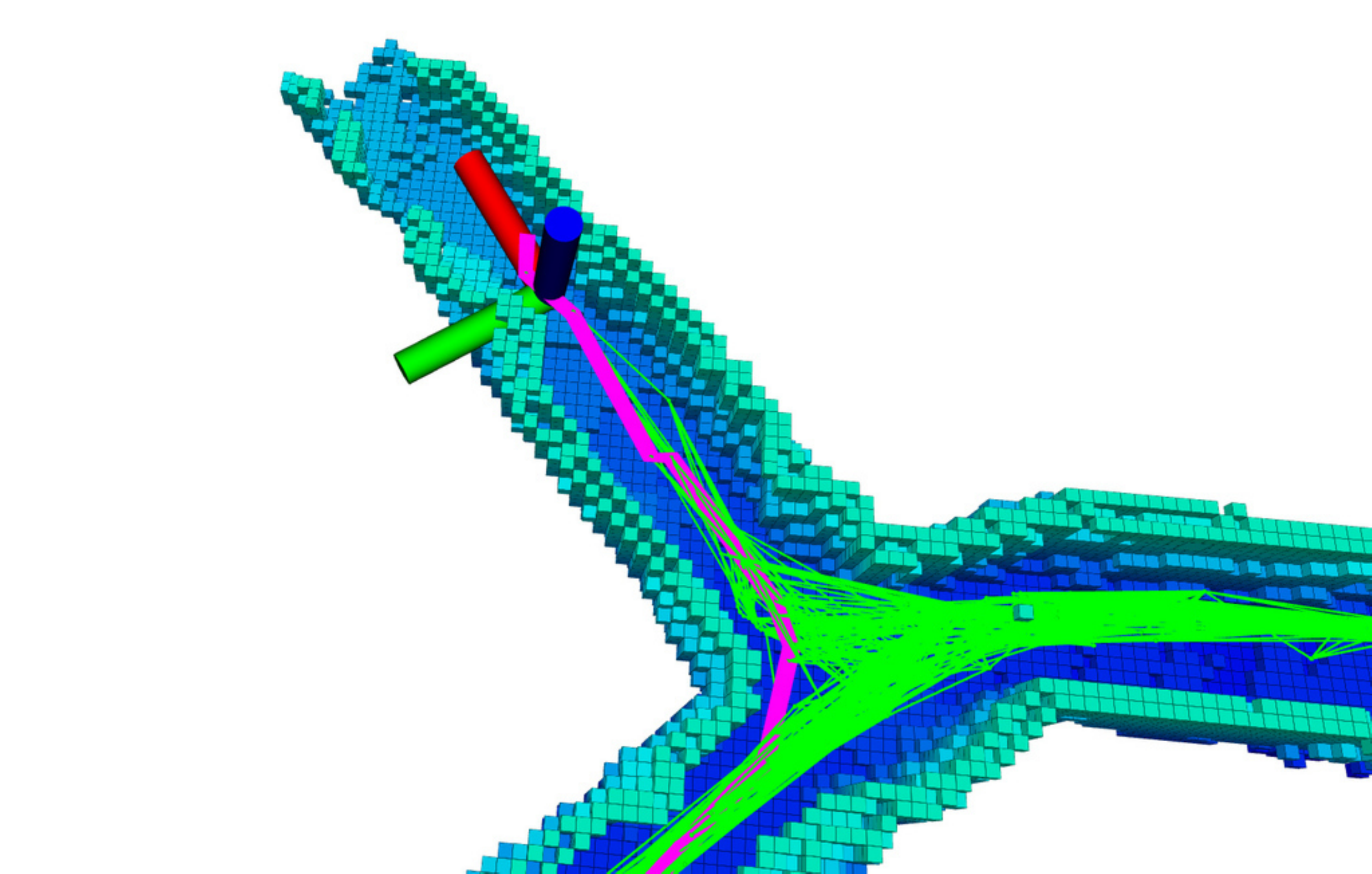}\label{fig:D02_0782s_narrow} }}
		\subfloat[]{{\includegraphics[width=.45\textwidth]{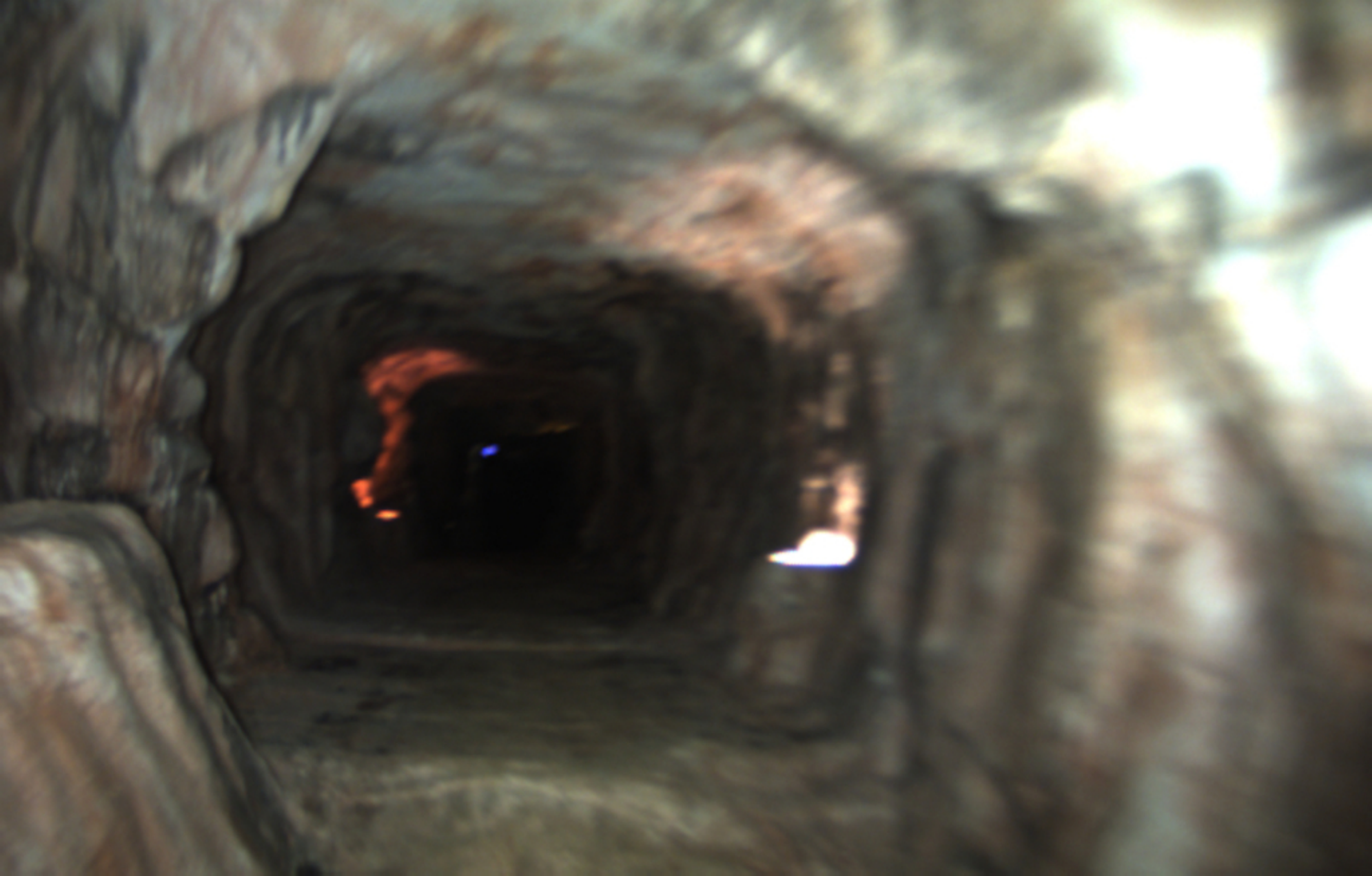}\label{fig:D02_0782s_narrow_map} }}
		\caption{Mobility limitations.}
		\label{fig:mobility_narrow_cave}
\end{figure}

The planner parameters were configured such that the agents would operate safely and not attempt to traverse highly confined spaces. One limitation to this approach, is that agents could not autonomously plan and traverse some areas, such as the utility corridor with low ceilings and narrow cave section, as shown in Figures \ref{fig:mobility_short_corridor} and \ref{fig:mobility_narrow_cave} respectively. The human supervisor was able to teleoperate the Spot through the cave section, and it autonomously traversed it when later exiting the cavern.

\subsection{Artifact Reports}

\begin{figure}[hbt!]
		\centering
		\subfloat[]{{\includegraphics[width=.24\textwidth]{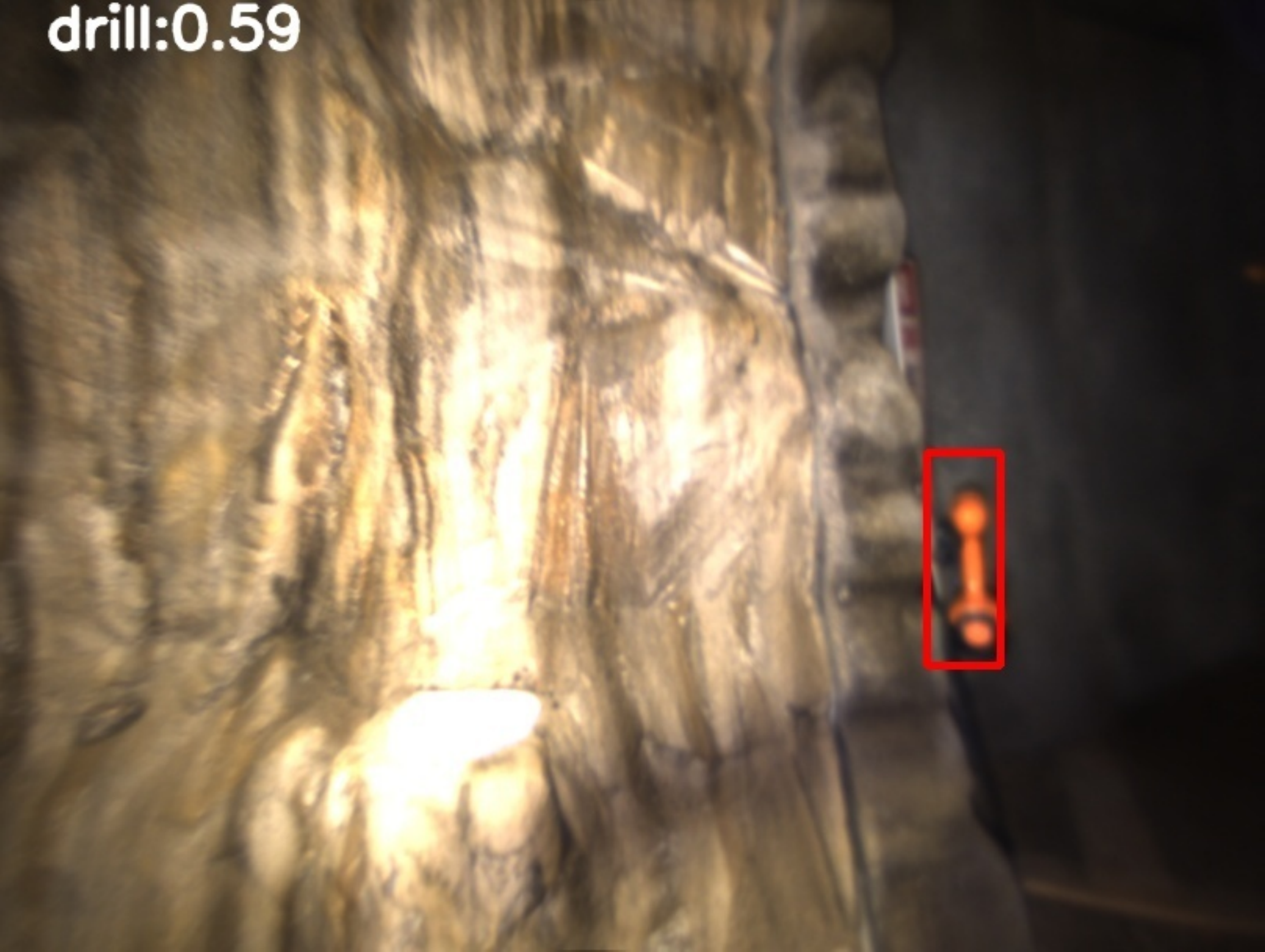}\label{fig:drill_mission} }}
		\subfloat[]{{\includegraphics[width=.24\textwidth]{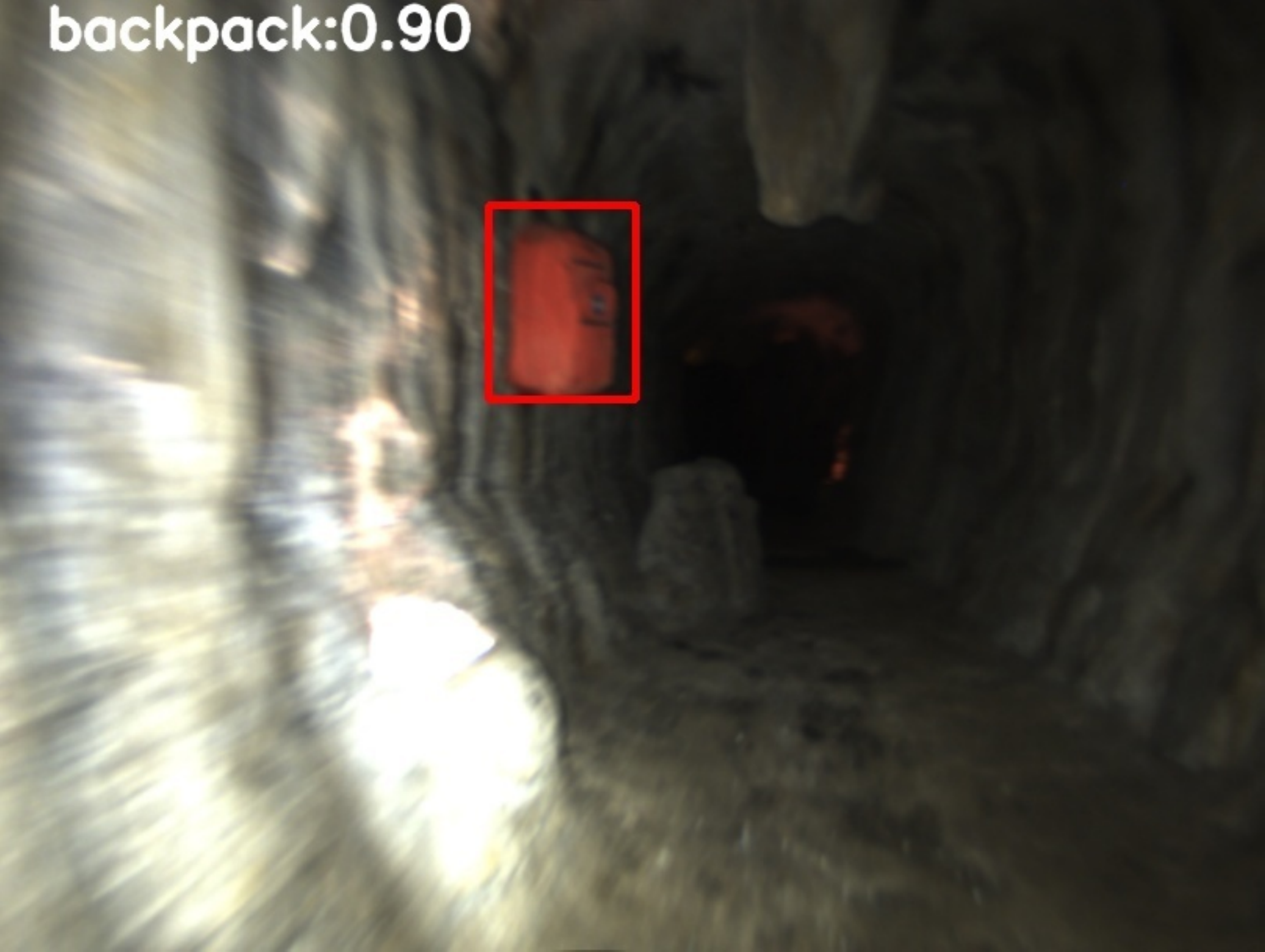}\label{fig:backpackcommon} }}
		\subfloat[]{{\includegraphics[width=.24\textwidth]{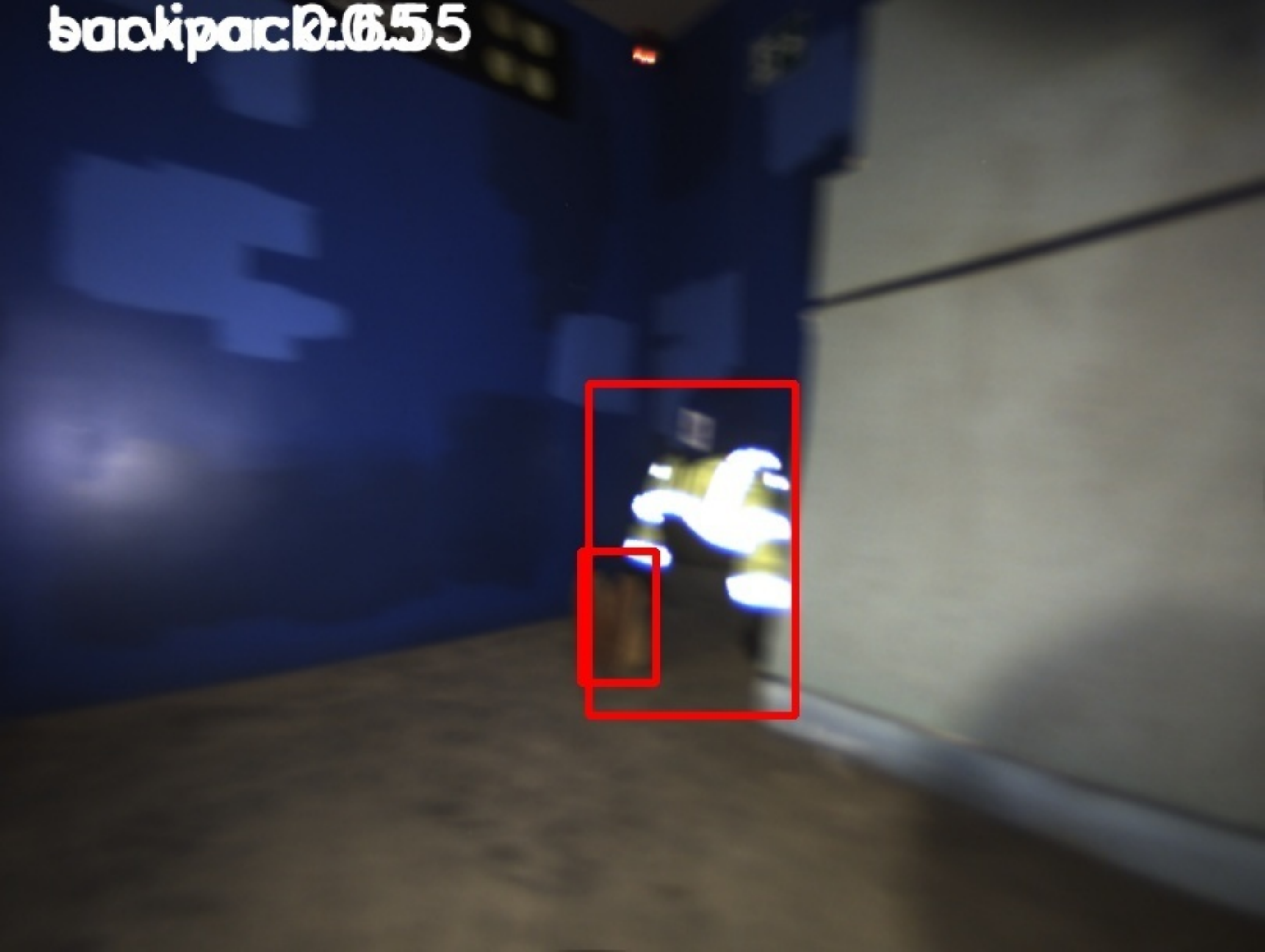}\label{fig:D01_4_6646_survivor1_blurry_backpack_occluded} }}
		\subfloat[]{{\includegraphics[width=.24\textwidth]{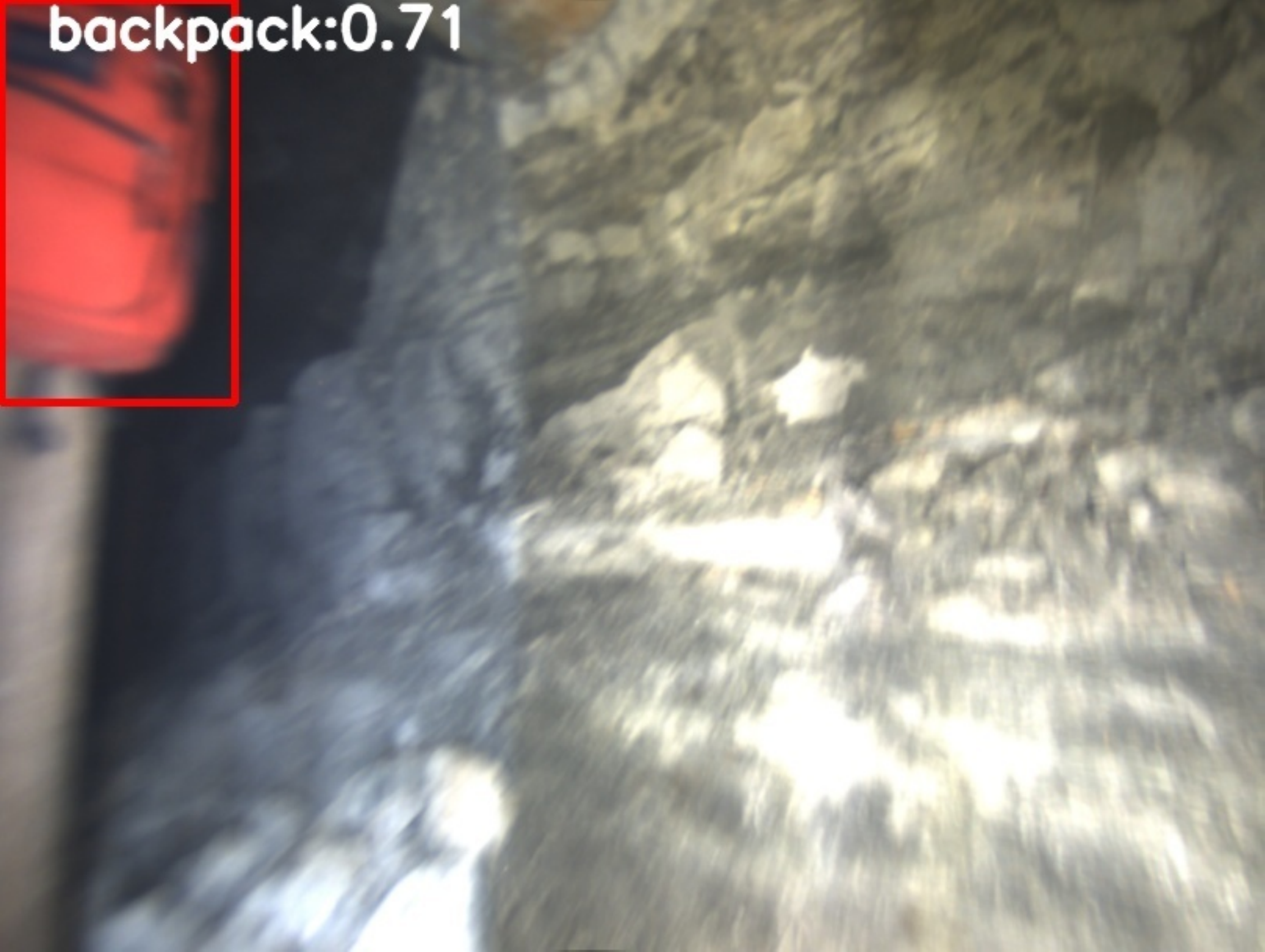}\label{fig:D01_14_8394_backpack1} }} \\
		\subfloat[]{{\includegraphics[width=.24\textwidth]{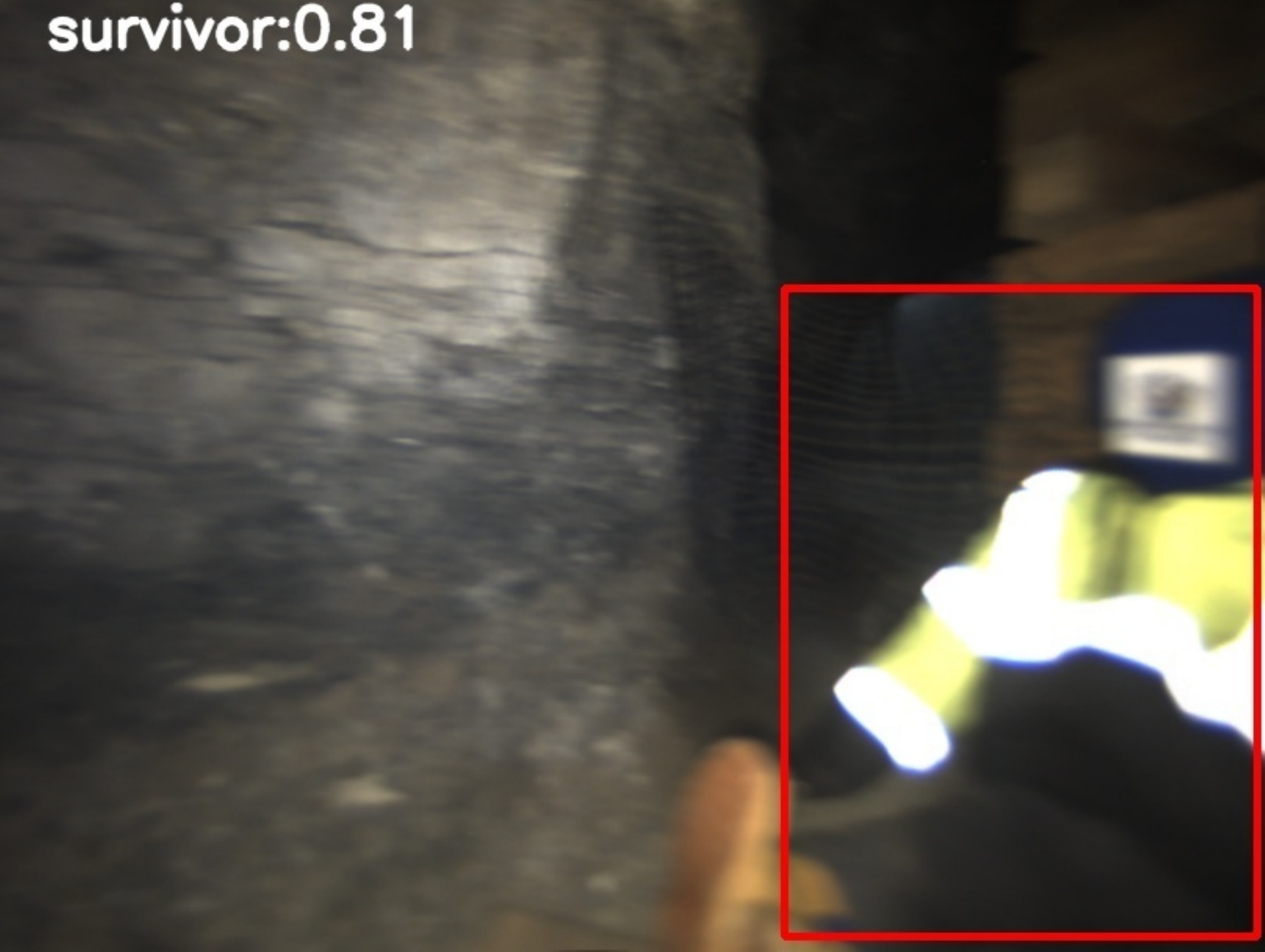}\label{fig:D01_15_8507_survivor2} }}
		\subfloat[]{{\includegraphics[width=.24\textwidth]{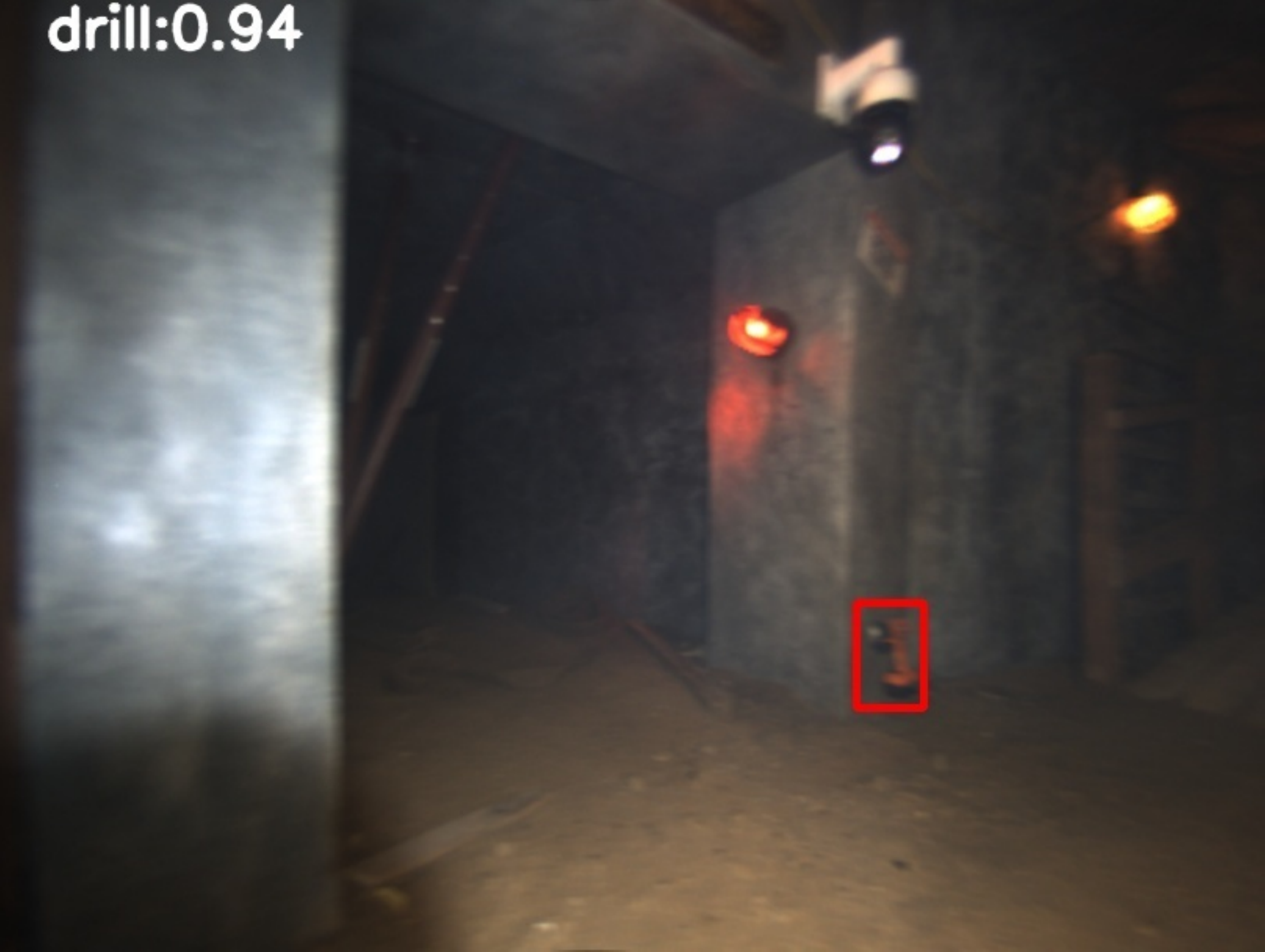}\label{fig:D01_16_8523_drill2} }}
		\subfloat[]{{\includegraphics[width=.24\textwidth]{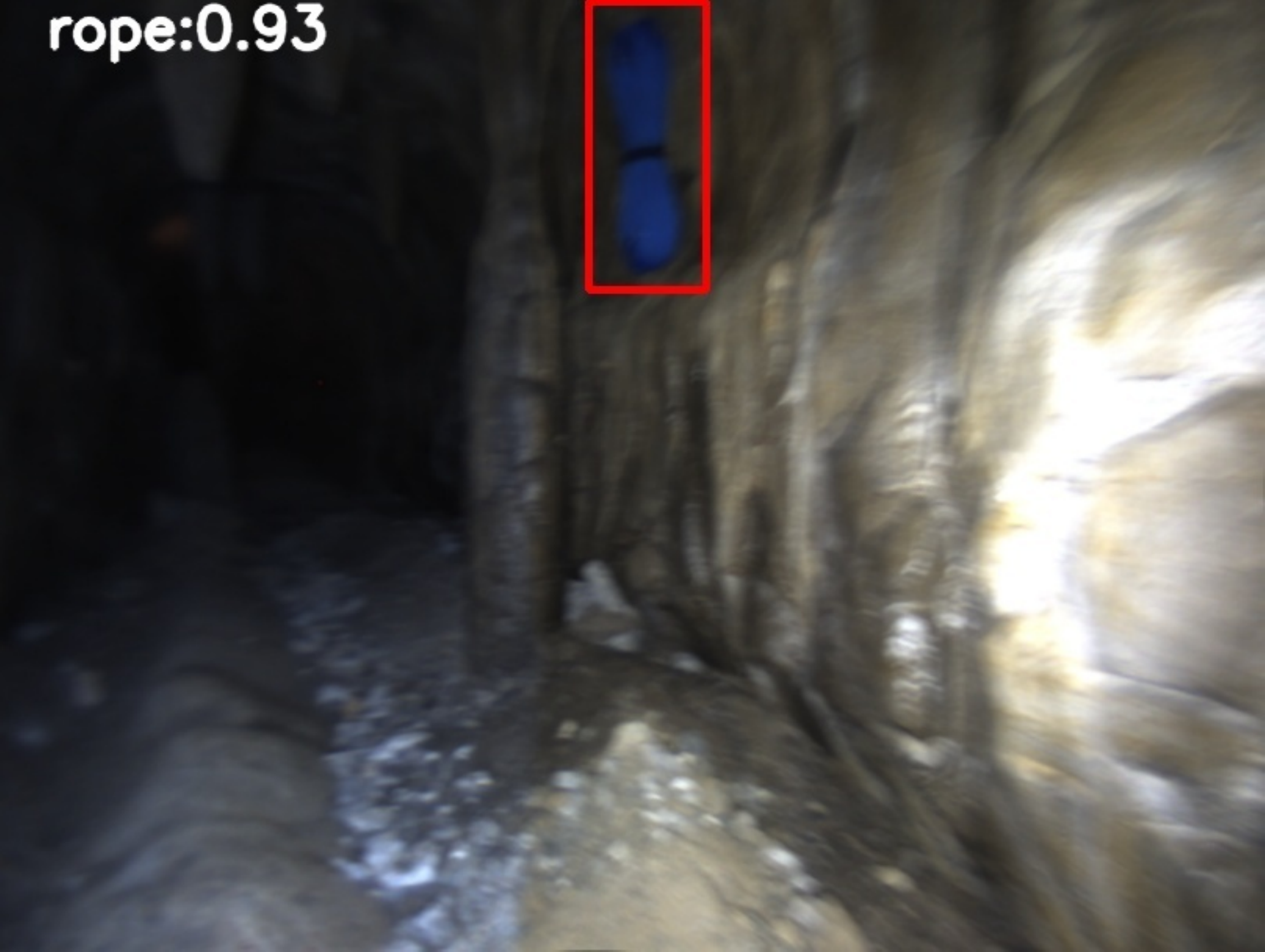}\label{fig:D01_19_8709_rope} }}
		\subfloat[]{{\includegraphics[width=.24\textwidth]{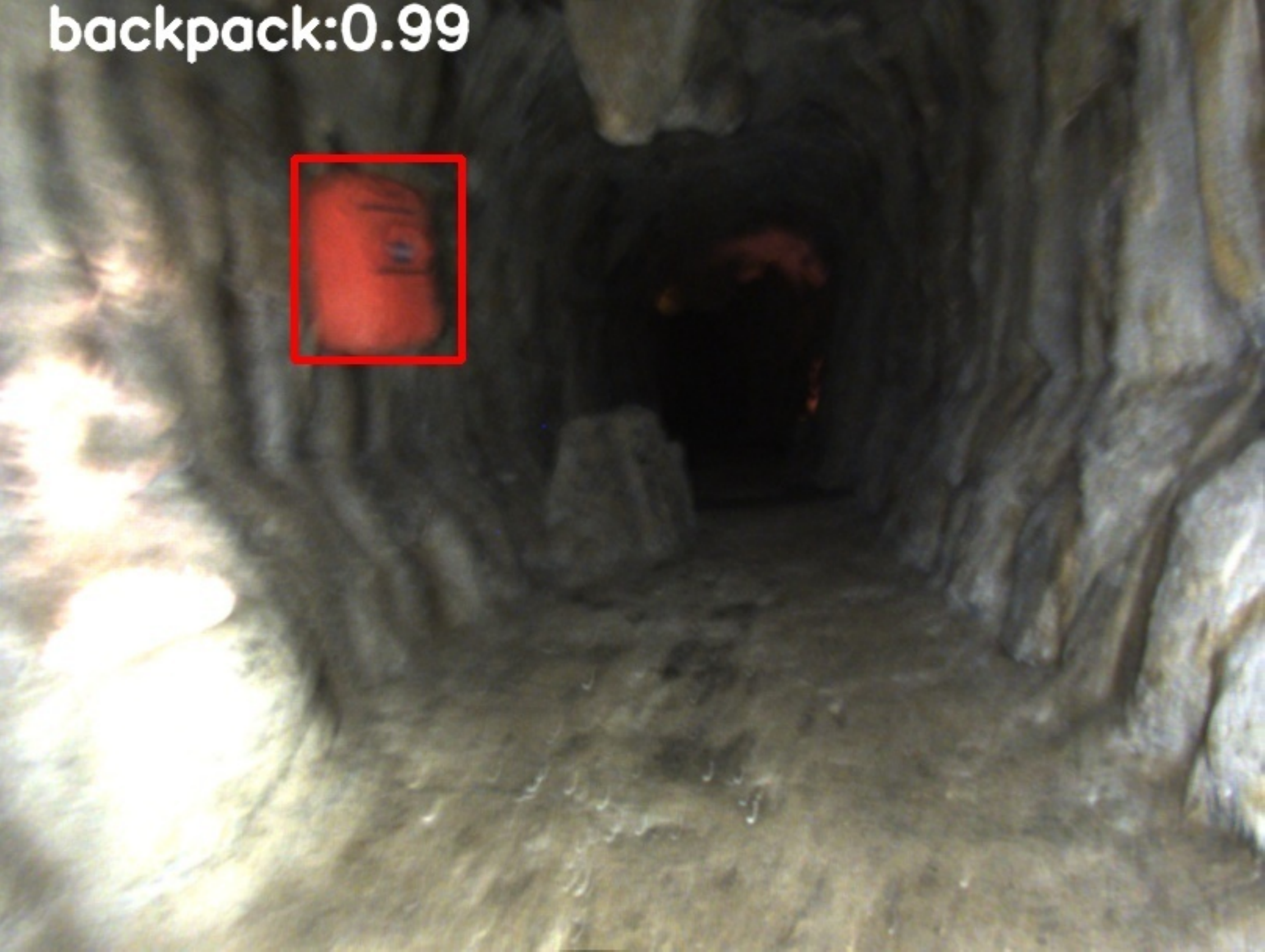}\label{fig:H02_2_6970__backpackcommon} }} \\
		\subfloat[]{{\includegraphics[width=.24\textwidth]{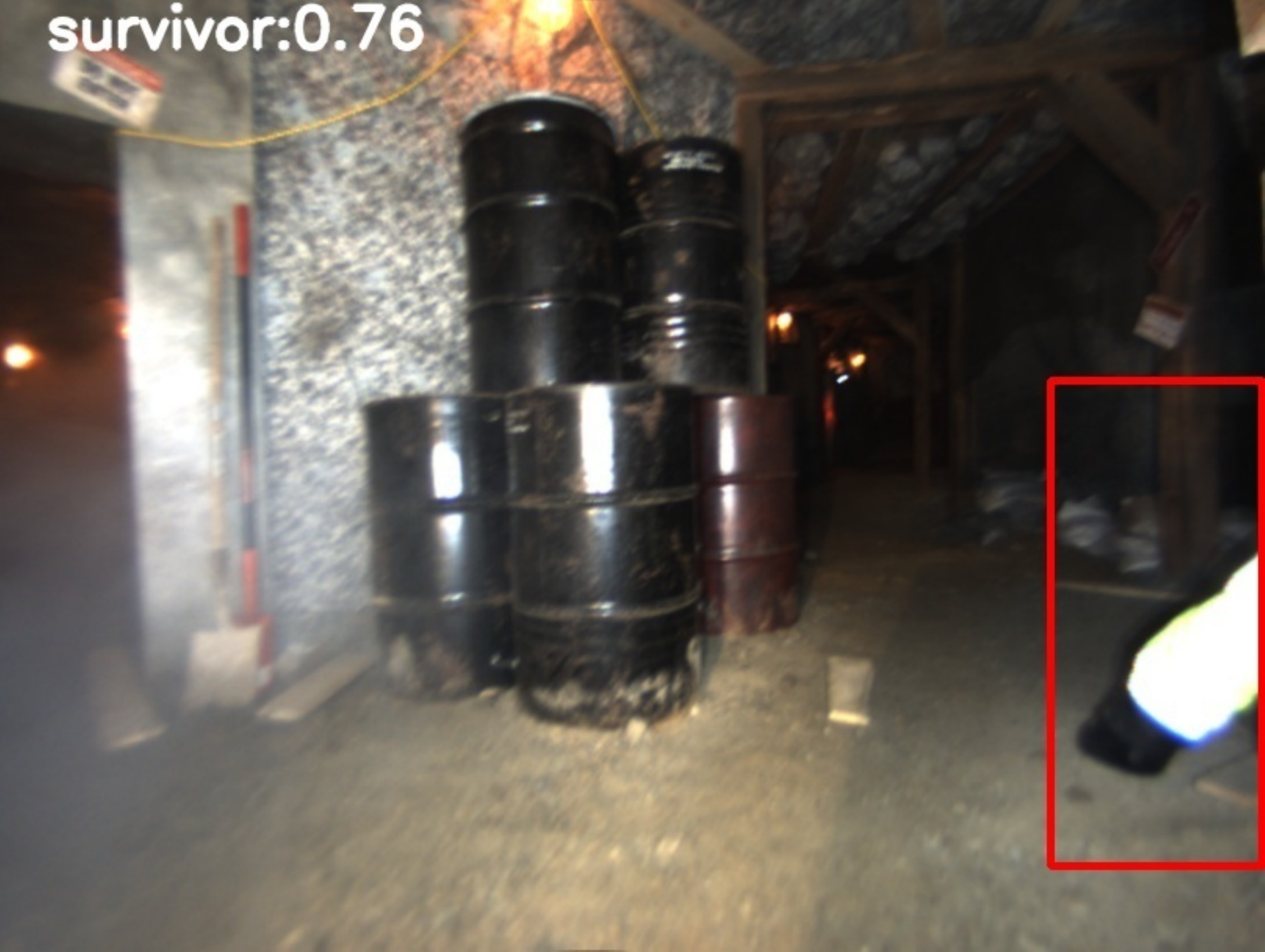}\label{fig:H01_2_7305_survivor3_sleeve} }}
		\subfloat[]{{\includegraphics[width=.24\textwidth]{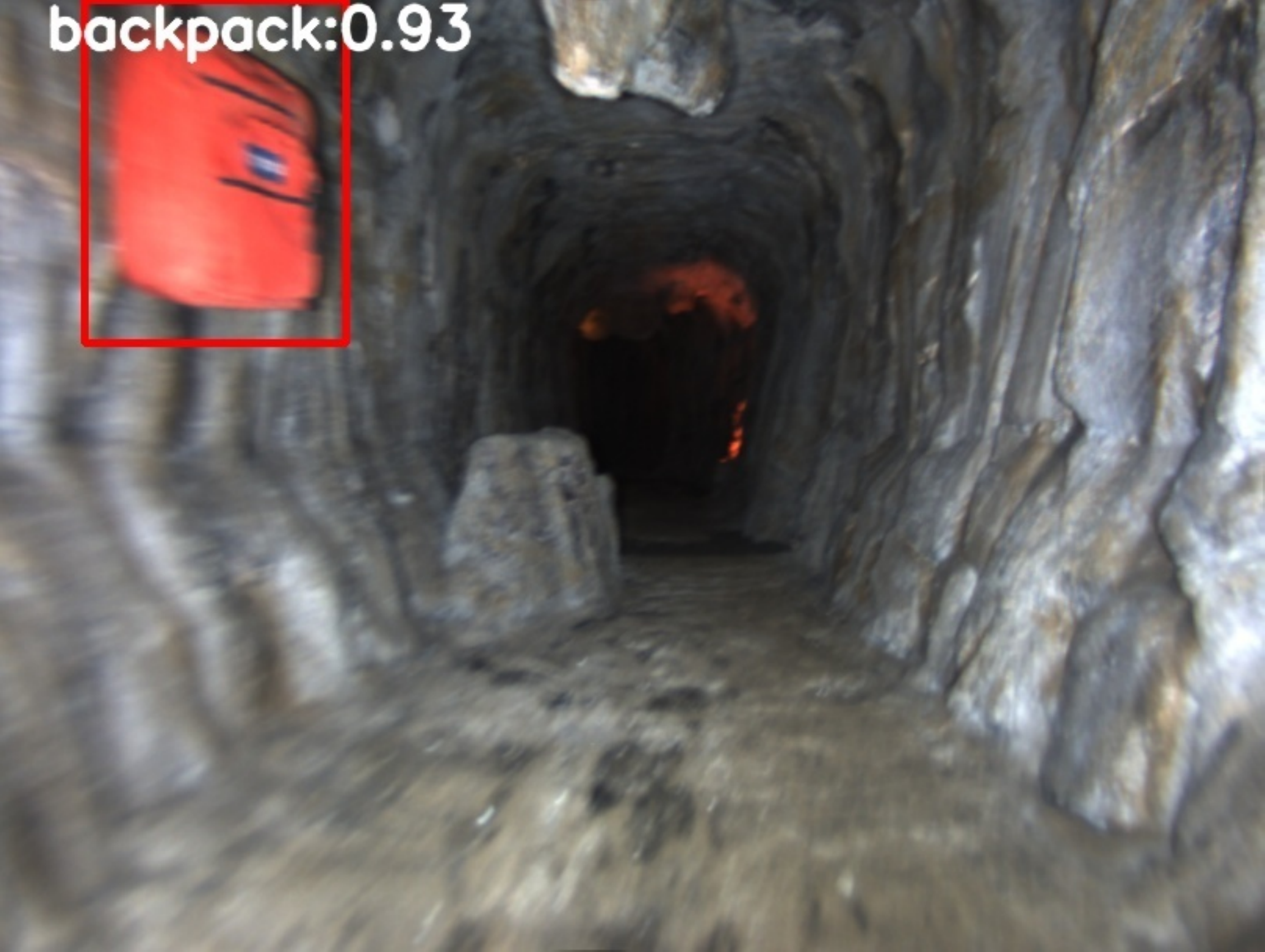}\label{fig:H01_5_8323_backpackcommon} }}
		\subfloat[]{{\includegraphics[width=.24\textwidth]{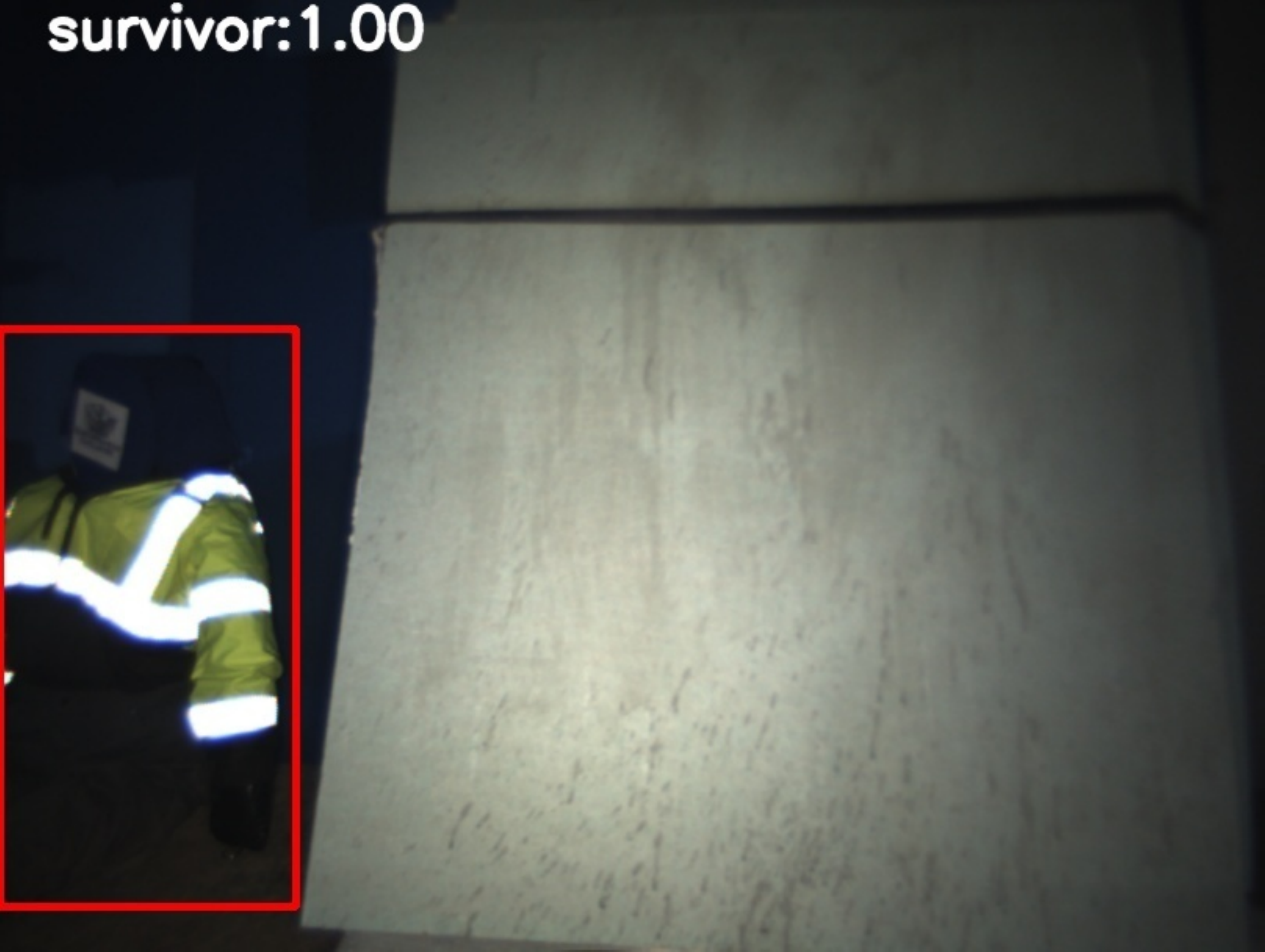}\label{fig:H01_7_8650_survivor1} }}
		\caption{All true positive reports of visual artifacts from autonomous artifact detection systems onboard remote agents D02 (a, b), D01 (c, d, e, f, g), H02 (h), and H01 (i, j, k), in order from mission start to mission end.}
		\label{fig:robot_artifact_reports_true_detections}
\end{figure}

\begin{figure}[hbt!]
		\centering
		\subfloat[]{{\includegraphics[width=.24\textwidth]{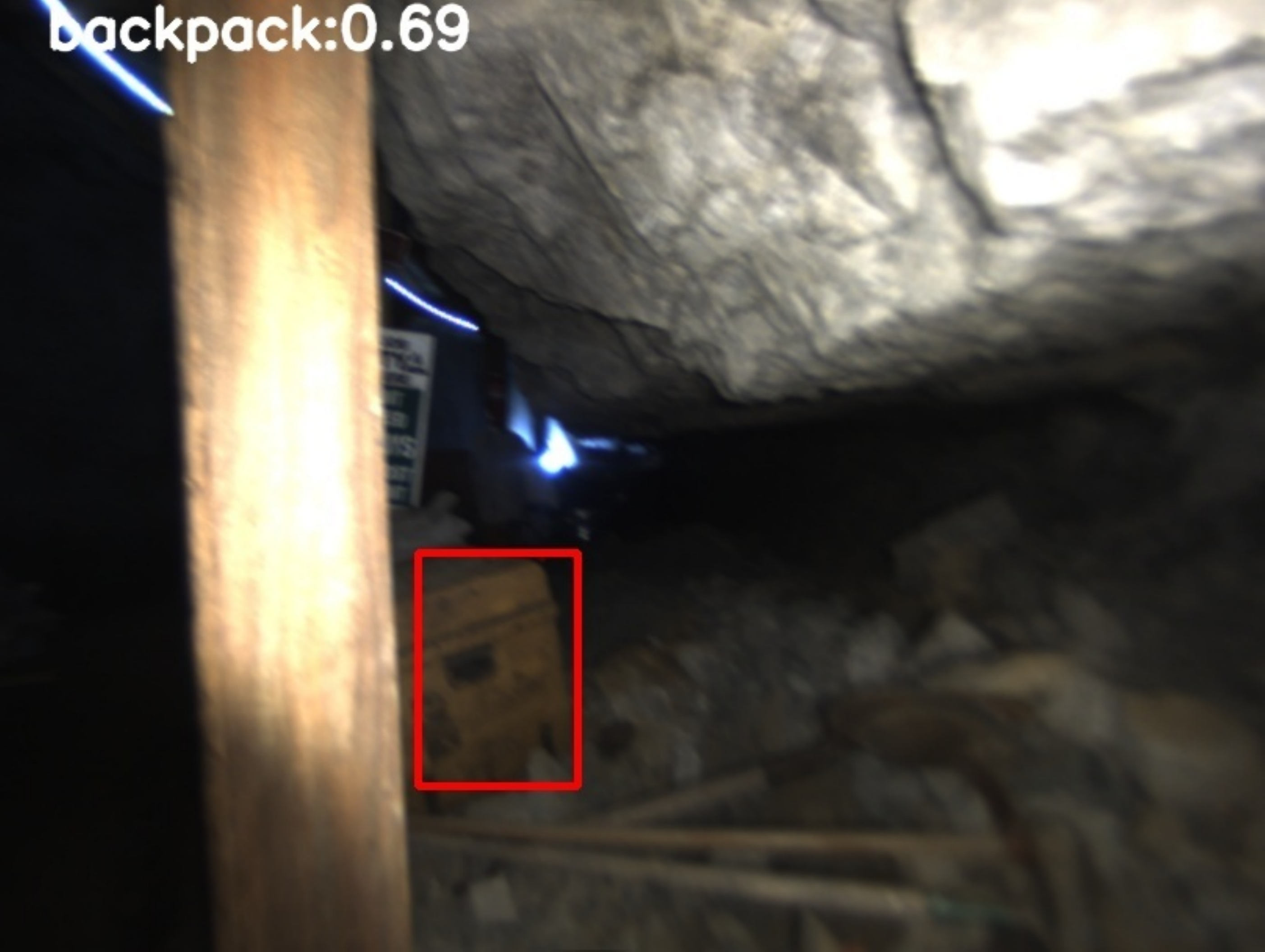}\label{fig:D02_5_7291_fake_backpack_brown_chest} }}
		\subfloat[]{{\includegraphics[width=.24\textwidth]{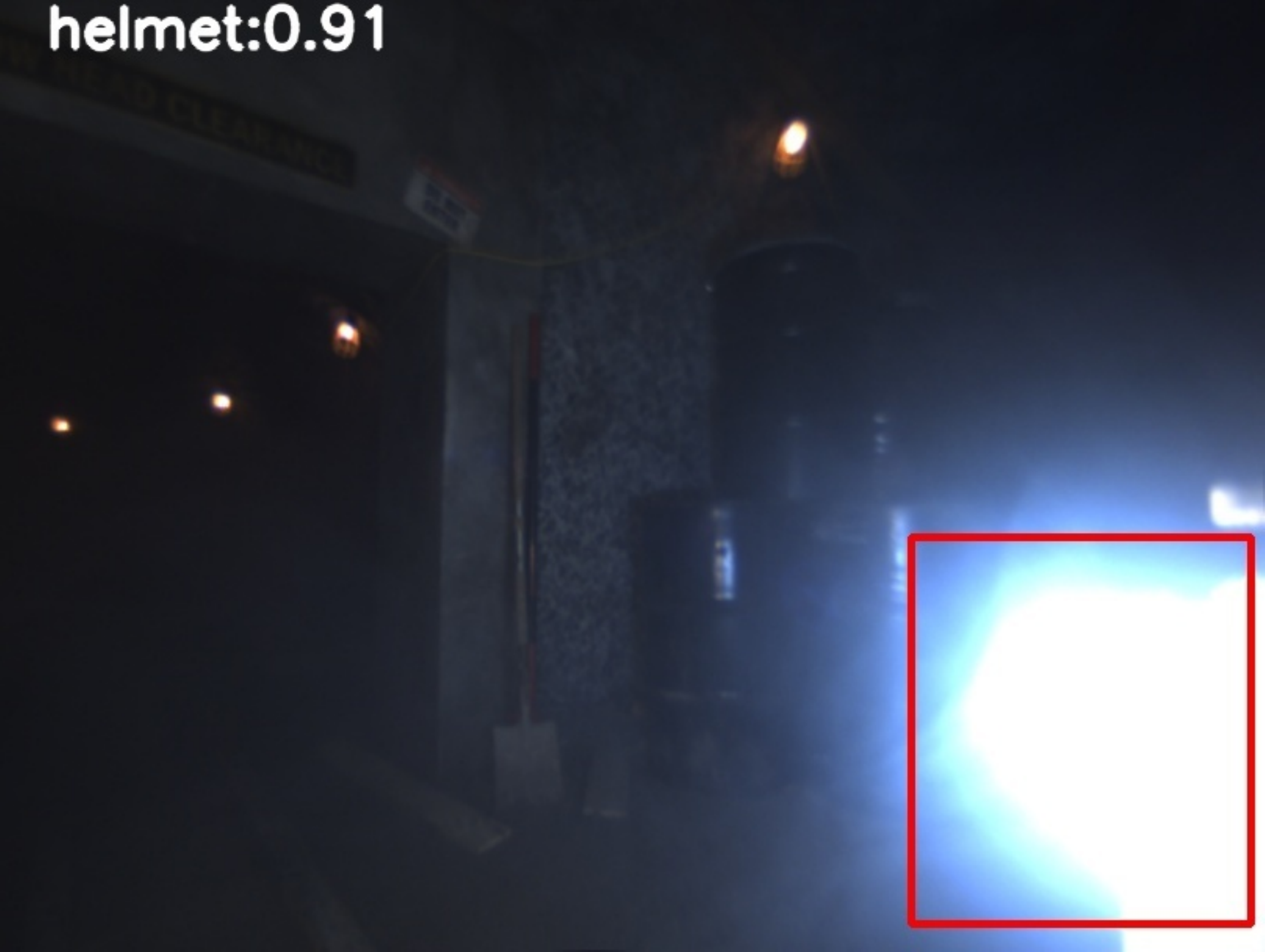}\label{fig:D02_6_7339_fake_helmet_white_light} }}
		\subfloat[]{{\includegraphics[width=.24\textwidth]{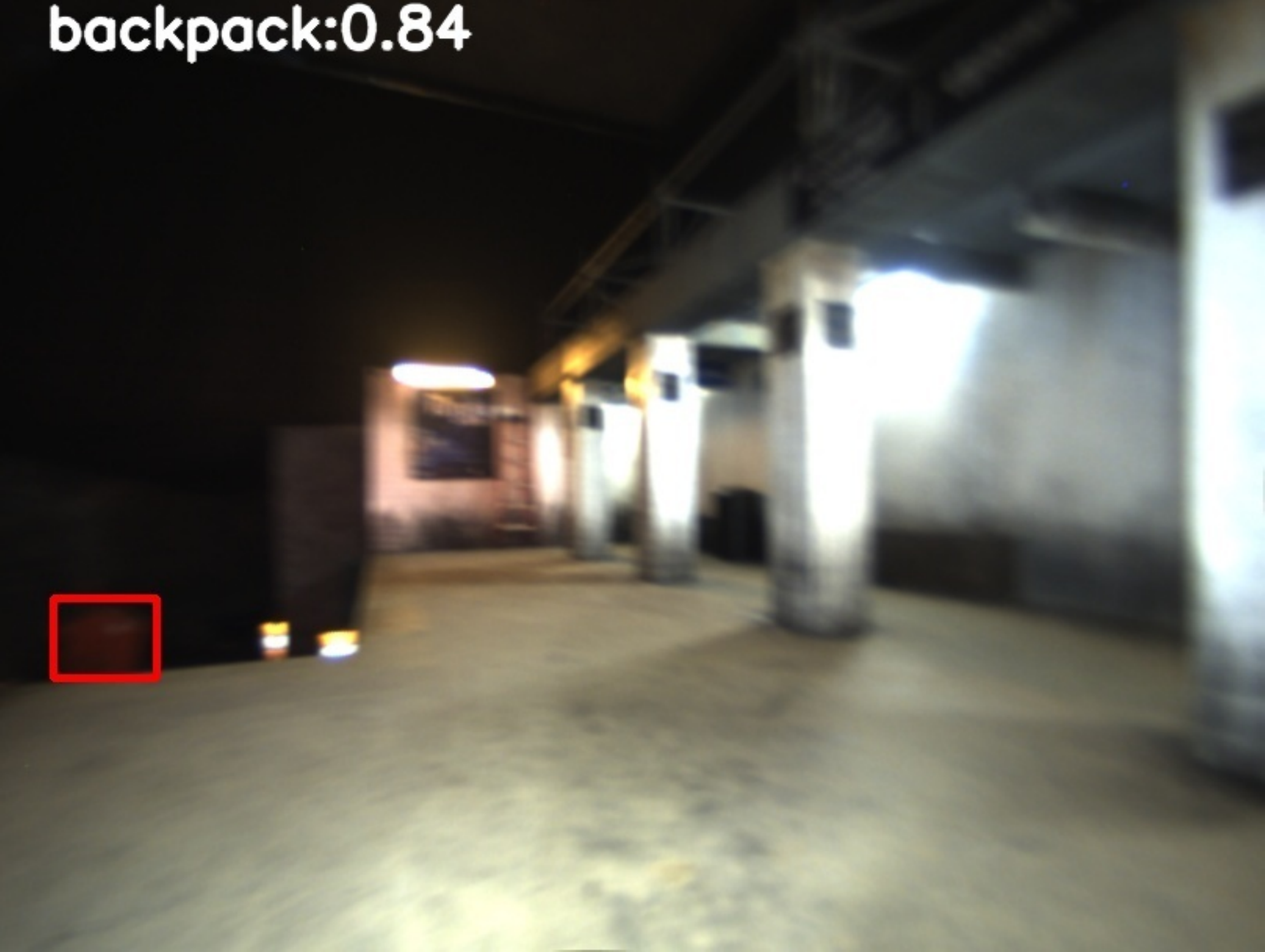}\label{fig:D02_10_8229_fake_backpack_red_box} }}
		\subfloat[]{{\includegraphics[width=.24\textwidth]{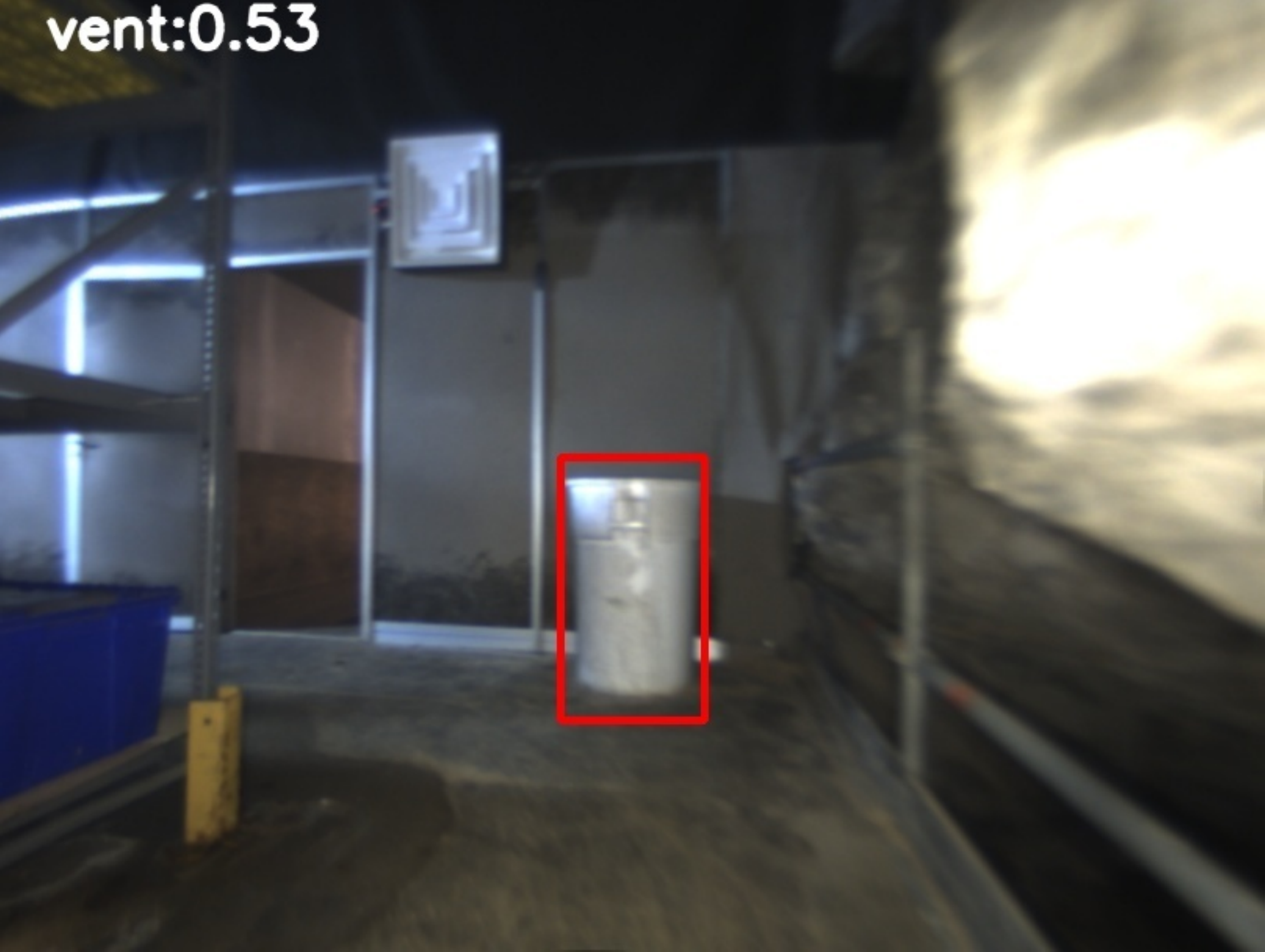}\label{fig:D02_14_9526_fake_vent_white_barrel} }} \\
		\subfloat[]{{\includegraphics[width=.24\textwidth]{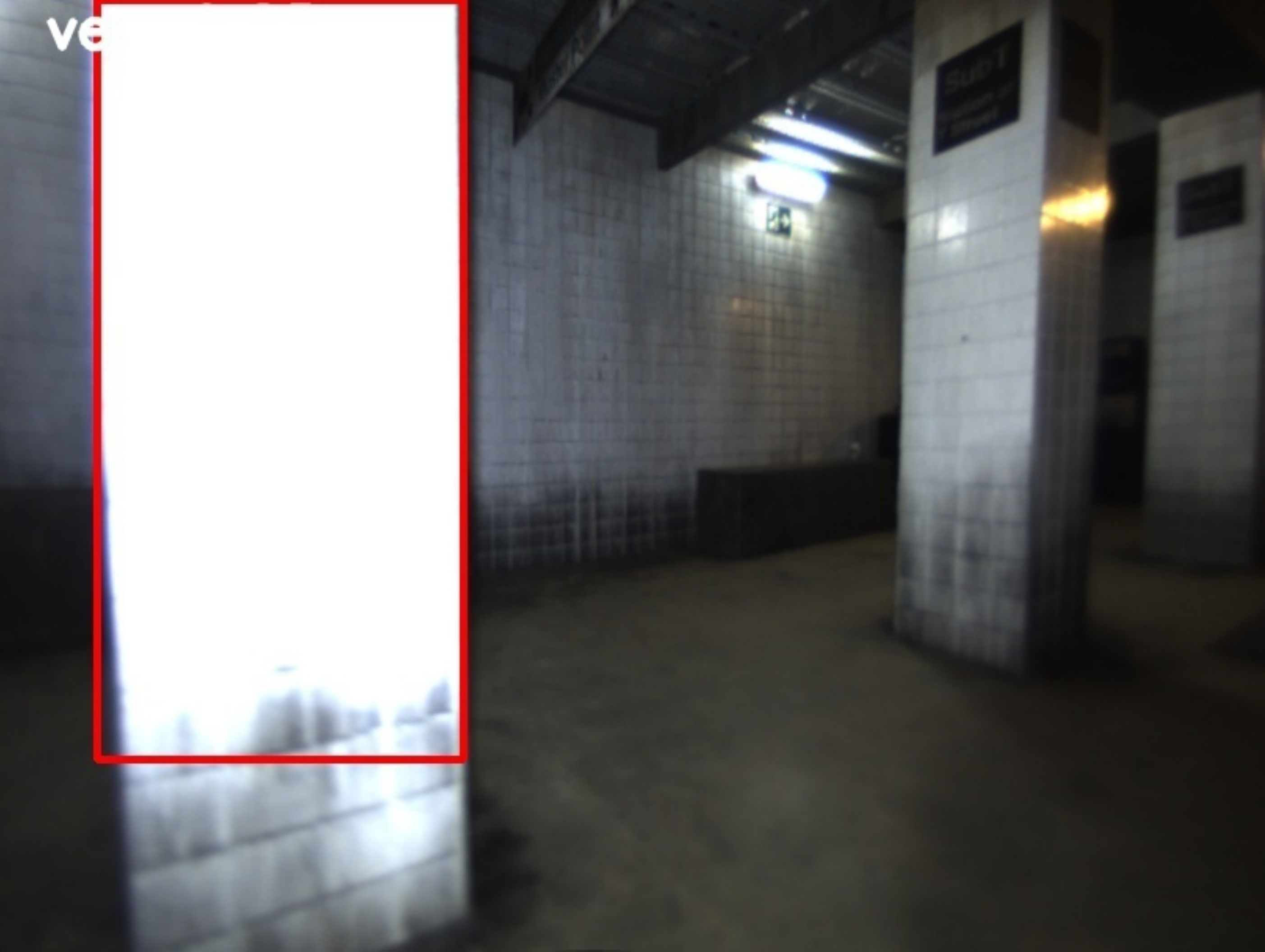}\label{fig:D01_6_6778_fake_vent_tile_subway_pillar} }}
		\subfloat[]{{\includegraphics[width=.24\textwidth]{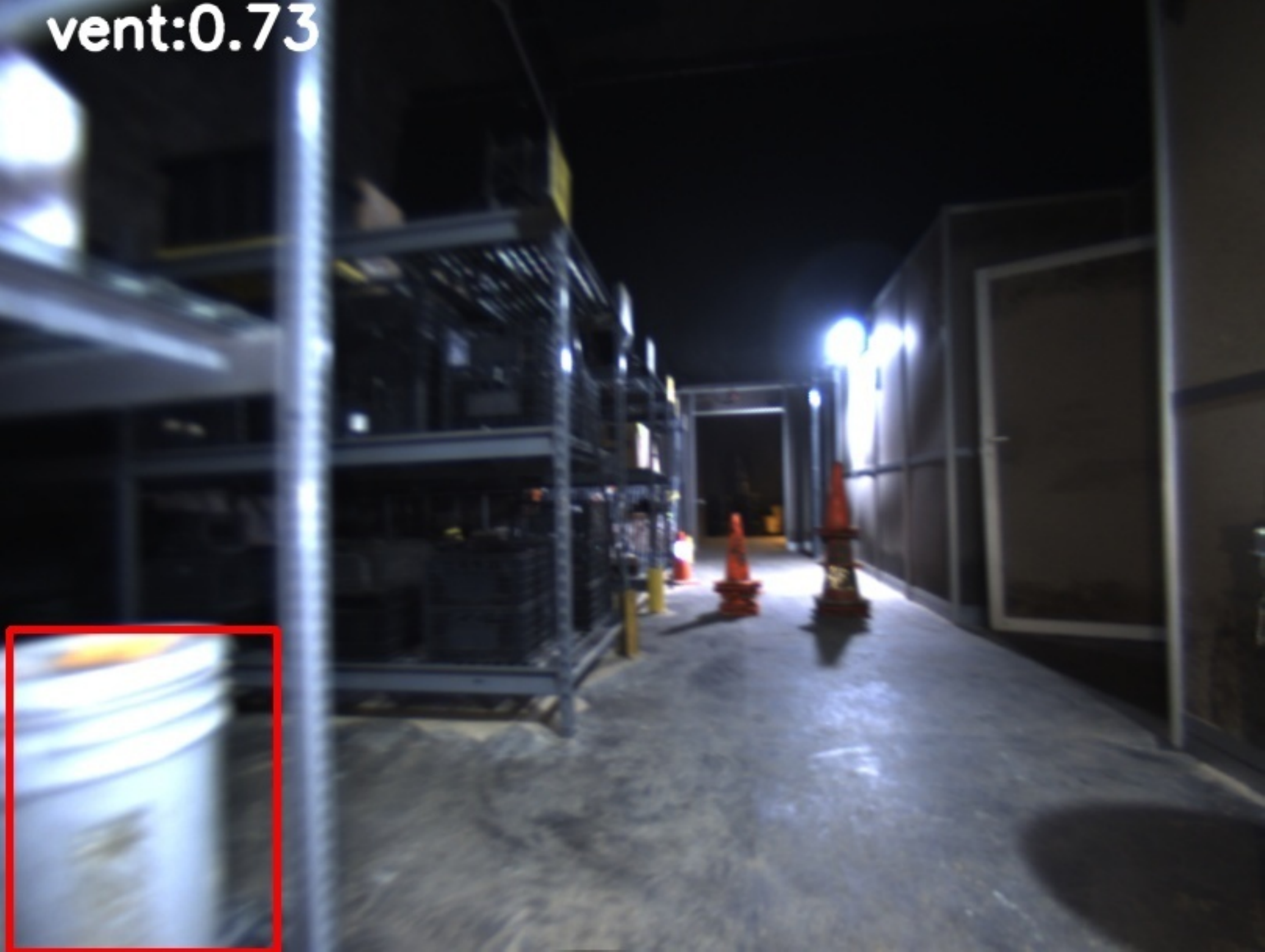}\label{fig:D01_7_7176_fake_vent_white_bucket} }}
		\subfloat[]{{\includegraphics[width=.24\textwidth]{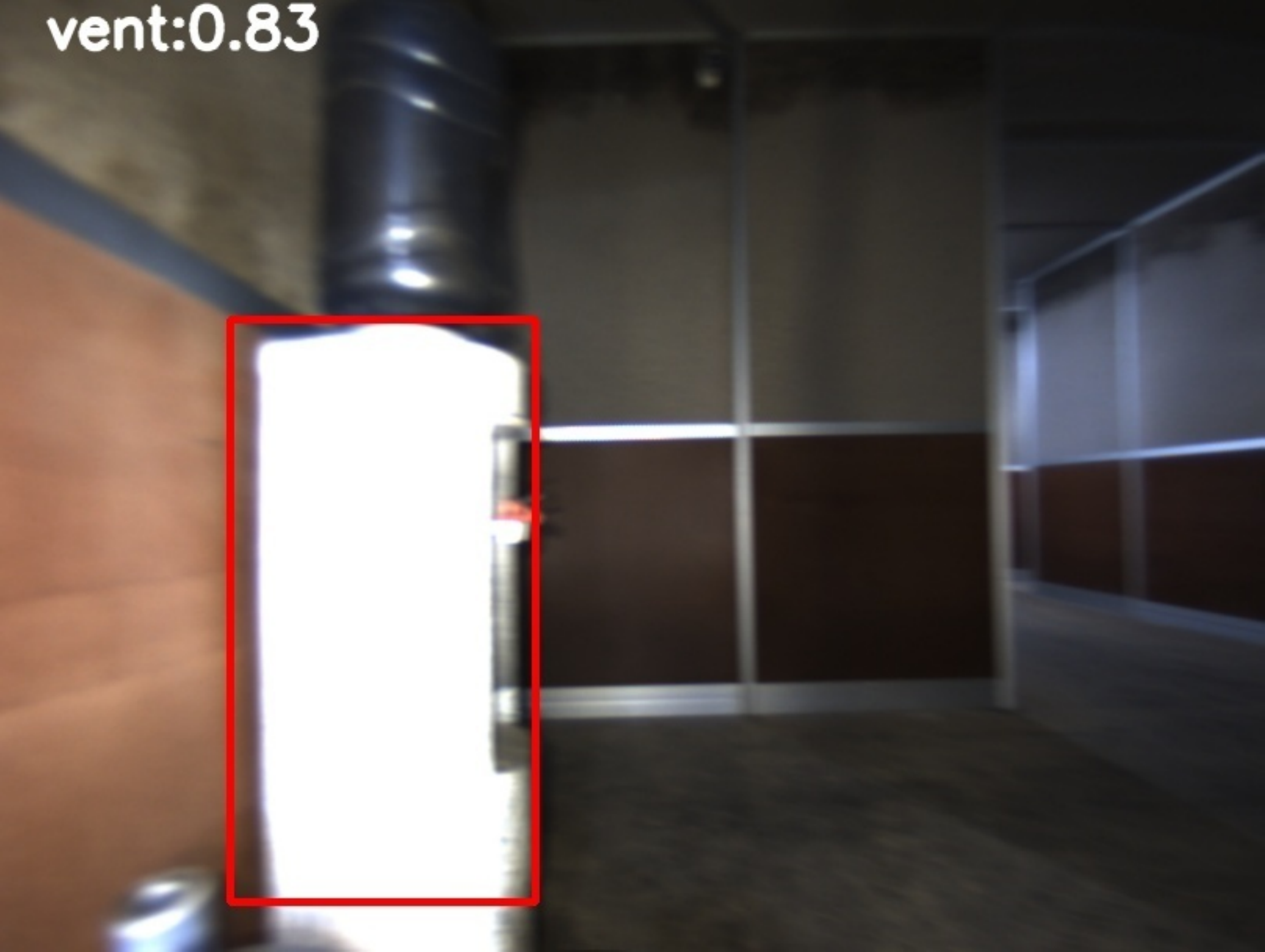}\label{fig:D01_8_7265_fake_vent_water_cooler} }}
		\subfloat[]{{\includegraphics[width=.24\textwidth]{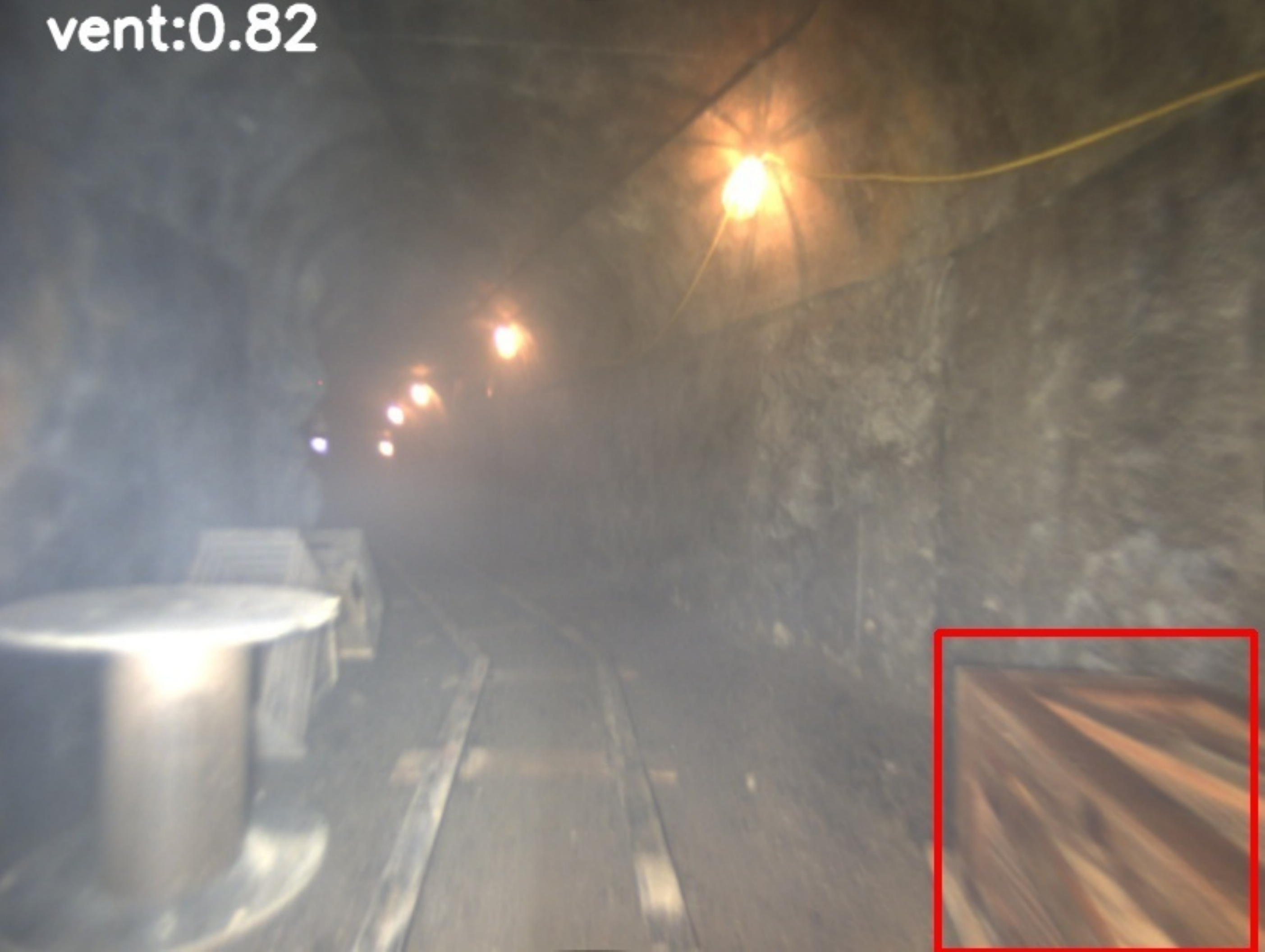}\label{fig:D01_11_8196_fake_backpack_wooden_box} }} \\
		\subfloat[]{{\includegraphics[width=.24\textwidth]{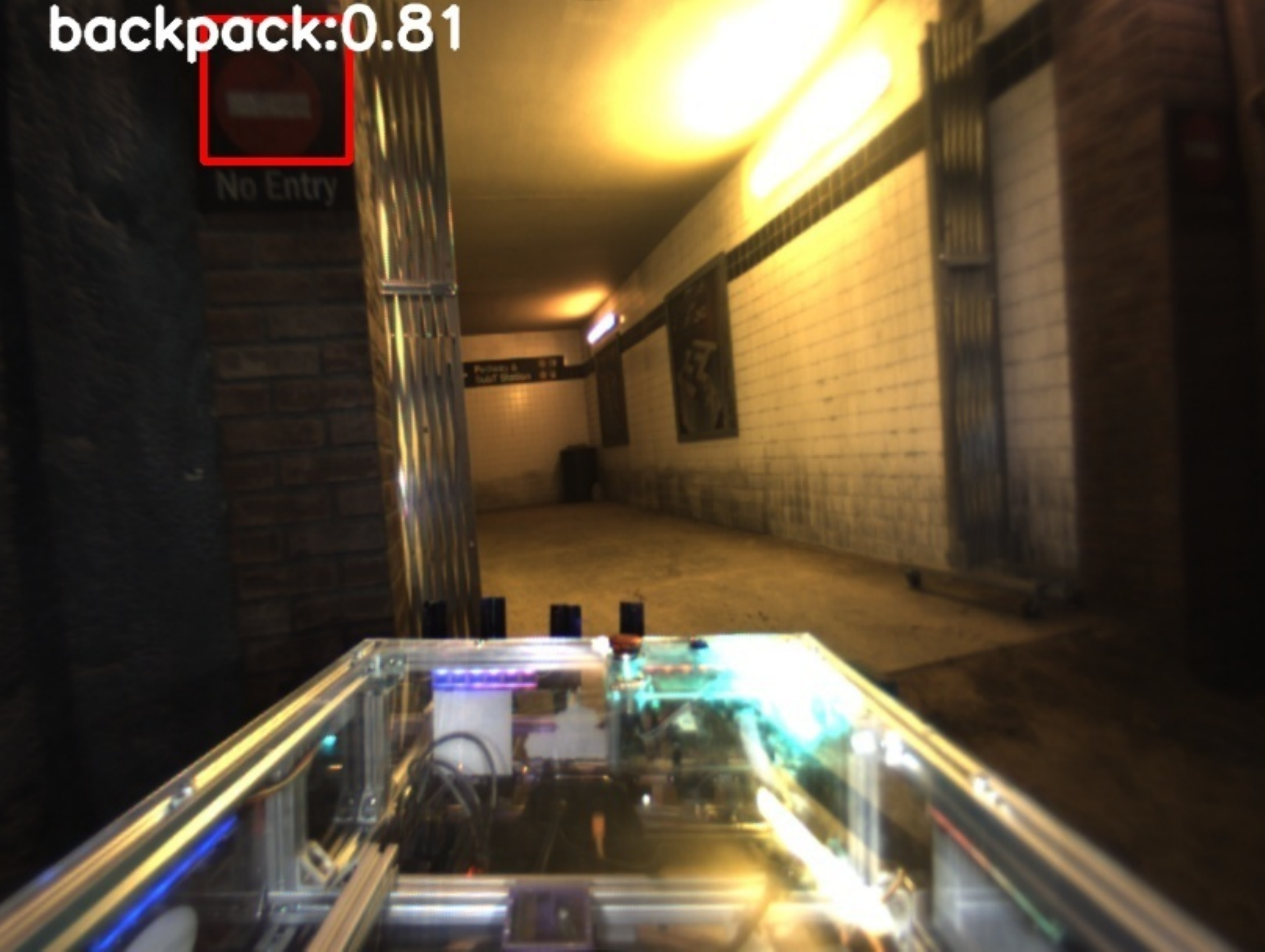}\label{fig:H02_3_6997_fake_backpack_no_entry_sign} }}
		\subfloat[]{{\includegraphics[width=.24\textwidth]{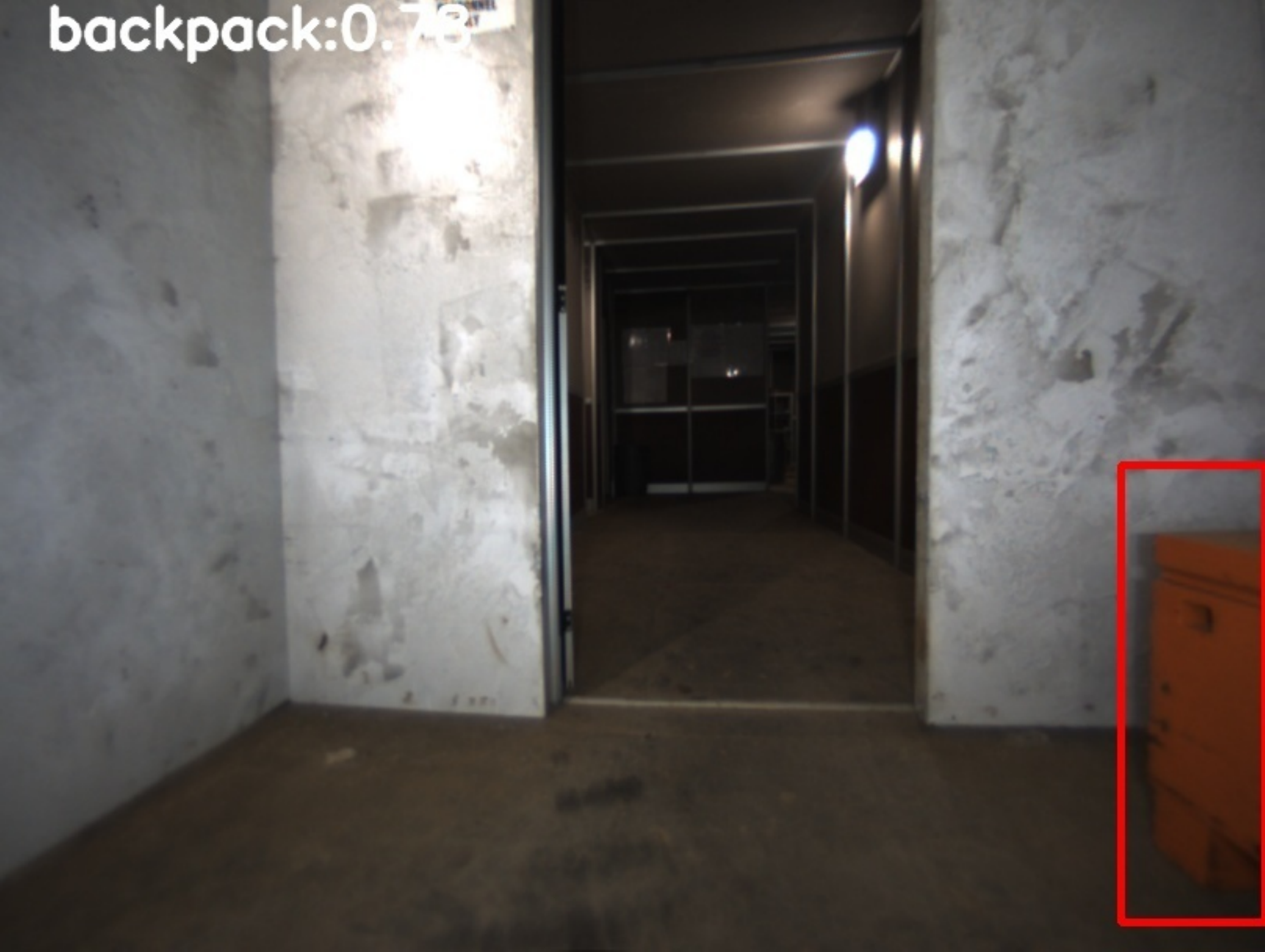}\label{fig:H01_6_8555_fake_backpack_orange_metal_box} }}
		\caption{All false positive reports of visual artifacts from autonomous artifact detection systems onboard remote agents D02 (a, b, c, d), D01 (e, f, g, h), H02 (i), and H01 (j), in order from mission start to mission end.}
		\label{fig:robot_artifact_reports_false_detections}
\end{figure}

All true positive \ref{fig:robot_artifact_reports_true_detections} and false positive artifact detections can be seen in Figure \ref{fig:robot_artifact_reports_false_detections}.

\subsection{Twenty-Five Explored Artifacts}
\label{ssec:sup_explored_artifacts}

This section focuses on all 25 artifacts that Team MARBLE agents were in the vicinity of during the Final Event Prize Run. To summarize, 18 of these artifacts were scored, one was not scored, and six were unreported. These three categories of artifacts are discussed in Section \ref{sssec:scored_artifacts}, Section \ref{sssec:missed_artifacts}, and Section \ref{sssec:unreported_artifacts}, respectively.

Each artifact has at least one attempt associated with it. The eighteen scored artifacts all end with a scored attempt (SA), and some will have multiple missed attempts (MA) before reaching a scored attempt. The one missed artifact was not scored, so it only has missed attempts associated with it. The six unreported artifacts have no attempts associated with them.

\subsubsection{Eighteen Scored Artifacts}
\label{sssec:scored_artifacts}

\textbf{L51 Drill (SA1)}: This drill was the first artifact scored, within 1 minute and 8 seconds of the mission start. D02 reported and scored the drill as it passed through the first junction of the course, splitting it into tunnel, urban, and cave corridors (SA1). The HS also saw the drill via D02 live FPV view and would have attempted to score it had D02 not automatically reported it.

\textbf{L53 Backpack (SA2)}: D02 quickly continued into the cave corridor, reporting and scoring the backpack (SA2). If D02 did not autonomously score the backpack, the HS may have scored it via D02 live FPV. Had that failed too, H02 and H01 accurately reported the backpack later in the mission and would have scored it.

\textbf{L55 Rope (SA3)}: Because D02 had difficulty traversing the narrow cave corridor, the HS manually teleoperated the agent through this section of the course, and in the process, saw the rope via D02 live FPV and reported it (SA3). D02 did not automatically detect the rope, likely due to poor lighting conditions.

\textbf{L26 Survivor (SA4)}: While D02 immediately explored the cave section, D01 began exploring the urban section and autonomously reported the survivor and scored it (SA4). Later in the mission, H01 accurately reported L26 and would have scored had D01 missed it at the beginning of the mission.

\textbf{L32 Survivor (SA5)}: The HS saw the survivor through D02 live FPV stream, submitted a manual report and scored the artifact (SA5). Later in the mission, accurate autonomous reports from D01 and H01 would have scored the artifact, had the HS had not already scored it.

\textbf{L08 Gas (SA6)}: D01 autonomously reported and scored the gas artifact as it traversed the urban environment (SA6).

\textbf{L31 Fire Extinguisher (SA7)}: During the middle of the mission, the HS teleoperated D01 through a foggy section of the tunnel environment. In this process, the HS saw the fire extinguisher through D01 live FPV stream, and scored the artifact via manual report (SA7). After this manual intervention, D01 went onto help score five more artifacts, L34, L38, L36, L40, and L67. The L31 fire extinguisher was also seen at the end of the mission when the HS was reviewing the D02 archived FPV images. 

\textbf{L34 Drill (SA8)}: After the HS teleoperated D01 through the fog, D01 autonomously reported and scored the drill (SA8). The HS also saw the drill through D01 live FPV stream, and would have reported the artifact manually had D01 not already scored it.

\textbf{L38 Fire Extinguisher (SA9)}: After teleoperating D01 through the fog, the HS saw the fire extinguisher though D01 live FPV stream and manually scored the artifact (SA9).

\textbf{L36 Cube (SA10)}: After the HS teleoperated D01 through the fog, D01 autonomously reported and scored the cube (SA10).

\textbf{L40 Backpack (SA11)}: After the HS teleoperated D01 through the fog, D01 autonomously reported and scored the backpack (SA11).

\textbf{L67 Rope (SA12)}: After the HS teleoperated D01 through the fog, D01 autonomously reported and scored the rope. The HS also saw the artifact though D01 live FPV stream, and would have manually reported the artifact had D01 not already scored it (SA12).

\textbf{L11 Cube (MA5, MA6, SA13)}: D01 and D02 both autonomously reported the cube, but were both missed attempts (MA5, MA6), with corresponding errors of 10.73m and 9.42m. The HS then used the location of the missed attempts to manually submit an adjusted location, scoring the cube (SA13) with an error of 1.83m.

\textbf{L22 Cell Phone (MA2, MA4, SA14)}: D01 autonomously reported the cell phone early in the mission, but was a missed attempt (MA2) with an error of 19.47m. H02 also autonomously reported the cell phone, but the report was a missed attempt (MA4) with an error of 7.77m. Near the end of the mission, H01 accurately localized the cell phone and scored the artifact (SA14), with an error of 4.06m.

\textbf{L47 Cell Phone (SA15)}: D02 autonomously reported the cell phone along the subway platform and scored it (SA15).

\textbf{L59 Cell Phone (MA1, MA3, MA10, SA16)}: Located in the left branch of the cave section, the cell phone was autonomously reported by D02, but was a missed attempt (MA1) with an error of 13.69m. The HS then manually submitted the cell phone with an adjusted location, but this was also a missed attempt (MA3) with an error of 7.74m. Later in the mission, a third missed attempt occurred (MA10), as D01 autonomously reported the cell phone with an error of 9.97m. Immediately after, the HS used the reported locations from D02 and D01 to submit another manual report with an adjusted location, and scored the artifact (SA16), with an error of 2.15m.

\textbf{L24 Gas (MA11, SA17)}: The gas was autonomously reported by H01, but with an error of 5.38m, resulted in a missed attempt (MA11). Soon after, H02 traversed the same space and autonomously reported the gas independently of H01. This report better estimated the position of the artifact, with an error of 2.55m, and scored (SA17) the gas artifact.

\textbf{L58 Helmet (SA18)}: The HS reviewed the D02 archived FPV images near the end of the mission, and saw the helmet in the cavern. The report was accurate to 1.74m and scored Team MARBLES 18th and final point of the mission (SA18). This point was made possible by the HS teleoperation through the narrow cave corridor.

\subsection{One Missed Artifact}
\label{sssec:missed_artifacts}

\begin{figure}[!htb]
		\centering
		\subfloat[]{{\includegraphics[width=.45\textwidth, trim={0 0 0 4cm},clip]{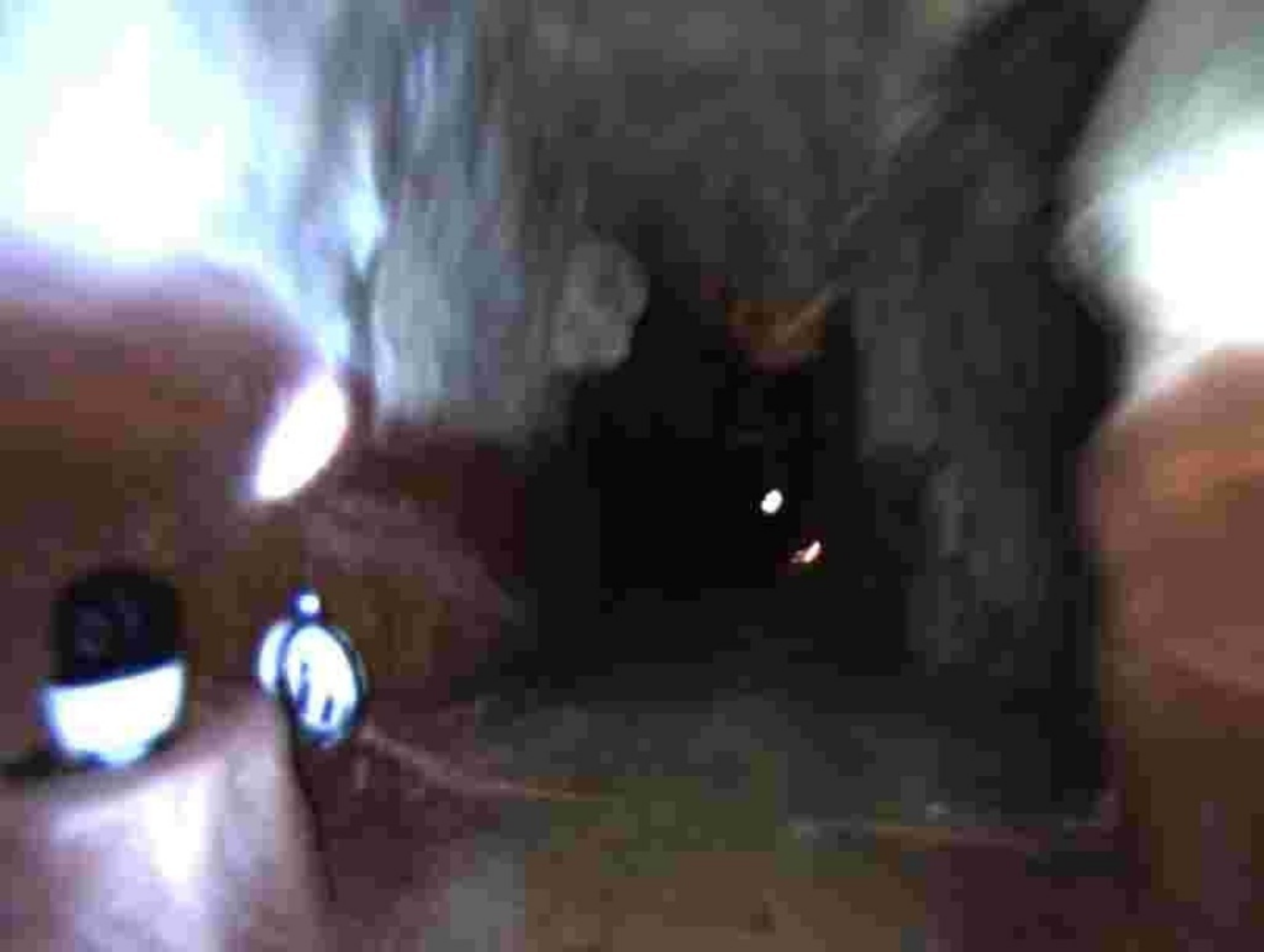}\label{f:Missed_Helmet} }}
		\subfloat[]{{\includegraphics[width=.45\textwidth]{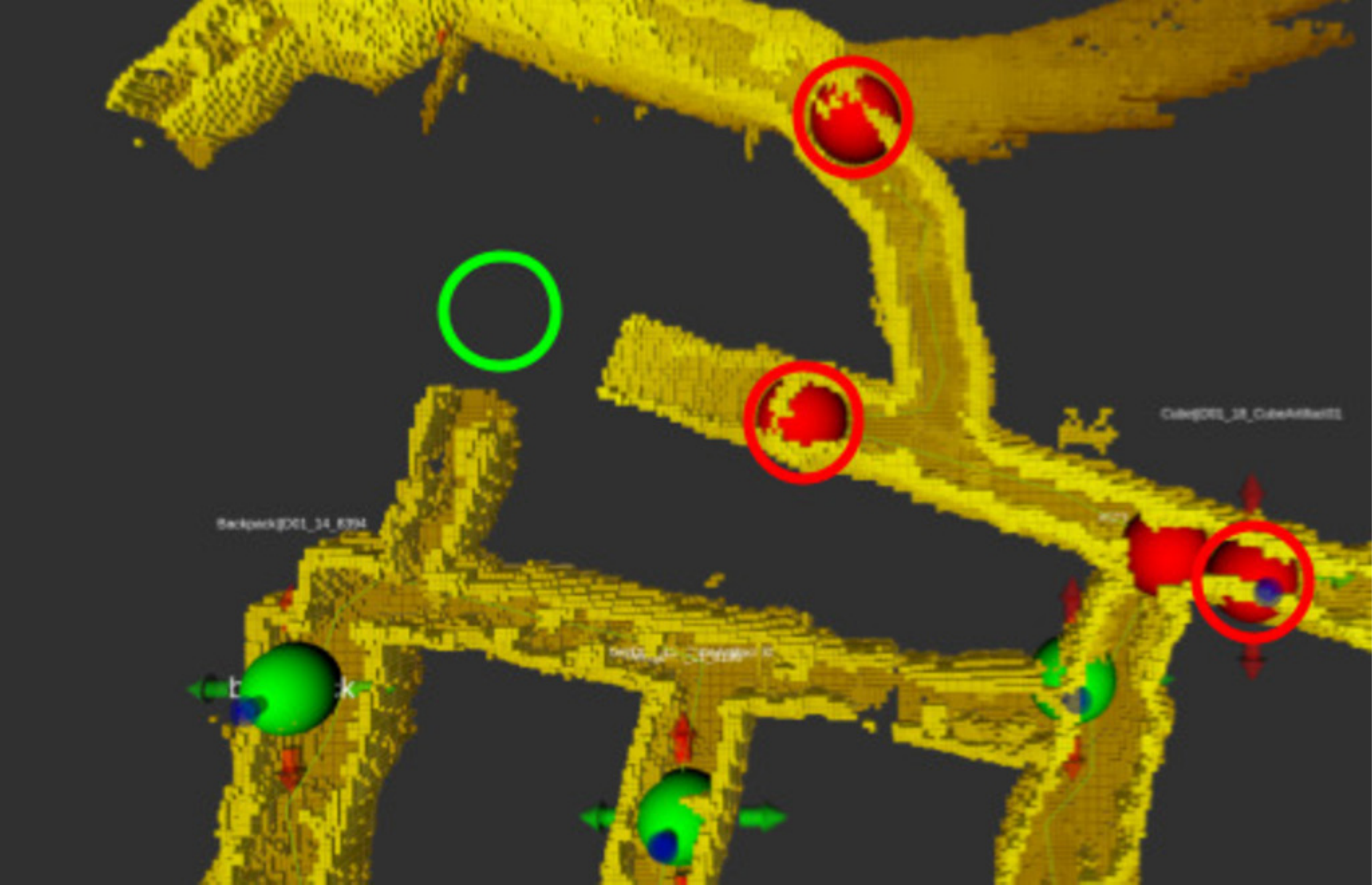}\label{f:Missed_Cube} }}
		\caption{Helmet FPV image transmitted to human supervisor, but missed due to workload (a). L64 Cube location, circled in green, and incorrect submissions, circled in red (b).}
		\label{fig:missed_artifacts}
\end{figure}

\textbf{L64 Cube (MA7, MA8, MA9)}: This cube artifact was located atop a steep slope found along the main corridor of the cave section. D01 autonomously reported the cube but was a missed attempt with an error of 8.24m (MA7). Based on that experience, the HS manually submitted two reports at adjacent locations, both of which were missed attempts (MA8, MA9), with errors of 12.23m and 20.15m. This artifact was never scored during the mission. Figure \ref{f:Missed_Cube} shows the failed score attempts circled in red, and the actual position circled in green.


\subsection{Six Unreported Artifacts}
\label{sssec:unreported_artifacts}

\textbf{L02 Vent}
L02 vent was not detected by the on-board artifact detection system, but was available in the human supervisor's FPV feed.  At approximately 20 minutes into the mission, D01 (Spot) was returning home due to a localization error noticed by the human supervisor.  While it was returning, the Supervisor turned attention to other robots, and did not scan the FPV feed for some time.  The robot stopped for 33 seconds and transmitted a series of images similar to Figure \ref{f:Missed_Vent_1}.  Then the robot moved forward and was stuck on the corner underneath the vent, due to the localization error, and this is when the Supervisor returned attention to the robot.  The FPV images were stored for later review by the Supervisor, but due to workload, these images were never reviewed during the mission, and thus the artifact remained unscored.

\textbf{L05 Vent}
Approximately 7 minutes prior to the end of the mission, D02 reported a vent and transmitted the image in Figure \ref{f:Missed_Vent_2} to the base station.  Unfortunately, due to workload and poor notification design in the GUI, the human supervisor never noticed the report, and thus it was not submitted.  Although the detection system identified the bucket as a vent, the Supervisor could easily see the actual vent above it and the reported position was 1.24m from the actual position.

\sidebyside
{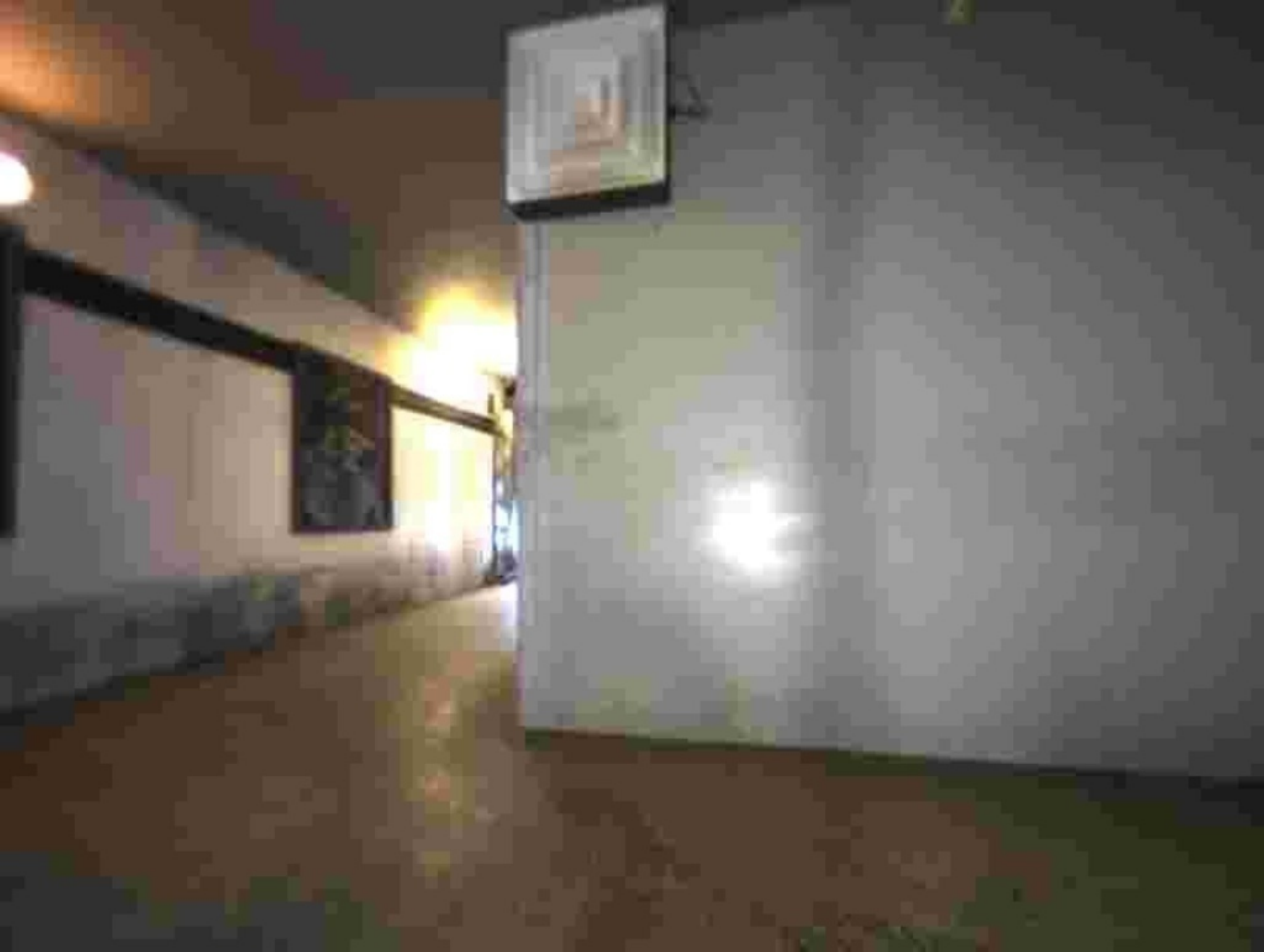}{Missed_Vent_1}{Vent seen by D01 while stopped and transmitted to human supervisor, but missed due to human supervisor workload.}
{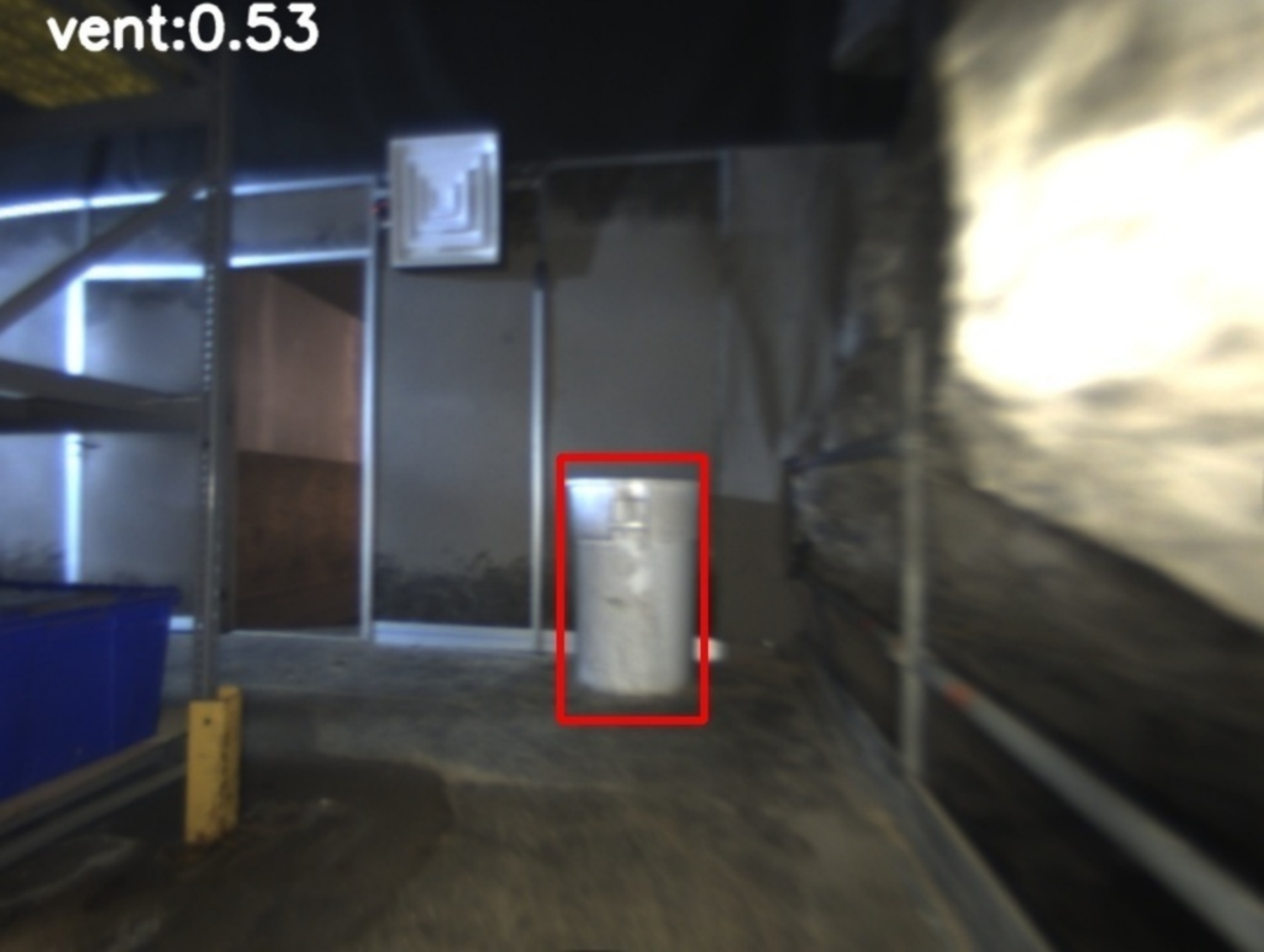}{Missed_Vent_2}{Vent detected and reported by D02 late in mission but missed by the human supervisor due to workload.}

\textbf{L62 Helmet}
L62 helmet located in the cave section near the tunnel intersection was seen via FPV, as in Figure \ref{f:Missed_Helmet} but not detected by the robot.  This was transmitted by D01 and also available for review by the human supervisor, but due to workload the images were not reviewed during the mission.

\textbf{L13 Gas}
Approximately 10 minutes prior to the end of the mission, D02 passed approximately within 1m of this gas artifact, but provided no reports.  Further analysis shows the only gas detection D02 had was a false report in an area with no CO2 nearby.  The only other robot to go near this gas was D01, but it only went near a doorway leading to the area where the gas was located, and had already reported and scored another gas artifact near that location, so if it did detect L13 it would have assumed it was the same as the previous artifact.

\textbf{L21 Vent}
This vent was in the subway platform area of the environment.  Both Spot robots viewed the vent with various cameras, but never detected it, likely due to the white wall background.  Additionally, communications to the base station were limited in this area, and none of the FPV images relayed to the human supervisor had the vent in view.  Interestingly, D01 did provide a false report of a vent in this area, and transmitted an image seen in Figure \ref{f:Incorrect_Vent}, which the human supervisor discarded as a false report.  According to the truth data, the location reported was 3.93m from the actual vent, and so would have scored if submitted. However, visual analysis indicates the vent position in the ground truth file appears 2m off, which would not have scored.

\textbf{L42 Fire Extinguisher}
This fire extinguisher was seen only with the right-facing camera on D01 for only a few frames, as seen in Figure \ref{f:Missed_Extinguisher}, which was not enough to trigger a detection and report.  The robot was outside of communications, so even if the front camera had seen it, it would not have been available for the human supervisor.

\sidebyside
{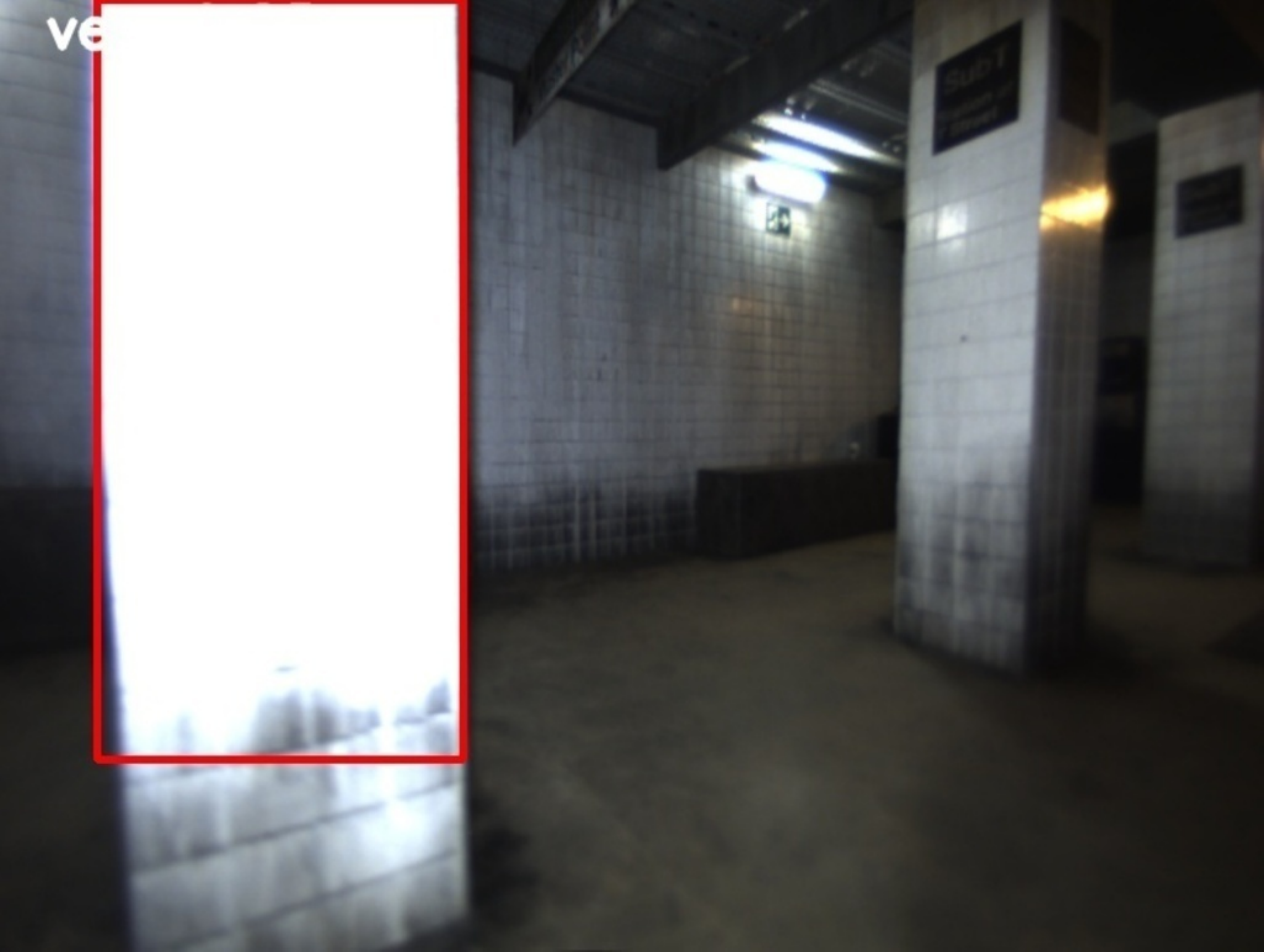}{Incorrect_Vent}{False vent reported close to actual vent by D01.}
{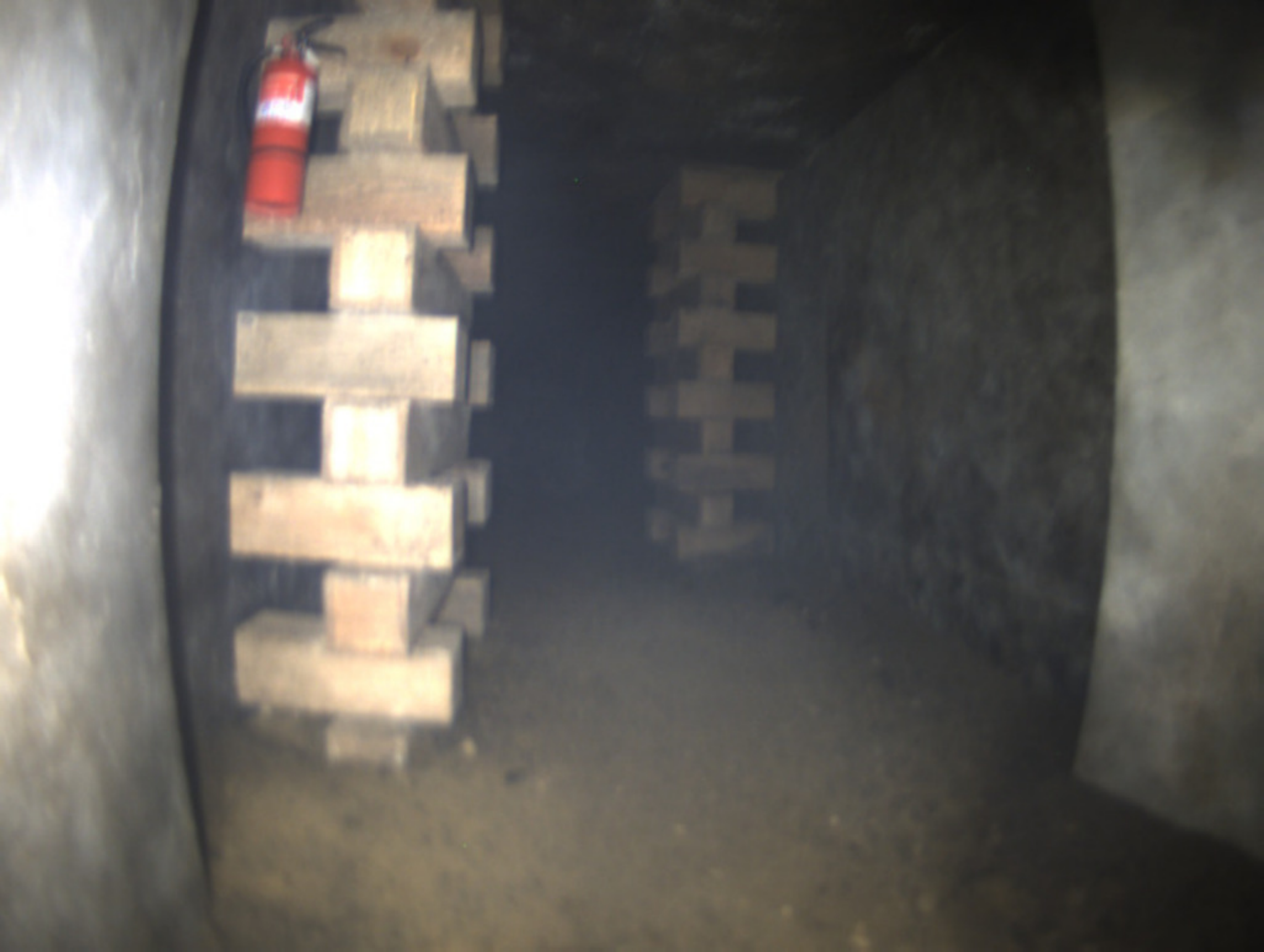}{Missed_Extinguisher}{Missed fire extinguisher only seen for a few frames by the right camera of D01.}

\subsection{Two False Artifacts}
\label{ssec:sup_false_artifacts}

This section focuses on two false artifacts that Team MARBLE reported, but were actually the result of false detections. Each false artifact has at least one false attempt (FA) associated with it. These false attempts are listed in Table \ref{tab:all_reports_robot_human_scorer}.

\textbf{Gas (FA1, FA2, FA5)}: Gas was autonomously reported by H02 early in the mission (FA1). The HS modified the location and manually reported again (FA2), but did not score. Later in the mission, D02 reported gas again in a very similar location as D01. This increased the confidence in the HS that gas was in the area, so the HS modified the location of D02 report, but this too did not score (FA4). It remains unknown why both agents detected elevated levels of CO$_2$, but it is likely that a source in that vicinity existed, even if it was not an gas artifact.

\textbf{Fire Extinguisher (FA3)}: Due to an unknown cause, the HS inadvertantly submitted the same report twice, which was a fire extinguisher at xyz-coordinates of (20.60, 20.59, -2.64). Because the fire extinguisher was scored by the last report, this report did not score, but simply wasted a report.

\textbf{Backpack (FA4)}: The image was not clear, but the HS attempted to report it, and it did not score.

\subsection{Tabulated Reports}
\label{ssec:sup_all_reports}

All reports that Team MARBLE submitted to DARPA are listed in Table \ref{tab:all_reports_robot_human_scorer}.

\begin{table}[hbt!]
\begin{center}
    \begin{tabular}{c c c c c c c c}
    \hline
    \textbf{Report} & \textbf{ID} & \textbf{Type} & \textbf{Error} & \textbf{Score} & \textbf{Time} & \textbf{Scorer} & \textbf{Assister}\\ 
    \textbf{} & \textbf{} & \textbf{} & \textbf{[m]} & \textbf{} & \textbf{[mm:ss]} & \textbf{} & \textbf{} \\ 
    \hline
    
    \textbf{SA1}    & 
    \textbf{L51}    &
    \raisebox{-0.4\totalheight}{\includegraphics[height=0.0325\textwidth]{figures/artifact_thumbnails/drill}} &   
    \textbf{0.57}   &
    \textbf{1}      &
    \textbf{01:08}  &
    \textbf{D02}    \\
    
    \textbf{SA2}    & 
    \textbf{L53}    &
    \raisebox{-0.4\totalheight}{\includegraphics[height=0.0325\textwidth]{figures/artifact_thumbnails/backpack}} &
    \textbf{2.23}   &
    \textbf{2}      &
    \textbf{01:23}  &
    \textbf{D02}    \\
    
    \textbf{SA3}    & 
    \textbf{L55}    &
    \raisebox{-0.4\totalheight}{\includegraphics[height=0.0325\textwidth]{figures/artifact_thumbnails/rope}} &
    \textbf{0.84}   &
    \textbf{3}      &
    \textbf{06:23}  &
    \textbf{HS\textsuperscript{\textdagger}} & 
    \textbf{D02 (live FPV stream)} \\ 
    
    \textbf{SA4}    & 
    \textbf{L26}    &
    \raisebox{-0.4\totalheight}{\includegraphics[height=0.0325\textwidth]{figures/artifact_thumbnails/survivor}} &
    \textbf{0.62}   &
    \textbf{4}      &
    \textbf{12:03}  &
    \textbf{D01}    \\
    
    MA1             &
    L59             &
    \raisebox{-0.4\totalheight}{\includegraphics[height=0.0325\textwidth]{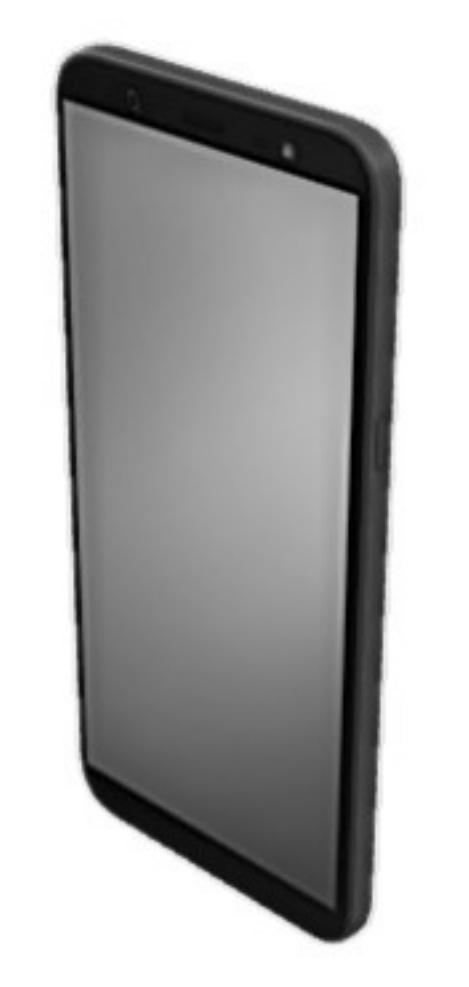}} &
    13.69           &
                    &
    14:09           &
    D02             \\
    
    MA2     &
    L22     &
    \raisebox{-0.4\totalheight}{\includegraphics[height=0.0325\textwidth]{figures/artifact_thumbnails/cellphone_bw}} &   
    19.47   &       
            &   
    14:13   &   
    D01     \\
    
    MA3     &
    L59     &   
    \raisebox{-0.4\totalheight}{\includegraphics[height=0.0325\textwidth]{figures/artifact_thumbnails/cellphone_bw}} &
    7.74    &      
            &   
    14:57   &   
    HS      & 
    D02 (MA1)       \\
    
    \textbf{SA5}    &
    \textbf{L32}    &   
    \raisebox{-0.4\totalheight}{\includegraphics[height=0.0325\textwidth]{figures/artifact_thumbnails/survivor}} &   
    \textbf{1.40}   &   
    \textbf{5}      &   
    \textbf{16:35}  &   
    \textbf{HS}     &
    \textbf{D02 (live FPV stream)}  \\
    
    \textbf{SA6}    &
    \textbf{L08}    &   
    \raisebox{-0.4\totalheight}{\includegraphics[height=0.0325\textwidth]{figures/artifact_thumbnails/gas}} &   
    \textbf{1.80}   &   
    \textbf{6}      &   
    \textbf{17:23}  &   
    \textbf{D01}    \\
    
    FA1     &
    ---     &   
    \raisebox{-0.4\totalheight}{\includegraphics[height=0.0325\textwidth]{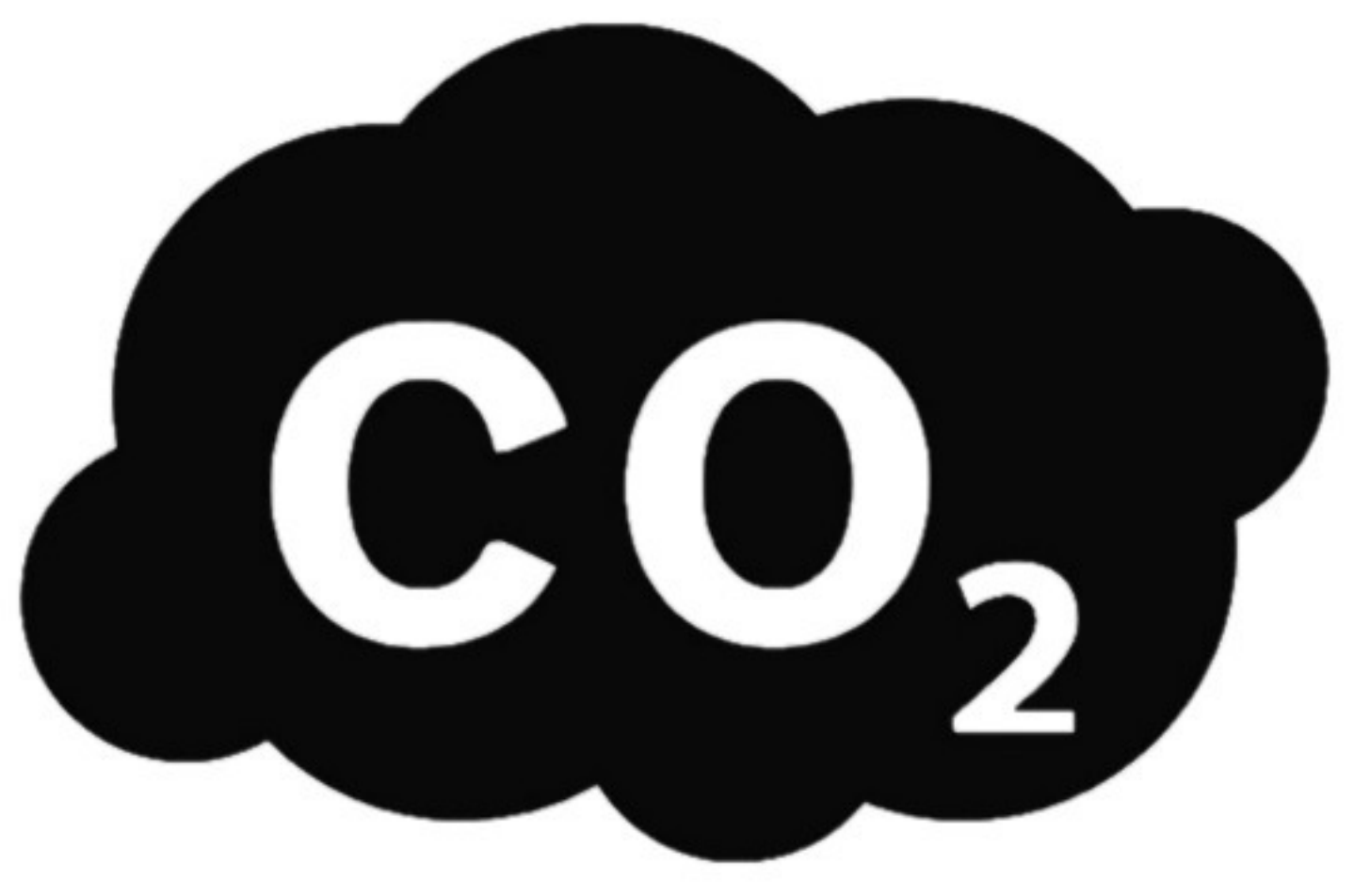}} &   
    ---     &       
            &   
    17:33   &   
    H02     \\
    
    FA2     &
    ---     &   
    \raisebox{-0.4\totalheight}{\includegraphics[height=0.0325\textwidth]{figures/artifact_thumbnails/gas_bw}} &   
    ---     &       
            &  
    17:59   &   
    HS & H02 (FA1)   \\
    
    MA4     &
    L22     &   
    \raisebox{-0.4\totalheight}{\includegraphics[height=0.0325\textwidth]{figures/artifact_thumbnails/cellphone_bw}} &
    7.77    &
            &   
    18:38   &
    H02     \\
    
    MA5     &
    L11     &
    \raisebox{-0.4\totalheight}{\includegraphics[height=0.0325\textwidth]{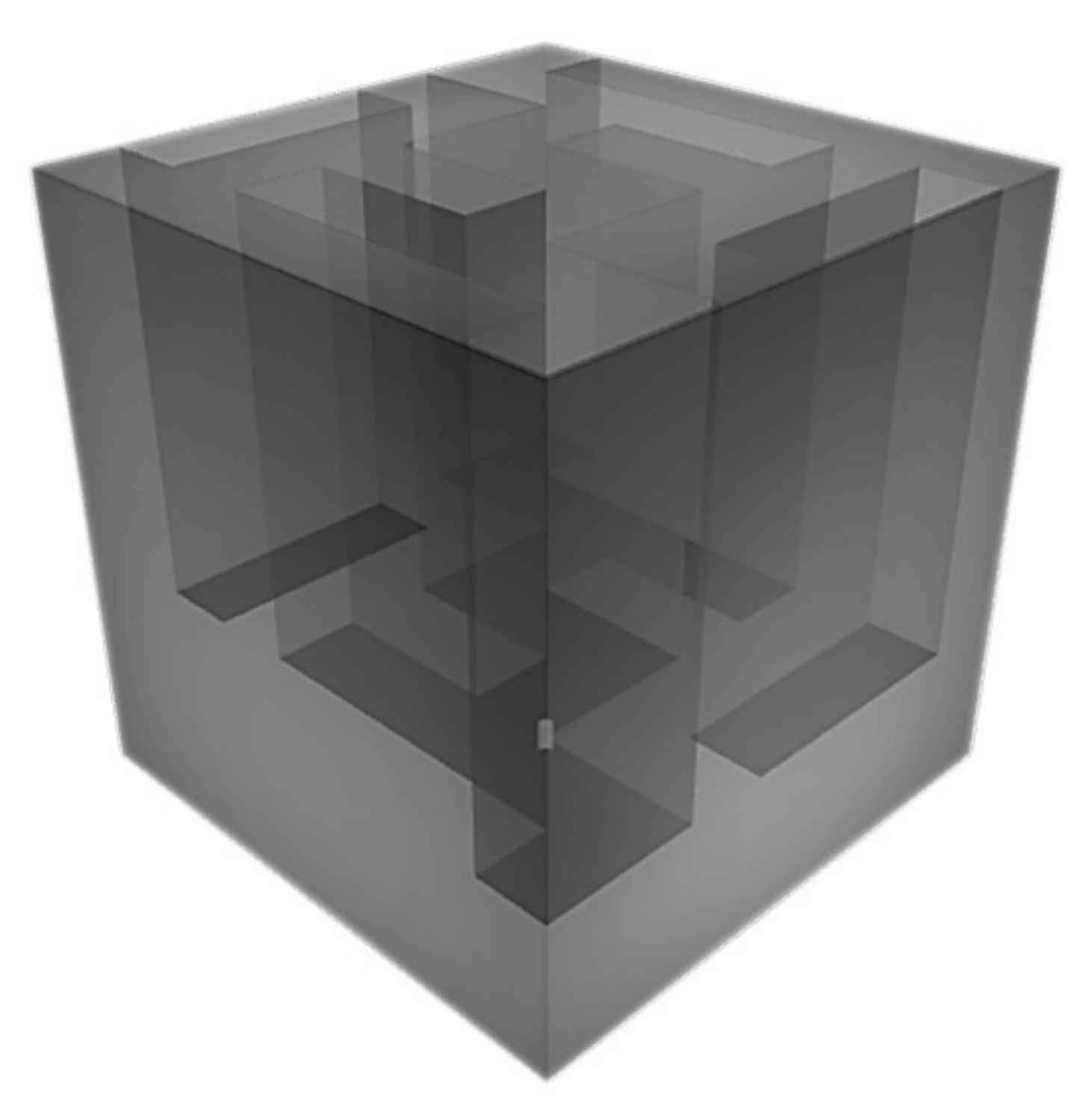}} &
    10.73   &       
            &   
    19:49   &   
    D01     \\
    
    \textbf{SA7}    &
    \textbf{L31}    &
    \raisebox{-0.4\totalheight}{\includegraphics[height=0.0325\textwidth]{figures/artifact_thumbnails/fire_extinguisher}} &   
    \textbf{1.31}   &   
    \textbf{7}      &   
    \textbf{28:08}  &   
    \textbf{HS\textsuperscript{\textdagger}}    & 
    \textbf{D01 (live FPV stream)}              \\
    
    FA3             &
    ---             &   
    \raisebox{-0.4\totalheight}{\includegraphics[height=0.0325\textwidth]{figures/artifact_thumbnails/fire_extinguisher}} &   
    ---             &   
                    &   
    33:38           &   
    HS              & 
                    \\
    
    \textbf{SA8}    &
    \textbf{L34}    &
    \raisebox{-0.4\totalheight}{\includegraphics[height=0.0325\textwidth]{figures/artifact_thumbnails/drill}} &   
    \textbf{1.43}   &   
    \textbf{8}      &
    \textbf{35:51}  &   
    \textbf{D01\textsuperscript{\textdagger}} \\
    
    \textbf{SA9}    &
    \textbf{L38}    &
    \raisebox{-0.4\totalheight}{\includegraphics[height=0.0325\textwidth]{figures/artifact_thumbnails/fire_extinguisher}} &   
    \textbf{2.82}   &
    \textbf{9}      &
    \textbf{36:53}  &
    \textbf{HS\textsuperscript{\textdagger}} &
    \textbf{D01 (live FPV stream)}           \\
    
    \textbf{SA10}   &
    \textbf{L36}    &
    \raisebox{-0.4\totalheight}{\includegraphics[height=0.0325\textwidth]{figures/artifact_thumbnails/cube}} &
    \textbf{3.94}   &
    \textbf{10}     &
    \textbf{37:08}  &
    \textbf{D01\textsuperscript{\textdagger}} \\
    
    \textbf{SA11}   &
    \textbf{L40}    &
    \raisebox{-0.4\totalheight}{\includegraphics[height=0.0325\textwidth]{figures/artifact_thumbnails/backpack}} &
    \textbf{1.40}   &
    \textbf{11}     &
    \textbf{37:58}  &
    \textbf{D01\textsuperscript{\textdagger}} \\
    
    \textbf{SA12}   &
    \textbf{L67}    &
    \raisebox{-0.4\totalheight}{\includegraphics[height=0.0325\textwidth]{figures/artifact_thumbnails/rope}} &
    \textbf{2.87}   &
    \textbf{12}     &
    \textbf{38:47}  &
    \textbf{D01\textsuperscript{\textdagger}}    \\
    
    FA4     &
    ---     &
    \raisebox{-0.4\totalheight}{\includegraphics[height=0.0325\textwidth]{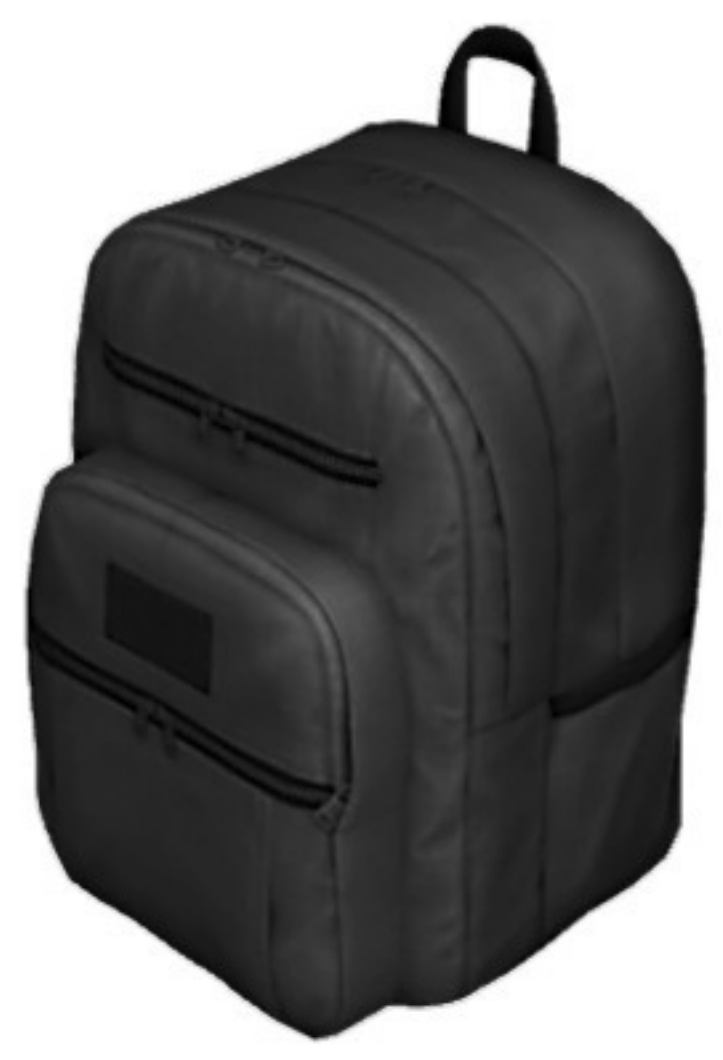}} &   
    ---     &
            &   
    45:18   &
    D02     \\
    
    FA5     &
    ---     &
    \raisebox{-0.4\totalheight}{\includegraphics[height=0.0325\textwidth]{figures/artifact_thumbnails/gas_bw}} &   
    ---     &
            &   
    46:44   &
    HS      &
    H02 (FA1), HS (FA2) \& D02    \\
    
    MA6     &
    L11     &   
    \raisebox{-0.4\totalheight}{\includegraphics[height=0.0325\textwidth]{figures/artifact_thumbnails/cube_bw}} &   
    9.42    &       
            &   
    47:31   &   
    D02     \\
    
    \textbf{SA13}   &
    \textbf{L11}    &   
    \raisebox{-0.4\totalheight}{\includegraphics[height=0.0325\textwidth]{figures/artifact_thumbnails/cube}} &   
    \textbf{1.83}   &   
    \textbf{13}     &   
    \textbf{47:53}  &   
    \textbf{HS}     & 
    \textbf{D01} (MA5), \textbf{D02} (MA6) \\
    
    MA7     &
    L64     &   
    \raisebox{-0.4\totalheight}{\includegraphics[height=0.0325\textwidth]{figures/artifact_thumbnails/cube_bw}} &
    8.24    &
            &  
    48:12   &
    D01     \\
    
    MA8     &
    L64     &   
    \raisebox{-0.4\totalheight}{\includegraphics[height=0.0325\textwidth]{figures/artifact_thumbnails/cube_bw}} & 
    12.23   &        
            &   
    48:42   &   
    HS      & 
    D01 (MA7) \\
    
    MA9     &
    L64     &
    \raisebox{-0.4\totalheight}{\includegraphics[height=0.0325\textwidth]{figures/artifact_thumbnails/cube_bw}} &   
    20.15   &        
            &   
    49:09   &   
    HS      & 
    D01 (MA7) \& HS (MA8)  \\
    
    \textbf{SA14}   &
    \textbf{L22}    &
    \raisebox{-0.4\totalheight}{\includegraphics[height=0.0325\textwidth]{figures/artifact_thumbnails/cellphone}} &
    \textbf{4.06}   &   
    \textbf{14}     &   
    \textbf{50:33}  &   
    \textbf{H01}    &
    \textbf{D01} (MA2), \textbf{H02} (MA4), D02 \\
    
    \textbf{SA15}   &
    \textbf{L47}    &
    \raisebox{-0.4\totalheight}{\includegraphics[height=0.0325\textwidth]{figures/artifact_thumbnails/cellphone}} &
    \textbf{4.00}   &
    \textbf{15}     &
    \textbf{50:45}  &
    \textbf{D02}    \\
    
    MA10    &
    L59     &
    \raisebox{-0.4\totalheight}{\includegraphics[height=0.0325\textwidth]{figures/artifact_thumbnails/cellphone_bw}} &
    9.07    &
            &   
    50:57   &
    D01     \\
    
    \textbf{SA16}   &
    \textbf{L59}    &   
    \raisebox{-0.4\totalheight}{\includegraphics[height=0.0325\textwidth]{figures/artifact_thumbnails/cellphone}} &
    \textbf{3.15}   &
    \textbf{16}     &
    \textbf{51:48}  &
    \textbf{HS}     & 
    \textbf{D02} (MA1), \textbf{HS} (MA3), \textbf{D01} (MA10) \\
    
    MA11    &
    L24     &
    \raisebox{-0.4\totalheight}{\includegraphics[height=0.0325\textwidth]{figures/artifact_thumbnails/gas_bw}} &   
    5.38    &        
            &   
    52:20   &   
    H01     \\
    
    \textbf{SA17}   &
    \textbf{L24}    &   
    \raisebox{-0.4\totalheight}{\includegraphics[height=0.0325\textwidth]{figures/artifact_thumbnails/gas}} &   
    \textbf{2.55}   &   
    \textbf{17}     &   
    \textbf{52:45}  &   
    \textbf{H02}    \\
    
    \textbf{SA18}   &
    \textbf{L58}    &   
    \raisebox{-0.4\totalheight}{\includegraphics[height=0.0325\textwidth]{figures/artifact_thumbnails/helmet}} &   
    \textbf{1.74}   &   
    \textbf{18}     &   
    \textbf{56:33}  &   
    \textbf{HS\textsuperscript{\textdagger}} & 
    \textbf{D02 (archived FPV images)}       \\ 
    
    \hline
    \end{tabular}
    \caption{List of all artifact reports submitted by Team MARBLE during the 60-minute Final Event Prize Run. Bolded entries represent reports that resulted in a score. There are three types of reports: a \textit{Scored Attempt (SA)}, a \textit{Missed Attempt (MA)} due to error exceeding 5m, and a \textit{False Attempt (FA)} due to a false positive detection. Listed next are artifact ID, artifact type, error, cumulative score, and time since mission start. The scorer is the agent that submitted the report and scored (or attempted to score), and the assister is the agent(s) that provided information that aided the scorer in scoring (or attempting to score). The reporting of these artifacts was completely autonomous, save artifacts scored by the HS as well as those with a (\textsuperscript{\textdagger}), denoting artifacts that were seen as a result of the HS temporarily teleoperating the agent into new areas of the course.}
    \label{tab:all_reports_robot_human_scorer}
\end{center}
\end{table}

\subsection{Field Deployments}
\label{ssec:sup_testing_and_validation}

Table \ref{tab:deployments} lists the dates and locations of all full-scale deployments, within the context of the events at the DARPA SubT Challenge.

\begin{table}[!htb]
\begin{center}
    \begin{tabular}{c c c c c}
    \hline
    \textbf{Date} & \textbf{Deployment} & \textbf{Environment} & \textbf{Onsite} & \textbf{Location}\\ 
    \hline
    \textit{Apr 7, 2019} & \textit{\textbf{STIX Event}} & \textit{Edgar Experimental Mine} &  & \textit{Idaho Springs, CO} \\
    \hline
    Jul 29, 2019    &   Pre-Tunnel 1    & Edgar Experimental Mine           &       &   Idaho Springs, CO \\
    Aug 1, 2019     &   Pre-Tunnel 2    & Edgar Experimental Mine           &       &   Idaho Springs, CO \\
    Aug 6, 2019     &   Pre-Tunnel 3    & Edgar Experimental Mine           &       &   Idaho Springs, CO \\
    Aug 9, 2019     &   Pre-Tunnel 4    & Edgar Experimental Mine           &       &   Idaho Springs, CO \\
    Aug 12, 2019    &   Pre-Tunnel 5    & Edgar Experimental Mine           &       &   Idaho Springs, CO \\
    \hline
    \textit{Aug 17, 2019} & \textit{\textbf{Tunnel Event}} & \textit{NIOSH Exp. \& Safety Research Mines} &  & \textit{Pittsburgh, PA} \\
    \hline
    Feb 4, 2020     &   Pre-Urban 1     & Geotech Warehouse                 &       &   Denver, CO \\
    Feb 8, 2020     &   Pre-Urban 2     & Geotech Warehouse                 &       &   Denver, CO \\
    Feb 12, 2020    &   Pre-Urban 3     & Geotech Warehouse                 &       &   Denver, CO \\
    \hline
    \textit{Feb 21, 2020} & \textit{\textbf{Urban Event}} & \textit{Satsop Nuclear Power Plant} &  & \textit{Elma, WA} \\
    \hline
    Aug 4, 2020     &   Pre-Cave 1      & Edgar Experimental Mine           &       &   Idaho Springs, CO \\
    Sep 17, 2020    &   Pre-Cave 2      & Eng. Center (L1)                  &   X   &   Boulder, CO \\
    Sep 19, 2020    &   Pre-Cave 3      & Eng. Center (L1)                  &   X   &   Boulder, CO \\
    \hline
    \textit{Sep 21, 2020} & \textit{\textbf{Cave Event*}} & \textit{Edgar Experimental Mine} &  & \textit{Idaho Springs, CO} \\
    \hline
    Jul 13, 2021    &   Pre-Final 1     & Folsom Parking Garage             &   X   &   Boulder, CO \\
    Jul 14, 2021    &   Pre-Final 2     & Folsom Parking Garage             &   X   &   Boulder, CO \\
    Jul 15, 2021    &   Pre-Final 3     & Edgar Experimental Mine           &       &   Idaho Springs, CO \\
    Aug 13, 2021    &   Pre-Final 4     & Folsom Parking Garage             &   X   &   Boulder, CO \\
    Aug 17, 2021    &   Pre-Final 5     & Eng. Center (LL \& Courtyard)     &   X   &   Boulder, CO \\
    Aug 19, 2021    &   Pre-Final 6     & Sust., Energy, and Env. Community &   X   &   Boulder, CO \\
    Aug 24, 2021    &   Pre-Final 7     & Eng. Center (LL \& Courtyard)     &   X   &   Boulder, CO \\
    Aug 26, 2021    &   Pre-Final 8     & Edgar Experimental Mine           &       &   Idaho Springs, CO \\
    Sep 1, 2021     &   Pre-Final 9     & Edgar Experimental Mine           &       &   Idaho Springs, CO \\
    Sep 8, 2021     &   Pre-Final 10    & Eng. Center (LL \& Courtyard)     &   X   &   Boulder, CO \\
    Sep 10, 2021    &   Pre-Final 11    & Eng. Center (LL \& Courtyard)     &       &   Boulder, CO \\
    Sep 12, 2021    &   Pre-Final 12    & Eng. Center (L2) + Rustandy       &   X   &   Boulder, CO \\
    Sep 14, 2021    &   Pre-Final 13    & Eng. Center (L2) + Rustandy       &   X   &   Boulder, CO \\
    \hline
    \textit{Sep 21, 2021} & \textit{\textbf{Final Event}} & \textit{Louisville, Megacavern} &  & \textit{Louisville, KY} \\
     \hline
    \end{tabular}
\caption{\label{tab:deployments} List of all full-scale field deployments. For the Final Event, Team MARBLE gave greater resources to system performance validation and human-robot teaming practice. This realized itself as  more frequent and diverse field deployments. For the circuit events, just two to three weeks were spent on field deployments, whereas for the Final Event, the team devoted two months. Instead of practicing in just one or two environments, the team was asked to perform in five unique environments. Selecting locations that were "onsite" University of Colorado Boulder campus enabled the team to be more nimble and operationally efficient. The (*) denotes that the Cave Event was a self-managed mock event.}
\end{center}
\end{table}

\end{document}